\documentclass{article}

\usepackage{microtype}
\usepackage{graphicx}
\usepackage{subfig}
\usepackage{booktabs} 

\usepackage{hyperref}

\usepackage[accepted]{icml2024}

\usepackage{amsmath}
\usepackage{amssymb}
\usepackage{mathtools}
\usepackage{amsthm}

\DeclareMathOperator*{\argmax}{argmax}

\usepackage[capitalize,noabbrev]{cleveref}

\theoremstyle{plain}

\theoremstyle{definition}

\theoremstyle{remark}

\usepackage[textsize=tiny]{todonotes}

\icmltitlerunning{}

\renewcommand{\vec}[1]{\mathbf{#1}}

\newcommand{\mtx}[1]{\mathbf{#1}}

\begin{document}

\twocolumn[
\icmltitle{How Do the Architecture and Optimizer Affect Representation Learning? \\ On the Training Dynamics of Representations in Deep Neural Networks}



\icmlsetsymbol{equal}{*}

\begin{icmlauthorlist}
\icmlauthor{Yuval Sharon}{yyy}
\icmlauthor{Yehuda Dar}{yyy}
\end{icmlauthorlist}

\icmlaffiliation{yyy}{Department of Computer Science, Ben-Gurion University of the Negev, Beer Sheva, Israel}

\icmlcorrespondingauthor{Yuval Sharon}{shyuva@post.bgu.ac.il}
\icmlcorrespondingauthor{Yehuda Dar}{ydar@bgu.ac.il}

\icmlkeywords{Machine Learning, ICML}

\vskip 0.3in
]



\printAffiliationsAndNotice{}  

\begin{abstract}
    In this paper, we elucidate how representations in deep neural networks (DNNs) evolve during training. Our focus is on overparameterized learning settings where the training continues much after the trained DNN starts to perfectly fit its training data. We examine the evolution of learned representations along the entire training process. We explore the representational similarity of DNN layers, each layer with respect to \textit{its own representations throughout the training process}. For this, we use two similarity metrics: (1) The centered kernel alignment (CKA) similarity; (2) Similarity of decision regions of linear classifier probes that we train for the DNN layers. We visualize and analyze the decision regions of the DNN output and the layer probes during the DNN training to show how they geometrically evolve. Our extensive experiments discover training dynamics patterns that can emerge in layers depending on the relative layer-depth, architecture and optimizer. Among our findings: (i) The training phases, including those related to memorization, are more distinguishable in SGD training than in Adam training, and for Vision Transformer (ViT) than for ResNet; (ii) Unlike ResNet, the ViT layers have synchronized dynamics of representation learning.    
\end{abstract}



\section{Introduction}
Deep neural networks (DNNs) have achieved remarkable practical success in various applications. The ability to learn representations for each of the DNN layers is an essential aspect that determines the overall DNN performance.  Yet, the inner workings of the representation learning process are still far from being sufficiently understood. An important question that we address in this paper is \textit{how do the training dynamics of the representations behave in each of the DNN layers?}

Usually, DNNs are \textit{overparameterized} models, namely, they have many more learnable parameters than training examples. Due to such overparameterization, classification DNNs are often trained to perfectly fit their training data, i.e., to correctly label all their training examples (according to the pairings in the given dataset). This perfect fitting property stands in sharp contrast to conventional machine learning guidelines that associate overfitting (and, hence, perfect fitting) with poor generalization performance. Yet, DNNs that perfectly fit their training data often show good generalization performance, raising foundational questions on their inner workings.

In this paper, we ask \textbf{how do the DNN architecture and optimization method affect the evolution of representations in various DNN layers during training?}
To address this question, we evaluate how similar are the representations that evolve at each layer \textit{along the entire training process}. Namely, for a layer of interest, we provide a detailed map that shows how similar are the representations of that layer at every two different time points (epochs) in the training. We believe that our evaluations provide the most extensive and detailed elucidation of representational similarity in the training process (as we discuss in Section \ref{sec:Related Work}, some previous works had a very limited evaluation of representational similarity in training with respect to only a particular training time-point).

In our evaluations, we use two different approaches to quantify representational similarity:
\begin{itemize}
    \item The centered kernel alignment (CKA) \citep{kornblith2019similarity}, which is a popular measure of representational similarity between layers of different DNNs or within the same DNN. 
    \item We propose a new approach to measure representational similarity by using linear classifier probes \citep{alain2017understanding} for the DNN layers during their training process. We use the linear classifier probes of the same layer at different training epochs and evaluate the similarity of their decision regions as a measure of similarity between the representations at these training epochs. Our layerwise decision regions similarity (DRS) approach allows interpretability by visualizing decision regions that participated in the similarity evaluation at different epochs (e.g., Figs.~\ref{fig:cifar10_resnet_first_layer_probe_vis_main_paper}, \ref{fig:ViT_CIFAR10_plane105_visual_for_main_paper}). 
\end{itemize}

Using our detailed representational similarity evaluation, we examine overparameterized learning settings where the training continues much after the DNN starts to perfectly fit its training data. By this, we elucidate the evolution of representations at the principal phases of training: 
\begin{itemize}
    \item Phase I: Learning of general representations that do not overfit.
    \item Phase II: Refining representations in order to memorize mislabeled and/or atypical examples and achieving perfect fitting of the entire training dataset.
    \item Phase III: Additional evolution of representations while maintaining the perfect fitting. This evolution might improve the representations but not necessarily. 
\end{itemize} 

Our representational similarity diagrams emphasize these three training phases as \textit{similarity blocks} where, in each, the representations are relatively similar but differ from the other training phases (thus the similarity blocks visually appear as lighter color blocks along the main diagonal of the similarity diagrams, e.g., see Fig.~\ref{fig:cka_comparison_vit_b_16_noise_0}).  
We further explore the training dynamics by the DRS approach and its visualization capability (e.g., Figs.~\ref{fig:cifar10_resnet_first_layer_probe_vis_main_paper}, \ref{fig:ViT_CIFAR10_plane105_visual_for_main_paper}).
All of these shed new light on representation learning in overparameterized models, including the following insights: 
\begin{enumerate}

\item 
Training phases appear as similarity blocks in more of the DNN layers,  and are more clearly identified (implying relatively-faster transition between phases and/or greater dissimilarity between representations of different training phases), in 
\begin{enumerate}
    \item ViT than in ResNet
    \item SGD training than in Adam training
\end{enumerate}

\item Training dynamics of representations are synchronized across the layers in ViT, but not in ResNet.
    
    \item Representations at the \textit{first layer}, in the perfect fitting regime of Adam training of a sufficiently wide DNN, are more similar to random representations (e.g., as at the initialized, untrained DNN) than to representations at the best early stopping epochs before perfect fitting starts. This phenomenon is consistently observed in Adam training of convolutional networks (such as ResNet or a two-layer model) but not for SGD and ViT. We show that the adaptive normalization of gradients by Adam is necessary for this phenomenon to occur.

        \item The decision regions of the DNN output and the deeper-layer probes are likely to become more fragmented during the training phase of memorizing atypical examples (in noiseless experiments), and then the fragmentation level stabilizes in the perfect fitting training phase (possibly with some reduction in fragmentation before the stabilization). This behavior is observed more clearly for ViT, also due to its synchronized training dynamics across its layers, showing that the implicit regularization of the optimizer during training can significantly depend on the DNN architecture.  

\end{enumerate}

\section{Related Work}
\label{sec:Related Work}


\subsection{Representational Similarity Throughout the Training Process}
\label{subsec:Related Work - Representational Similarity Throughout the Training Process}

\citet{raghu2017svcca} proposed the SVCCA metric for representational similarity and used it for several analyses, including showing how the representational similarity between the model during training and the model at the end of training evolves during training of a convolutional model and a ResNet model. \citet{liu2021autofreeze} showed the evolution of the SVCCA metric during 4 epochs of a BERT model training.  
\citet{gotmare2018closer} analyze CCA representational similarity among few training time-points in a VGG model training (see more details in Appendix \ref{appendix:subsec:Related Work - Representational Similarity Throughout the Training Process}),  but they do not show or analyze the detailed evolution of the representational similarity during the entire training process.
In this paper, we examine the evolution of representational similarity much more extensively than previous works, we consider various new settings (such as for a ViT architecture). 
In contrast to previous works, \textit{in our paper here we examine the representational similarity in much more detail, enabling to understand the representational similarity of various pairs of epochs, throughout a long training process with a small constant learning rate that does not affect the natural training dynamics. Moreover, we focus here on understanding how the representational similarity evolves in overparameterized learning where perfect fitting does not prevent good generalization.} 

In Appendix \ref{appendix:subsec:Related Work - Representational Similarity between Trained and Untrained Models}, we discuss the related works by \citet{kornblith2019similarity,chowers2023what,hermann2020what} who studied the similarity between trained and untrained models.

\subsection{Training Phases and Memorization}
\label{subsec:Related Work - Training Phases and Memorization}

\citet{zhang2017understanding} showed that DNNs have the ability to perfectly fit (i.e., memorize) pure noise. \citet{arpit2017closer} showed that the training process learns real data patterns before memorizing the noise; and that even without artificially adding noise to real data, the training examples can vary from easy to hard in their learning difficulty -- unlike in pure noise where all examples are similarly hard to learn. 
\citet{liu2020early} showed that learning of classifiers from training data with \textit{label noise} has two phases. First, an early phase where the training examples with correct labels dominate the learning (due to dominating the overall loss gradient), which yields a relatively high prediction accuracy also for the training examples with the wrong labels (this prediction accuracy is according to the true labels that are unknown in the training for these examples). Then, the next phase memorizes the wrong labels, enabling the model to perfectly fit the training data at the expense of the high prediction accuracy (w.r.t.~the true labels) it had earlier in training.

\citet{maini2023can} showed that memorization of mislabeled or atypical training examples (identified by a method by \citet{jiang2021characterizing}) is usually not limited to the last nor other specific layers of the DNN. They showed that such memorization can be induced by a small set of neurons dispersed across the DNN layers, and that this neuron set can vary due to learning problem, training procedure, DNN architecture, etc. They based their claims using layer-wise gradient evaluation, rewinding/retraining individual layers, and optimization-based search for neurons critical for memorization.

Differently from these previous works, we elucidate the training phases and memorization from the perspective of representations at intermediate layers of the DNNs.

\subsection{Inductive Bias of DNN Architectures}

\citet{raghu2021do} analyzed the representational similarity and differences between ResNet and ViT \textit{fully trained} models, but not during their training like we do here.
\citet{somepalli2022can} studied the inductive bias of various DNNs by evaluating similarity between decision regions of the \textit{prediction output of fully-trained DNNs}, measuring their fragmentation level, and visualizing them. Specifically, they showed that convolutional architectures have less fragmented decision regions compared to architectures such as ViT. \textit{We significantly extend their analysis technique and insights, as we here study decision regions' evolution during the training process and for intermediate layers using probes.}

\subsection{The Implicit Regularization of the Adam Optimizer}

\citet{kunstner2023noise,cattaneo2024on} analyzed the implicit regularization of the Adam optimizer, which significantly differs from SGD due to the adaptive gradient normalization in Adam. We discuss the work by \citet{cattaneo2024on} in more detail in Section \ref{subsec:First Layer in Convolutional DNNs} together with our results that contribute to better understanding of the implicit regularization of Adam on the representation learning dynamics.

\section{Training Dynamics of Representational Similarity using CKA}
\label{sec:Training Dynamics of Representational Similarity via CKA}

\subsection{The CKA Measure: Definition and Formulation}
\label{subsec:The CKA Measure: Definition and Formulation}

Quantifying similarity of representations in intermediate DNN layers is known as a challenging task due to the inconsistent and intricate role the various neurons have in making the representations. The CKA was proposed by \citet{kornblith2019similarity} as a robust method for measuring similarity of representations; indeed, the CKA (based on a linear kernel) is a popular, leading choice for analyzing representations in DNNs \citep[e.g.,][]{nguyen2021do,raghu2021do}. 

Consider two classification DNNs, denoted as the functions $f,g:\mathcal{X}\rightarrow\mathcal{Y}$ where $\mathcal{X}$  is the input domain and $\mathcal{Y}$ is the output domain (set of possible labels).
We denote the mapping from the input to the representation of the $L^{\sf th}$ layer (i.e., the activations of the $L^{\sf th}$ layer) of a DNN $f$ as $f_L : \mathcal{X}\rightarrow\mathbb{R}^{p_L}$. Here, $p_L$ is the number of neurons in layer $L$ and, hence, also the dimension of the layer's representation when organized as a column vector.

The CKA is computed for two layers of possibly different DNNs, namely, $f_L: \mathcal{X}\rightarrow\mathbb{R}^{p_L}$ and $g_{L'}: \mathcal{X}\rightarrow\mathbb{R}^{p_{L'}}$ based on a set of $m$ inputs examples $\mathcal{S}_m =\{\vec{x}_1,...,\vec{x}_m\}\subset\mathcal{X}$. 
For these examples, the representations are computed and organized in the following matrices: $\mtx{F} \triangleq \left[ f_L\left(\vec{x}_1\right),...,f_L\left(\vec{x}_m\right)\right]^T$ and $\mtx{G} \triangleq \left[ g_{L'}\left(\vec{x}_1\right),...,g_{L'}\left(\vec{x}_m\right)\right]^T$.
Then,  $\mtx{K}\triangleq \mtx{F} \mtx{F}^T $ and $\mtx{L}\triangleq \mtx{G} \mtx{G}^T $ are the Gram matrices of the representations (separately for $f_L$, $g_{L'}$) whose components quantify the representation similarity between pairs of examples. 
These Gram matrices further go through centering of their rows and columns, namely, computing $\mtx{K}' = \mtx{H}\mtx{K}\mtx{H}$ and $\mtx{L}' = \mtx{H}\mtx{L}\mtx{H}$ where $\mtx{H}\triangleq \mtx{I}_{m\times m} - \frac{1}{m} \mtx{1}_{m\times m}$ is the centering matrix, $\mtx{I}_{m\times m}$ is the $m\times m$ identity matrix, and $\mtx{1}_{m\times m}$ is an $m\times m$ matrix of ones. Then, the HSIC is computed as the normalized dot product of the vectorized forms of $\mtx{K}'$, $\mtx{L}'$: ${\sf HSIC}_0 (f_L, g_{L'} ; \mathcal{S}_m) = {\sf vec}\left(\mtx{K}'\right) \cdot {\sf vec}\left(\mtx{L}'\right) / {(m-1)}$
where $\mtx{K}'$ and $\mtx{L}'$ were defined above based on $f_L$ and $g_{L'}$, respectively, and the set of examples $\mathcal{S}_m$. 
Importantly, the HSIC measure is not affected by permutation of neurons and orthogonal transformations of the representations. The important property of invariance to uniform scaling of the representations is added by the following normalization that defines the CKA measure: 
\begin{align}
    \label{eq:CKA definition}
    &{\sf CKA} (f_L, g_{L'} ; \mathcal{S}_m) = \\ \nonumber
    &~~ \frac{{\sf HSIC}_0 (f_L, g_{L'} ; \mathcal{S}_m)}{\sqrt{{\sf HSIC}_0 (f_L, f_L ; \mathcal{S}_m)} \sqrt{{\sf HSIC}_0 (g_{L'}, g_{L'} ; \mathcal{S}_m)}}. 
\end{align}

\subsection{The Proposed CKA Evaluation During Training}
\label{subsec:The Proposed CKA Evaluation During the Training Process}

\begin{figure*}[t]
  \centering
    \subfloat[]{
    \includegraphics[width=0.13\textwidth]{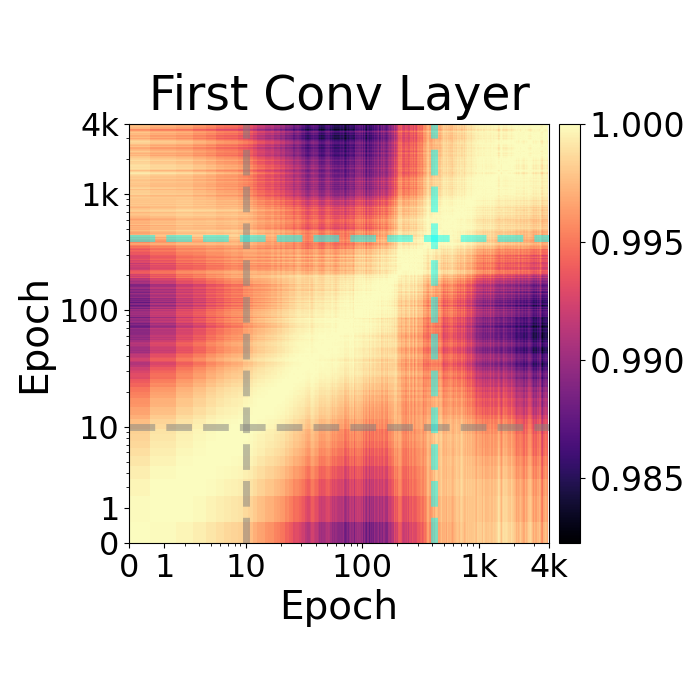}
    \label{Resnet18_k_64_first_conv_layer_noiseless}}
    \subfloat[]{
    \includegraphics[width=0.13\textwidth]{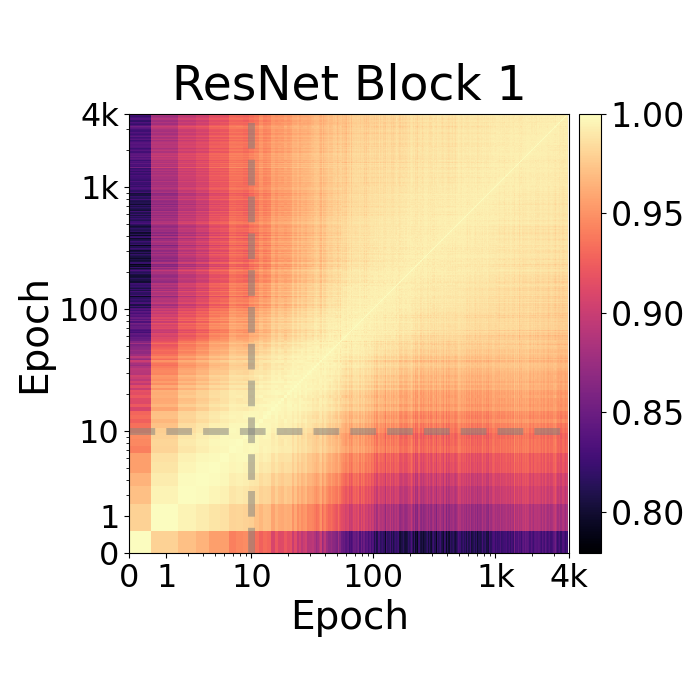}
    \label{Resnet18_k_64_block_1_noiseless}
    }
    \subfloat[]{
    \includegraphics[width=0.13\textwidth]{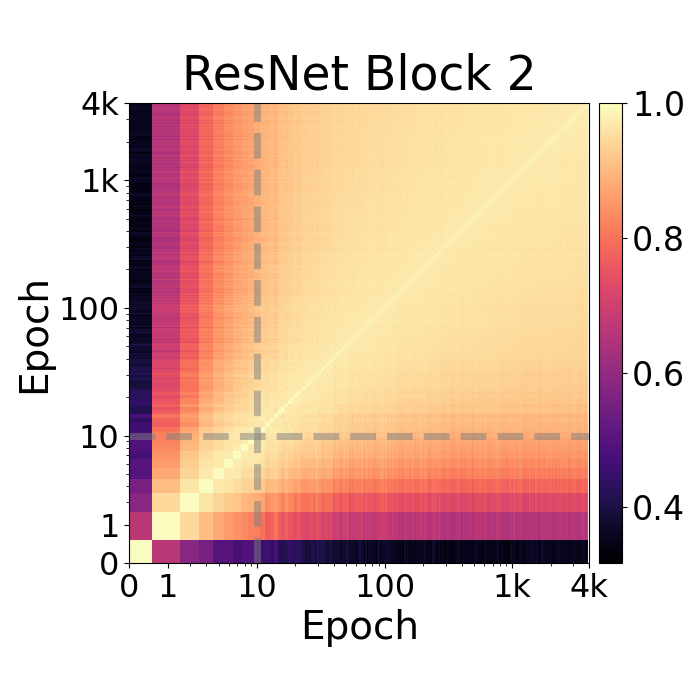}
    \label{Resnet18_k_64_block_2_noiseless}}
    \subfloat[]{
    \includegraphics[width=0.13\textwidth]{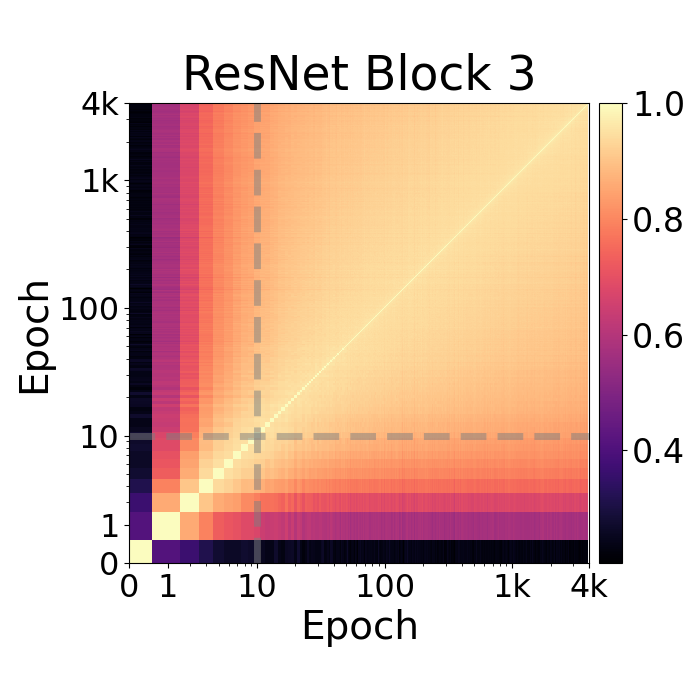}
    \label{Resnet18_k_64_block_3_noiseless}}
    \subfloat[]{
    \includegraphics[width=0.13\textwidth]{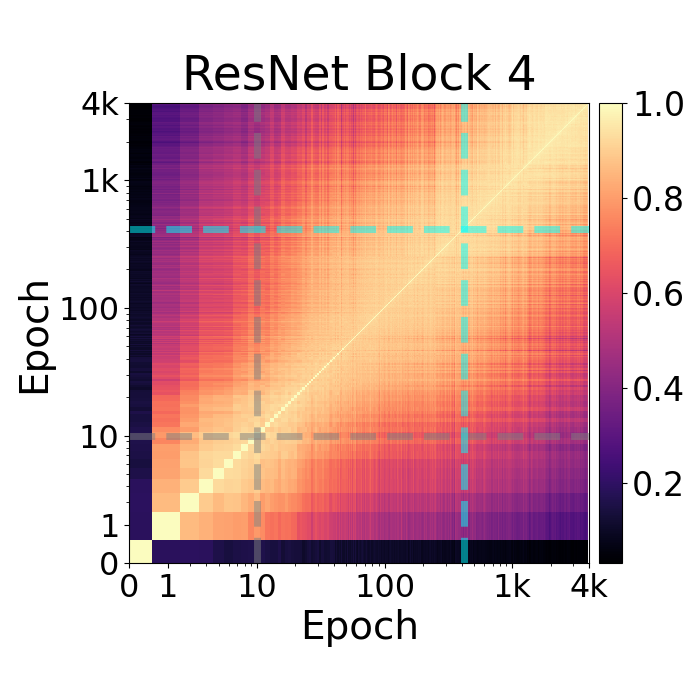} 
    \label{Resnet18_k_64_block_4_noiseless}}
     \subfloat[]{
    \includegraphics[width=0.13\textwidth]{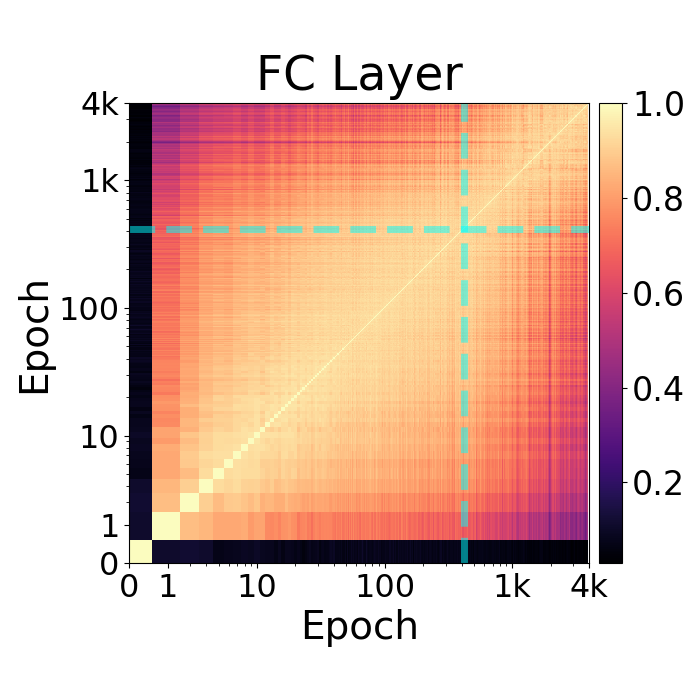}
    \label{Resnet18_k_64_FC_noiseless}}
    \subfloat[]{
    \includegraphics[width=0.13\textwidth]{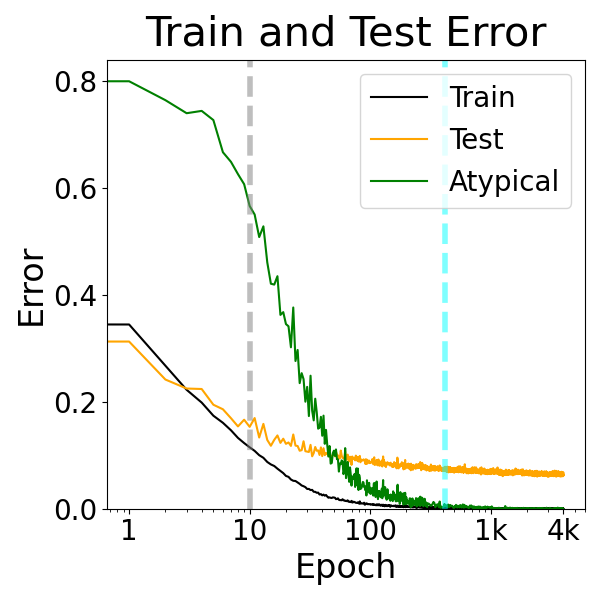}\label{ResNet18_k_64_cifar10_noiseless_error_curve}
    }
  \caption{CKA evaluations for ResNet-18, trained using Adam on CIFAR-10 without label noise. Each of the (a)-(f) subfigures shows the CKA representational similarity of a specific layer in the ResNet-18 during its training. ResNet Block $j$ refers to the representation at the end of the $j^{\sf th}$ block of layers in the ResNet-18 architecture. (g) shows train (of all the train dataset), atypical train examples, and test errors during training. Dashed lines denote transition from training phase I to II (gray) and training phase II to III (cyan).}
  \label{fig:Resnet18_k_64_noise_0}
\end{figure*}

In this section we examine the evolving representations of the DNN layers throughout the training. The similarity of representations in each layer are computed with respect to the representations of the same layer, at different time points (epochs) in the training process. 
This can be formulated as follows. The mapping from the input to the representation of the $L^{\sf th}$ layer at training epoch $t\in\{0,\dots,3999\}$ is denoted as $f_L^{(t)}: \mathcal{X}\rightarrow\mathbb{R}^{p_L}$, where $p_L$ is the dimension of the representation of layer $L$. 
Then, we compute the CKA for $f_L^{(t)}$ and $f_L^{(t')}$ for every two training epochs $t,t'\in\mathcal{T}\subset\{0,\dots,3999\}$, where $\mathcal{T}$ is a relatively dense subset of training epochs from the entire training process  (we did this grid downsampling due to the high computational and storage demands, and even then, our epoch grid is relatively dense w.r.t.~the presentation of the results using a logarithmic epoch axis; the details on the epoch grids are provided in Appendix \ref{appendix:sec:Additional Details on the Experiment Settings}). Namely, for a layer $L$, we create a 2D diagram (heatmap) of CKA values, ${\sf CKA} (f_L^{(t)}, f_L^{(t')} ; \mathcal{S}_m)$ for $t,t'\in\{0,\dots,3999\}$.

Our experiments mainly consider ResNet-18 and ViT-B/16 DNNs, training using Adam and SGD, and CIFAR-10 (without and with 20\% label noise) and SVHN datasets. The ResNet-18 was trained also on the Tiny ImageNet dataset. Unless otherwise specified, models were trained for 4000 epochs with the Adam optimizer at a constant learning rate of 0.0001 (without weight decay) that ensures smooth and natural evolution of the training process (this was also the leading setting in the deep double descent study by \citet{nakkiran2021deep}). We show results for SGD with/without momentum (e.g., Figs.~\ref{fig:sgd_0_0001_Resnet18_k_64_noise_0}, \ref{fig:sgd_0_0001_Resnet18_k_64_noise_20}, \ref{fig:sgd_0_01_Resnet18_k_64_noise_20}), and for Adam with weight decay (Fig.~\ref{fig:wd_0_001_Resnet18_k_64_noise_20}). Standard data augmentation was applied. Random initialization was applied, epoch 0 corresponds to the initialized DNN before its training starts. See Appendix \ref{appendix:sec:experiment_summary} and Table \ref{table:cka_experiment_summary} for the summary of all experiment settings. 

The resulting CKA diagrams are provided in Figs.~\ref{fig:Resnet18_k_64_noise_0}, \ref{fig:cka_comparison_vit_b_16_noise_0} and in Appendix Figs.~\ref{fig:cka_comparison_resnet18_k_64_noise_20}-\ref{fig:vit_svhn_sgd_cka_comparison_noise_0}.  
Note that the CKA of two training epochs, from the same training process, is symmetric, ${\sf CKA} (f_L^{(t)}, f_L^{(t')}; \mathcal{S}_m)={\sf CKA} (f_L^{(t')}, f_L^{(t)}; \mathcal{S}_m)$. 
Naturally, the similarity is maximal along the main diagonal that connects the lower-left and upper-right corners, as values along this diagonal correspond to similarity of a layer at a specific epoch with itself. Usually, the similarity is relatively high around the main diagonal, as representations usually evolve relatively smooth throughout the training process.

Each of the 2D diagrams is for a specific layer in the DNN and can be read according to its rows. Each row is associated with a different training epoch to consider its representation as a reference to evaluate the similarity for the entire training process, which corresponds to the horizontal axis of the diagram.
E.g., the first (bottom) row of the diagram shows how the representations throughout the entire training process are similar to the random representations at the initialization. 
With the CKA diagrams, we provide error graphs of test data (in yellow line), train data (all the train data, in black line), and the atypical train examples (in green line, available for selected experiments without artificially added label noise). We explain in Appendix \ref{appendix:subsec:Atypical Inputs Setting} how the atypical train examples were found.

The dashed lines denote the (estimated) transition points between the training phases. The gray dashed line denotes the transition from Phase I of learning general features to Phase II of memorizing mislabeled/atypical examples; this transition point occurs towards the arrival of the mislabeled/atypical error graph to its steepest decrease slope (due to the intricacy of these real training dynamics, the \textit{exact} transition point between Phase I and Phase II does not follow a clear mathematical condition for all the experiments and therefore we manually annotate it according to the CKA diagrams). 
The cyan dashed line denotes when perfect fitting of the entire training dataset starts, namely, the transition to Phase III of training; we estimate this transition point as the epoch at which the train error (of the entire dataset) starts to be below 0.001 (because accurate perfect fitting of zero train error is rare for real datasets and iterative training using mini-batches; note that the beginning of the similarity block may not necessarily exactly align with the arrival to zero train error, this behavior agrees with the deep double descent analysis by \citet{nakkiran2021deep} where the test error peak is not necessarily exactly aligned with the arrival to zero train error).

In the following, we discuss the various findings that stem from these CKA diagrams. 

\subsection{Analysis of the CKA Results}
\label{subsec:Analysis of the CKA Results}

\begin{figure*}[]
\centering
\subfloat[]{
\includegraphics[width=0.133\textwidth]{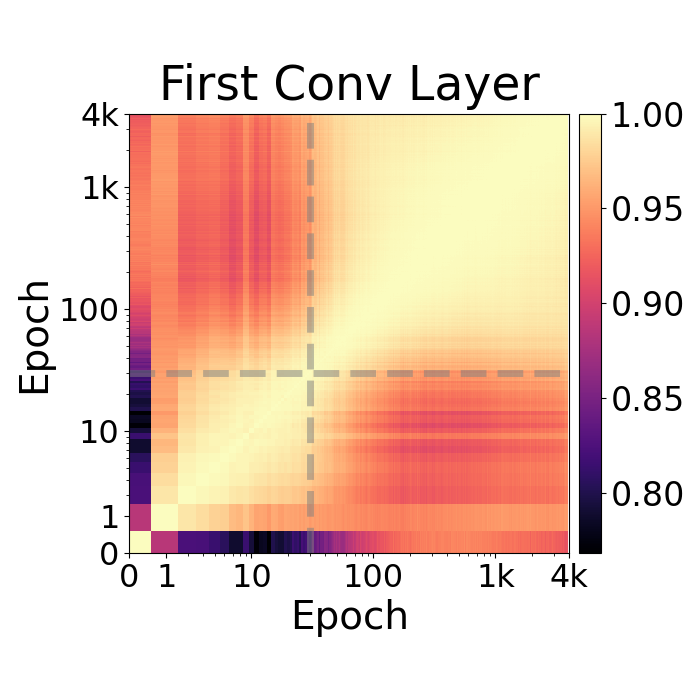}
\label{vit_first_conv_layer_noise_0}}
\subfloat[]{
\includegraphics[width=0.133\textwidth]{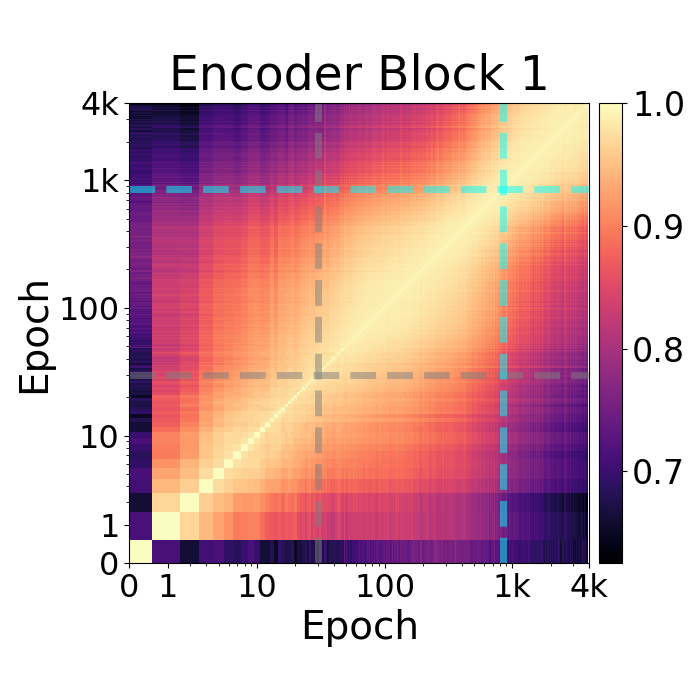}
\label{Encoder_block_1_noise_0}}
\subfloat[]{
\includegraphics[width=0.133\textwidth]{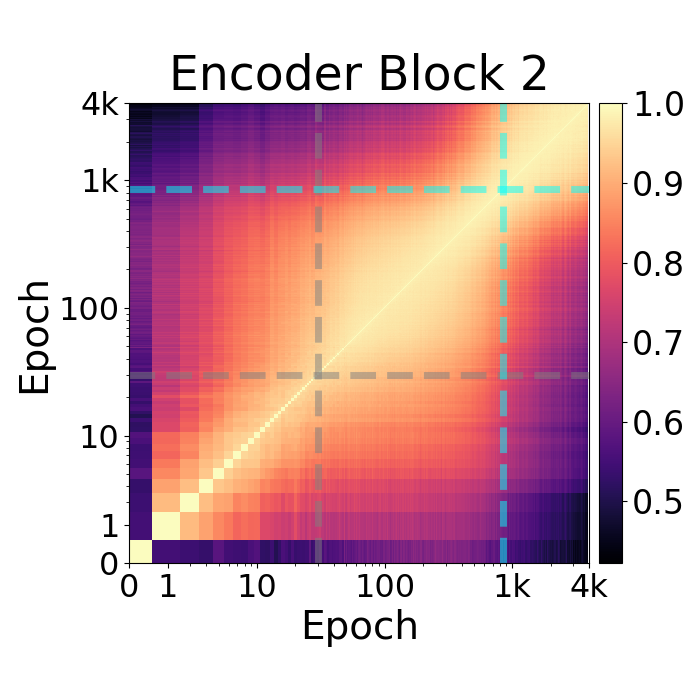}
\label{Encoder_block_2}}
\subfloat[]{
\includegraphics[width=0.133\textwidth]{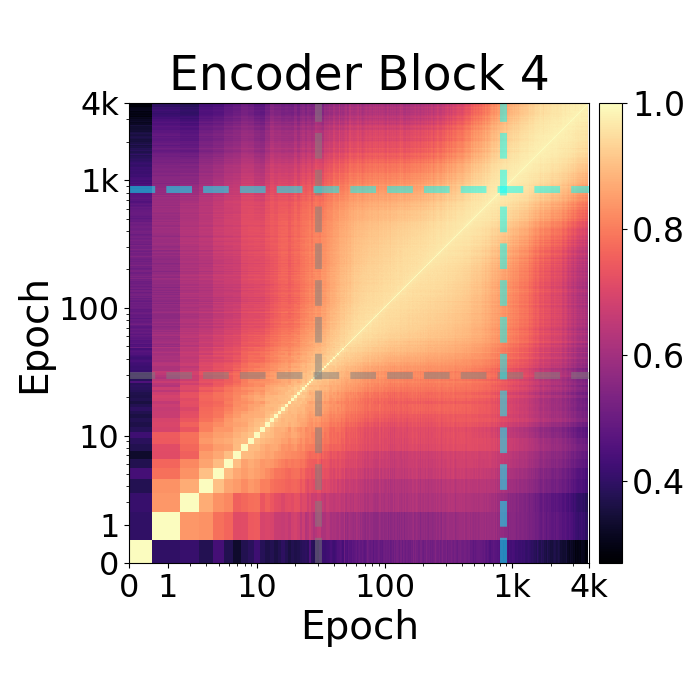}
\label{Encoder_block_4_noise_0}}
\subfloat[]{
\includegraphics[width=0.133\textwidth]{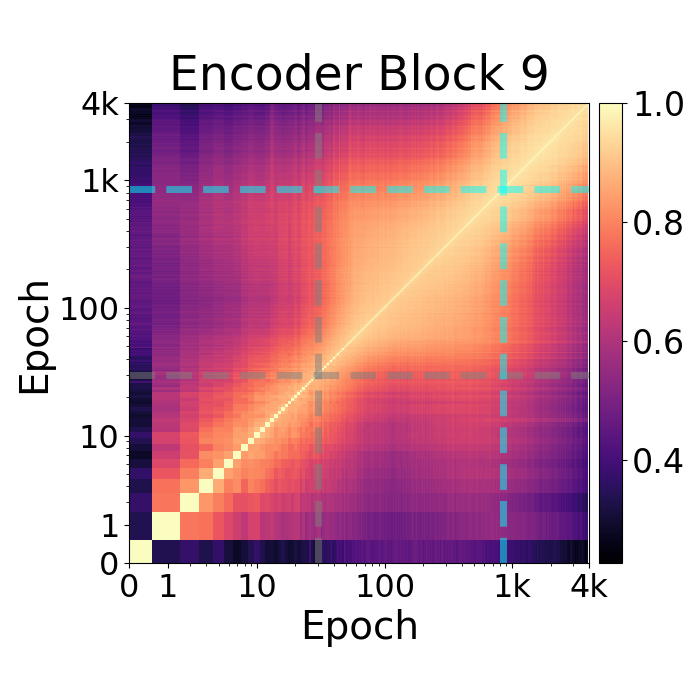}
\label{Encoder_block_9_noise_0}}
\subfloat[]{
\includegraphics[width=0.133\textwidth]{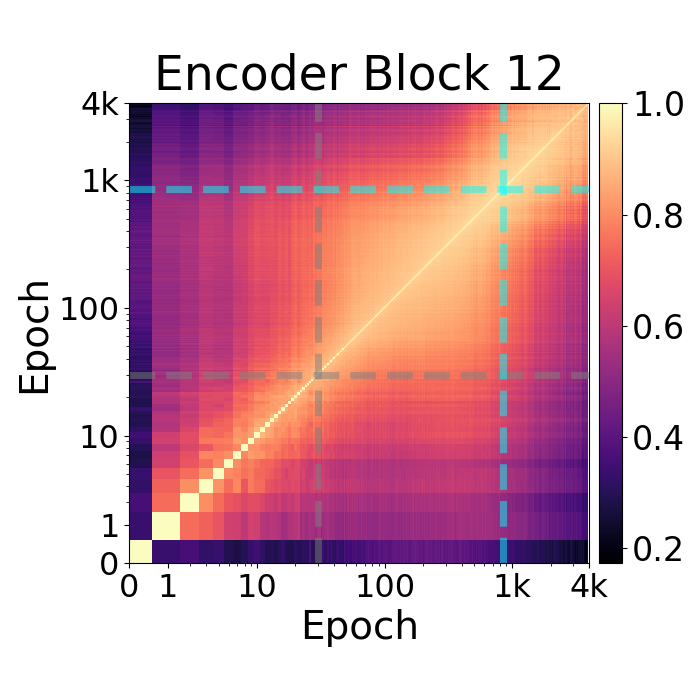}
\label{Encoder_block_12_noise_0}}
\subfloat[]{
\includegraphics[width=0.133\textwidth]{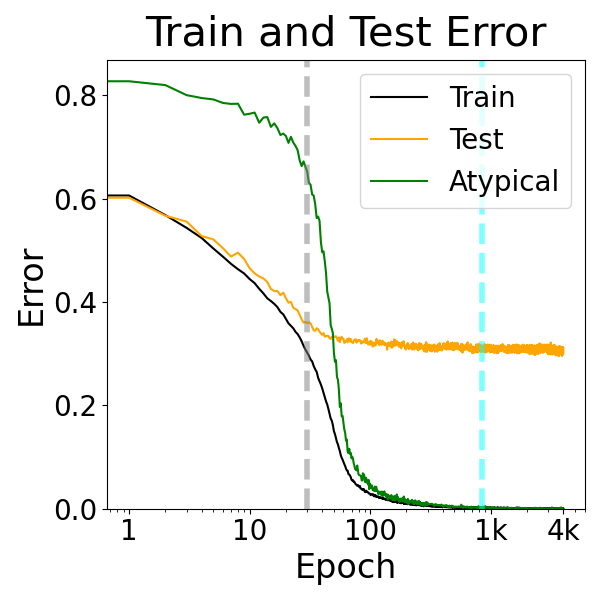}
\label{vit_train_test_error_noise_0}}
\caption{CKA evaluations for ViT-B/16, trained using Adam on CIFAR-10 without label noise.}
\label{fig:cka_comparison_vit_b_16_noise_0}
\end{figure*}



Representations can have one or two transition points in their training evolution, creating \textit{similarity blocks} of relatively lighter color around the main diagonal of the similarity diagram. Different layers can differ in the number of similarity blocks as well as the range of epochs in which the similarity blocks emerge. 
We observe that \textbf{similarity blocks may be aligned with the training phase transitions in some of the DNN layers, but not in all of them}; this gives a new perspective on the gradient-based analysis by \citet{maini2023can} that memorization can be due to some (possibly arbitrary) layers or neurons in the DNN. 
For example, the similarity blocks are aligned with the training phases in the first conv layer and the fourth ResNet Block output for ResNet-18 training (Figs.~\ref{Resnet18_k_64_first_conv_layer_noiseless}, \ref{Resnet18_k_64_block_4_noiseless}); and in the outputs of Encoder Blocks 4-12 in ViT training (see Figs.~\ref{fig:cka_comparison_vit_b_16_noise_0}, \ref{fig:cka_comparison_vit_b_16_noise_20}).

Our results show that similarity blocks can be more pronounced (i.e., reflecting relatively-quick change that causes higher dissimilarity of representations before and after a time point in training) depending on the DNN architecture and optimizer -- we will now explain this in more detail.

\subsubsection{ResNet vs.~ViT}
\paragraph{Training phase transitions are much more abrupt and synchronized across layers, and similarity blocks are more clearly identified, in ViT than in ResNet.} The similarity blocks are usually visibly clearer in ViT than in ResNet. Specifically, in ViT training the similarity blocks are well observed for both noiseless (Fig.~\ref{fig:cka_comparison_vit_b_16_noise_0}, \ref{fig:svhn_vit_b_16_noise_0_k_64_adam_lr_0.0001_momentum_0_bs_128_20250112}) and noisy (Fig.~\ref{fig:cka_comparison_vit_b_16_noise_20}) datasets. 
Remarkably, our results show that the \textit{training dynamics of representations is highly synchronized across the ViT layers}; this is due to the strong effect of skip connections in ViT, as demonstrated by \citet{raghu2021do} for representational similarity of fully trained ViTs -- we, therefore, significantly extend their finding by showing that the entire representation learning process is synchronized.
These synchronized dynamics are also aligned with the training phases, implying that memorization of atypical/noisy examples and perfect fitting can more clearly explain the representation learning evolution in ViT than in ResNet.

\subsubsection{Adam vs.~SGD}
\paragraph{Training phase transitions are much more abrupt, and similarity blocks are more clearly identified, in SGD training than in Adam training.}
Training ResNet-18 with SGD (e.g., Figs.~\ref{fig:sgd_0_0001_Resnet18_k_64_noise_20}, \ref{fig:sgd_0_01_Resnet18_k_64_noise_20}, \ref{fig:cifar10_resnet18_noise_0_k_64_sgd_lr_0.001_momentum_0_bs_128}) has clearer similarity blocks and transition points between them than in training with Adam (e.g., Figs.~\ref{fig:Resnet18_k_64_noise_0}, \ref{fig:cka_comparison_resnet18_k_64_noise_20}). 
This implies that memorization of atypical/noisy examples has a greater effect on representation learning dynamics in SGD  than in Adam training. 
In Appendix \ref{appendix:sec:Adam vs SGD Direct Comparison by Representational Similarity} we directly compare Adam and SGD training processes using CKA and discuss the results.

\subsubsection{First Layer in Convolutional Networks}
\label{subsec:First Layer in Convolutional DNNs}
\paragraph{In the perfect fitting regime of training with Adam, representations at the first layer are more similar to random representations than to good early-stopping representations.}
The CKA diagrams of the first layer in ResNet-18 trained using Adam without/with label noise (see Figs.~\ref{Resnet18_k_64_first_conv_layer_noiseless}, \ref{cka_comparison_resnet18_k_64_noise_20__first_conv_layer}, \ref{Resnet18_k_64_first_conv_layer_noise_10} for CIFAR-10; Figs.~\ref{svhn_Resnet18_k_64_adam_first_conv_layer_noiseless} for SVHN) show higher CKA values than the other layers, suggesting that representations of the first layer do not change much compared to their random initialization. Yet, within this range of high CKA values, the diagrams show that in the perfect fitting regime of training the first layer representation is more similar to the random representations at the initialization and early in training rather than to representations in the beneficial early-stopping range of training (i.e., at epochs where the test error is relatively low and perfect fitting did not start yet).

Note that the similarity to the random representations is not only to the \textit{specific} initialization of the training process, but also to other random representations unrelated to the specific training process; we examine this further in Appendix \ref{appendix:subsec:CKA Similarity of the First Layer Representations to Random Representations} and Fig.~\ref{fig:resnet18_conv1_similarity_with_diff_init}.

For the SGD optimizer (e.g., Figs.~\ref{sgd_0_0001_Resnet18_k_64_first_conv_layer_noise_0}, \ref{sgd_0_0001_Resnet18_k_64_first_conv_layer_noise_20}, \ref{sgd_0_01_Resnet18_k_64_first_conv_layer_noise_20}) the first layer representations behave somewhat differently than for Adam, although they still have very high CKA similarity during the entire training process. The lack of this phenomenon in SGD training suggests that its existence may require adaptive normalization of gradients such as implemented in Adam (and SGD does not have). 

It was shown recently by \citet{cattaneo2024on} that, in contrast to SGD, the implicit regularization of Adam can significantly change throughout the training -- e.g., implicit anti-regularization that late in training turns into regularization. Accordingly, we believe that our observations here on the first layer representational behavior in Adam training can be potentially due to changes in implicit regularization. Specifically, as discussed by \citet{cattaneo2024on}, when the componentwise gradient magnitudes become lower than the Adam's numerical stability parameter $\epsilon$, the normalizing adaptivity of gradients becomes less effective. 
Indeed, we show in Figs.~\ref{fig:first_layer_grad_adam_low_eps_conv1} that our first-layer phenomenon does not exist in Adam training with a large $\epsilon$ ($0.01$ instead of the standard $10^{-8}$) that effectively disables the adaptive gradient normalization (see gradient histogram in Fig.~\ref{fig:first_layer_grad_adam_low_eps}); this further shows the implicit role that the adaptive gradient normalization has on representation learning dynamics.
Importantly, \citet{cattaneo2024on} focus on mathematical analysis of the learned model parameters and their gradients (and not representational similarity) and their empirical results for DNNs use full batch Adam; in contrast, we focus here on layerwise representational similarity in realistic settings of mini-batch Adam, therefore, our findings here consider more intricate DNN training procedures and significantly add to those by \citet{cattaneo2024on}.

The representations of the first layer have increased similarity to the random initialization at the perfect fitting regime only \textbf{if the DNN is wide enough} (Figs.~\ref{Resnet18_k_32_first_conv_layer}, \ref{Resnet18_k_42_first_conv_layer}); we discuss this in more detail in Appendix \ref{appendix:subsec:The Effect of DNN Width and Weight Decay on the First Layer Representations}. The phenomenon is observed not only for ResNet-18, but also for a two-layer convolutional model (trained on 1k examples from Fashion MNIST, Fig.~\ref{fig:simple_model_fashion_mnist}). Accordingly, we believe this phenomenon may occur in Adam training of overparameterized (sufficiently wide) convolutional networks.

\section{Task-Specific Representations: Decision Regions Similarity of Linear Classifier Probes of Layers During Training}
\label{sec:Decision Regions Similarity of Linear Classifier Probes of Layers During Training}

\begin{figure*}[t]
\centering
\subfloat[]{
\includegraphics[width=0.15\textwidth]{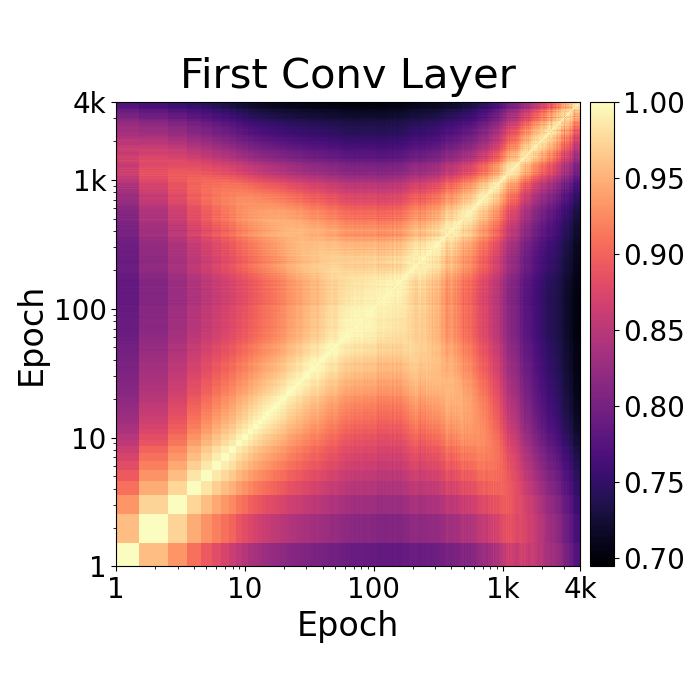}
\label{lp_resnet_noise_0_first_conv_layer}}
\subfloat[]{
\includegraphics[width=0.15\textwidth]{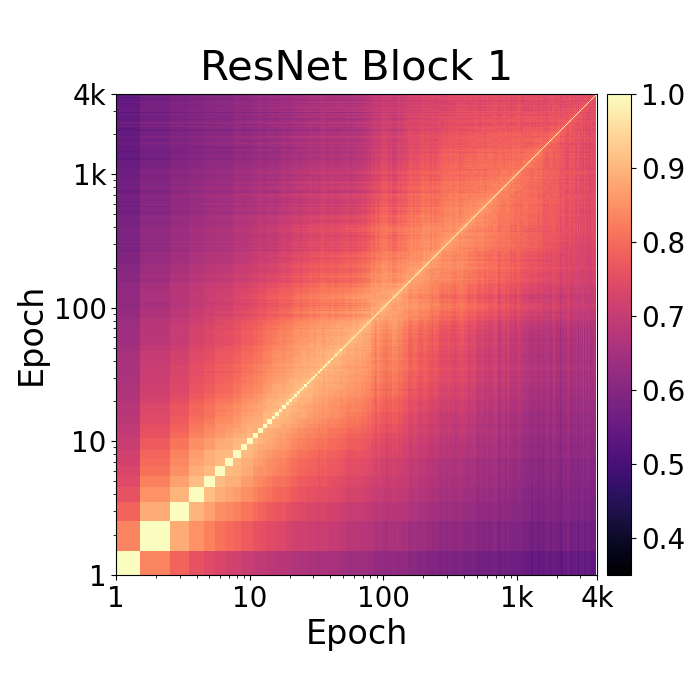}
\label{lp_resnet_noise_0_block_1}}
\subfloat[]{
\includegraphics[width=0.15\textwidth]{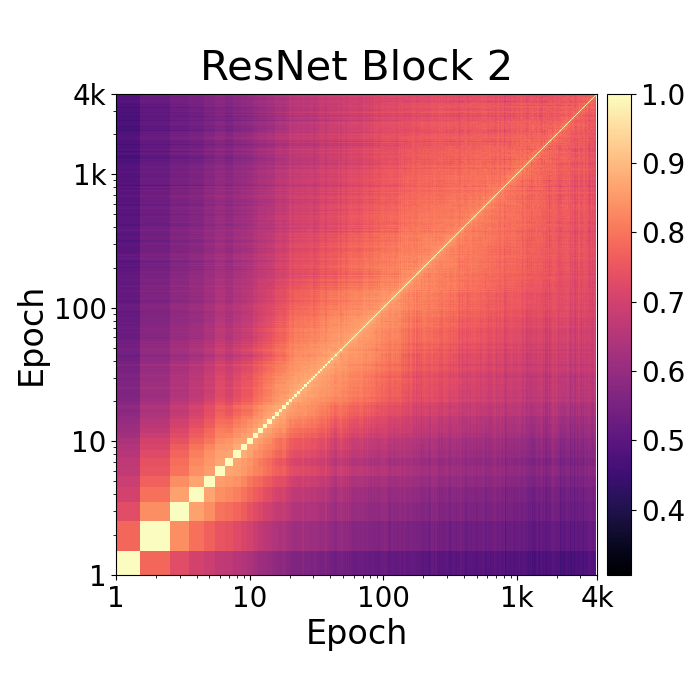}
\label{lp_resnet_noise_0_block_2}}
\subfloat[]{
\includegraphics[width=0.15\textwidth]{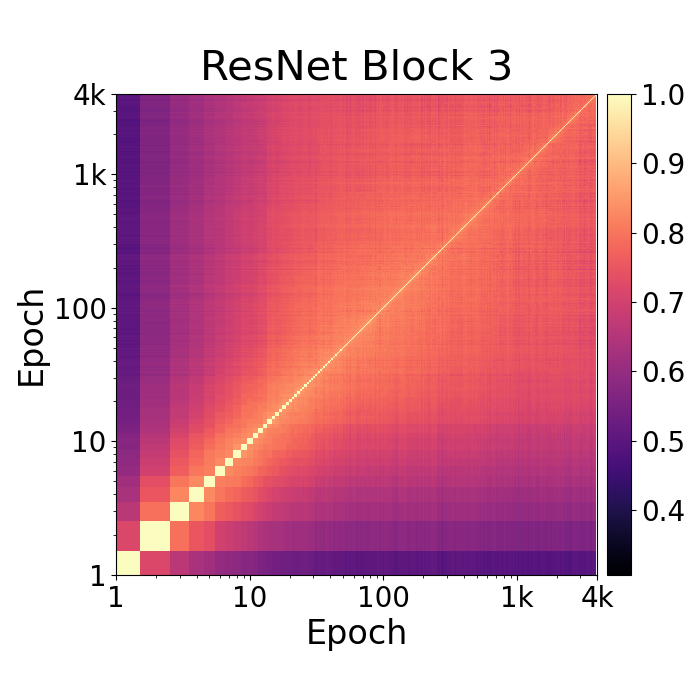}
\label{lp_resnet_noise_0_block_3}}
\subfloat[]{
\includegraphics[width=0.15\textwidth]{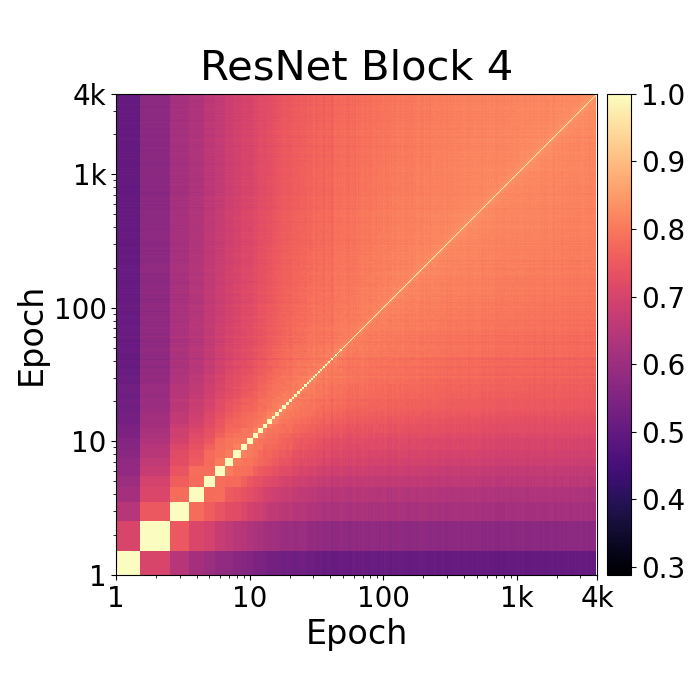}
\label{lp_resnet_noise_0_block_4}}
\caption{Decision regions similarity of linear classifier probes. Evaluations for ResNet-18 trained on CIFAR-10 without label noise. Each subfigure shows the DRS of a specific layer in the ResNet-18 during its training. The error curves are provided in Fig.~\ref{ResNet18_k_64_cifar10_noiseless_error_curve}.}
\label{fig:lp_resnet_noise_0}
\end{figure*}

\begin{figure*}[t]
    \centering
        \includegraphics[width=0.8\textwidth]{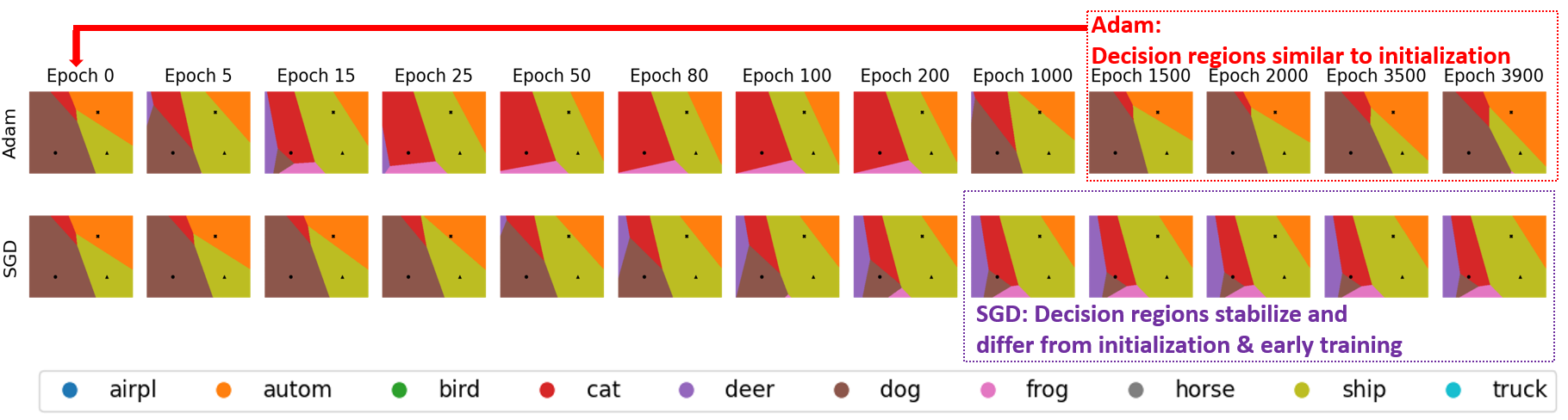}
\caption{Visualization of decision regions of the linear probe of the first convolution layer in ResNet-18 (trained on CIFAR-10).}
\label{fig:cifar10_resnet_first_layer_probe_vis_main_paper}
\end{figure*}

Following our CKA-based insights, we now turn to a significantly different approach to analyze training dynamics of learned representations in DNN layers. 
In this section, we propose a new analysis approach that elucidates the task-specific similarity among representations during training. 

\subsection{Training Linear Classifier Probes for Layers Throughout the Training Process}
\label{subsec:Training Linear Classifier Probes for Layers Throughout the Training Process}

From the complete training process, we define a relatively dense subset of training epochs $\mathcal{T}\subset\{0,1,\dots,3999\}$ for which we examine various layers in the trained DNN. We denote again that for layer $L$ at epoch $t\in\mathcal{T}$, the mapping from the input to the layer's representation is $f_L^{(t)}:\mathcal{X}\rightarrow\mathbb{R}^{p_L}$. 
We use $f_L^{(t)}$ (i.e., the DNN segment from the input to layer $L$) as a fixed feature map, and add a trainable linear layer on top of it to act as a probe. Namely, we train the weights $\mtx{W}\in\mathbb{R}^{|\mathcal{Y}|\times p_L}$ for the linear classifier probe 
\begin{equation}
    \label{eq:linear classifier probe - score vector - formulation}
    \widetilde{{\sf Probe}}_L^{(t)} \left({f_L^{(t)}\left(\vec{x}\right)}\right) = {\sf softmax}\left({\mtx{W}}f_L^{(t)}\left(\vec{x}\right) \right)
\end{equation}
where $\mathcal{Y}$ is the label domain of the original classification task.
The training of the linear probe is completely separate from the training of the original DNN that $f_L^{(t)}$ is taken from. Moreover, linear probes at different training epochs (of the original DNN) are trained separately and do not share their weights $\mtx{W}$. See more details on the probe training in Appendix \ref{appendix:subsec:DRS Details}.  Note that the linear probe layer in (\ref{eq:linear classifier probe - score vector - formulation}) maps the $p_L$-dimensional representation of $f_L^{(t)}$ to a score vector of classes (these classes are as in the classification task of the original DNN, and the probe's score vector also goes through a softmax function). The predicted class label from $\mathcal{Y}$ is the one with the maximal softmax-based class score, namely, by denoting the $j^{\sf th}$ class score as $\widetilde{{\sf Probe}}_L^{(t)}\left({f_L^{(t)}\left(\vec{x}\right)}\right)[j]$ we have 
\vspace{-0.1in}
\begin{equation}
    \label{eq:linear classifier probe - score vector - formulation}
     {\sf Probe}_L^{(t)} \left({f_L^{(t)}\left(\vec{x}\right)}\right) = 
      \argmax_{j\in\{1,\dots,|\mathcal{Y}|\}} \widetilde{{\sf Probe}}_L^{(t)}\left({f_L^{(t)}\left(\vec{x}\right)}\right)[j].
\end{equation}

\subsection{Measuring Similarity of Probe Decision Regions}
\label{subsec:Measuring Similarity of Decision Regions of Linear Classifier Probes}

\begin{figure*}[]
    \centering
        \includegraphics[width=0.87\textwidth]{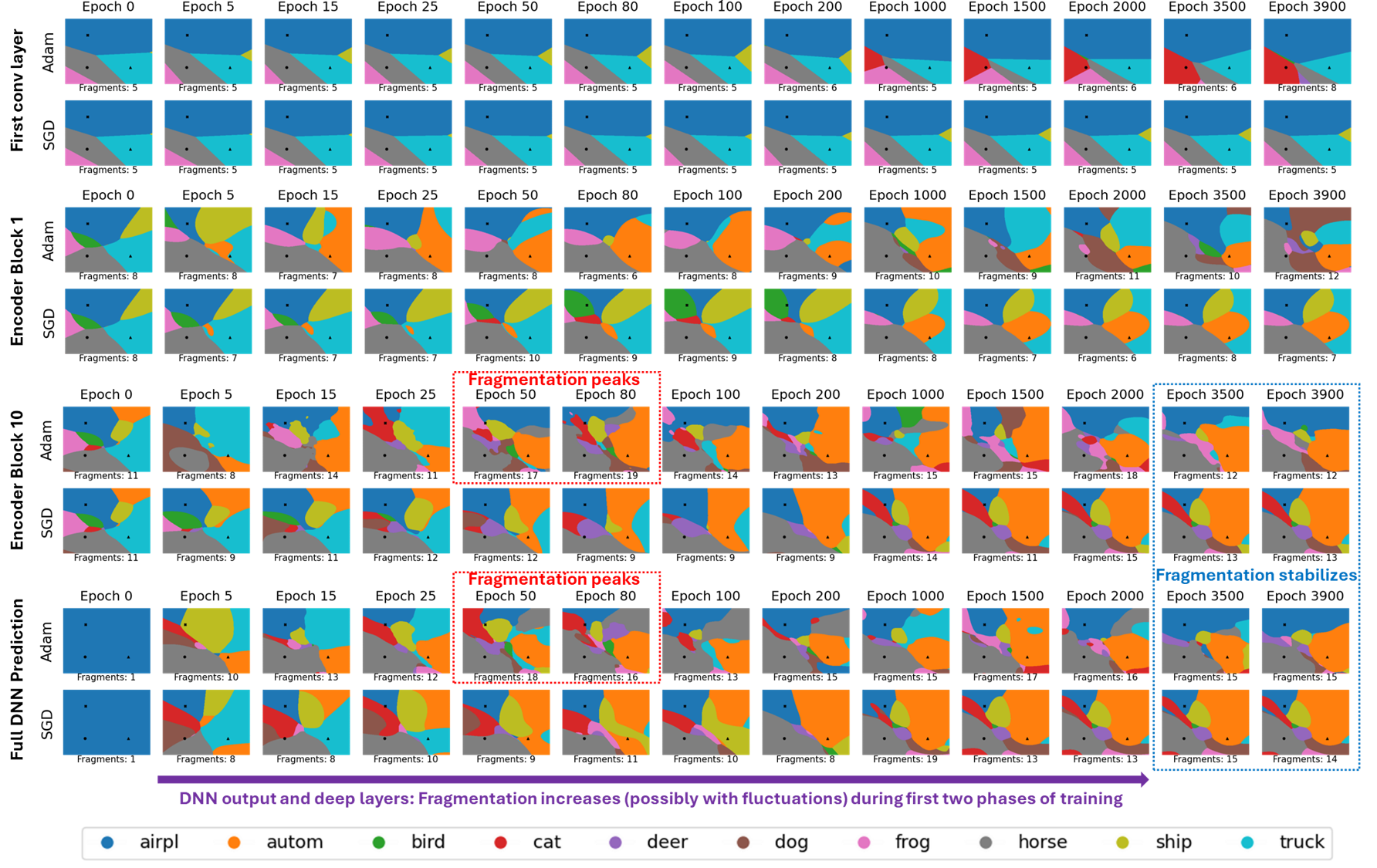}
\caption{Visualization of decision regions of the layerwise linear probes and full DNN prediction in ViT-B/16 (trained on CIFAR-10).}
   \label{fig:ViT_CIFAR10_plane105_visual_for_main_paper}
\end{figure*}

Linear classifier probes are often used for analyzing the quality of the learned features by evaluating their usefulness for the original \citep{alain2017understanding,raghu2021do} or another classification task \citep{hermann2020what}. In this paper, we are interested in understanding the evolution of representations during training; accordingly, we propose here a new probing-based similarity measure that reflects the overall representation. Specifically, we propose to compare representations of the same layer at two different training epochs by evaluating the \textit{similarity of the decision regions of the linear classifier probes} of the layer at the respective training epochs. 

This Decision Region Similarity (DRS) approach considers the representations in a task-specific manner, as the probes train to classify the layer representations to the same set of classes the original DNN has. The DRS approach is importantly different from the CKA that does not explicitly use the specific task in the representation analysis.

Evaluating similarity of decision regions was proposed by \citet{somepalli2022can} for comparing \textit{fully-trained} DNNs in their standard classification functionality of the DNN prediction output (i.e., not its layers and without probes). Here, we extend this decision-region similarity evaluation to the completely different analysis task of comparing linear probes of intermediate layers during their training. 

\subsubsection{The DRS Computation Procedure}
\label{subsubsec:The DRS Computation Procedure}
Given two linear classifier probes of the same layer $L$ at different training epochs $t, t'$, we numerically estimate the relative portion of the input space for which their predictions agree. As \citet{somepalli2022can}, we define $N_Q = 500$ planes that reside in the input space, each plane is defined by randomly choosing three training examples and computing the plane that they span. See more details in Appendix \ref{appendix:subsec:DRS Details}. The $j^{\sf th}$ plane, $j\in\{1,\dots,N_Q\}$, is denoted as $\mathcal{Q}_j\subset\mathcal{X}$. $\mathcal{Q}_j^{\sf discrete}$ is a discrete, finite version of $\mathcal{Q}_j$, which is defined for the numerical evaluation based on a uniformly-spaced 2D grid of $|\mathcal{Q}_j^{\sf discrete}|=2500$ points. Each $\vec{x}\in\mathcal{Q}_j^{\sf discrete}$ is an image, sampled from the plane, that can be given to  $f_L^{(t)}$ and $f_L^{(t')}$ as input. Then, the similarity of the decision regions of the probes of layer $L$ at epochs $t, t'$ is 
\begin{align}
    \label{eq:decision regions similarity of linear probes decision regions}
    & {\scriptstyle {\sf DRS} \left(f_L^{(t)},f_L^{(t')}\right) \triangleq} \\ \nonumber
    & {\scriptstyle   \frac{ \sum_{j=1}^{N_Q} \sum_{\vec{x}\in\mathcal{Q}_j^{\sf discrete}} \mathbb{I}\left({\sf Probe}_L^{(t)} \left({f_L^{(t)}\left(\vec{x}\right)}\right) = {\sf Probe}_L^{(t')} \left({f_L^{(t')}\left(\vec{x}\right)}\right)\right) }{N_Q\cdot|\mathcal{Q}_j^{\sf discrete}|}}
\end{align}
where $\mathbb{I}({\sf condition})$ is an indicator function that returns 1 if the ${\sf condition}$ is satisfied, and 0 otherwise. 
The ${\sf DRS} \left(f_L^{(t)},f_L^{(t')}\right)$ value is between 0 and 1, where a higher value reflects higher similarity and agreement of the decision regions of the linear probes.

Our proposed idea here is that \textit{similar representations lead to similar decision regions of the respective linear classifier probes}. 
In this approach, the intricacy of comparing the representations of different models (from different training epochs) is first addressed by the training of linear classifier probes, which interpret the representational structure to achieve their own training objectives. Clearly, the functionality that the probes learn reflects and depends on the representational structure of their inputs. It should be noted though that there is some level of inherent randomness in the training process of the probes. 
Comparing the decision regions of the probes is a way to compare their functionality and, by that, to compare the task-specific representational structure they get as their inputs.  
Importantly, the DRS is an interpretable approach, as the decision regions can be visualized and and further elucidate the representation learning dynamics beyond the CKA capabilities.

\subsubsection{Analysis of the DRS Results}
\label{subsec:Analysis of the Linear Classifier Probes Results}

In Figs.~\ref{fig:lp_resnet_noise_0}, \ref{fig:lp_resnet_noise_20} -\ref{fig:DRS_cifar_resnet_noise_0_sgd}  we show DRS diagrams, each corresponds to a different layer in the DNN. 
For a start, note that the DRS evaluations in Figs.~\ref{fig:lp_resnet_noise_0}, \ref{fig:lp_resnet_noise_20} support insights that we got from the CKA in Figs.~\ref{fig:Resnet18_k_64_noise_0}, \ref{fig:cka_comparison_resnet18_k_64_noise_20}. Specifically, recall Section \ref{subsec:First Layer in Convolutional DNNs} and note in Figs.~\ref{lp_resnet_noise_0_first_conv_layer}, \ref{lp_resnet_noise_20_first_conv_layer} that towards and in the perfect fitting phase of ResNet training, the first layer becomes more similar to random representations as at the initialization.  Yet, there are two main differences here compared to the CKA results. First, the DRS values have a wider range than in the CKA for the first layer, suggesting that indeed there is a significant training dynamics of task-specific features in the first layer. Second, the DRS shows that the similarity to random representations towards and in the perfect fitting phase is temporary and later when the training ends the first layer representations may differ from random representations.

In Figs.~\ref{fig:cifar10_resnet_first_layer_probe_vis_main_paper}, \ref{fig:ViT_CIFAR10_plane105_visual_for_main_paper}, \ref{fig:cifar10_resnet_first_layer_probe_vis_appendix_1}, \ref{fig:cifar10_resnet_first_layer_probe_vis_appendix_2},  \ref{fig:ViT_CIFAR10_plane_visual_for_appendix_1}, \ref{fig:resnet_full_vis_triplet_494}  we visualize the evolution of decision regions during training at different layer probes, as well as the DNN prediction output. Each of the visualized planes shows the geometrical evolution of the decision regions, demonstrating when during training the decisions change more, less or even stabilize (at least approximately). These visualizations also show geometrical properties of the decision regions, e.g., decision regions size and fragmentation level (i.e., number of fragments per visualized plane segment).
In Figs.~\ref{fig:fragmentation_score_resnet}, \ref{fig:fragmentation_score_vit} we complement the visualizations by providing quantitative evaluations of the average number of fragments per plane at various layers and DNN output during training (see Appendix~\ref{appendix:sec:Quantitative Evaluation of Decision Region Fragmentation of Layer Probes and DNN Output} for more details).

In Figs.~\ref{fig:cifar10_resnet_first_layer_probe_vis_main_paper}, \ref{fig:ViT_CIFAR10_plane105_visual_for_main_paper}, \ref{fig:cifar10_resnet_first_layer_probe_vis_appendix_1}, \ref{fig:cifar10_resnet_first_layer_probe_vis_appendix_2},  \ref{fig:ViT_CIFAR10_plane_visual_for_appendix_1}, \ref{fig:resnet_full_vis_triplet_494} we observe that the decision regions evolve significantly different depending on the optimizer and DNN architecture -- reflecting the different implicit regularization that different optimizers and architectures may have. In Figs.~\ref{fig:cifar10_resnet_first_layer_probe_vis_main_paper}, \ref{fig:cifar10_resnet_first_layer_probe_vis_appendix_1}, \ref{fig:cifar10_resnet_first_layer_probe_vis_appendix_2}, for Adam, we observe the first layer behavior where late in training the decision regions approximately revert to their random initialization form; in contrast, in SGD, the first layer evolution includes two main phases where late in training the decision regions stabilize and differ from their initialization. 

The quantitative fragmentation evaluations (Figs.~\ref{fig:fragmentation_score_resnet}, \ref{fig:fragmentation_score_vit}) and the visualizations (Figs.~\ref{fig:ViT_CIFAR10_plane105_visual_for_main_paper}, \ref{fig:ViT_CIFAR10_plane_visual_for_appendix_1}) show that the decision regions of the DNN output and the deeper-layer probes may become more fragmented during the training phase of memorizing atypical examples (without artificially added label noise), and then the fragmentation level stabilizes in the perfect fitting training phase (possibly with some reduction in fragmentation before the stabilization). This behavior is observed more clearly for ViT, also due to its synchronized training dynamics across its layers --- emphasizing a significant aspect of implicit regularization on representation learning dynamics in ViT compared to ResNet.

\section{Conclusion}
In this paper we have elucidated the training dynamics of representations in deep learning. We have used the CKA similarity measure and proposed a new similarity measure based on decision regions of linear classifier probes. Our extensive experimental evaluations provide a highly detailed description of the representation evolution during training. By this, we provide new insights on representation learning dynamics and how they depend on the architecture (ResNet vs.~ViT) and optimizer (Adam vs.~SGD). We believe that the new insights and empirical analysis tools of this paper can pave the way for more studies on related research questions.





\section*{Broader Impact}
\label{sec:broader impact}
This paper presents a foundational research of the training process of DNNs. As such, the understanding and insights in this paper may reduce to some extent the extensive trial and error engineering in the common practice and, by that, may reduce the amount of time and energy needed for developing new deep learning models. Moreover, foundational insights, as in this paper, can potentially contribute to better interpretability of deep learning models.

\bibliography{main}
\bibliographystyle{icml2024}


\appendix

\renewcommand\thefigure{\thesection.\arabic{figure}}    
\setcounter{figure}{0}  

\renewcommand\thetable{\thesection.\arabic{table}}    
\setcounter{table}{0} 

\counterwithin*{figure}{section}
\counterwithin*{table}{section}

 \section*{Appendices}

\section{Related Work: Additional Discussion}
\label{appendix:sec:Related Work A Detailed Overview and Discussion}

This Appendix extends the Related Work section of the main paper (Section \ref{sec:Related Work}). 

The effect of DNN width on similarity of representations was examined by \citet{morcos2018insights} using the CCA metric, and by \citet{kornblith2019similarity,nguyen2021do} using the CKA metric. In contrast to these works, our focus is on elucidating the representational similarity during the training process. In the following, we discuss the existing literature which is more related to our work.

\subsection{Representational Similarity Throughout the Training Process}
\label{appendix:subsec:Related Work - Representational Similarity Throughout the Training Process}

\citet{raghu2017svcca} proposed the SVCCA metric for representational similarity and used it for several analyses, including showing how the representational similarity between the model during training and the model at the end of training evolves during training of a convolutional model and a ResNet model. In addition, \citet{liu2021autofreeze} showed the evolution of the SVCCA metric during 4 epochs of a BERT model training.  In this paper, we examine the evolution of representational similarity much more extensively than previous works, we consider various new settings (such as for a ViT architecture) and, most importantly, \textit{we use it for the first time as a tool for analyzing training that continues much after perfect fitting.} 

\citet{gotmare2018closer} examined the loss landscape (specifically, connectivity of local minima) and representational similarity (using the CCA  metric \citep{raghu2017svcca}) for various training heuristics from the common practice, such as learning rate restarts, warmup, and distillation. Their results include evaluation of CCA representational similarity between a VGG model at training iteration 0 and the specific training iteration at the end of their warmup phase (if applied); their evaluation reflects high similarity for the first convolutional layer of the model. They also evaluate the CCA similarity between the model at the end of training  and at the intermediate training iteration after the warmup. Importantly, the work by \citet{gotmare2018closer} does not show or analyze the detailed evolution of the representational similarity during the entire training process. In contrast, \textit{in our paper here we examine the representational similarity in much more detail, enabling to understand the representational similarity of various pairs of epochs, throughout a long training process with a small constant learning rate that does not affect the natural training dynamics. Moreover, we focus here on understanding how the representational similarity evolves in overparameterized learning where perfect fitting does not prevent good generalization.}

\subsection{Representational Similarity between Trained and Untrained Models}
\label{appendix:subsec:Related Work - Representational Similarity between Trained and Untrained Models}

\citet{kornblith2019similarity} examined the CKA similarity between trained and untrained convolutional DNNs. They showed that the representations in the first layers of the DNN are similar to those of the corresponding layers in the untrained DNN. Their CKA values for the first layer of a 10-layer convolutional model on CIFAR-10 was slightly above 0.8. In this paper (e.g., Figs.~\ref{Resnet18_k_64_first_conv_layer_noiseless}, \ref{cka_comparison_resnet18_k_64_noise_20__first_conv_layer}, \ref{Resnet18_k_64_first_conv_layer_noise_10}), we show that the first layer can have CKA similarity (nearly) 1 in the perfect fitting regime of training.

\citet{chowers2023what} propose to analyze the first layer in convolutional DNNs using the energy spectrum of the learned filters in the PCA basis of the input image patches. Their results show a typical form of the learned energy spectrum which is consistent over different convolutional architectures and datasets but very different compared to its random initialization at the beginning of training; we disprove the generality of this claim in this paper, by using the CKA similarity metric and showing that the representations of the first layer of a ResNet model can be very similar to the representations at the random initialization. This difference is probably due to our training setting that allows long training in the perfect fitting regime and avoids early stopping (whereas their training is much shorter; for example, their Fig.~18c in \citep{chowers2023what} shows a training process of 200 epochs that stops approximately when perfect fitting starts). 

\citet{hermann2020what} use synthetic datasets (with controllable feature relevance) to study which features a trained model uses and how the choice of the features relates to their relevance to the task and to their linear-decodability from representations in an untrained model. Moreover, they show that trained models have relatively higher representational similarity (quantified by the neuroscientific method of \citet{kriegeskorte2008representational}) to untrained models than to models trained on another task. Importantly, the representational similarity between trained and untrained models which is reported by \citet{hermann2020what} is moderate, and even quite low, in contrast to our results in this paper that shows much higher similarity levels according to other similarity metrics.

Importantly, in addition to our extensive CKA evaluations, we propose a new representational similarity measure based on similarity of the decision regions of linear classifier probes. Both the CKA and the new probing-based measure lead to new insights on representation learning that are not available in the existing literature.

\section{Additional Details on the Experiment Settings}
\label{appendix:sec:Additional Details on the Experiment Settings}

\subsection{Software Packages}
\label{appendix:subsec:Assets and Software Packages}

For the datasets, we used CIFAR-10 and Tiny ImageNet, which are publicly available:
\begin{itemize}
\item \textbf{CIFAR-10:} Available at \url{https://www.cs.toronto.edu/~kriz/cifar.html}. The CIFAR-10 dataset includes 10 classes, each with 5000 training examples, and a total of 50,000 training examples.
\item \textbf{Tiny ImageNet:} A subset of 200 classes out of the ImageNet dataset. 500 training examples per class, and a total of 100,000 training examples. We used a downsampled version of $32\times32\times3$ pixels as inputs to our models. 
\end{itemize}

For software, we utilized several packages:
\begin{itemize}
\item \textbf{PyTorch:} Used for training the DNNs and running computations on GPUs. PyTorch is distributed under the Apache Contributor License Agreement (CLA). More information can be found at \url{https://pytorch.org/}.
\item \textbf{torch-cka:} Used for CKA computation, incorporating technical methods from the package. The package is available at \url{https://github.com/AntixK/PyTorch-Model-Compare} and is licensed under the MIT License.
\item \textbf{dbViz:} For decision region similarity (DRS) experiments, we used methods introduced in \citep{somepalli2022can} with code available at \url{https://github.com/somepago/dbViz}, licensed under the Apache 2.0 License.
\end{itemize}

\subsection{Models}
\label{appendix:subsec:models}
\textbf{ResNet-18}: We used the (Preactivation) ResNet-18 architecture as introduced in \citep{he2016deep}. The ResNet-18 structure includes an initial convolutional layer followed by four ResNet blocks, each consisting of two BatchNorm-ReLU-Convolution layers. The layer widths for these four ResNet blocks are defined as [k, 2k, 4k, 8k] where $k$ is a width parameter (positive integer), and $k=64$ corresponds to the standard ResNet-18 layout.
Similarly to the deep double descent research in \citep{nakkiran2021deep}, we consider several values of $k$ to examine the effect of width and parameterization level on the ResNet-18 model. The ResNet-18 implementation was as in \citep{nakkiran2021deep,somepalli2022can} that originate in \url{https://github.com/kuangliu/pytorch-cifar}.

\textbf{ResNet-50}: For the experiment on ResNet-50 in Fig.~\ref{fig:Resnet50_k_64_noise_0} we used a standard ResNet-50 implementation that originates in \url{https://github.com/kuangliu/pytorch-cifar} and follows \citep{he2016deep}. No changes were applied on the model, including not on its standard width. 

\textbf{ViT-B/16}: In our implementation, we utilized the Vision Transformer (ViT) architecture, specifically the ViT-B/16 model, as described in \citep{dosovitskiy2020image}. The ViT-B/16 structure begins with an initial patch embedding layer, followed by a sequence of 12 transformer blocks.
We used the ViT-B/16 implementation of PyTorch.

\subsection{CKA Experimental Details}
\label{appendix:subsec:CKA Details}
The CKA evaluations in this paper are based on two different datasets: CIFAR-10 and Tiny ImageNet.

\begin{itemize}
    \item \textbf{CIFAR-10 and SVHN:}
    \begin{itemize}
        \item The CKA evaluations used a total of 2048 test examples, divided into 4 batches of 512 samples each.
        \item Each batch contained almost an equal number of examples from each class.
        \item The average CKA score was calculated over all batches to ensure robust and reliable evaluation.
        \item In our experiments, we did not find a significant difference when compared to using a larger set of examples. 
    \end{itemize}
    
    \item \textbf{Tiny ImageNet:}
    \begin{itemize}
        \item The CKA evaluations used a total of 10,000 test examples.
        \item \textbf{No Label Noise:} As described previously, the evaluations used 5 batches of 2000 samples each, i.e., a total of 10,000 test examples.
        \item \textbf{20\% Label Noise:} For Tiny ImageNet with 20\% label noise, the evaluation used a single batch of 2000 samples, i.e., a total of 2000 test examples.
    \end{itemize}
\end{itemize}

The CKA evaluations considered the following epoch grid - all epochs from 0 to 300, every third epoch between 300 to 900, and every fifth epoch from 900 to 3999.

\subsection{Atypical Inputs Setting}
\label{appendix:subsec:Atypical Inputs Setting}
In selected experiments on datasets without artificially added label noise, we evaluate the DNN classification erroron \emph{atypical examples} from its train dataset. The atypical examples are found as follows:
\begin{itemize}
    \item \textbf{CIFAR-10:} We use the C-score defined by~\citet{jiang2021characterizing} and set the threshold to $0.5$. All images with $\mathrm{C\text{-}score} > 0.5$ are considered atypical.
    \item \textbf{SVHN:} We use the loss-based C-score introduced by~\citet{wu2020curricula}, adjusting its loss threshold to match the proportion of atypical inputs found for CIFAR-10. 
\end{itemize}

\subsection{Decision Regions Similarity (DRS) Experimental Details}
\label{appendix:subsec:DRS Details}
For evaluating the Decision Region Similarity (DRS) between two models, we used the following procedure:

\begin{enumerate}
    \item First, we defined a grid of epochs from the 4000 training epochs where we sampled the DRS score.
    \begin{itemize}
        \item For ResNet-18, the grid included all epochs from 0 to 300, every third epoch between 300 to 900, and every fifth epoch from 900 to 3999.
        \item For ViT-B/16, the grid was the same, except for epochs 900-4000 where we sampled every tenth epoch.
    \end{itemize}

    \item For each epoch, we used the model from that epoch to extract the representations of the training set for the layer of interest. We then trained a linear classifier probe on top of the representations of the training set (all of the 50,000 train examples of CIFAR-10) and used the probe to predict the labels of the training set. All linear classifiers were trained with the Adam optimizer with a learning rate of 0.0001, cross-entropy loss, and for 10 epochs.

    \item We randomly chose 500 triplets of training examples (these triplets were constant for all DRS experiments). Then, for each triplet, we computed the plane that they span (as mathematically explained in \citep{somepalli2022can} and based on their code). The \( j^{\text{th}} \) plane spanned by those 3 samples, denoted as \( \mathcal{Q}_j \subset \mathcal{X} \), is a 2D plane in the input space. We then discretized each plane \( \mathcal{Q}_j \) into a uniformly-spaced 2D grid of \( |\mathcal{Q}_j^{\text{discrete}}| = 2500 \) points.

    \item The DRS was then computed as the average agreement of the linear classifier probes on the discretized planes, according to Eq.~(\ref{eq:decision regions similarity of linear probes decision regions}).
\end{enumerate}

\section{Summary of Experiment Settings}
\label{appendix:sec:experiment_summary}

\begin{itemize}
    \item \textbf{CKA experiments:} Please refer to Table \ref{table:cka_experiment_summary} for a summary of our CKA experiments, describing the datasets, architectures, optimizers, width parameters $k$, label noise percentage, and the corresponding figures of the results.
    \item \textbf{DRS experiments:} Generating the Decision Region Similarity (DRS) heatmaps is highly computationally demanding, requiring significant time and compute resources. Therefore, we limited the DRS evaluations to only a subset of the of the experiment settings. Please refer to Table \ref{table:cka_experiment_summary} for the relevant figures.
\end{itemize}

\begin{table*}[ht]
\centering
\begin{tabular}{l l l l l l l}
\textbf{Dataset} & \textbf{Architecture} & \textbf{Optimizer} & \textbf{Width P. ($k$) } & \textbf{\% Label Noise} & \textbf{CKA Fig.~\#} & \textbf{DRS Fig.~\#}\\ \hline
CIFAR-10  & ResNet-18 & Adam & 64 & 0 & \ref{fig:Resnet18_k_64_noise_0} & \ref{fig:lp_resnet_noise_0} \\
 &  &  &  & 10 & \ref{fig:Resnet18_k_64_noise_10} & \\
 &  &  &  & 20 & \ref{fig:cka_comparison_resnet18_k_64_noise_20} & \ref{fig:lp_resnet_noise_20} \\
 &  &  & 52 & 20 & \ref{fig:Resnet18_k_52_noise_20} & \\
 &  &  & 42 & 20 & \ref{fig:Resnet18_k_42_noise_20} & \\
 &  &  & 32 & 20 & \ref{fig:Resnet18_k_32_noise_20} & \\
 &  & SGD (0.001) & 64 & 0 & \ref{fig:cifar10_resnet18_noise_0_k_64_sgd_lr_0.001_momentum_0_bs_128} & \\
 &  & SGD+Mom. & 64 & 20 & \ref{fig:sgd_0_0001_Resnet18_k_64_noise_20} & \\
 &  & SGD+Mom. & 64 & 0 & \ref{fig:sgd_0_0001_Resnet18_k_64_noise_0} & \\
  &  & SGD (0.01) & 64 & 20 & \ref{fig:sgd_0_01_Resnet18_k_64_noise_20} & \\
 &  & Adam (wd) & 64 & 0 & \ref{fig:wd_0_001_Resnet18_k_64_noise_20} & \\
& ResNet-50 & Adam & N.A & 20 & \ref{fig:Resnet50_k_64_noise_0} & \\
 & ViT-B/16 & Adam & N.A & 0 & \ref{fig:cka_comparison_vit_b_16_noise_0} & \ref{fig:DRS_cifar10_vit_noise_0}\\
  & ViT-B/16 & SGD (0.001) & N.A & 0 & \ref{fig:vit_cifar10_sgd_cka_comparison_noise_0} & \ref{fig:DRS_cifar10_vit_noise_0_sgd}\\
 & ViT-B/16 & Adam & N.A & 20 & \ref{fig:cka_comparison_vit_b_16_noise_20} & \\
Tiny ImageNet & ResNet-18 & Adam & 64 & 0 & \ref{fig:Resnet18_k_64_noise_0_tin} & \\
 &  &  &  & 20 & \ref{fig:Resnet18_k_64_tiny_imagenet_noise_20} & \\
 Tiny ImageNet $1/2$ & ResNet-18 & Adam & 64 & 20 & \ref{fig:Resnet18_k_64_tiny_imagenet_noise_20_half_train_set} & \\
 SVHN & ResNet-18 & Adam & 64 & 0 & \ref{fig:svhn_Resnet18_k_64_adam_noise_0} & \ref{fig:DRS_svhn_resnet_noise_0_Adam} \\
& &  SGD (0.001) & 64 & 0 &  \ref{fig:svhn_Resnet18_k_64_sgd_noise_0} & \ref{fig:DRS_svhn_resnet_noise_0_SGD} \\
& ViT-B/16 & Adam & N.A & 0 & \ref{fig:vit_svhn_Adam_cka_comparison_noise_0} & \\
& & SGD (0.001) & N.A & 0 & \ref{fig:vit_svhn_sgd_cka_comparison_noise_0} &  \\
\hline
\end{tabular}
\caption{Summary of the CKA experiments conducted. Note that here Adam (wd) stands for training with Adam optimizer with weight decay regularization of 0.001, and Tiny ImageNet $1/2$ stands for the experminet where we used only half of the train samples of Tiny ImageNet. In addition, SGD (0.01) refers to Fig.~\ref{fig:sgd_0_01_Resnet18_k_64_noise_20}, where we used SGD optimizer with constant learning rate of 0.01 and no momentum.}
\label{table:cka_experiment_summary}
\end{table*}

\section{Additional Experimental Results}
\label{appendix:sec:Additional Experimental Results}

\begin{figure*}[]
  \centering
    \subfloat[]{
    \includegraphics[width=0.13\textwidth]{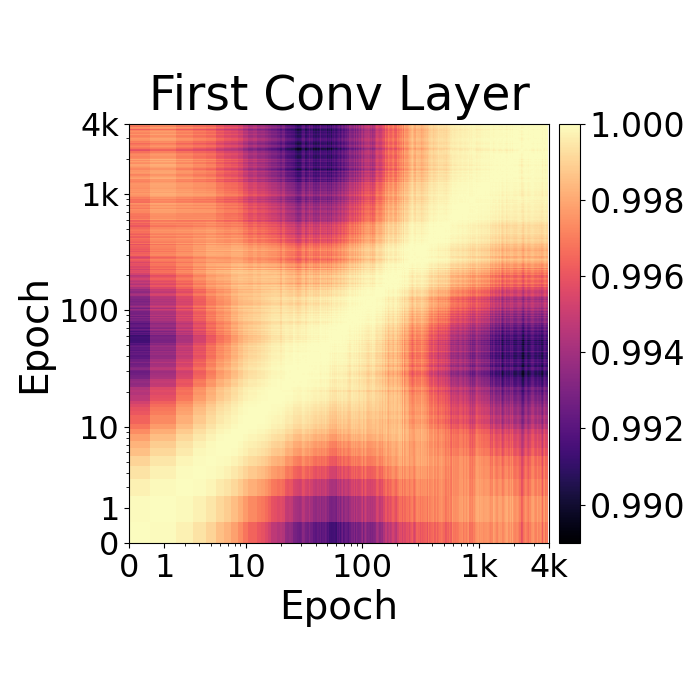}
    \label{cka_comparison_resnet18_k_64_noise_20__first_conv_layer}}
    \subfloat[]{
    \includegraphics[width=0.13\textwidth]{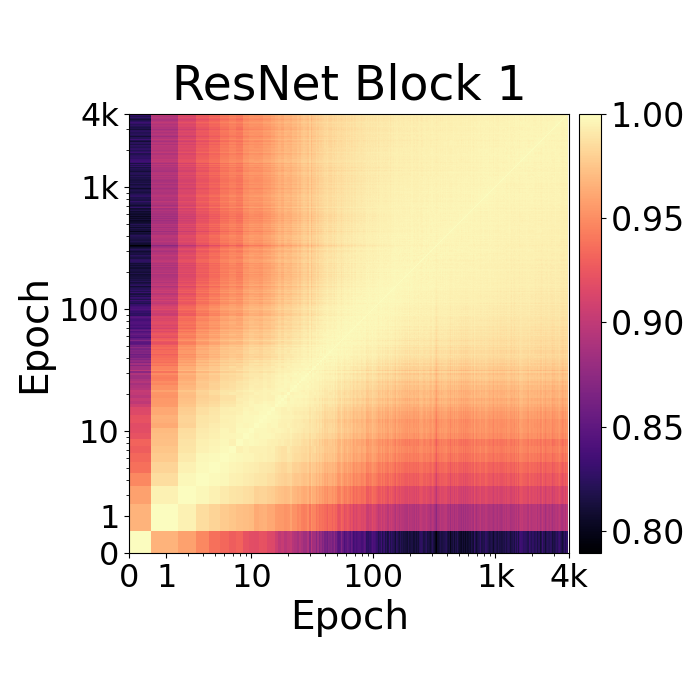}
    \label{cka_comparison_resnet18_k_64_noise_20__Resnet_block_1}
    }
    \subfloat[]{
    \includegraphics[width=0.13\textwidth]{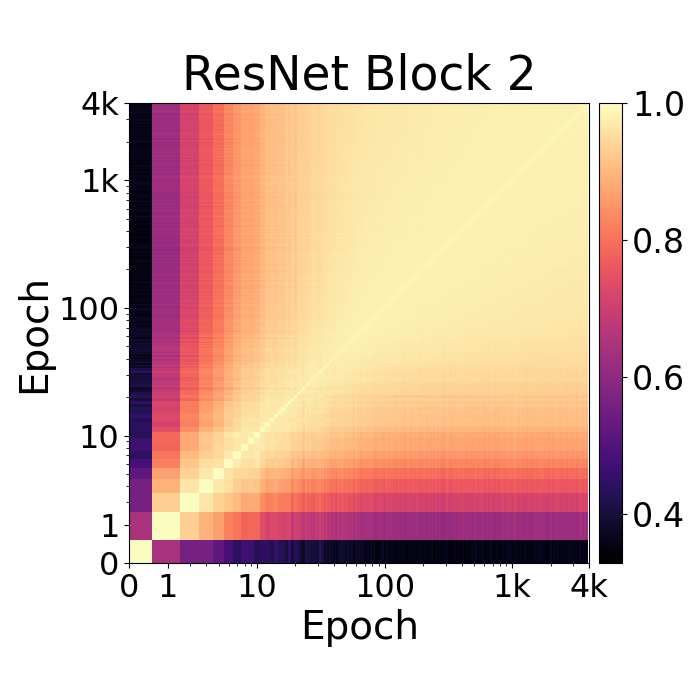}
    \label{cka_comparison_resnet18_k_64_noise_20__Resnet_block_2}}
    \subfloat[]{
    \includegraphics[width=0.13\textwidth]{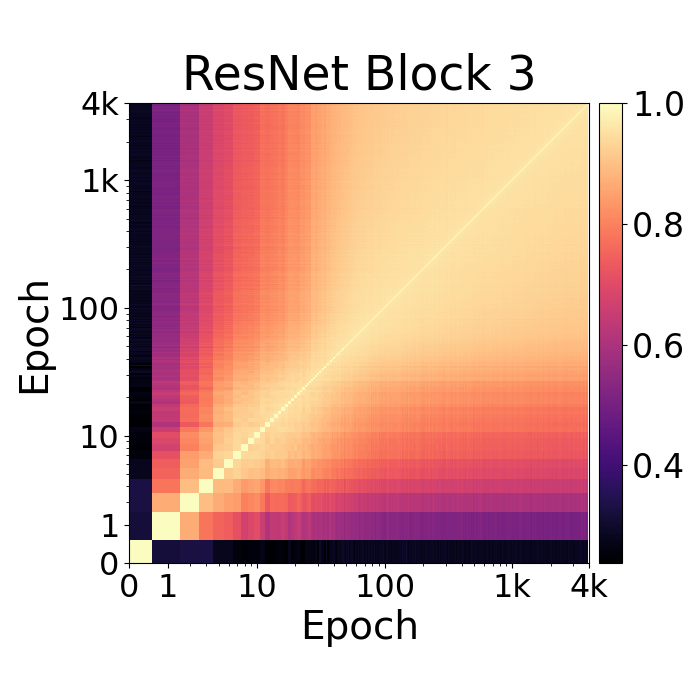}
    \label{cka_comparison_resnet18_k_64_noise_20__Resnet_block_3}}
    \subfloat[]{
    \includegraphics[width=0.13\textwidth]{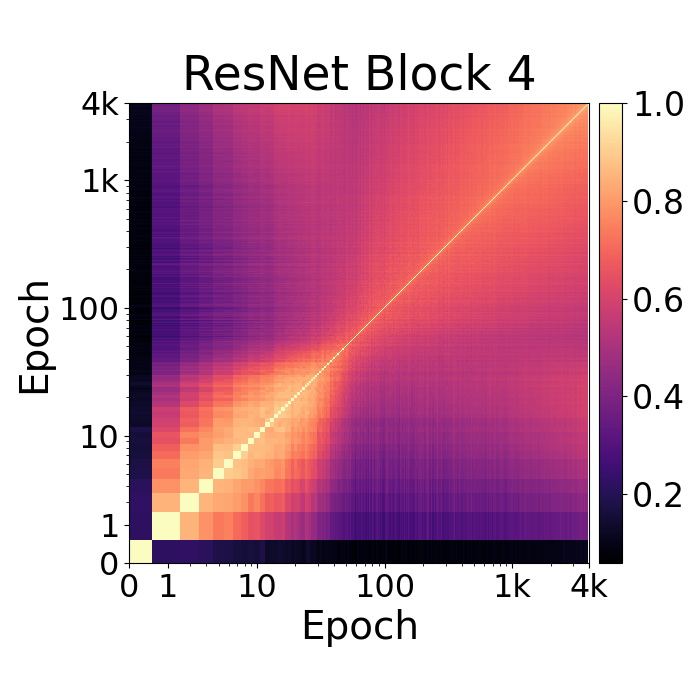} 
    \label{cka_comparison_resnet18_k_64_noise_20__Resnet_block_4}}
     \subfloat[]{
    \includegraphics[width=0.13\textwidth]{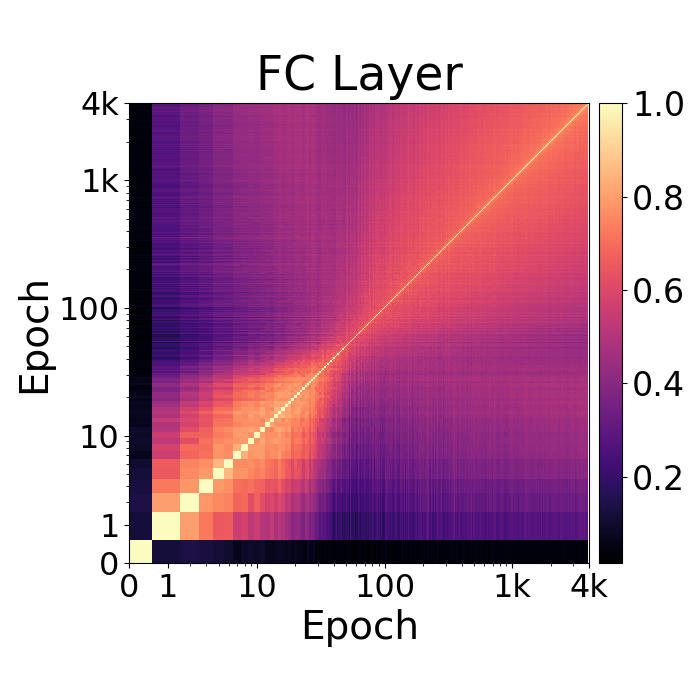}
    \label{cka_comparison_resnet18_k_64_noise_20__Resnet_FC}}
    \subfloat[]{
    \includegraphics[width=0.13\textwidth]{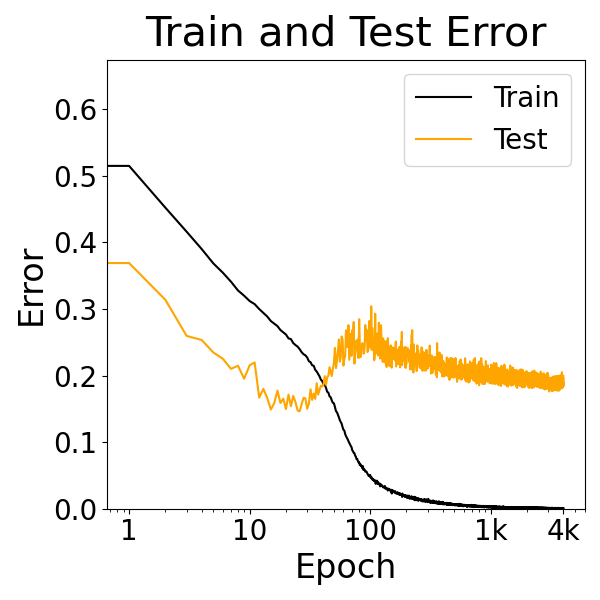}\label{ResNet18_k_64_cifar10_noise_20_error_curve}
    }  
  \caption{CKA evaluations for ResNet-18 trained on CIFAR-10 with 20\% label noise.}
\label{fig:cka_comparison_resnet18_k_64_noise_20}
\end{figure*}

\begin{figure*}[t]
      \centering
    \subfloat[]{
    \includegraphics[width=0.13\textwidth]{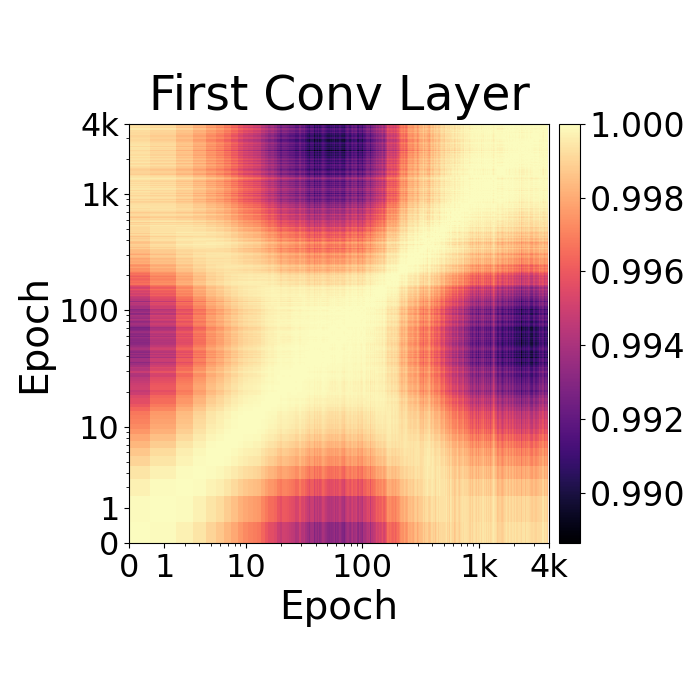}
    \label{Resnet18_k_64_first_conv_layer_noise_10}}
    \subfloat[]{
    \includegraphics[width=0.13\textwidth]{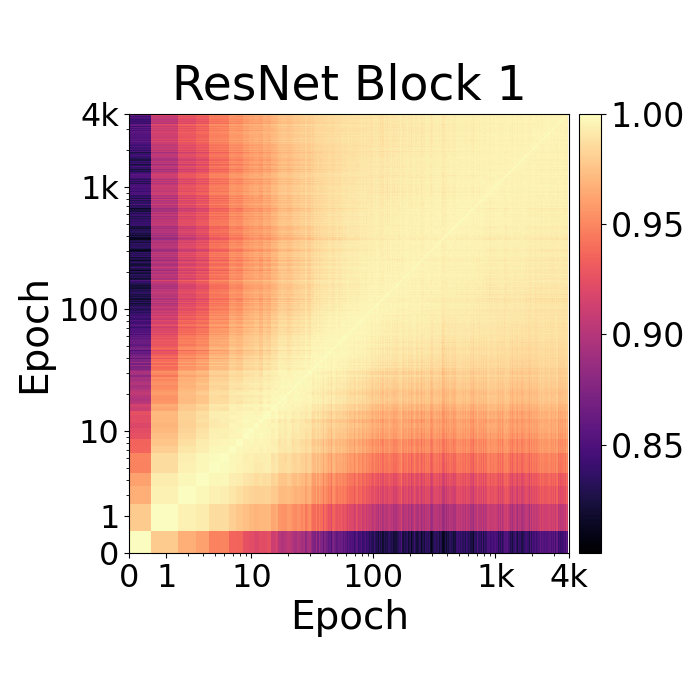}
    \label{Resnet18_k_64_block_1_noise_10}
    }
    \subfloat[]{
    \includegraphics[width=0.13\textwidth]{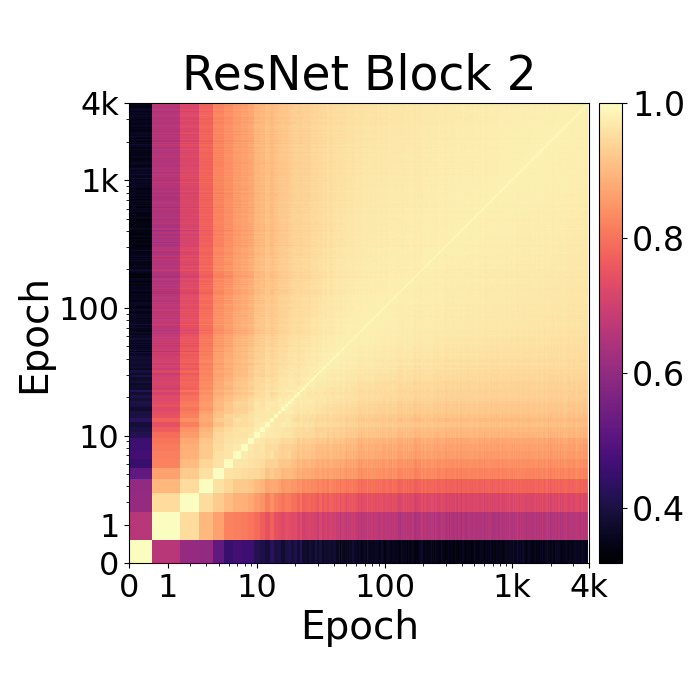}
    \label{Resnet18_k_64_block_2_noise_10}}
    \subfloat[]{
    \includegraphics[width=0.13\textwidth]{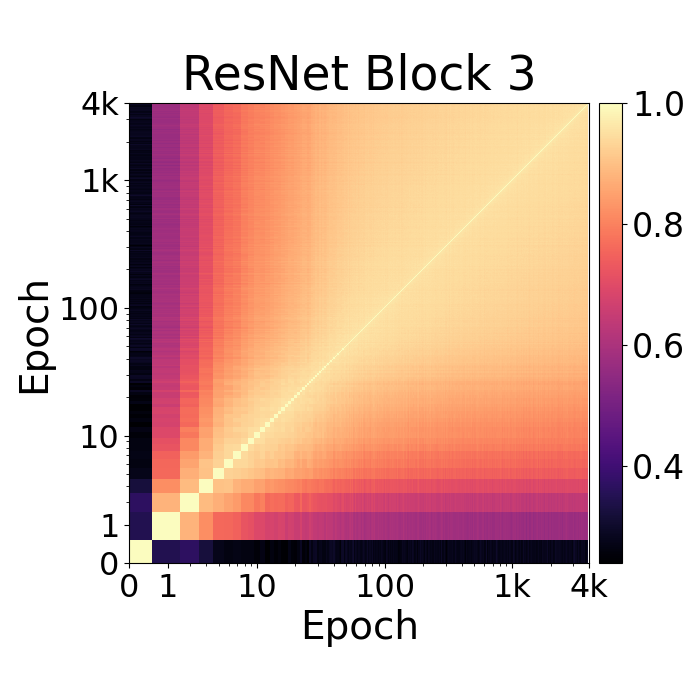}
    \label{Resnet18_k_64_block_3_noise_10}}
    \subfloat[]{
    \includegraphics[width=0.13\textwidth]{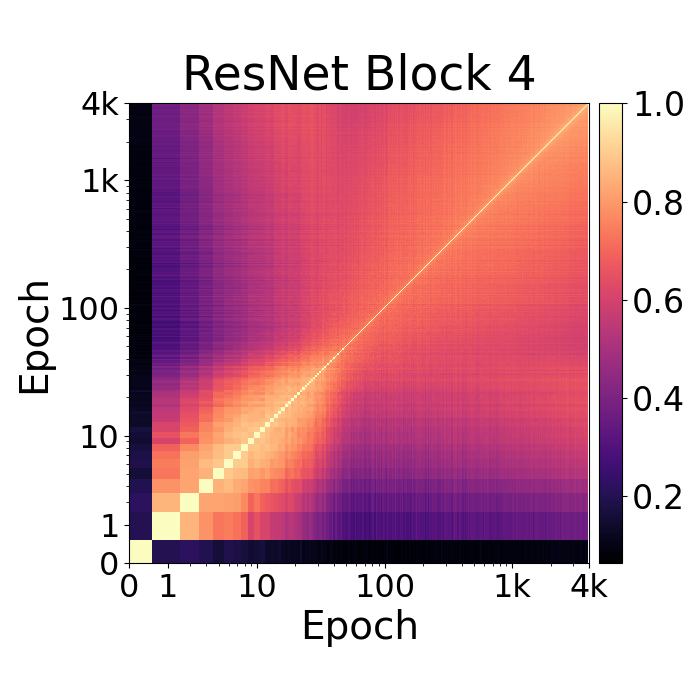} 
    \label{Resnet18_k_64_block_4_noise_10}}
     \subfloat[]{
    \includegraphics[width=0.13\textwidth]{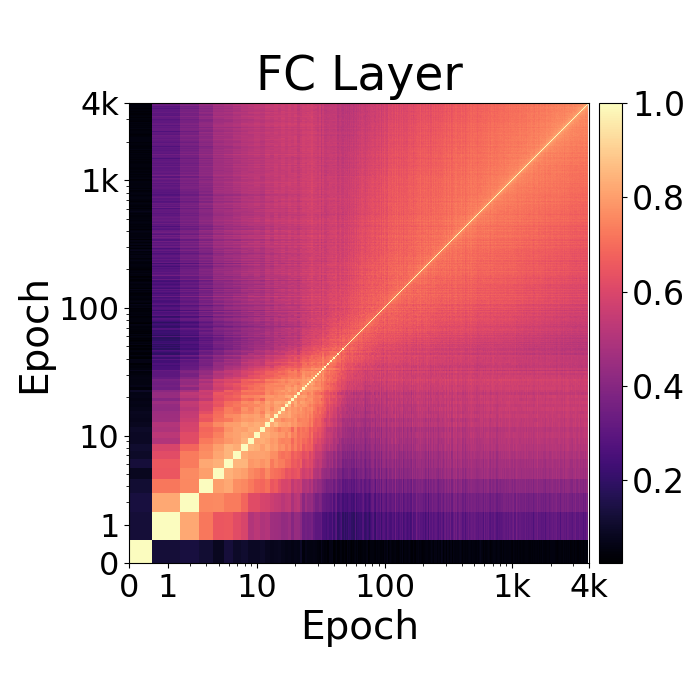}
    \label{Resnet18_k_64_FC_noise_10}}
    \subfloat[]{
    \includegraphics[width=0.13\textwidth]{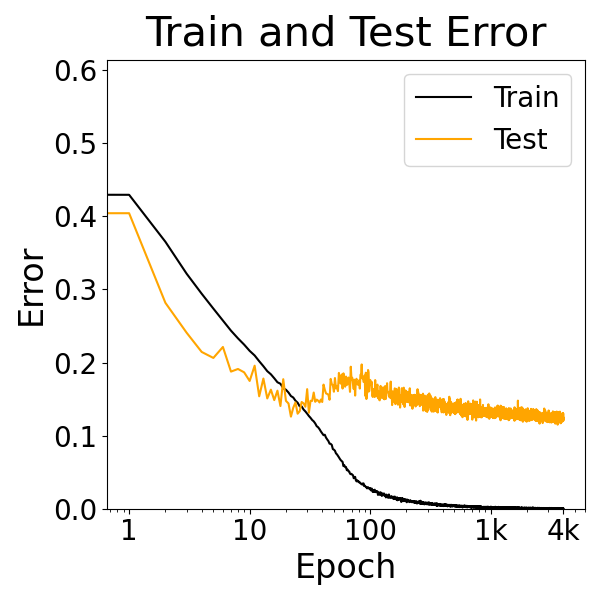}\label{Resnet18_k_64_error_curve_noise_10}
    }
  \caption{CKA evaluations for ResNet-18 trained on CIFAR-10 with 10\% label noise. Each of the (a)-(f) subfigures shows the CKA representational similarity of a specific layer in the ResNet-18 during its training. ResNet Block $j$ refers to the representation at the end of the $j^{\sf th}$ block of layers in the ResNet-18 architecture. (g) shows the test and train errors during training.}
  \label{fig:Resnet18_k_64_noise_10}
\end{figure*}

\begin{figure*}[t]
  \centering
      \subfloat[]{
    \includegraphics[width=0.13\textwidth]{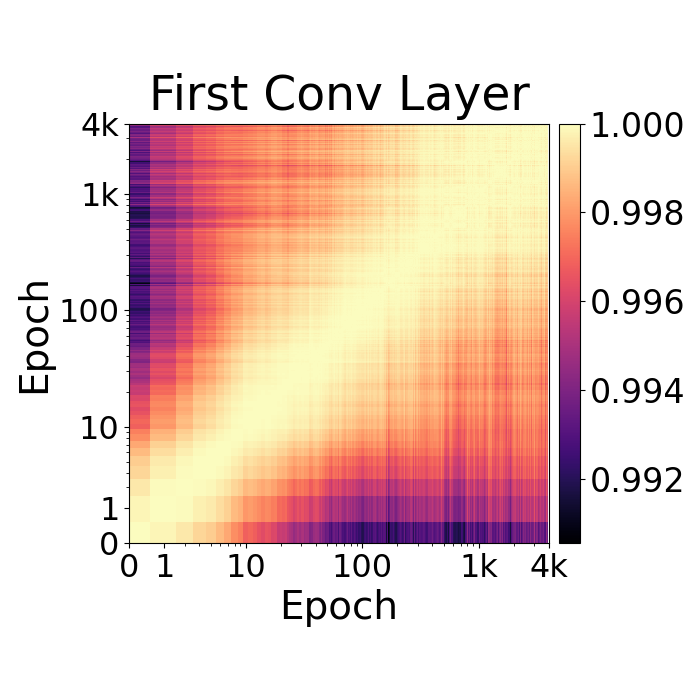}
    \label{Resnet18_k_32_first_conv_layer}}
    \subfloat[]{
    \includegraphics[width=0.13\textwidth]{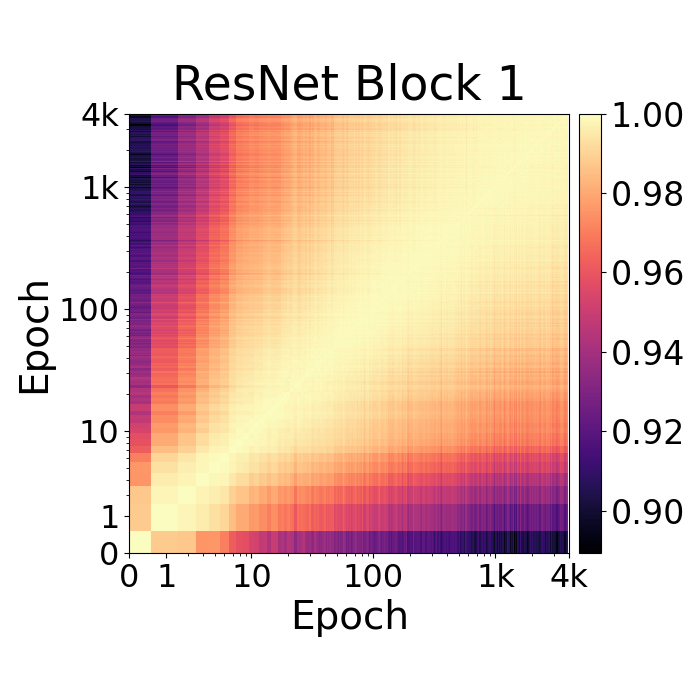}
    \label{Resnet18_k_32_block_1}
    }
    \subfloat[]{
    \includegraphics[width=0.13\textwidth]{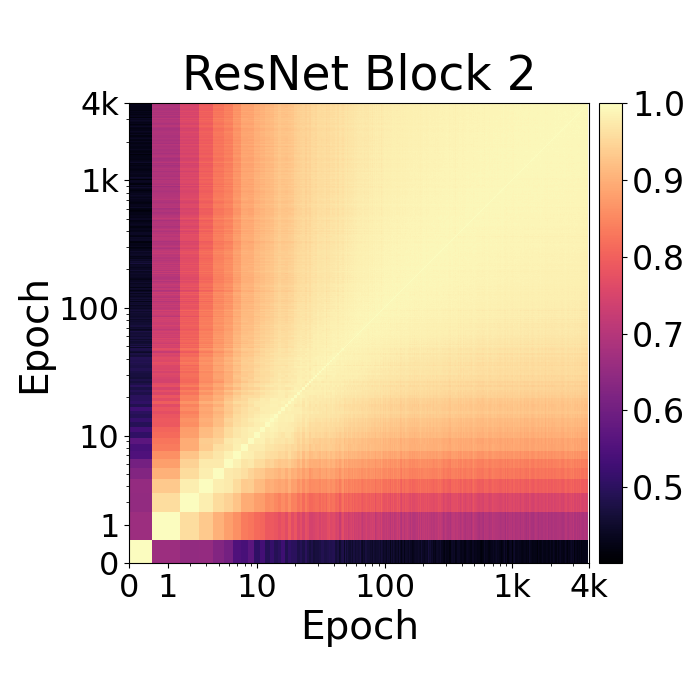}
    \label{Resnet18_k_32_block_2}}
    \subfloat[]{
    \includegraphics[width=0.13\textwidth]{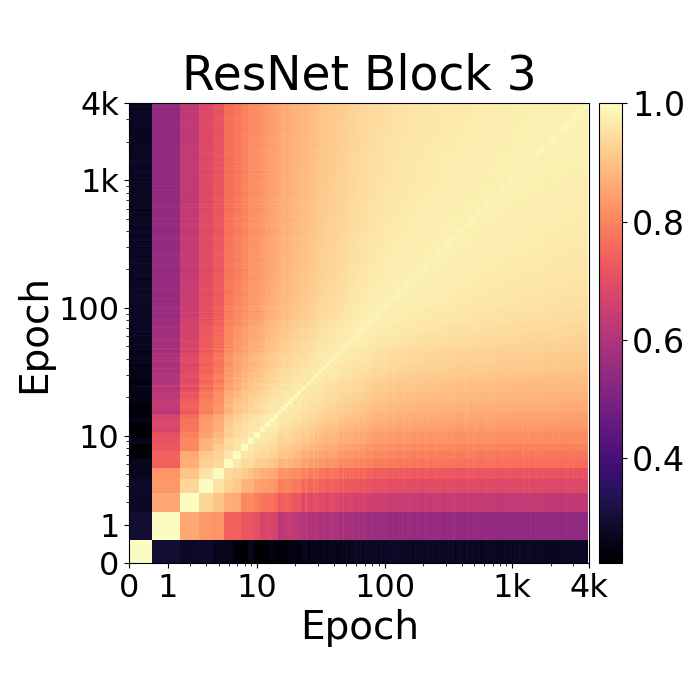}
    \label{Resnet18_k_32_block_3}}
    \subfloat[]{
    \includegraphics[width=0.13\textwidth]{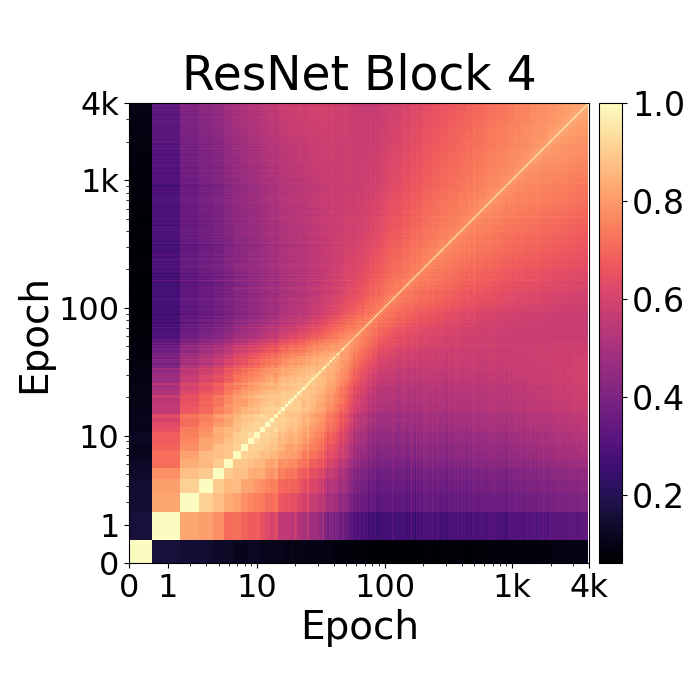} 
    \label{Resnet18_k_32_block_4}}
     \subfloat[]{
    \includegraphics[width=0.13\textwidth]{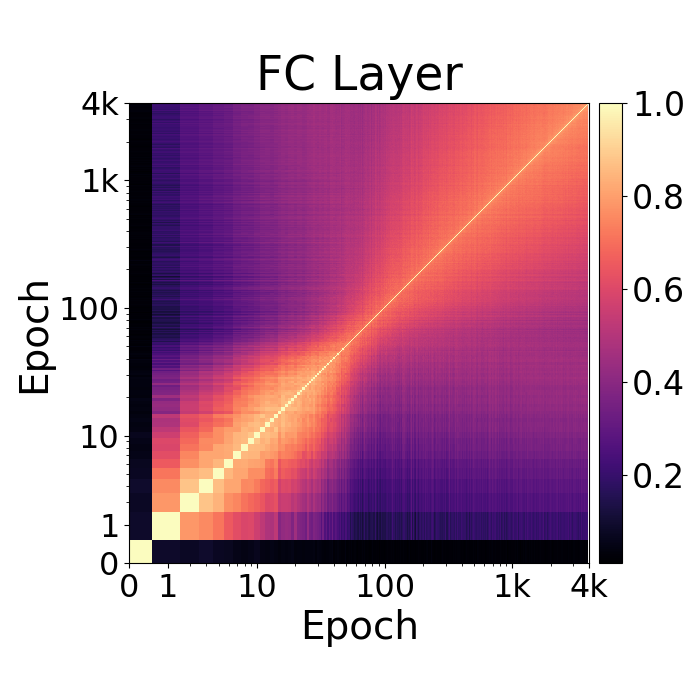}
    \label{Resnet18_k_32_FC}}
    \subfloat[]{
    \includegraphics[width=0.13\textwidth]{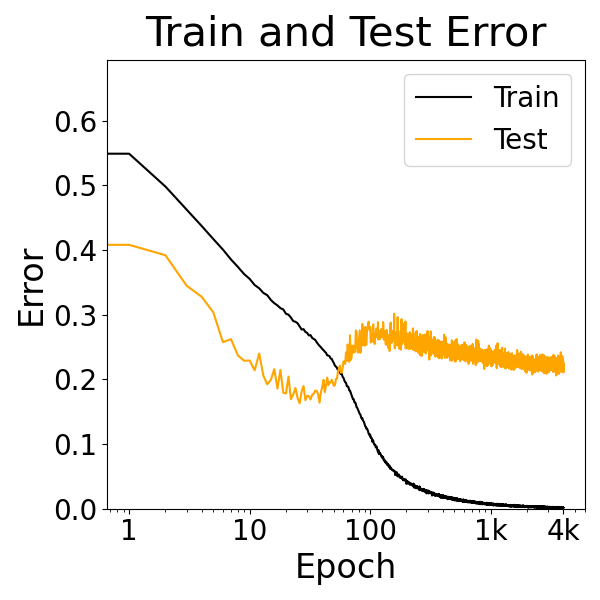}\label{Resnet18_k_32_error_curve}
    }
  \caption{CKA evaluations for ResNet-18 of \textbf{width parameter $k=32$} (which is narrower than the standard ResNet-18 width of $k=64$), trained on CIFAR-10 with 20\% label noise. Each of the (a)-(f) subfigures shows the CKA representational similarity of a specific layer in the ResNet-18 during its training. (g) shows the test and train errors during training.}
  \label{fig:Resnet18_k_32_noise_20}
\end{figure*}

\begin{figure*}[t]
 \centering
      \subfloat[]{
    \includegraphics[width=0.13\textwidth]{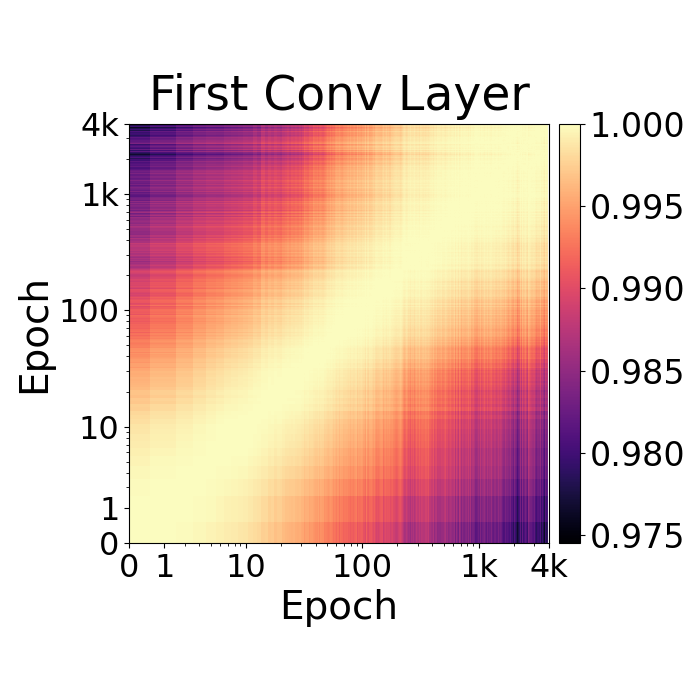}
    \label{Resnet18_k_42_first_conv_layer}}
    \subfloat[]{
    \includegraphics[width=0.13\textwidth]{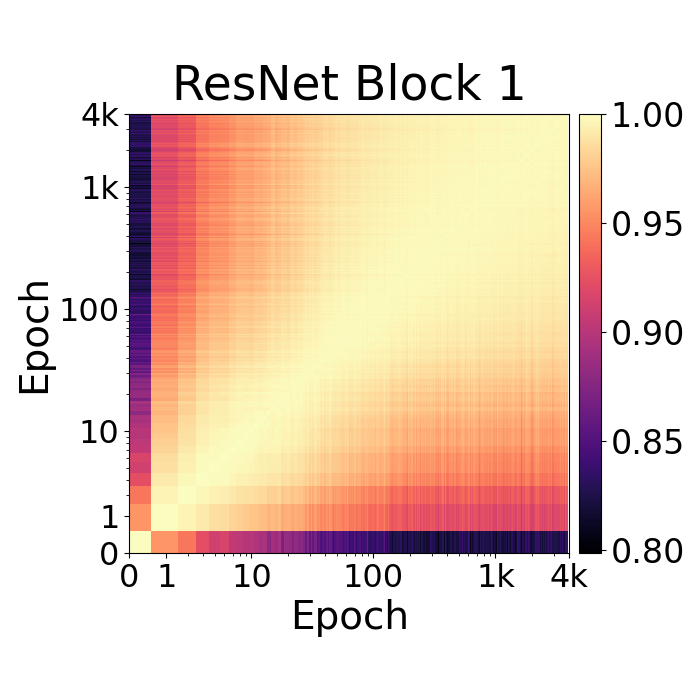}
    \label{Resnet18_k_42_block_1}
    }
    \subfloat[]{
    \includegraphics[width=0.13\textwidth]{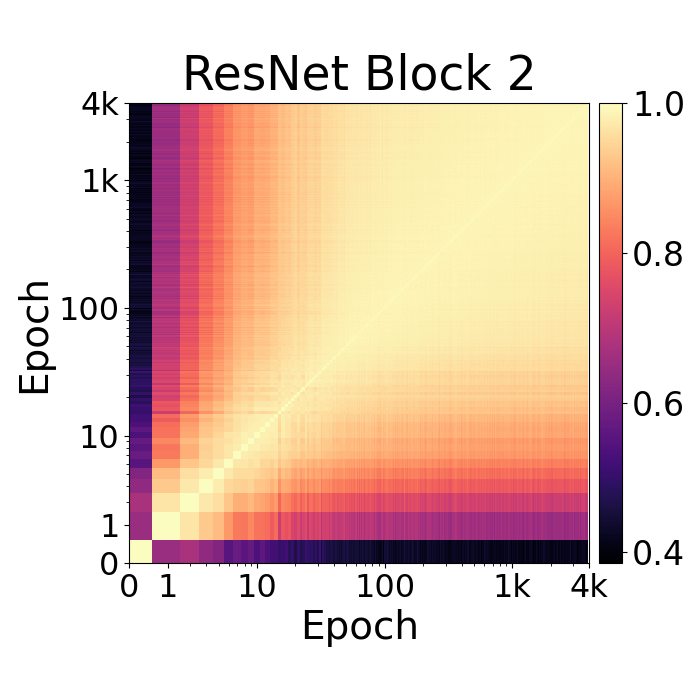}
    \label{Resnet18_k_42_block_2}}
    \subfloat[]{
    \includegraphics[width=0.13\textwidth]{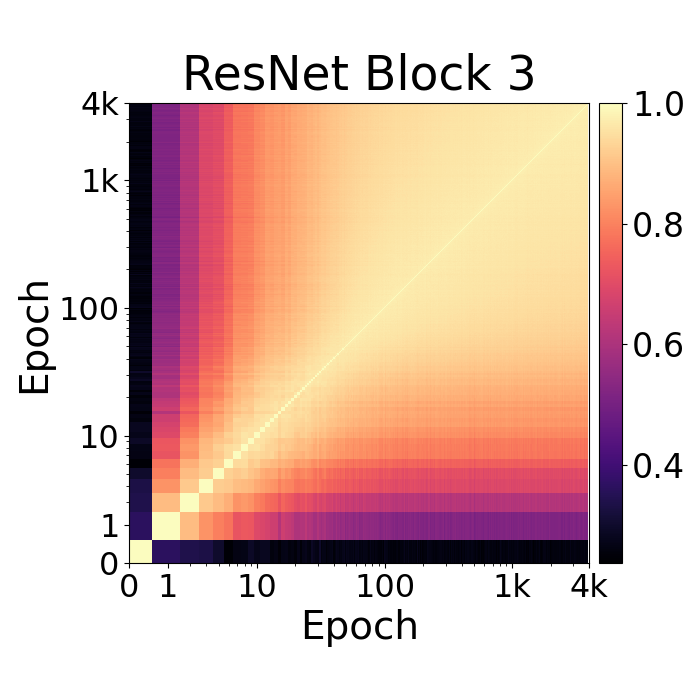}
    \label{Resnet18_k_42_block_3}}
    \subfloat[]{
    \includegraphics[width=0.13\textwidth]{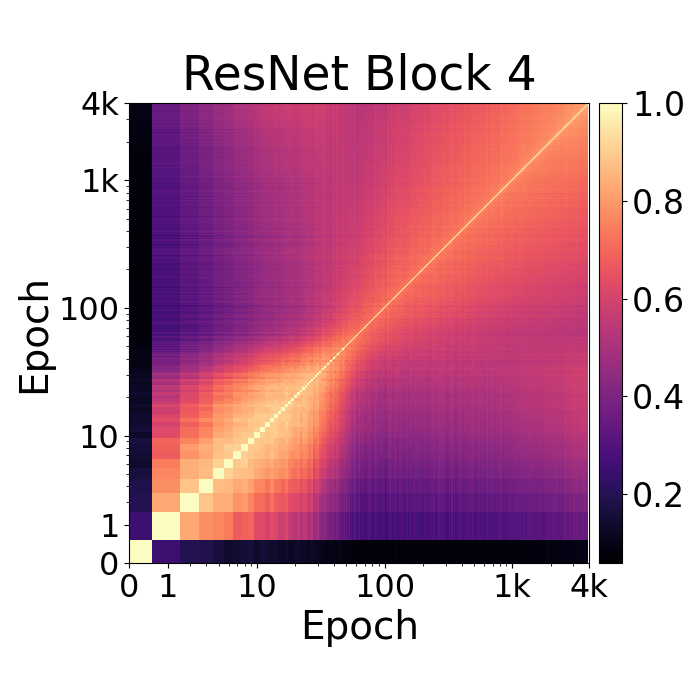} 
    \label{Resnet18_k_42_block_4}}
     \subfloat[]{
    \includegraphics[width=0.13\textwidth]{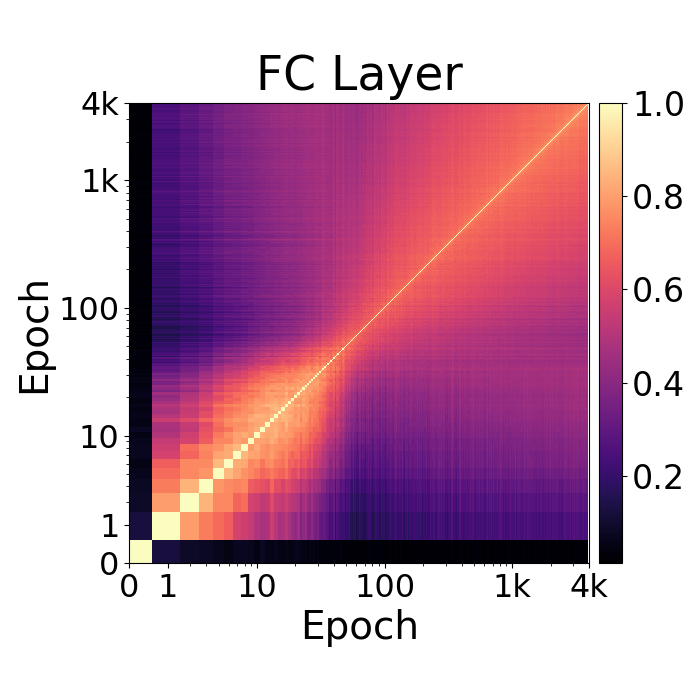}
    \label{Resnet18_k_42_FC}}
    \subfloat[]{
    \includegraphics[width=0.13\textwidth]{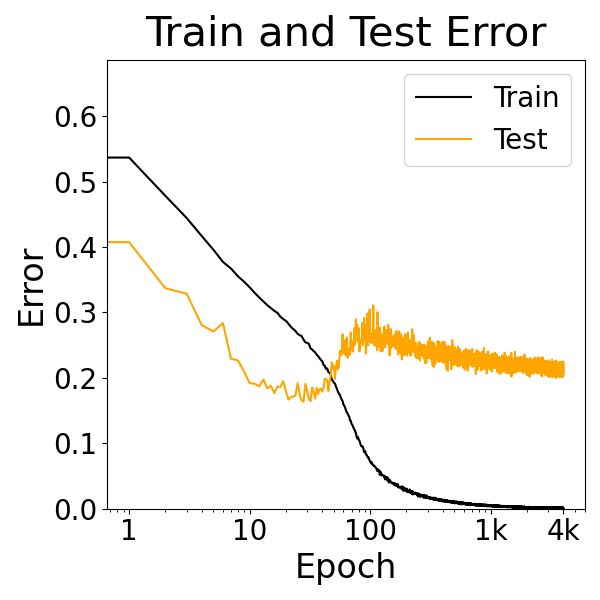}\label{Resnet18_k_42_error_curve}
    }
  \caption{CKA evaluations for ResNet-18 of \textbf{width parameter $k=42$} (which is narrower than the standard ResNet-18 width of $k=64$), trained on CIFAR-10 with 20\% label noise. Each of the (a)-(f) subfigures shows the CKA representational similarity of a specific layer in the ResNet-18 during its training. (g) shows the test and train errors during training.}
  \label{fig:Resnet18_k_42_noise_20}
\end{figure*}

\begin{figure*}[t]
 \centering
      \subfloat[]{
    \includegraphics[width=0.13\textwidth]{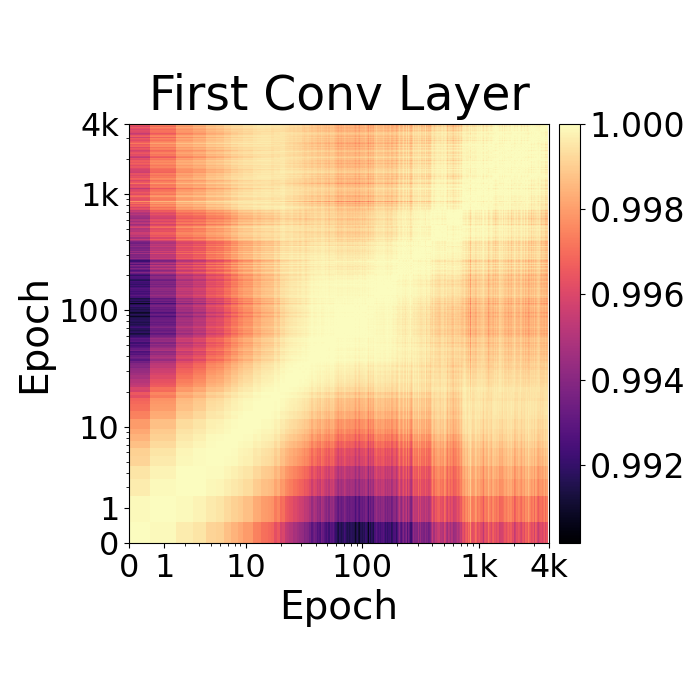}
    \label{Resnet18_k_52_first_conv_layer}}
    \subfloat[]{
    \includegraphics[width=0.13\textwidth]{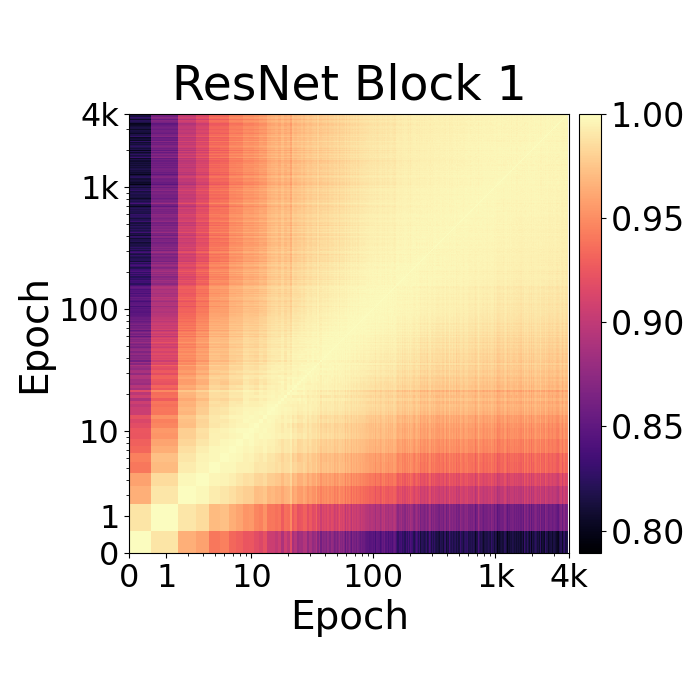}
    \label{Resnet18_k_52_block_1}
    }
    \subfloat[]{
    \includegraphics[width=0.13\textwidth]{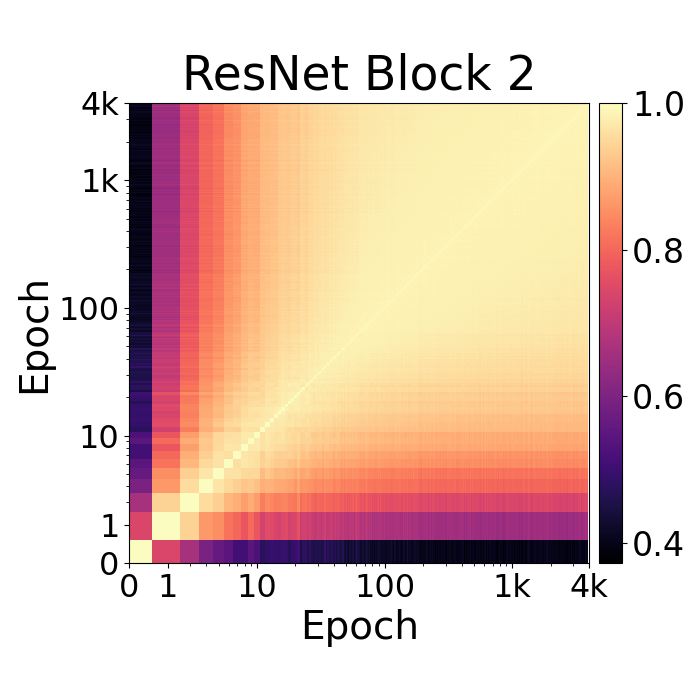}
    \label{Resnet18_k_52_block_2}}
    \subfloat[]{
    \includegraphics[width=0.13\textwidth]{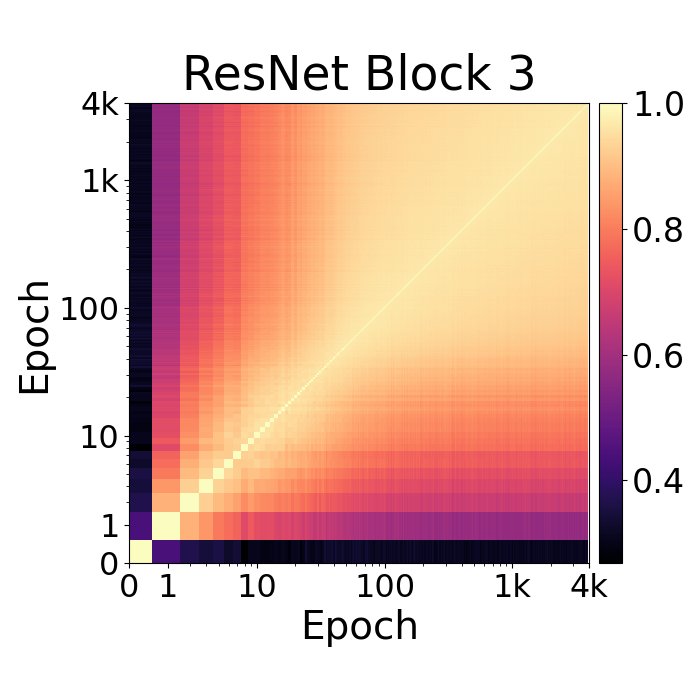}
    \label{Resnet18_k_52_block_3}}
    \subfloat[]{
    \includegraphics[width=0.13\textwidth]{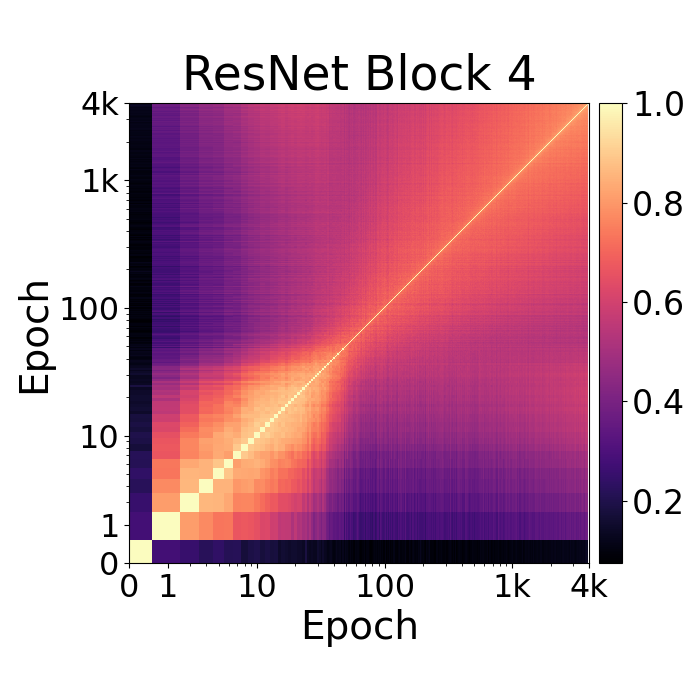} 
    \label{Resnet18_k_52_block_4}}
     \subfloat[]{
    \includegraphics[width=0.13\textwidth]{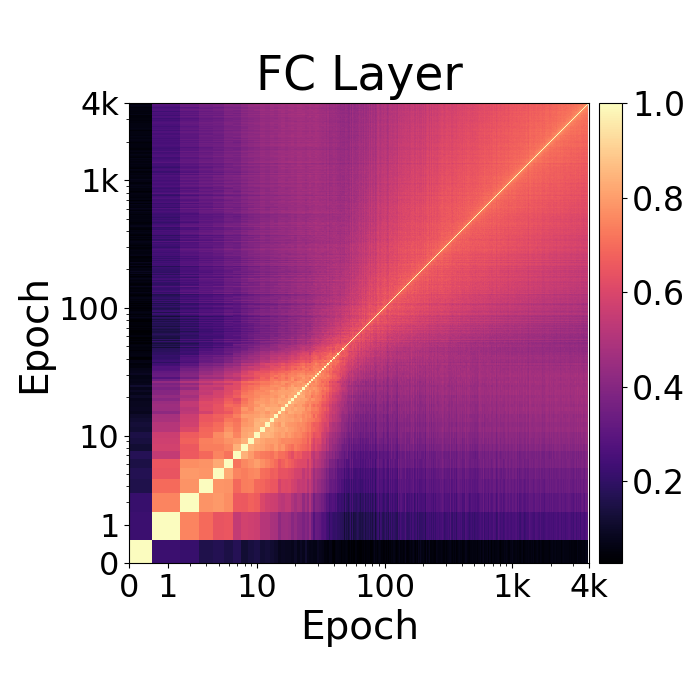}
    \label{Resnet18_k_52_FC}}
    \subfloat[]{
    \includegraphics[width=0.13\textwidth]{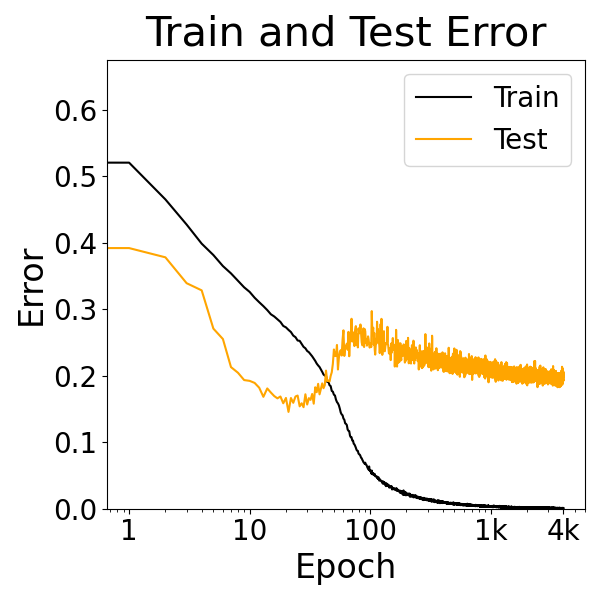}\label{Resnet18_k_52_error_curve}
    }
  \caption{CKA evaluations for ResNet-18 of \textbf{width parameter $k=52$} (which is narrower than the standard ResNet-18 width of $k=64$), trained on CIFAR-10 with 20\% label noise. Each of the (a)-(f) subfigures shows the CKA representational similarity of a specific layer in the ResNet-18 during its training. (g) shows the test and train errors during training.}
  \label{fig:Resnet18_k_52_noise_20}
\end{figure*}

\begin{figure*}[t]
 \centering
      \subfloat[]{
    \includegraphics[width=0.13\textwidth]{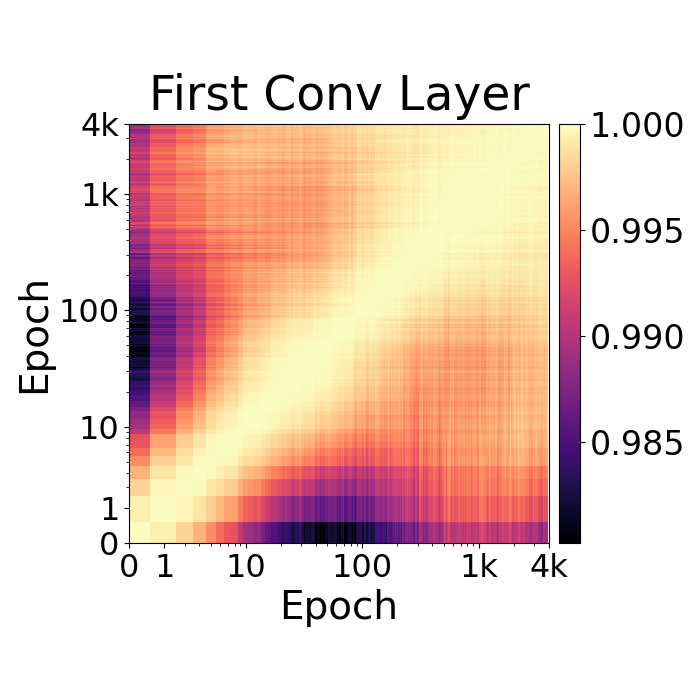}
    \label{Resnet18_k_64_first_conv_layer_tin_noise_0}}
    \subfloat[]{
    \includegraphics[width=0.13\textwidth]{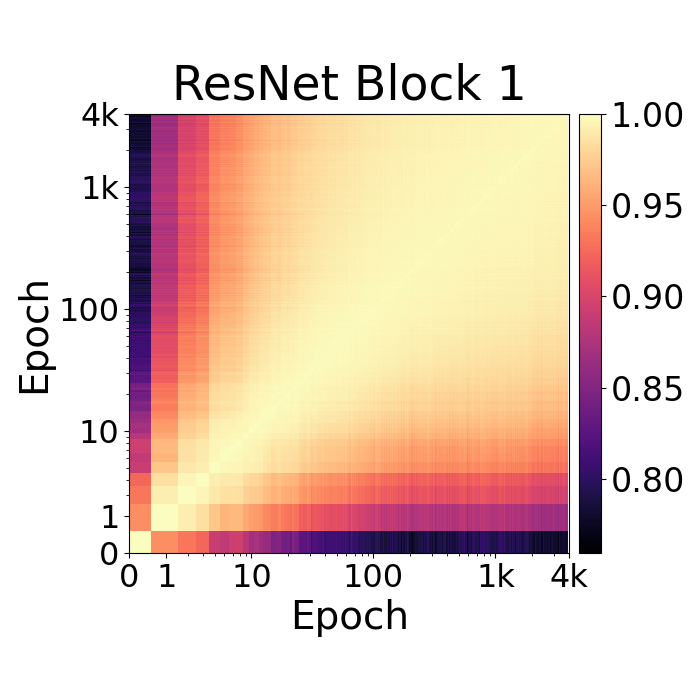}
    \label{Resnet18_k_64_block_1_tin_noise_0}
    }
    \subfloat[]{
    \includegraphics[width=0.13\textwidth]{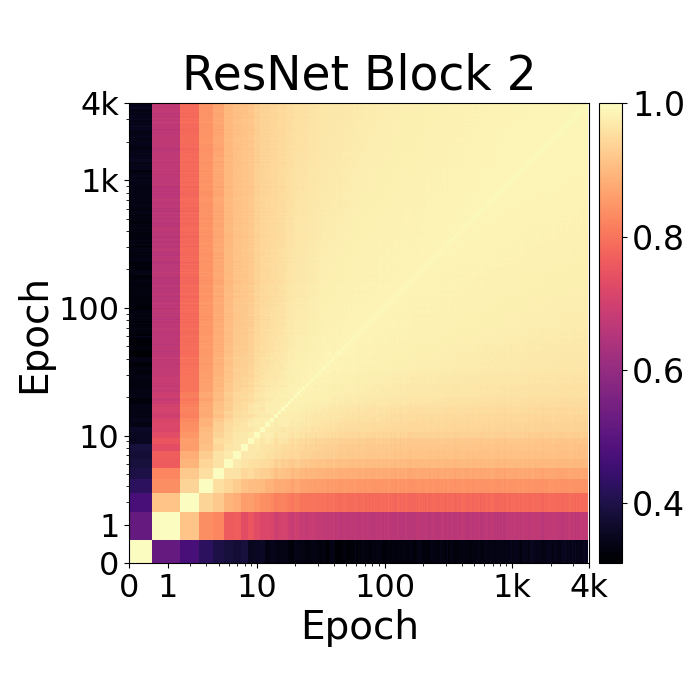}
    \label{Resnet18_k_64_block_2_tin_noise_0}}
    \subfloat[]{
    \includegraphics[width=0.13\textwidth]{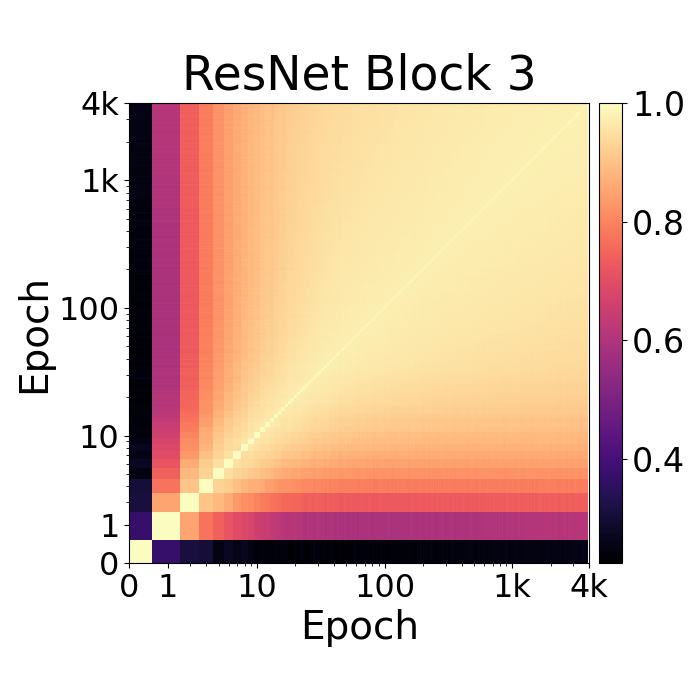}
    \label{Resnet18_k_64_block_3_tin_noise_0}}
    \subfloat[]{
    \includegraphics[width=0.13\textwidth]{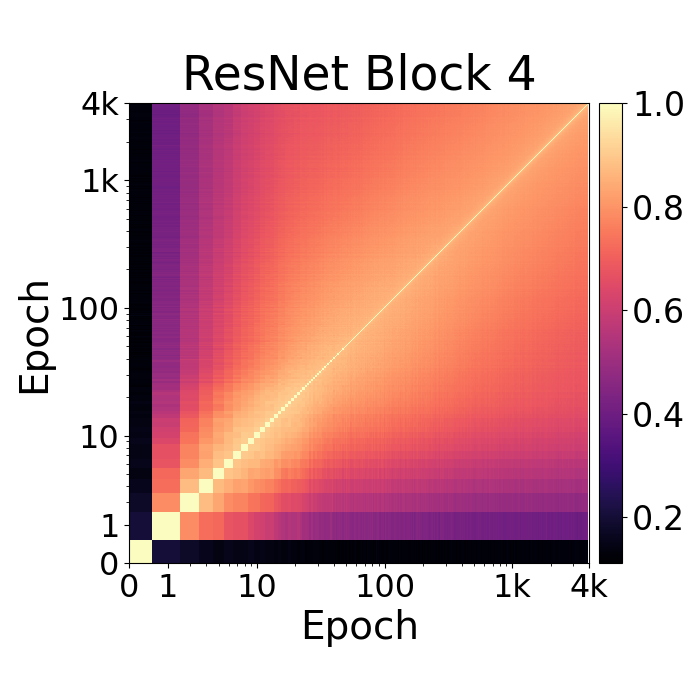} 
    \label{Resnet18_k_64_block_4_tin_noise_0}}
     \subfloat[]{
    \includegraphics[width=0.13\textwidth]{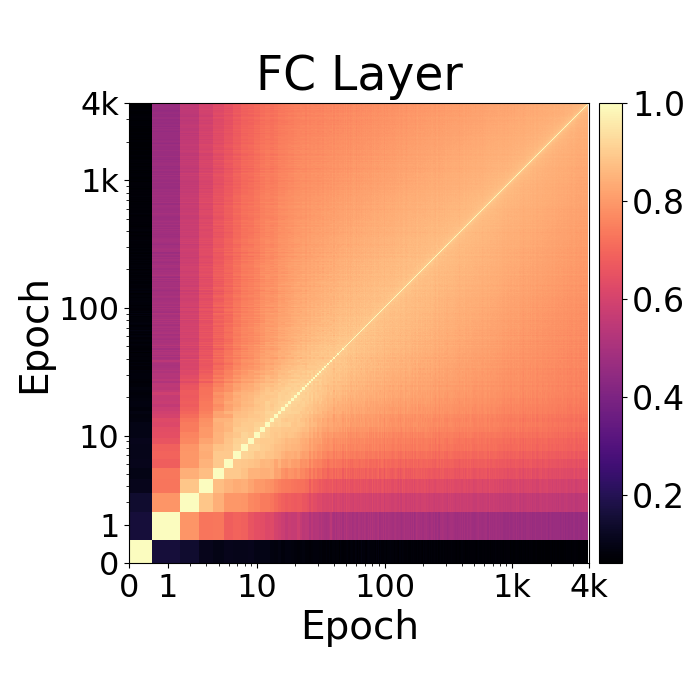}
    \label{Resnet18_k_64_FC_tin_noise_0}}
    \subfloat[]{
    \includegraphics[width=0.13\textwidth]{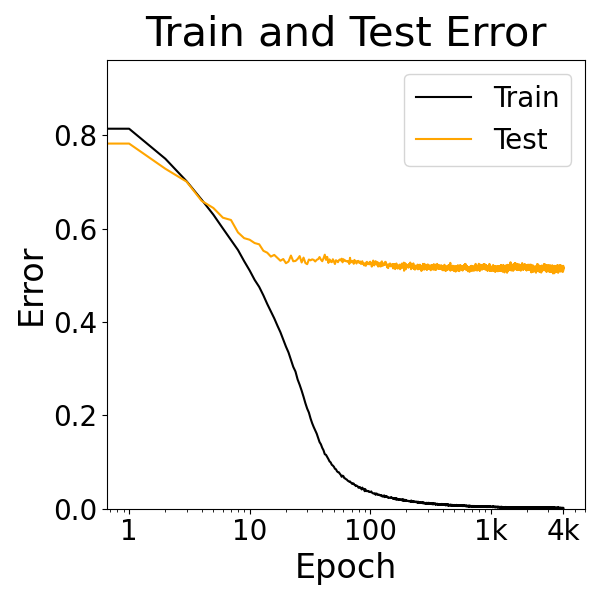}\label{Resnet18_k_64_error_curve_tin_noise_0}
    }
  \caption{CKA evaluations for ResNet-18 trained on \textbf{Tiny ImageNet} (image size $32\times32\times3$) without label noise. }
  \label{fig:Resnet18_k_64_noise_0_tin}
\end{figure*}

\begin{figure*}[t]
 \centering
      \subfloat[]{
    \includegraphics[width=0.13\textwidth]{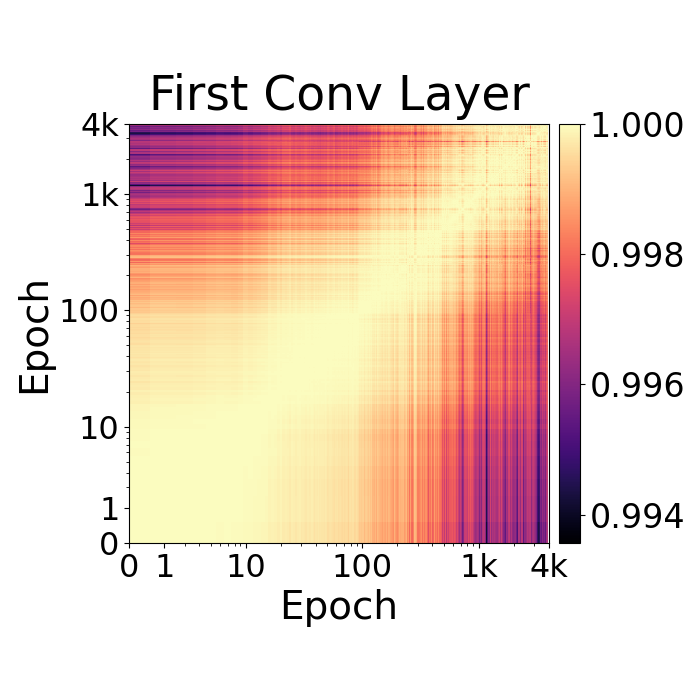}
    \label{Resnet50_k_64_first_conv_layer_noise_20}}
    \subfloat[]{
    \includegraphics[width=0.13\textwidth]{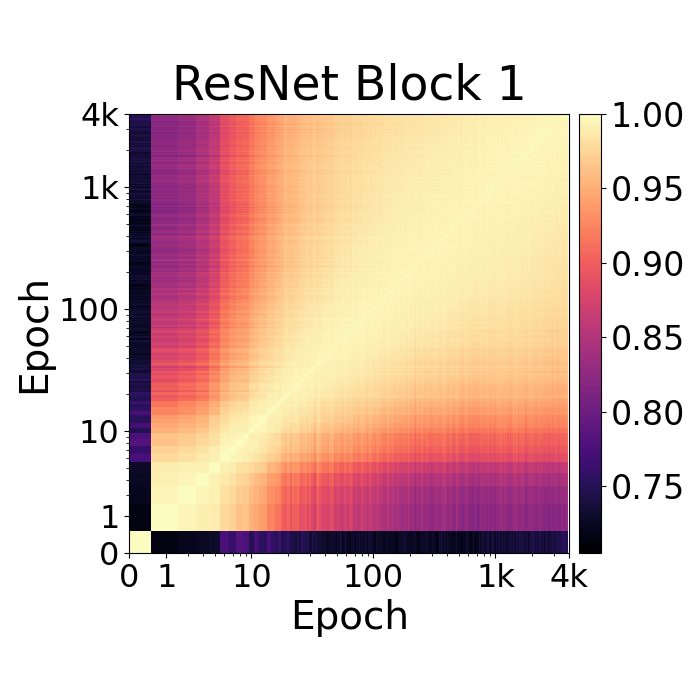}
    \label{Resnet50_k_64_block_1_noise_20}
    }
    \subfloat[]{
    \includegraphics[width=0.13\textwidth]{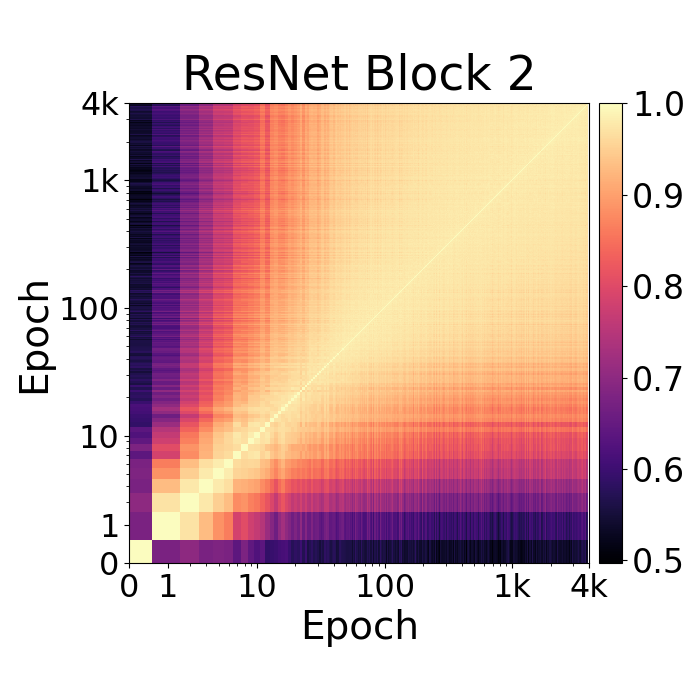}
    \label{Resnet50_k_64_block_2_noise_20}}
    \subfloat[]{
    \includegraphics[width=0.13\textwidth]{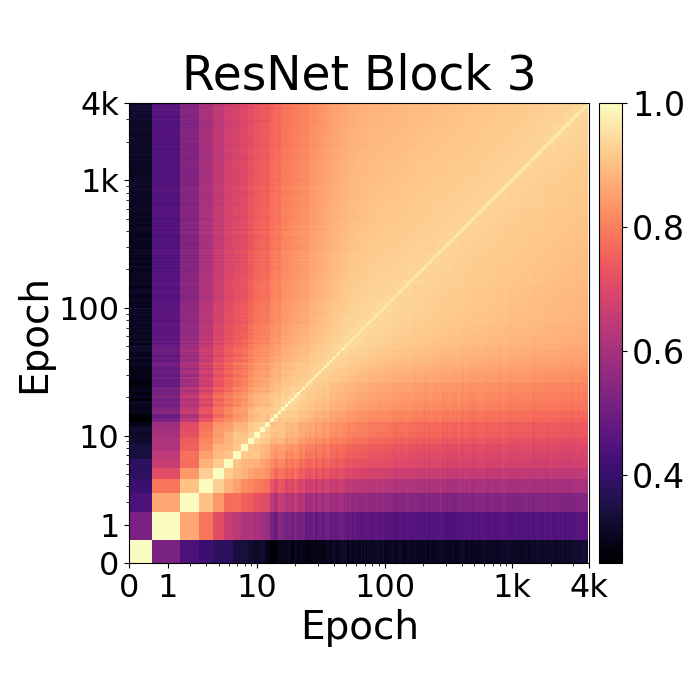}
    \label{Resnet50_k_64_block_3_noise_20}}
    \subfloat[]{
    \includegraphics[width=0.13\textwidth]{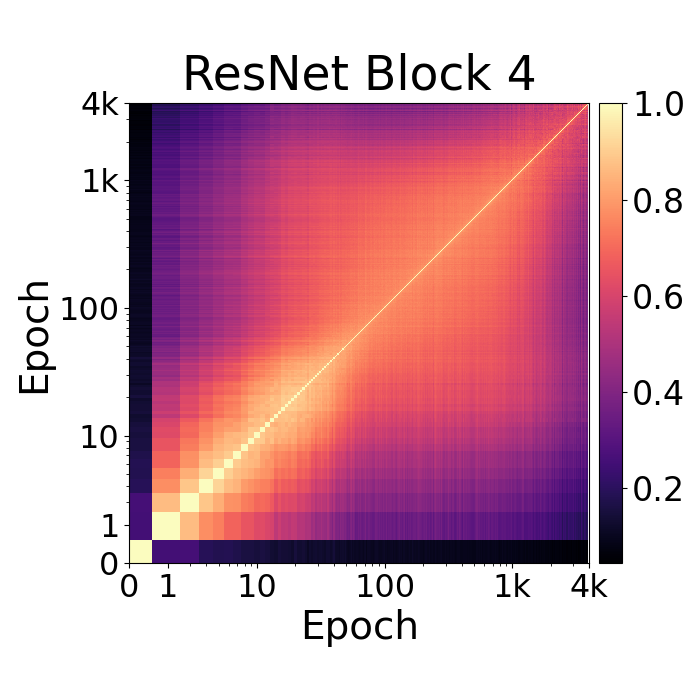} 
    \label{Resnet50_k_64_block_4_noise_20}}
     \subfloat[]{
    \includegraphics[width=0.13\textwidth]{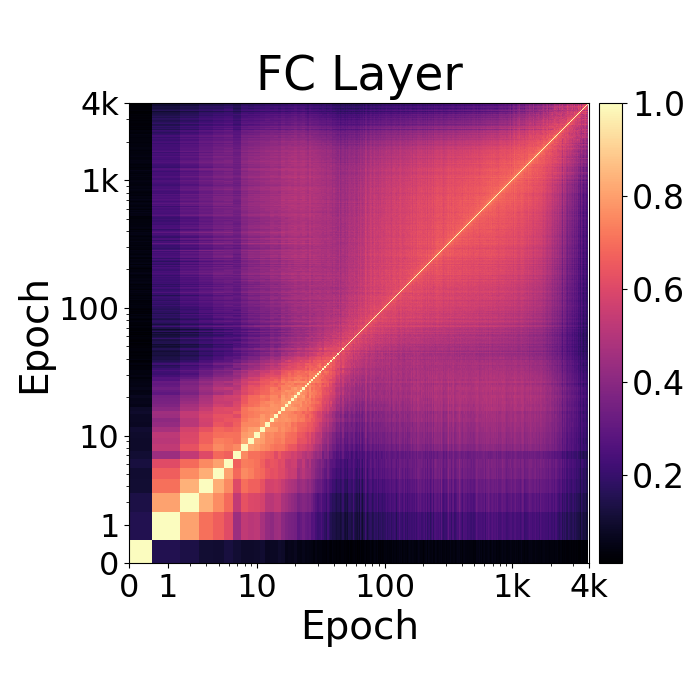}
    \label{Resnet50_k_64_FC_noise_20}}
    \subfloat[]{
    \includegraphics[width=0.13\textwidth]{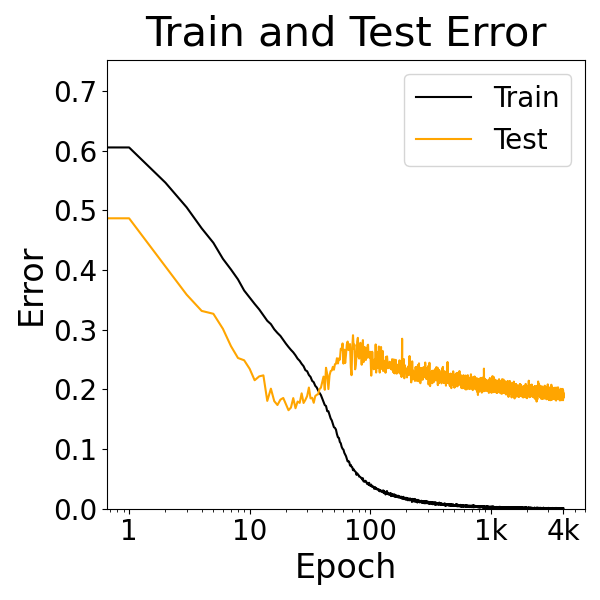}\label{Resnet50_k_64_error_curve_noise_20}
    }
  \caption{CKA evaluations for ResNet-50, trained on CIFAR-10 with 20\% label noise.}
  \label{fig:Resnet50_k_64_noise_0}
\end{figure*}

\begin{figure*}[t]
  \centering
    \subfloat[]{
    \includegraphics[width=0.13\textwidth]{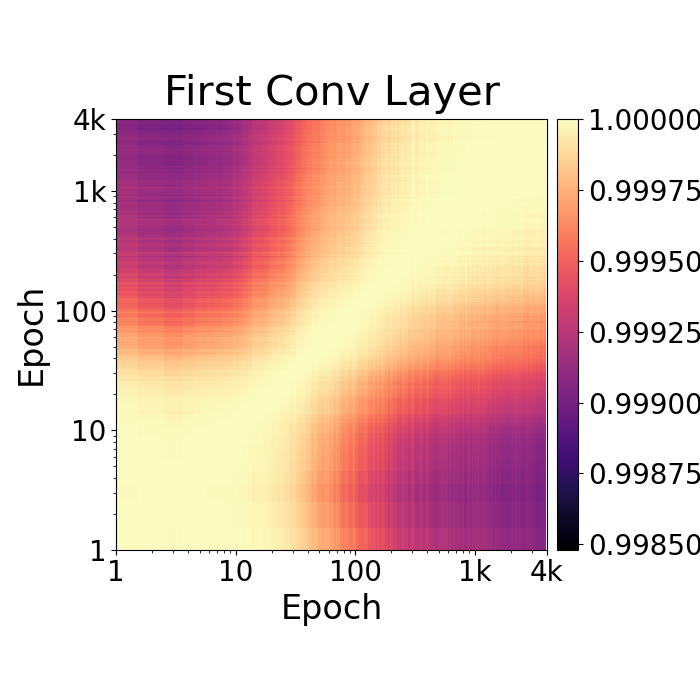}
    \label{sgd_0_0001_Resnet18_k_64_first_conv_layer_noise_0}}
    \subfloat[]{
    \includegraphics[width=0.13\textwidth]{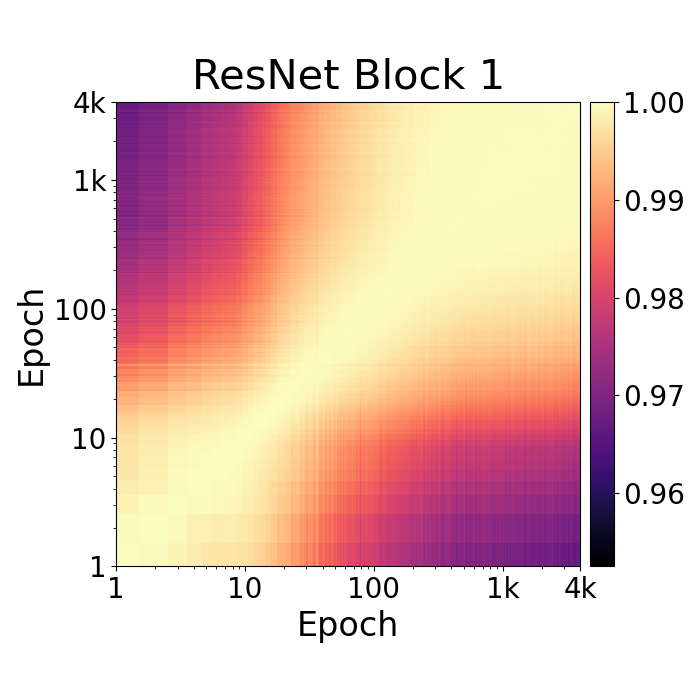}
    \label{sgd_0_0001_Resnet18_k_64_block_1_noise_0}
    }
    \subfloat[]{
    \includegraphics[width=0.13\textwidth]{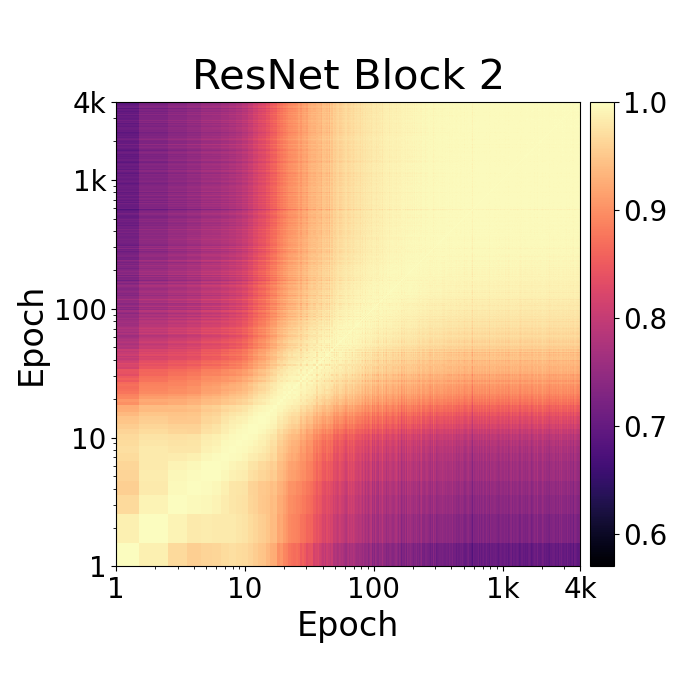}
    \label{sgd_0_0001_Resnet18_k_64_block_2_noise_0}}
    \subfloat[]{
    \includegraphics[width=0.13\textwidth]{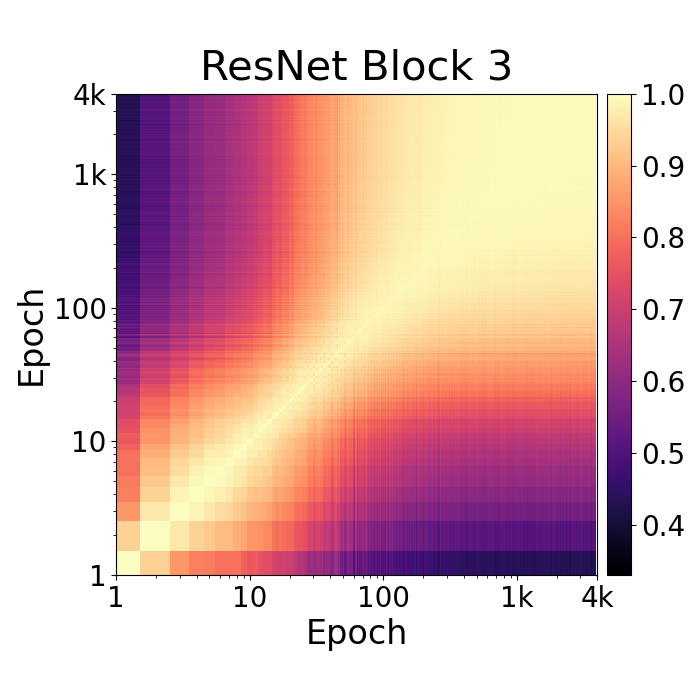}
    \label{sgd_0_0001_Resnet18_k_64_block_3_noise_0}}
    \subfloat[]{
    \includegraphics[width=0.13\textwidth]{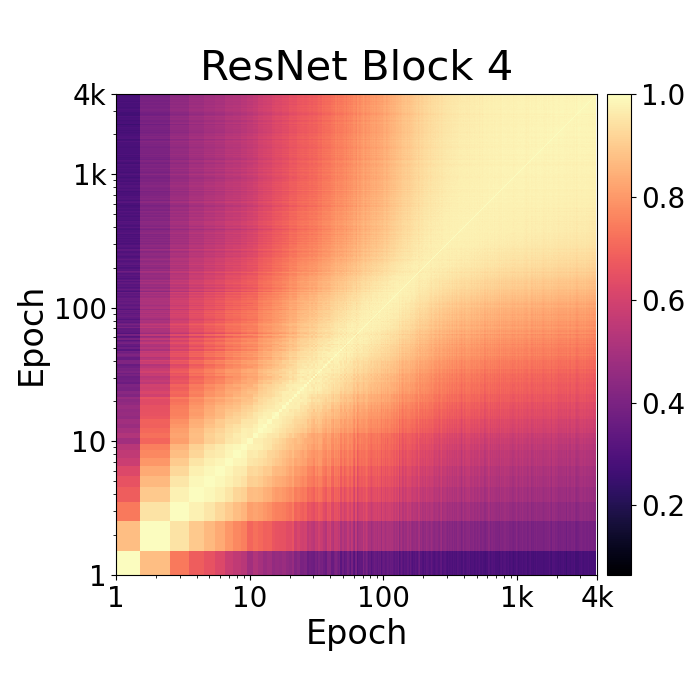}
    \label{sgd_0_0001_Resnet18_k_64_block_4_noise_0}}
     \subfloat[]{
    \includegraphics[width=0.13\textwidth]{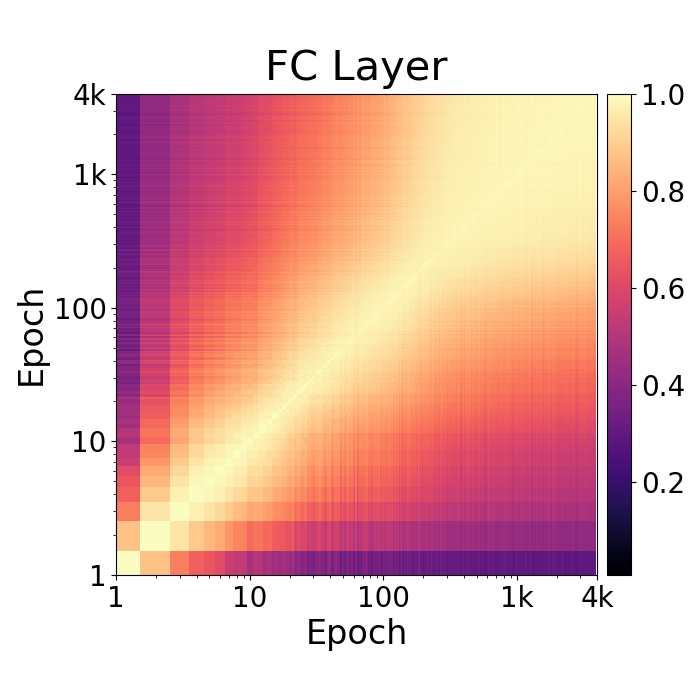}
    \label{sgd_0_0001_Resnet18_k_64_FC_noise_0}}
    \subfloat[]{
    \includegraphics[width=0.13\textwidth]{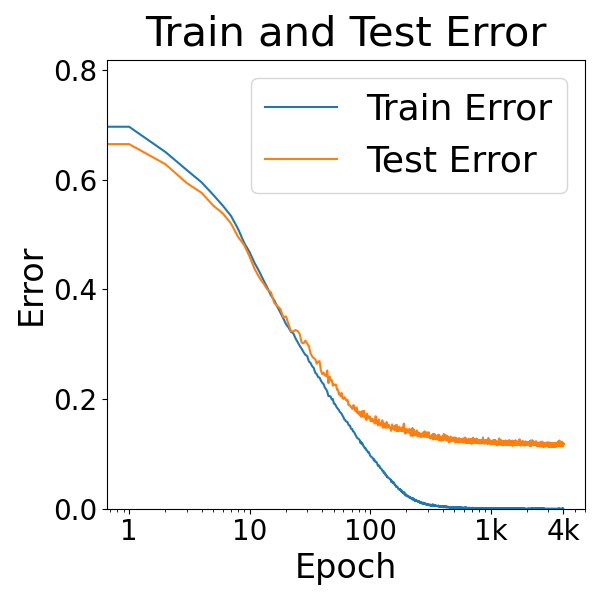}
    \label{sgd_0_0001_Resnet18_k_64_error_curve_noise_0}}
  \caption{CKA evaluations for ResNet-18 trained on CIFAR-10 without label noise. Here, the optimizer is \textbf{SGD} with constant learning rate 0.0001 and \textbf{momentum} of 0.9.}
  \label{fig:sgd_0_0001_Resnet18_k_64_noise_0}
\end{figure*}

\begin{figure*}[t]
  \centering
    \subfloat[]{
    \includegraphics[width=0.13\textwidth]{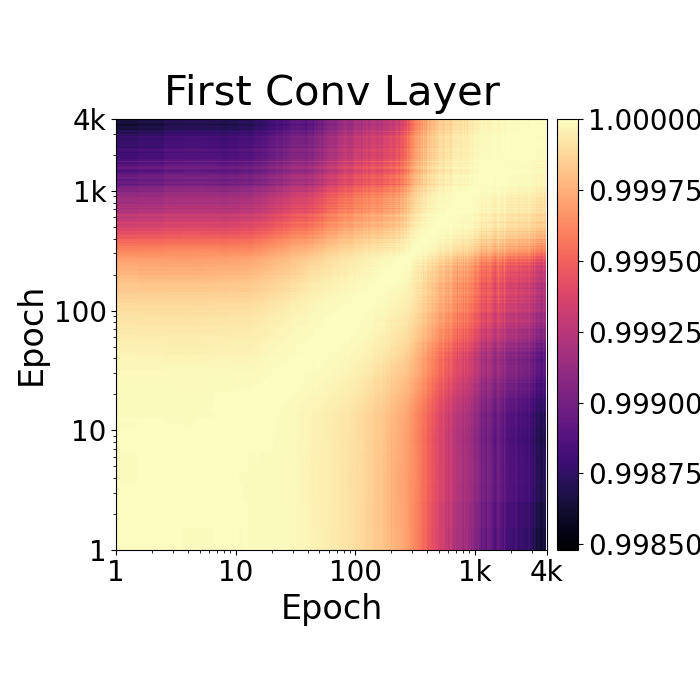}
    \label{sgd_0_0001_Resnet18_k_64_first_conv_layer_noise_20}}
    \subfloat[]{
    \includegraphics[width=0.13\textwidth]{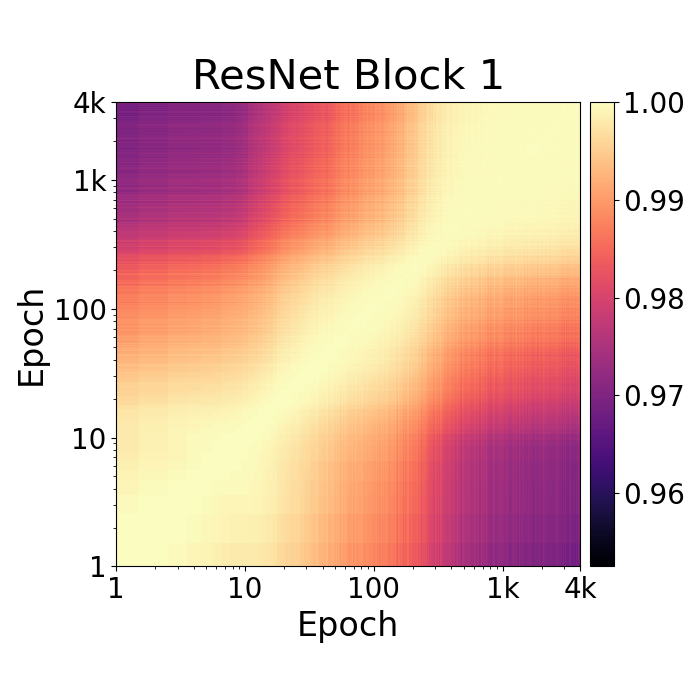}
    \label{sgd_0_0001_Resnet18_k_64_block_1_noise_20}
    }
    \subfloat[]{
    \includegraphics[width=0.13\textwidth]{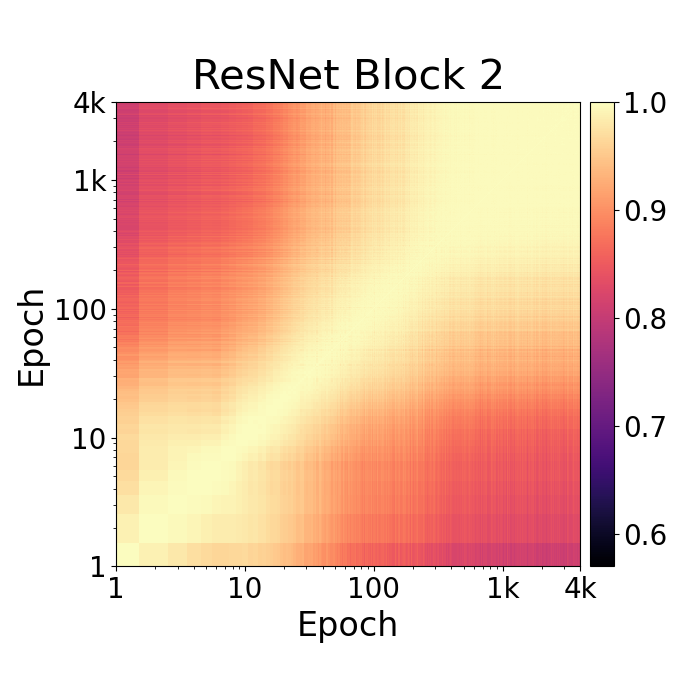}
    \label{sgd_0_0001_Resnet18_k_64_block_2_noise_20}}
    \subfloat[]{
    \includegraphics[width=0.13\textwidth]{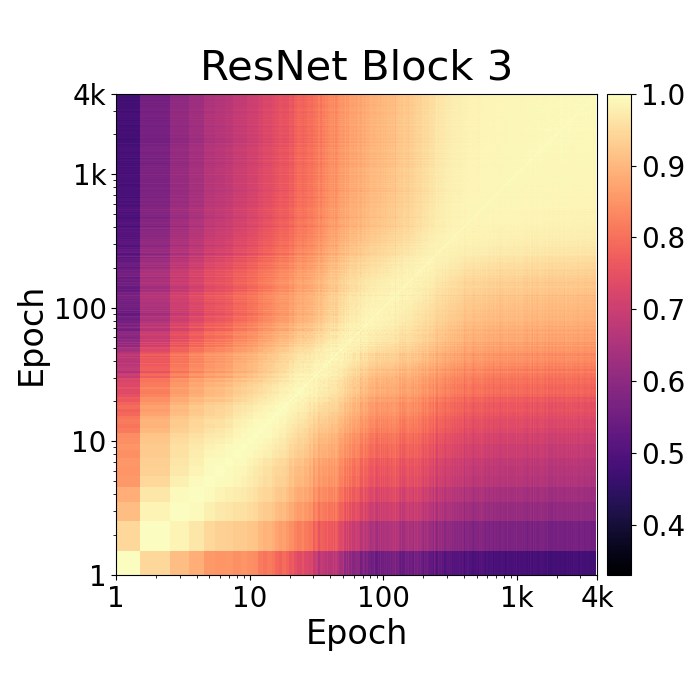}
    \label{sgd_0_0001_Resnet18_k_64_block_3_noise_20}}
    \subfloat[]{
    \includegraphics[width=0.13\textwidth]{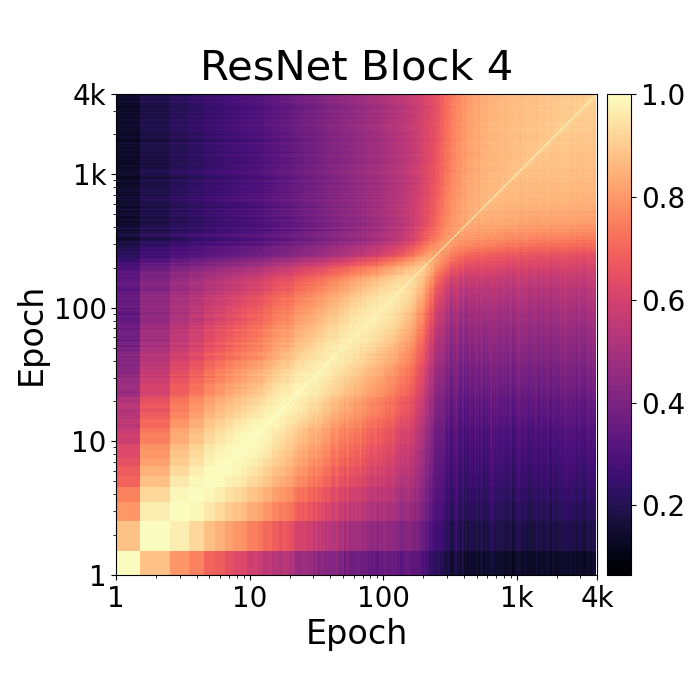}
    \label{sgd_0_0001_Resnet18_k_64_block_4_noise_20}}
     \subfloat[]{
    \includegraphics[width=0.13\textwidth]{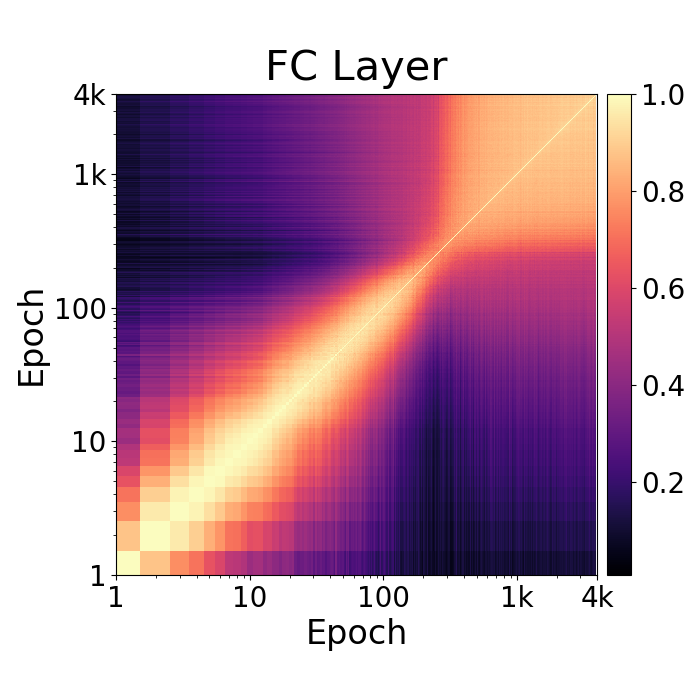}
    \label{sgd_0_0001_Resnet18_k_64_FC_noise_20}}
    \subfloat[]{
    \includegraphics[width=0.13\textwidth]{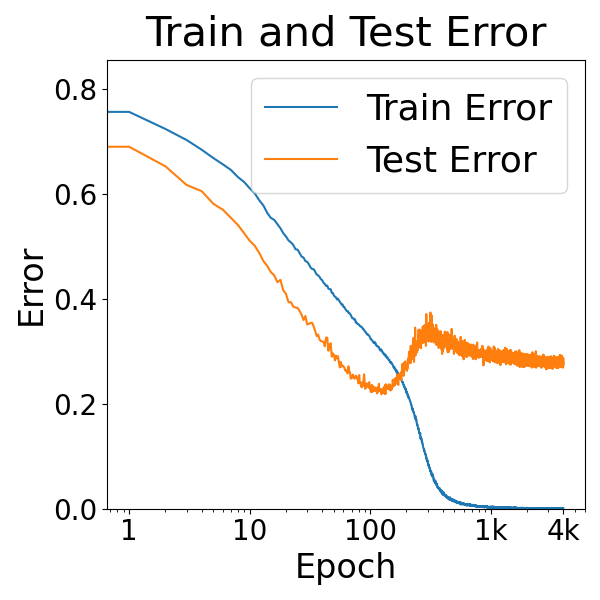}
    \label{sgd_0_0001_Resnet18_k_64_error_curve_noise_20}}
  \caption{CKA evaluations for ResNet-18 trained on CIFAR-10 with 20\% label noise. Here, the optimizer is \textbf{SGD} with constant learning rate 0.0001 and \textbf{momentum} of 0.9.}
  \label{fig:sgd_0_0001_Resnet18_k_64_noise_20}
\end{figure*}

\begin{figure*}[t]
  \centering
    \subfloat[]{
    \includegraphics[width=0.13\textwidth]{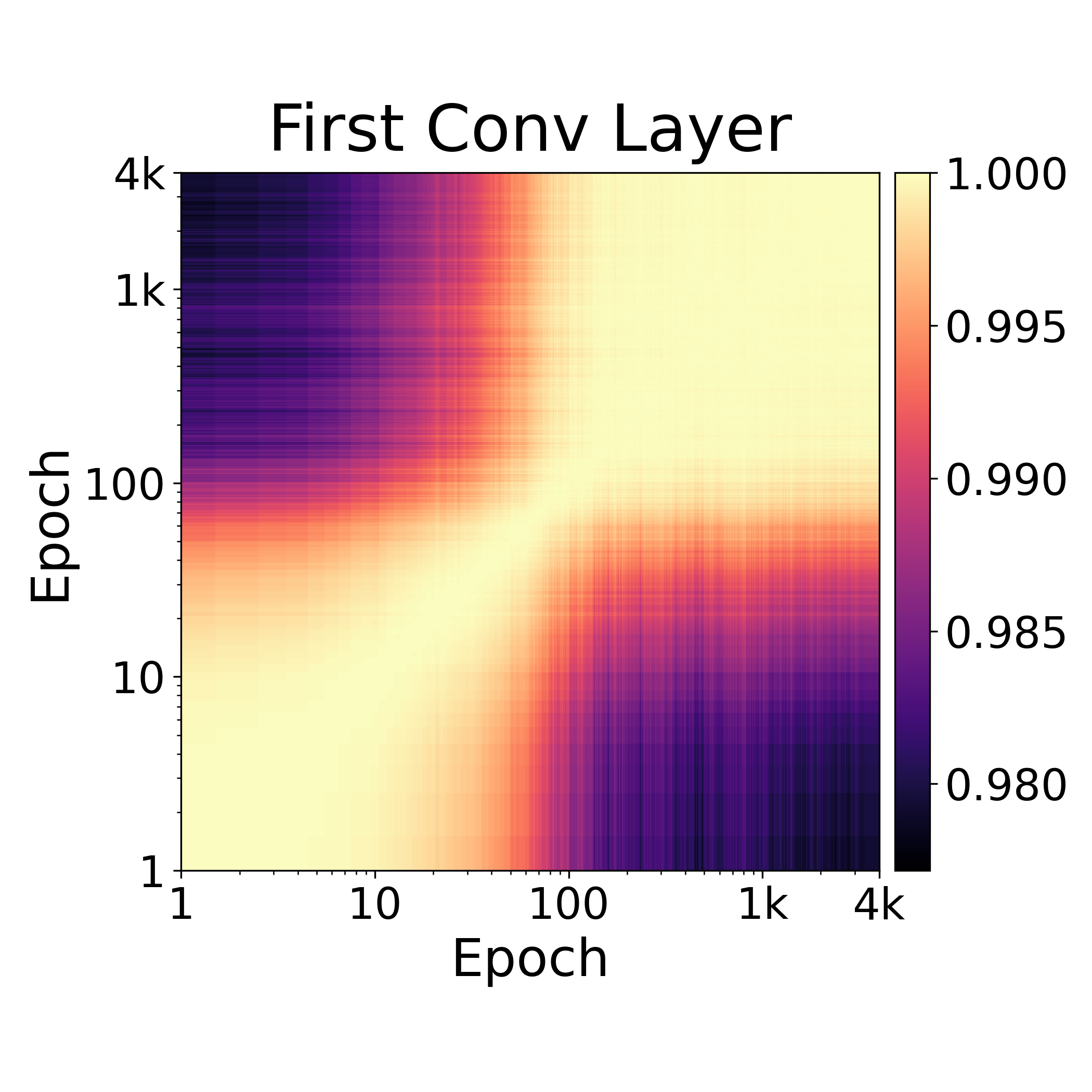}
    \label{sgd_0_01_Resnet18_k_64_first_conv_layer_noise_20}}
    \subfloat[]{
    \includegraphics[width=0.13\textwidth]{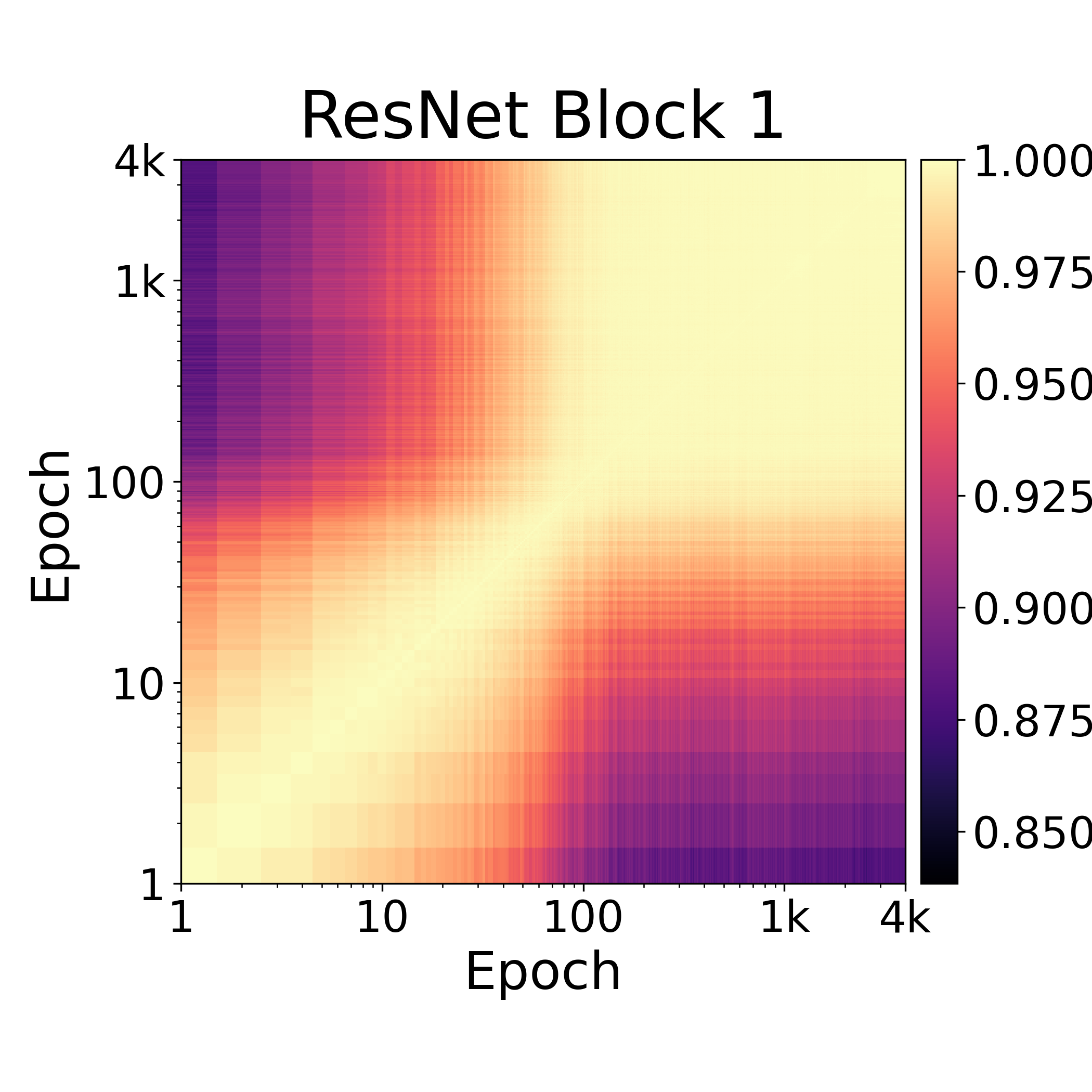}
    \label{sgd_0_01_Resnet18_k_64_block_1_noise_20}
    }
    \subfloat[]{
    \includegraphics[width=0.13\textwidth]{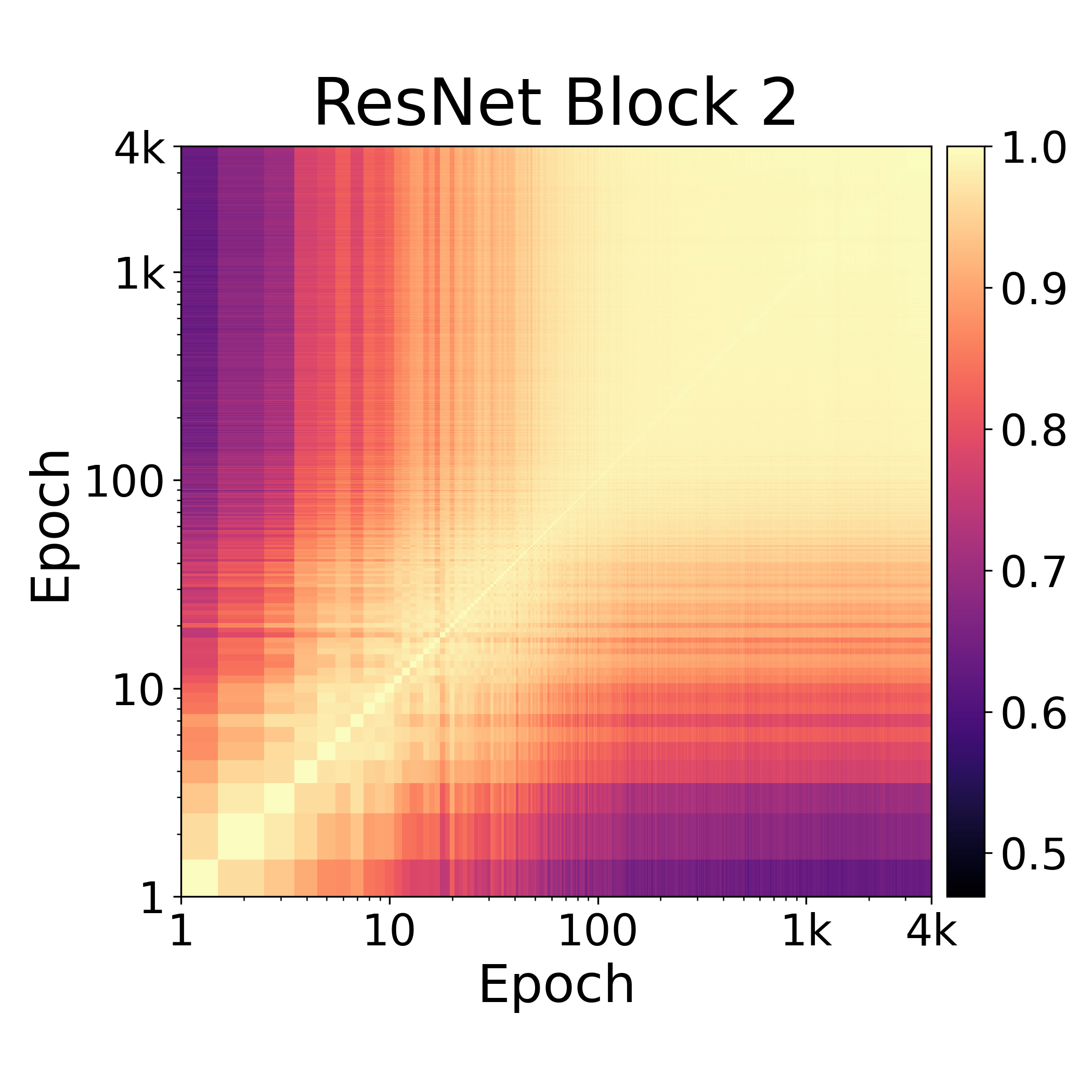}
    \label{sgd_0_01_Resnet18_k_64_block_2_noise_20}}
    \subfloat[]{
    \includegraphics[width=0.13\textwidth]{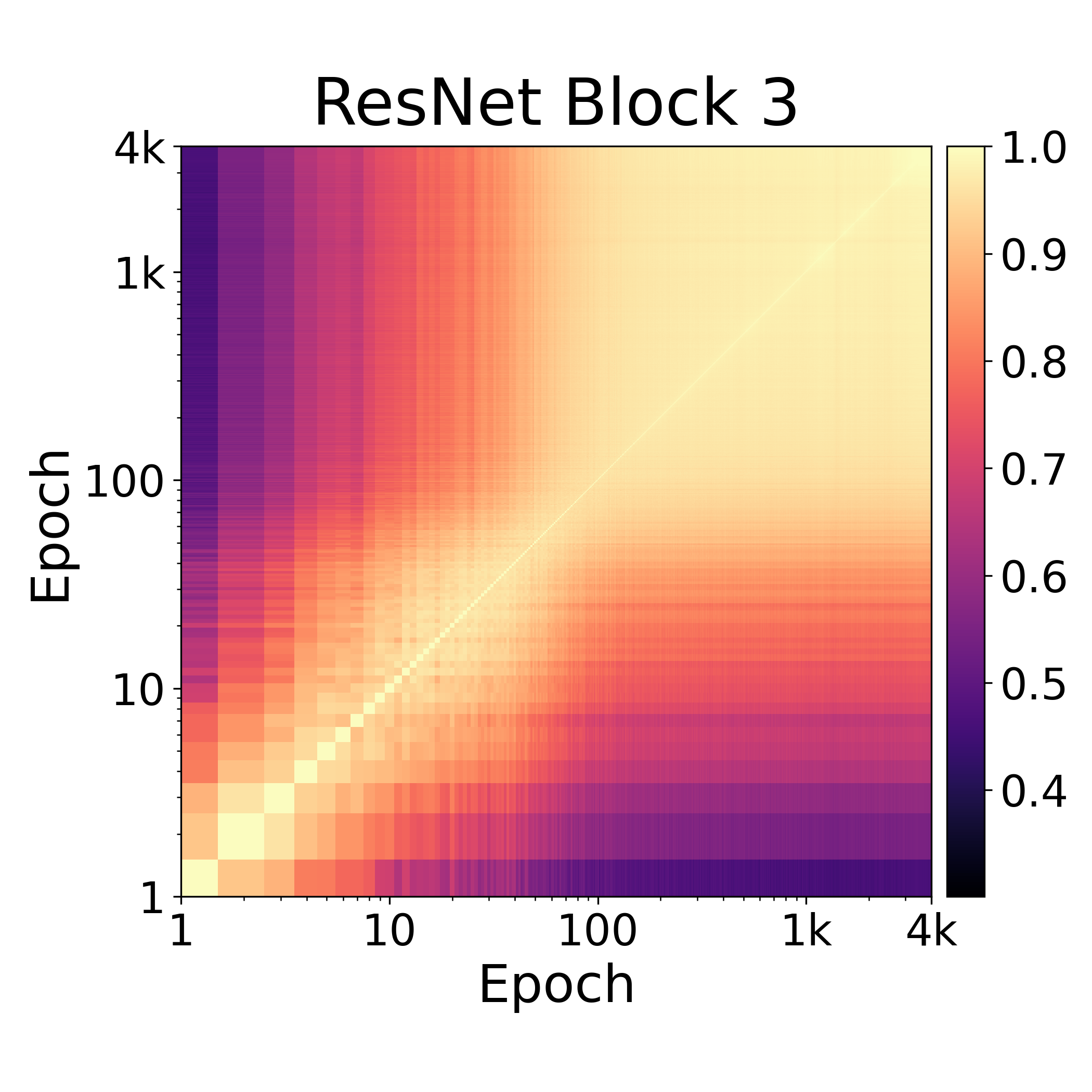}
    \label{sgd_0_01_Resnet18_k_64_block_3_noise_20}}
    \subfloat[]{
    \includegraphics[width=0.13\textwidth]{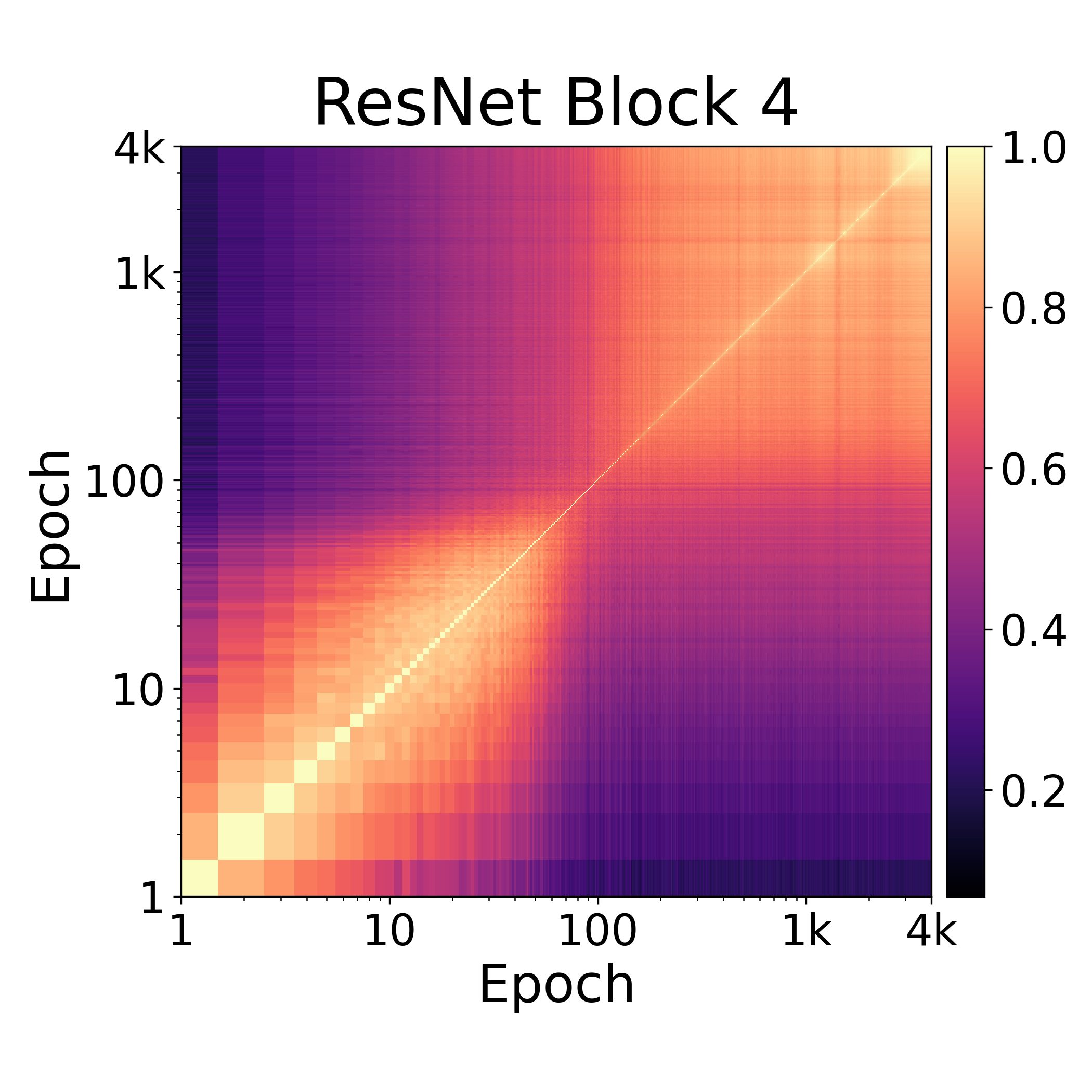}
    \label{sgd_0_01_Resnet18_k_64_block_4_noise_20}}
     \subfloat[]{
    \includegraphics[width=0.13\textwidth]{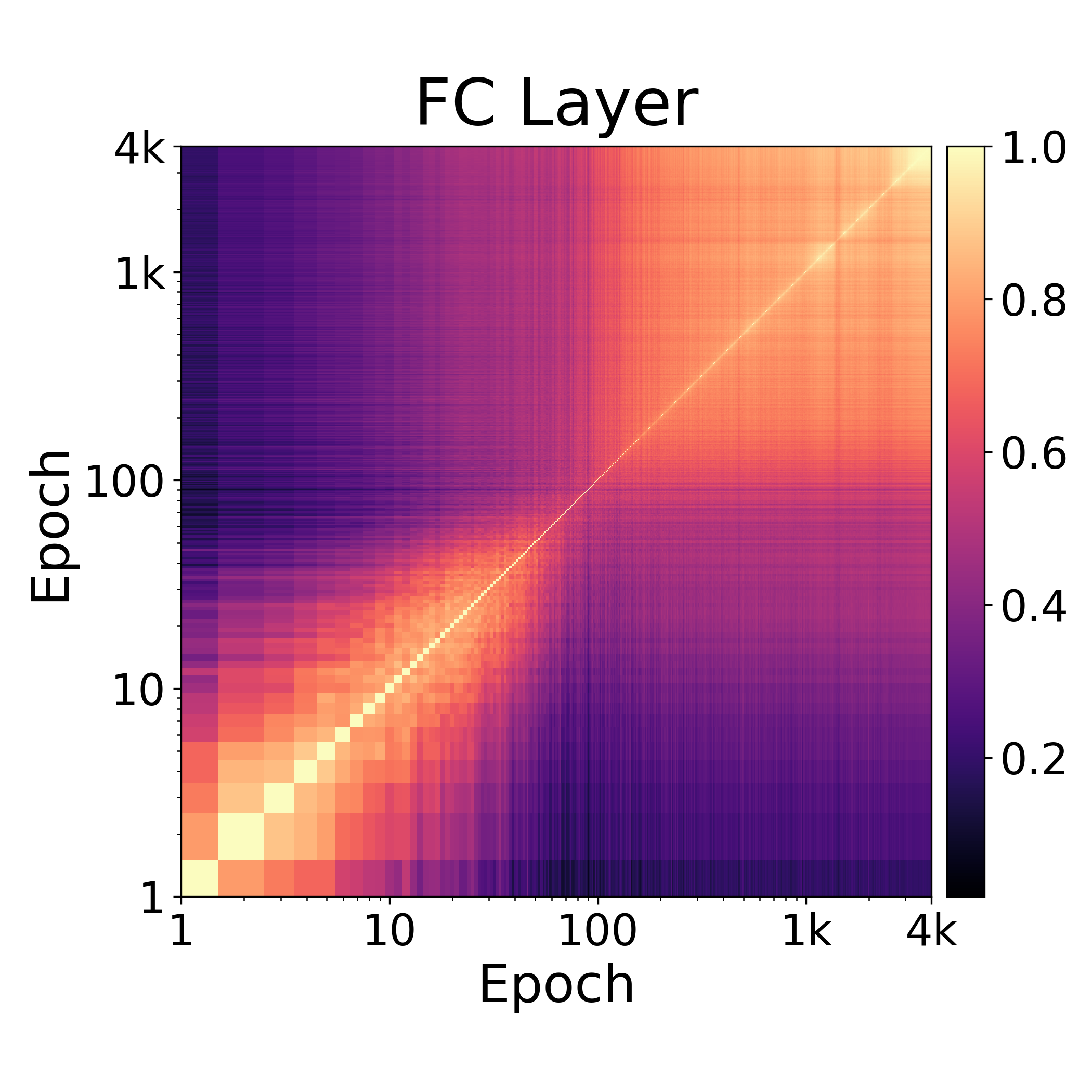}
    \label{sgd_0_01_Resnet18_k_64_FC_noise_20}}
    \subfloat[]{
    \includegraphics[width=0.13\textwidth]{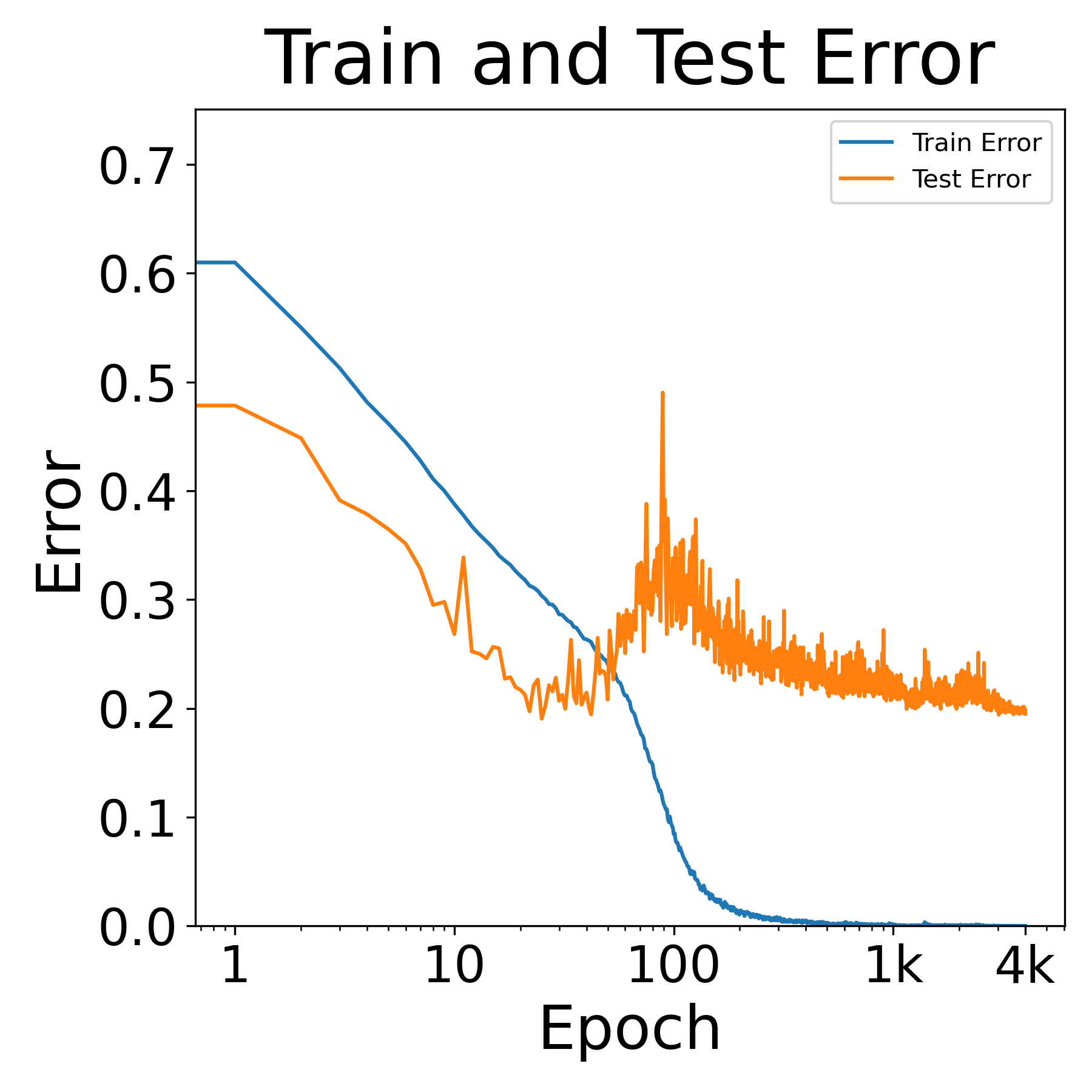}
    \label{sgd_0_01_Resnet18_k_64_error_curve_noise_20}}
  \caption{CKA evaluations for ResNet-18 trained on CIFAR-10 with 20\% label noise. Here, the optimizer is \textbf{SGD} with constant learning rate 0.1 \textbf{without momentum}.}
  \label{fig:sgd_0_01_Resnet18_k_64_noise_20}
\end{figure*}

\begin{figure*}[t]
  \centering
    \subfloat[]{
    \includegraphics[width=0.13\textwidth]{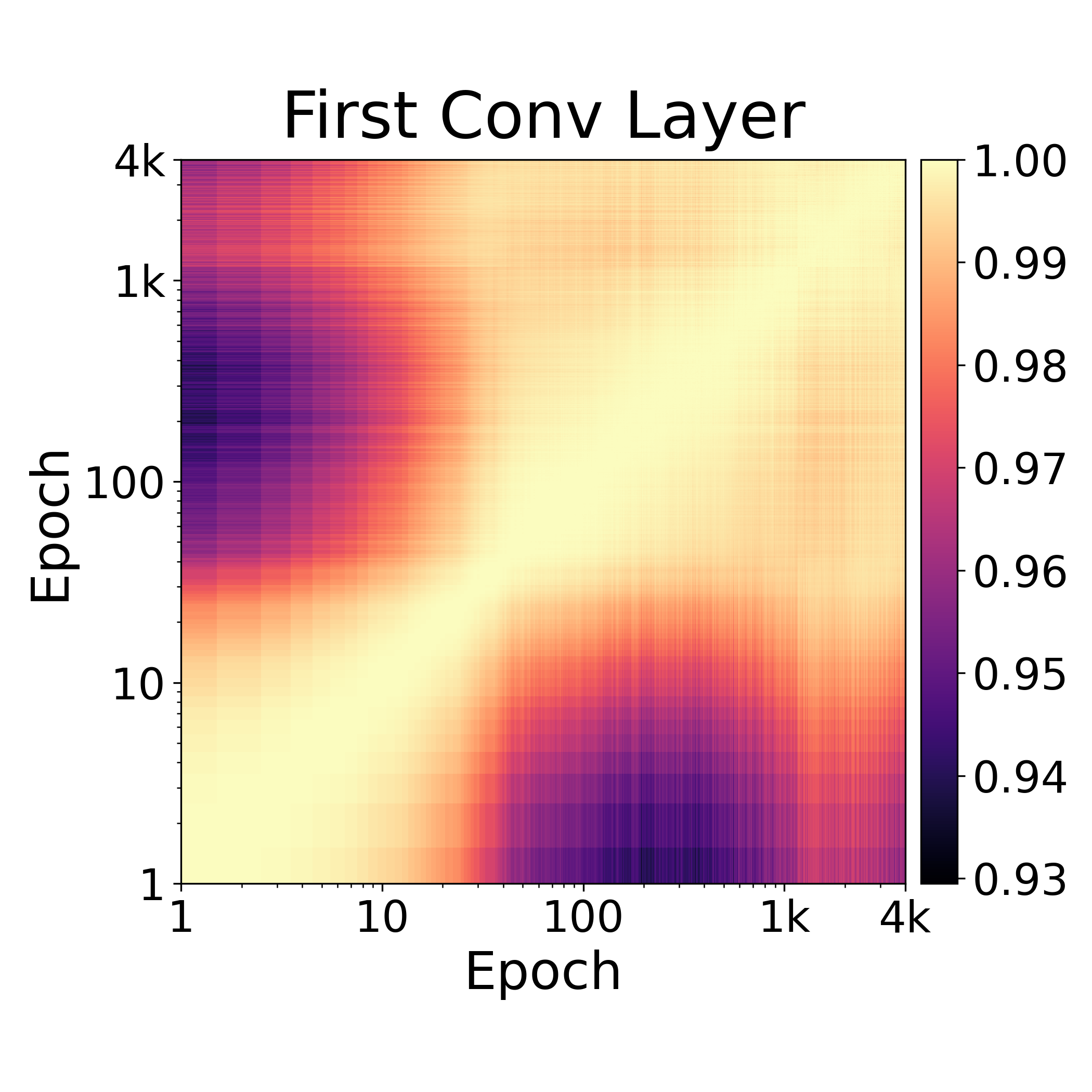}
    \label{wd_0_001_Resnet18_k_64_first_conv_layer_noise_20}}
    \subfloat[]{
    \includegraphics[width=0.13\textwidth]{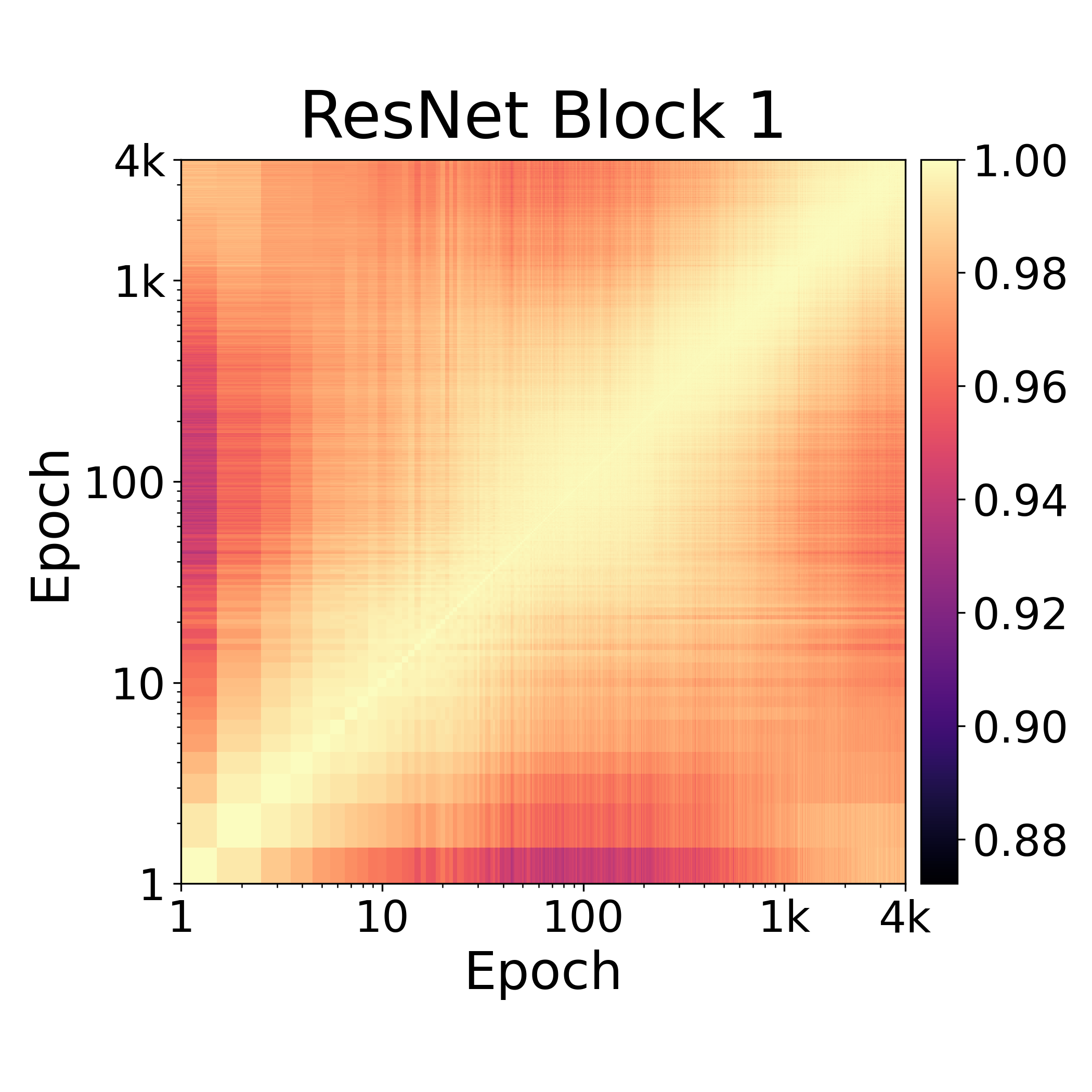}
    \label{wd_0_001_Resnet18_k_64_block_1_noise_20}
    }
    \subfloat[]{
    \includegraphics[width=0.13\textwidth]{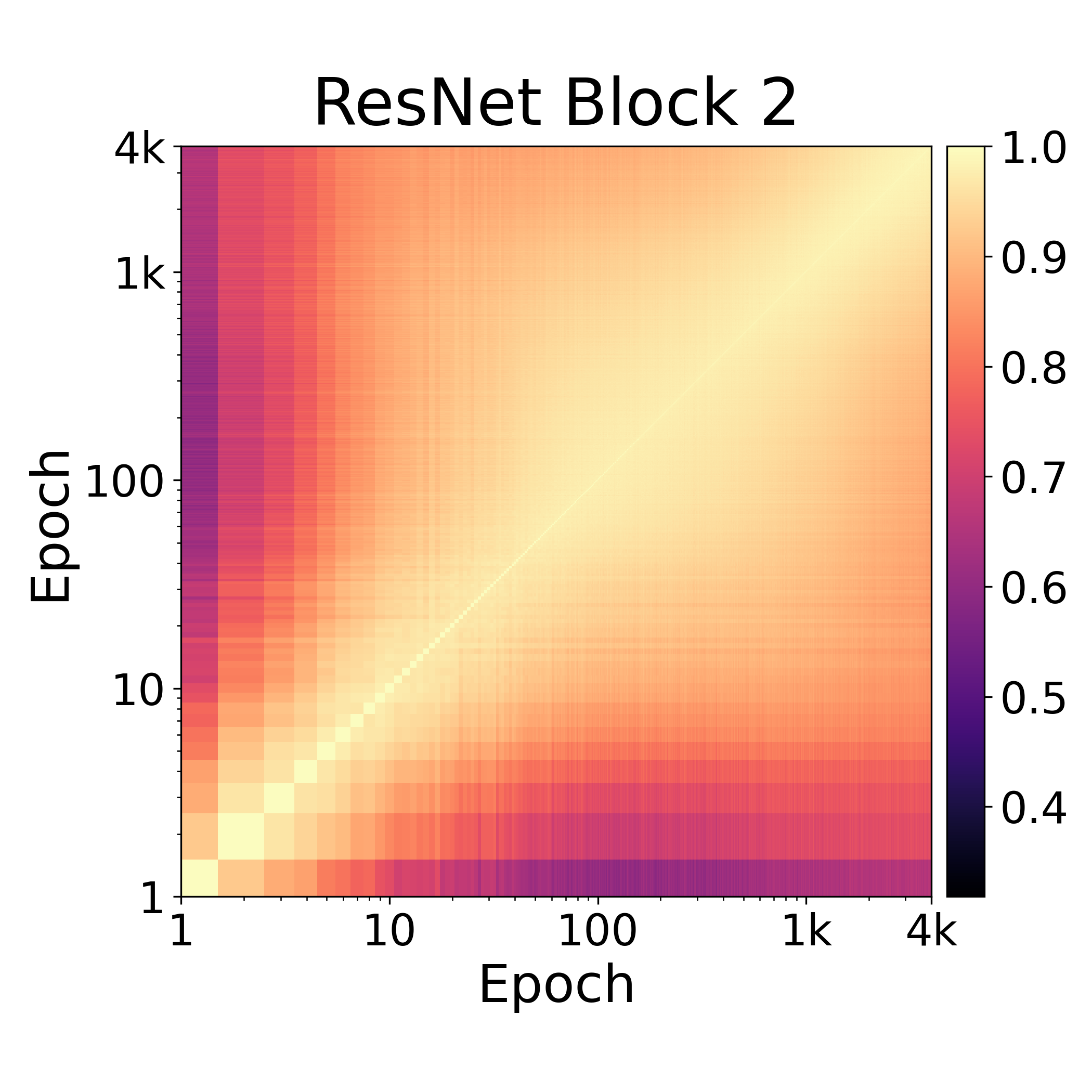}
    \label{wd_0_001_Resnet18_k_64_block_2_noise_20}}
    \subfloat[]{
    \includegraphics[width=0.13\textwidth]{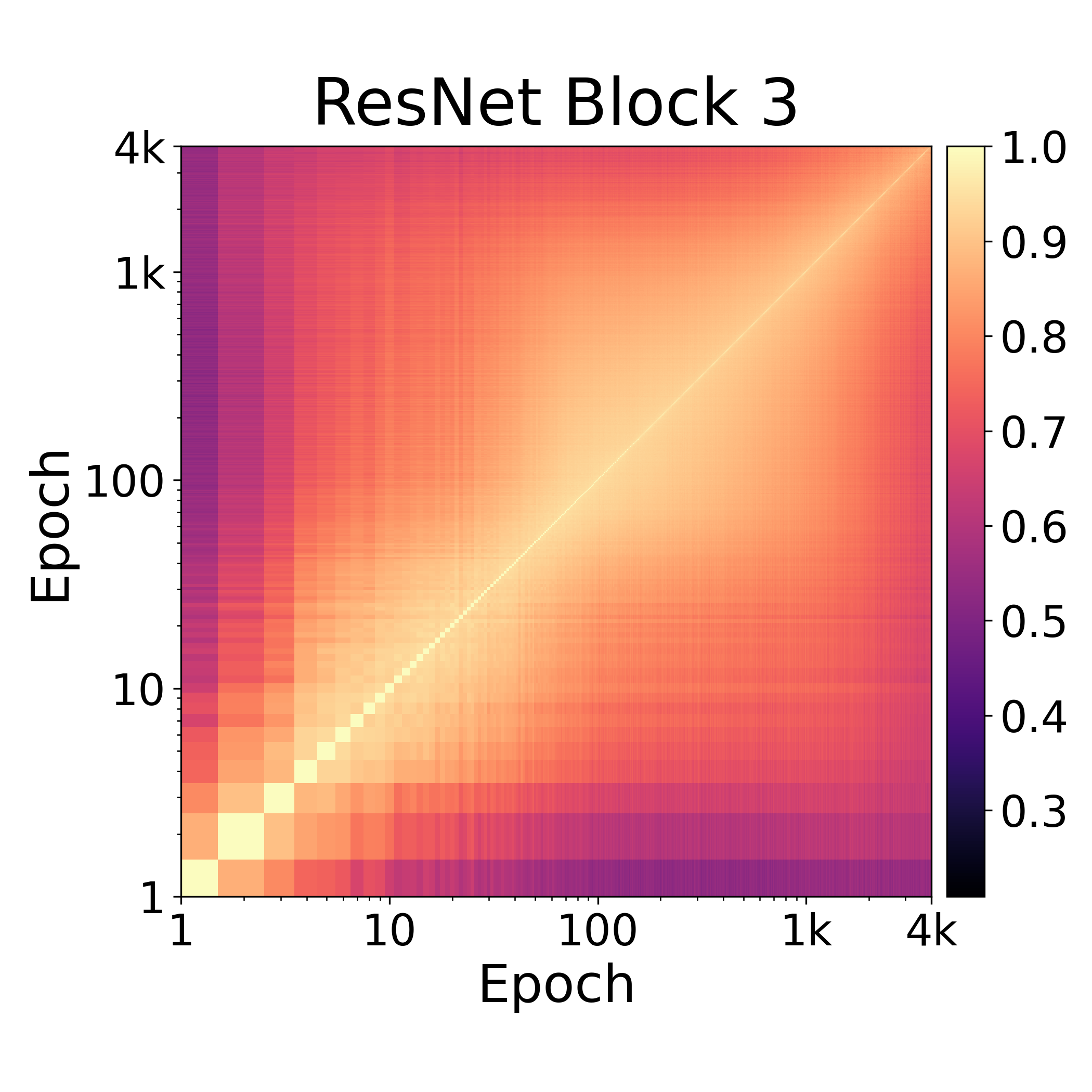}
    \label{wd_0_001_Resnet18_k_64_block_3_noise_20}}
    \subfloat[]{
    \includegraphics[width=0.13\textwidth]{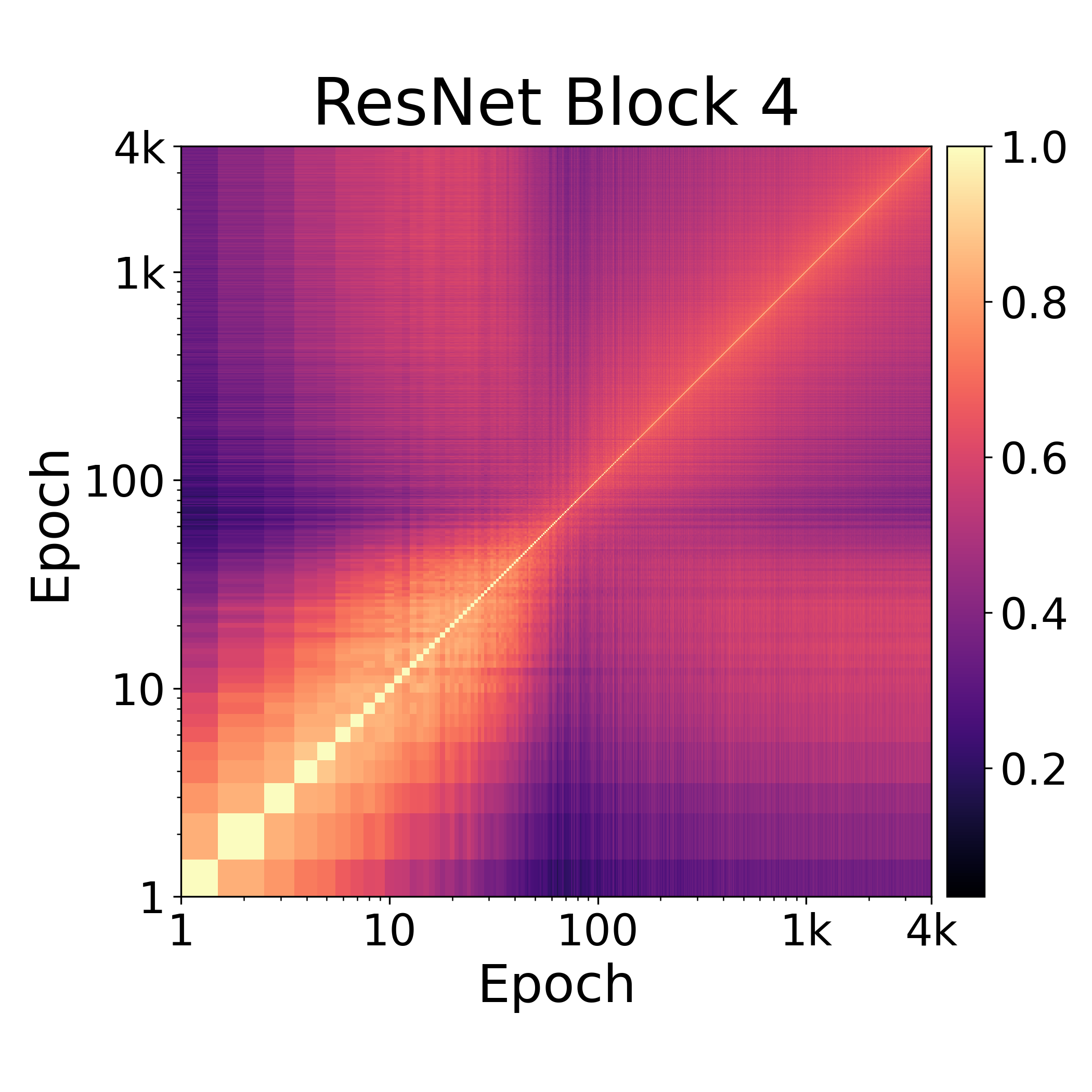}
    \label{wd_0_001_Resnet18_k_64_block_4_noise_20}}
     \subfloat[]{
    \includegraphics[width=0.13\textwidth]{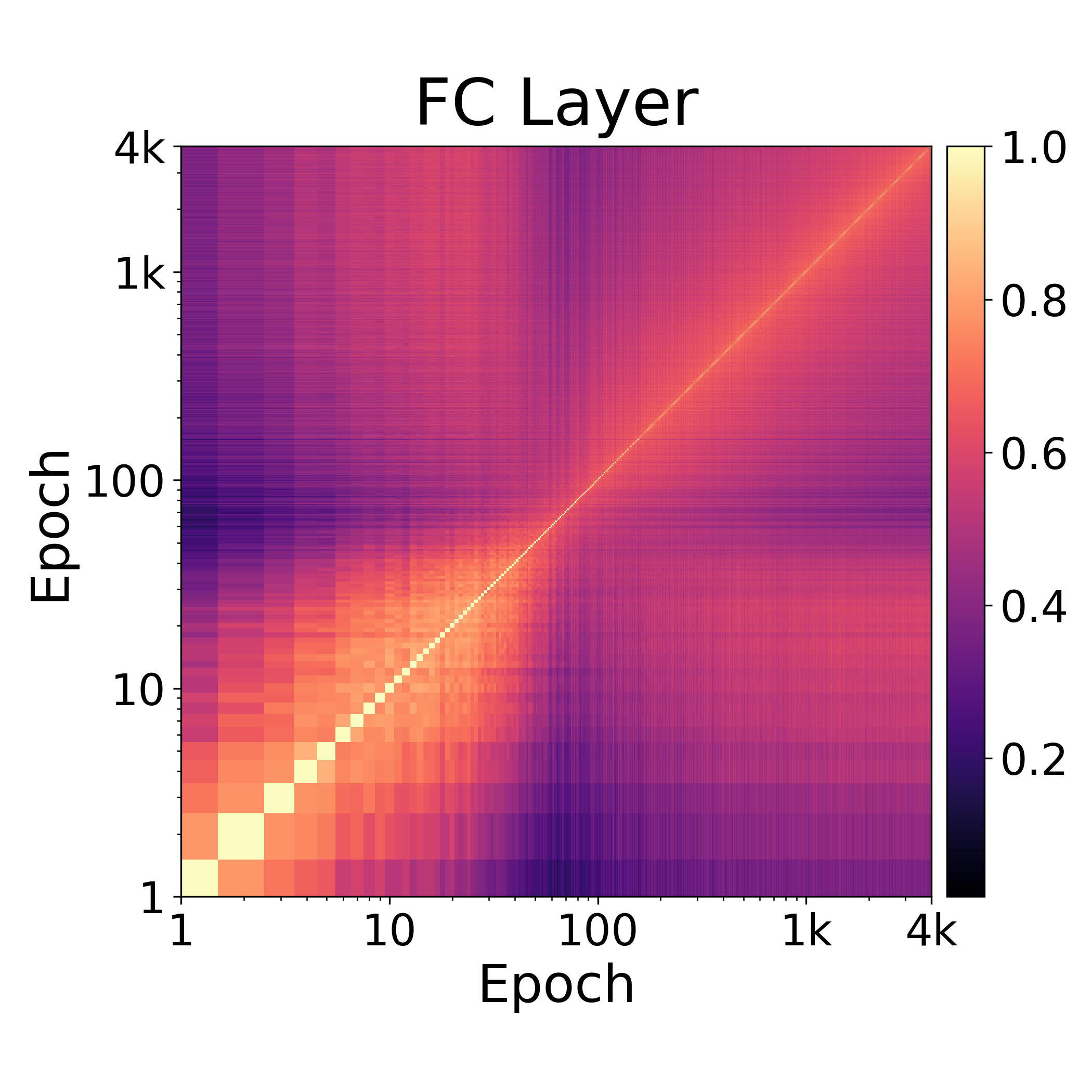}
    \label{wd_0_001_Resnet18_k_64_FC_noise_20}}
    \subfloat[]{
    \includegraphics[width=0.13\textwidth]{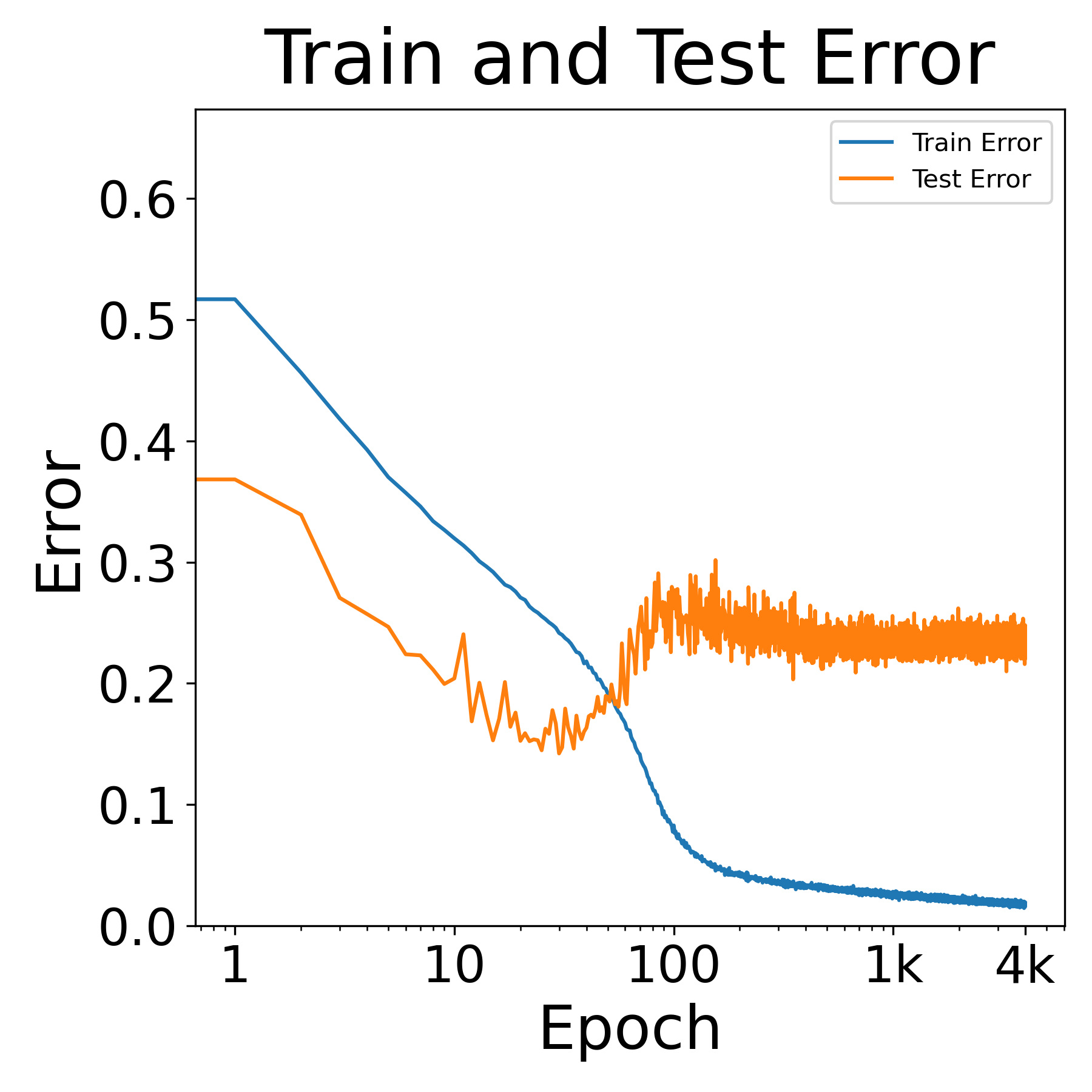}
    \label{wd_0_001_Resnet18_k_64_error_curve_noise_20}}
  \caption{CKA evaluations for ResNet-18 trained on CIFAR-10 with 20\% label noise. Here, the optimizer is \textbf{Adam with weight decay of 0.001}.}
  \label{fig:wd_0_001_Resnet18_k_64_noise_20}
\end{figure*}

\begin{figure*}[]
\centering
\subfloat[]{
\includegraphics[width=0.24\textwidth]{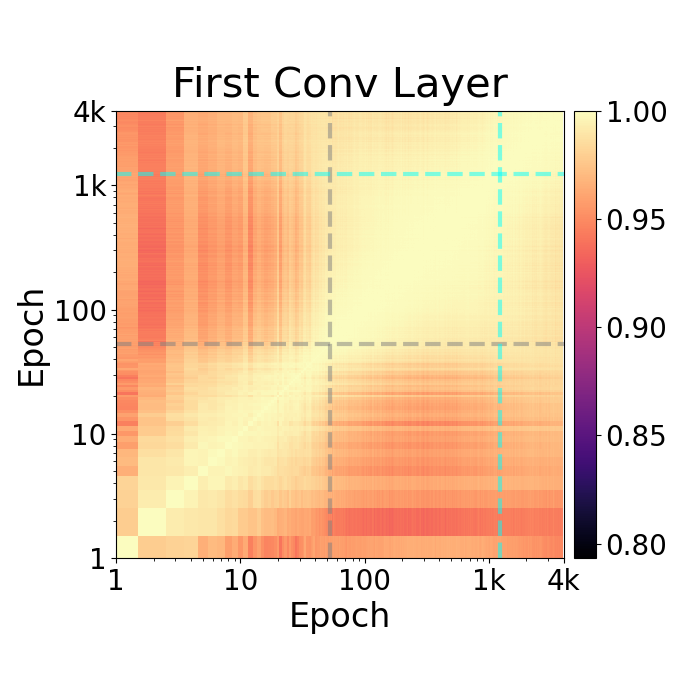}
\label{vit_first_conv_layer}}
\subfloat[]{
\includegraphics[width=0.24\textwidth]{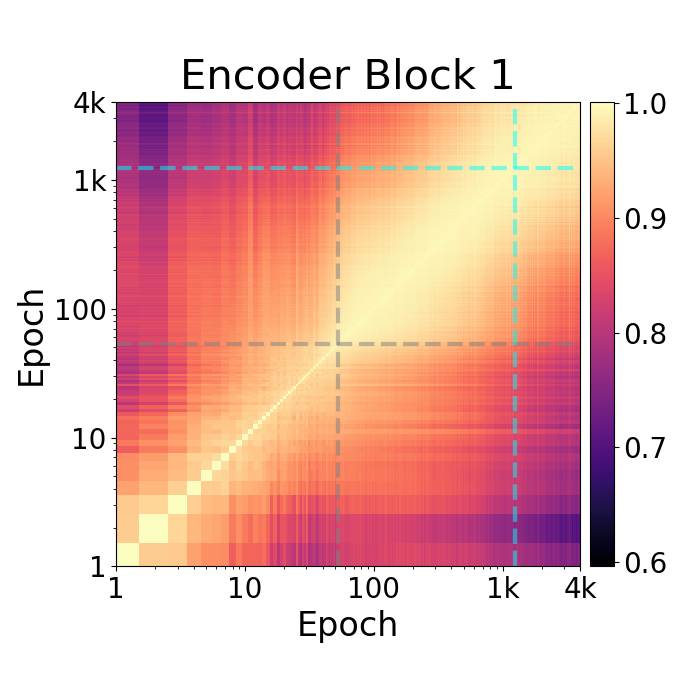}
\label{Encoder_block_1}}
\subfloat[]{
\includegraphics[width=0.24\textwidth]{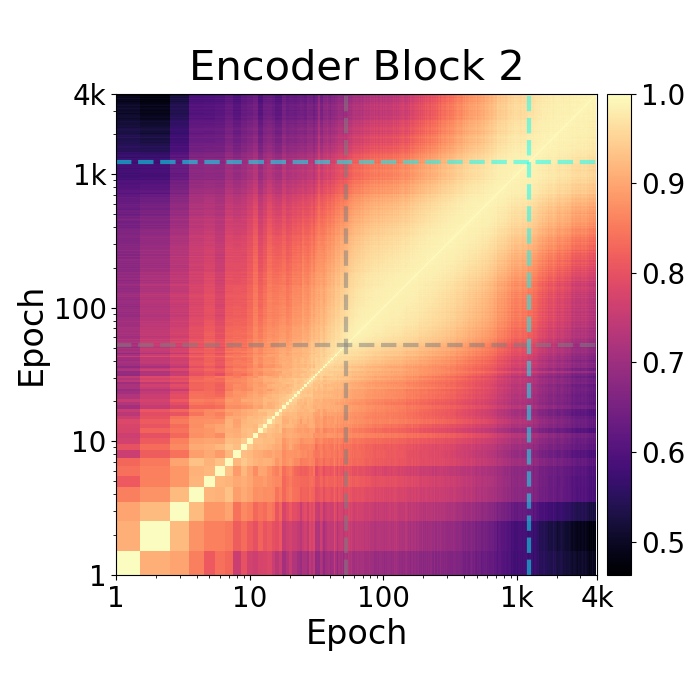}
\label{Encoder_block_2}}
\subfloat[]{
\includegraphics[width=0.24\textwidth]{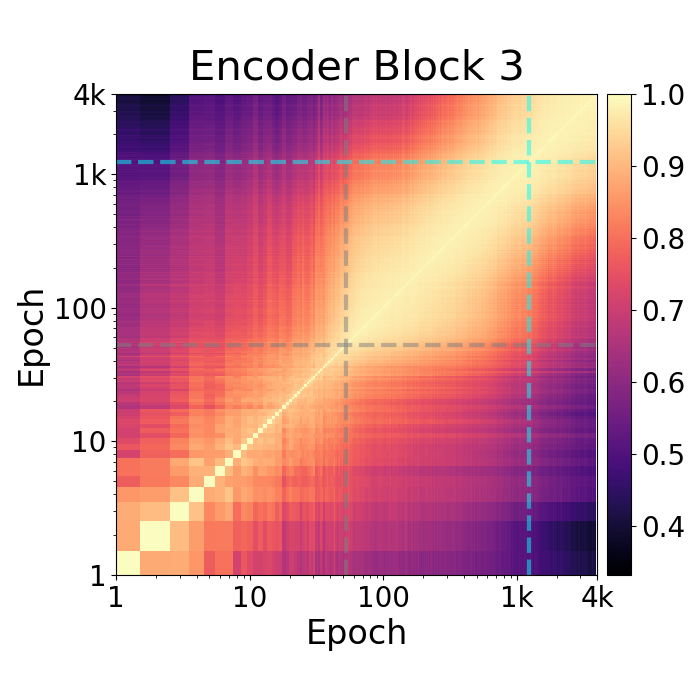}
\label{Encoder_block_3}}
\\[-3ex]
\subfloat[]{
\includegraphics[width=0.24\textwidth]{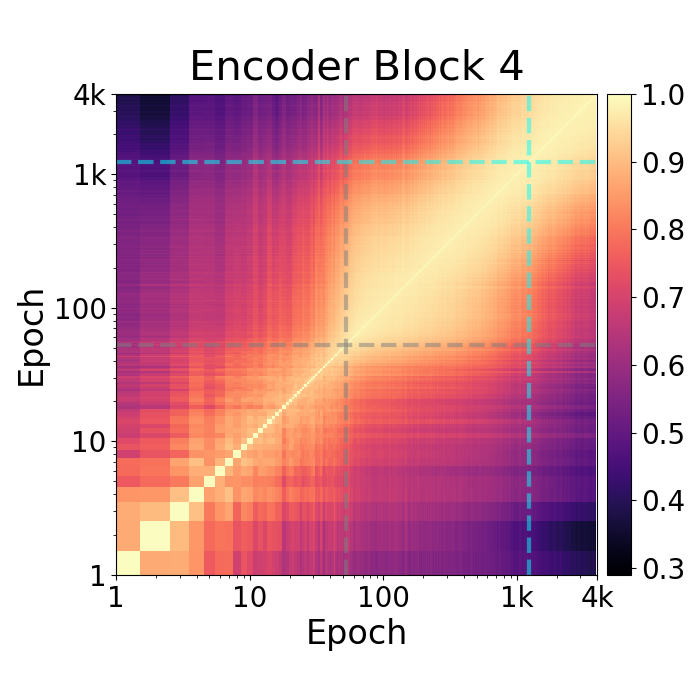}
\label{Encoder_block_4}}
\subfloat[]{
\includegraphics[width=0.24\textwidth]{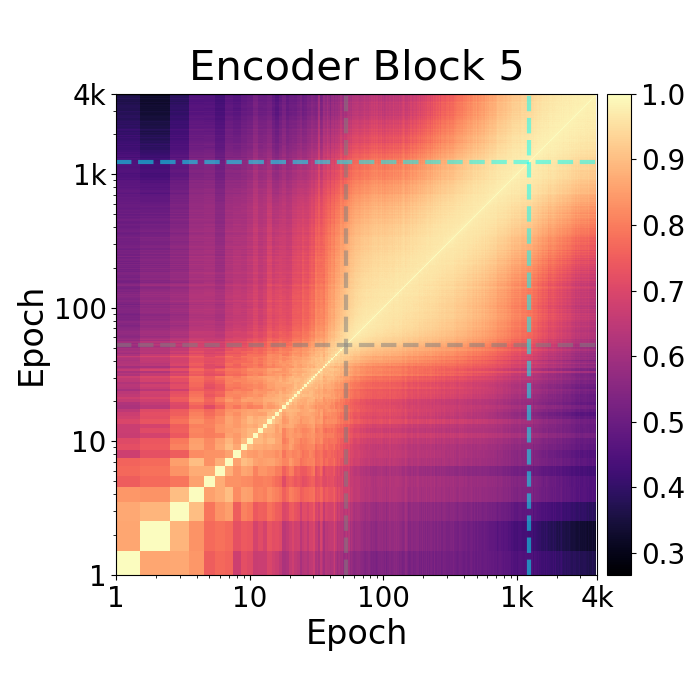}
\label{Encoder_block_5}}
\subfloat[]{
\includegraphics[width=0.24\textwidth]{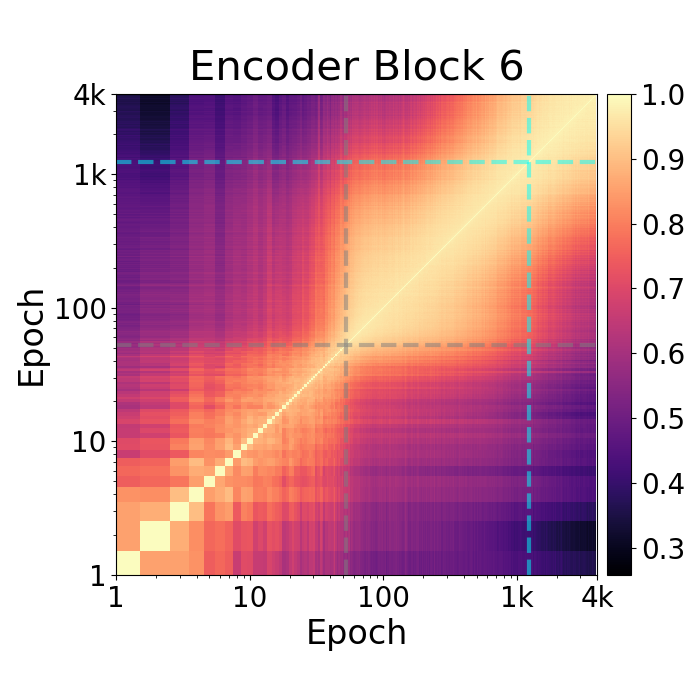}
\label{Encoder_block_6}}
\subfloat[]{
\includegraphics[width=0.24\textwidth]{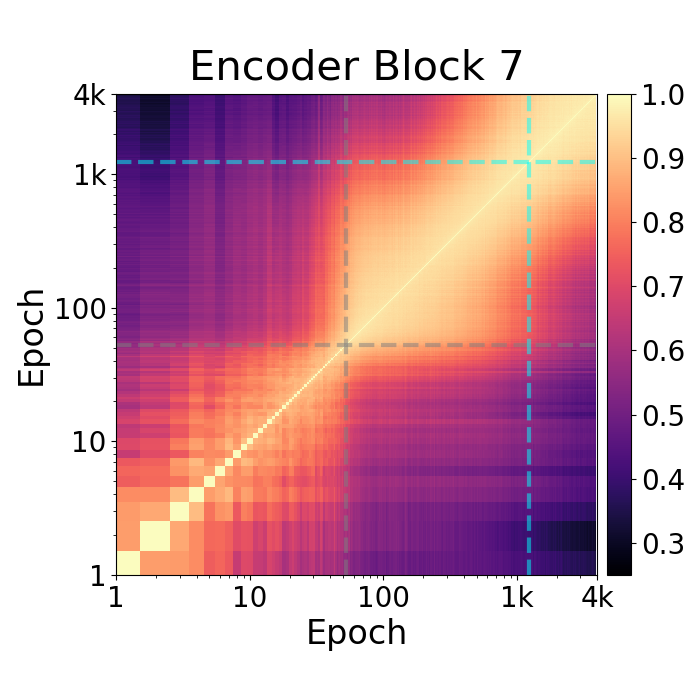}
\label{Encoder_block_7}}
\\[-3ex]
\subfloat[]{
\includegraphics[width=0.24\textwidth]{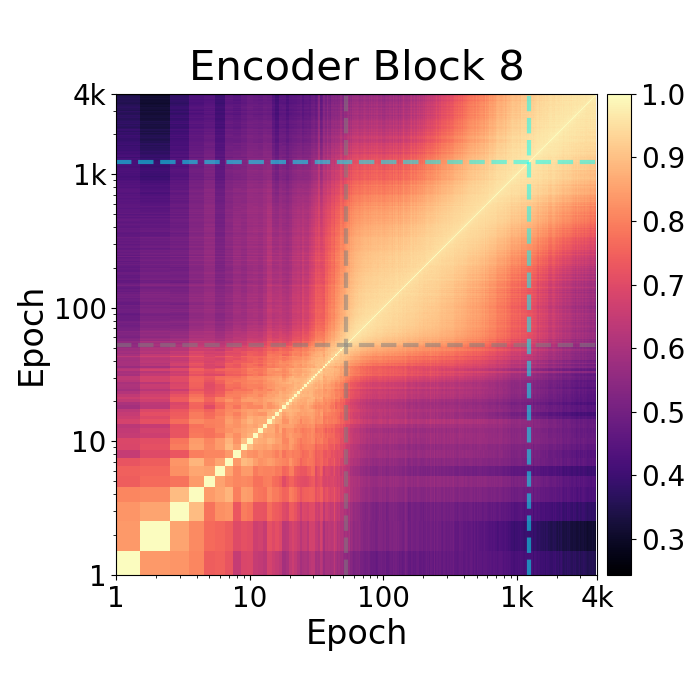}
\label{Encoder_block_8}}
\subfloat[]{
\includegraphics[width=0.24\textwidth]{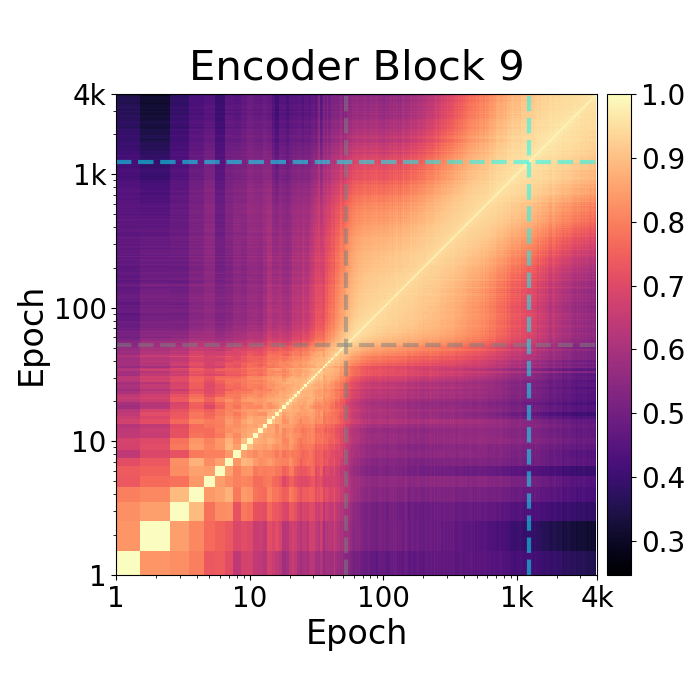}
\label{Encoder_block_9}}
\subfloat[]{
\includegraphics[width=0.24\textwidth]{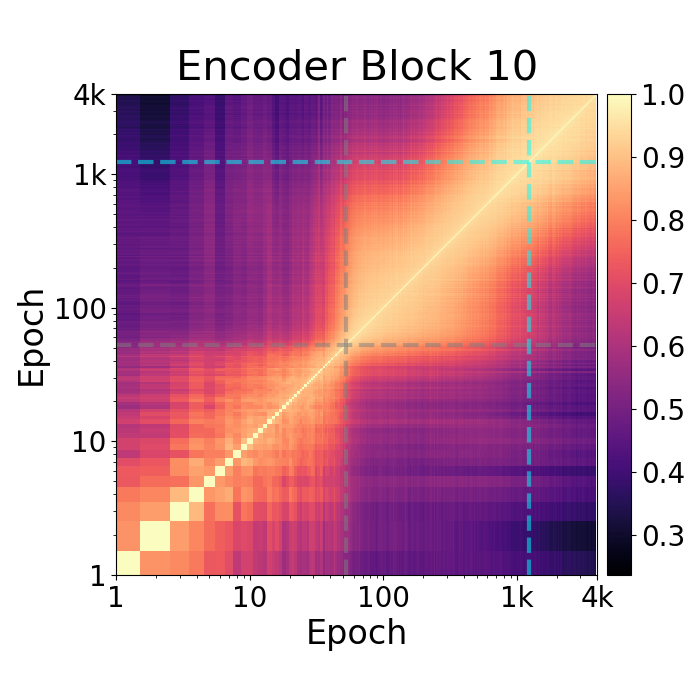}
\label{Encoder_block_10}}
\subfloat[]{
\includegraphics[width=0.24\textwidth]{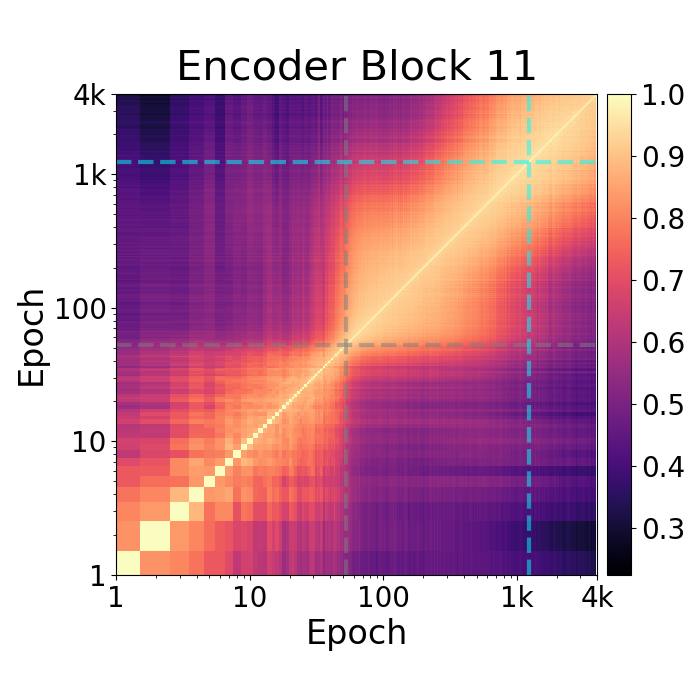}
\label{Encoder_block_11}}
\\[-3ex]
\subfloat[]{
\includegraphics[width=0.24\textwidth]{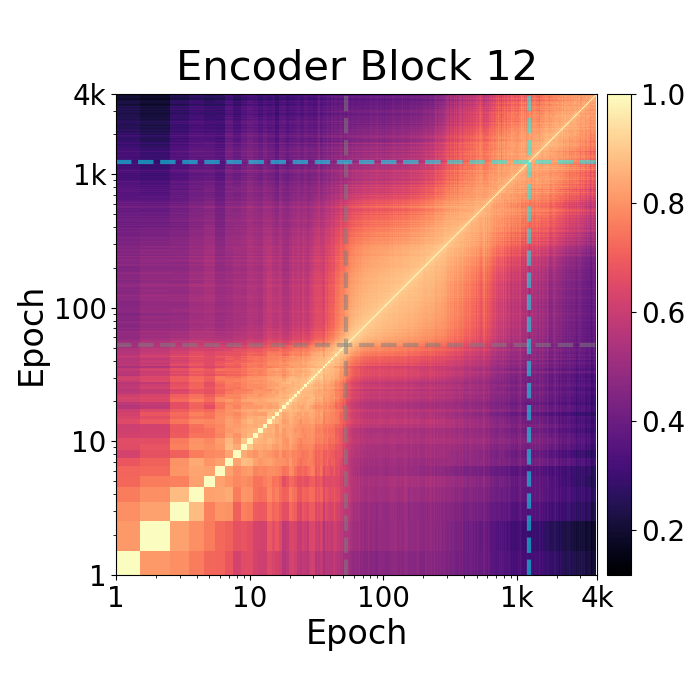}
\label{Encoder_block_12}}
\subfloat[]{
\includegraphics[width=0.23\textwidth]{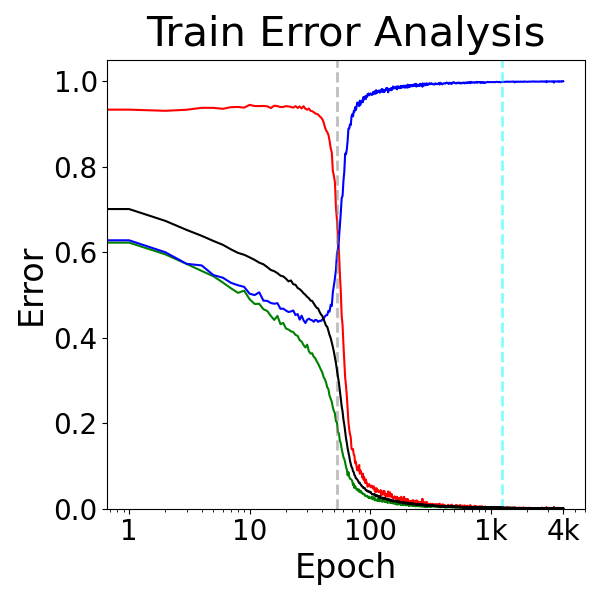}
\label{vit_with_noise_err_breakdown}}
\subfloat[]{
\includegraphics[width=0.28\textwidth]{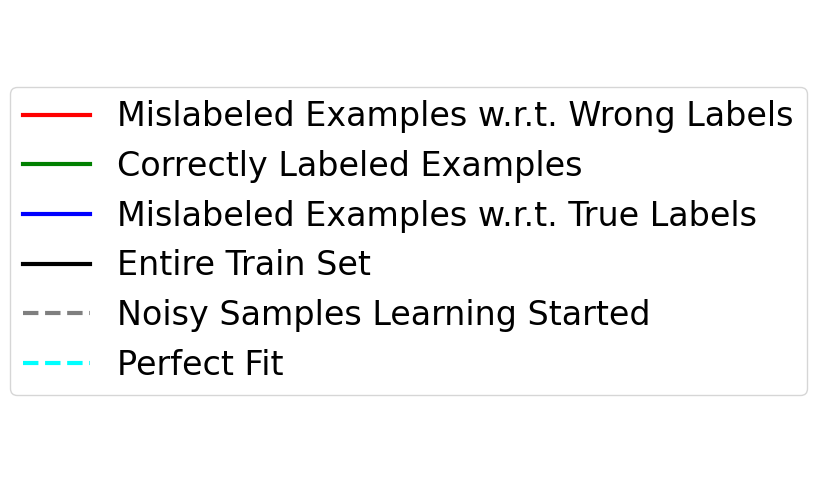}
\label{vit_with_noise_err_breakdown_legends}}
\subfloat[]{
\includegraphics[width=0.23\textwidth]{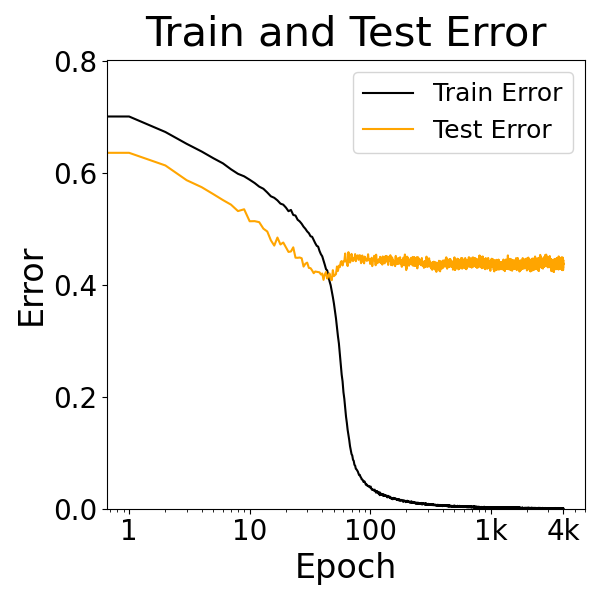}
\label{vit_train_test_error}}
\caption{CKA evaluations for ViT-B/16, trained on CIFAR-10 with 20\% label noise.}
\label{fig:cka_comparison_vit_b_16_noise_20}
\end{figure*}

\subsection{CKA Similarity of the First Layer Representations to Random Representations}
\label{appendix:subsec:CKA Similarity of the First Layer Representations to Random Representations}

\begin{figure*}[t]
    \centering
        \subfloat[]{
        \includegraphics[width=0.24\textwidth]{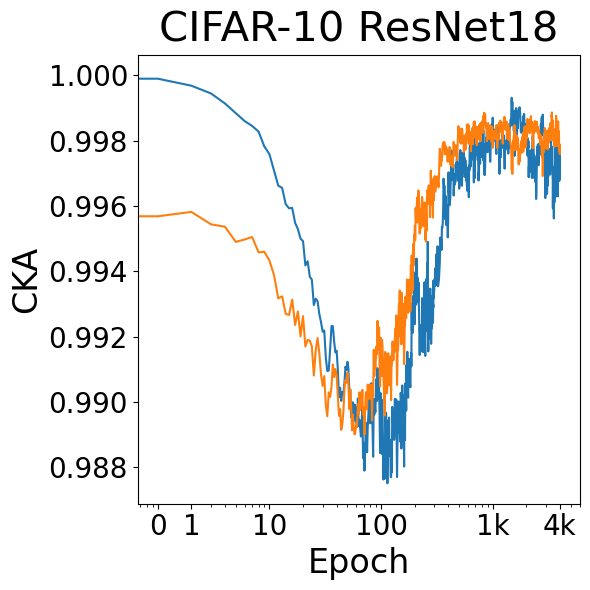}
        \label{diff_init_plot_conv1_resnet18_cifar10}}
        \subfloat[]{
        \includegraphics[width=0.24\textwidth]{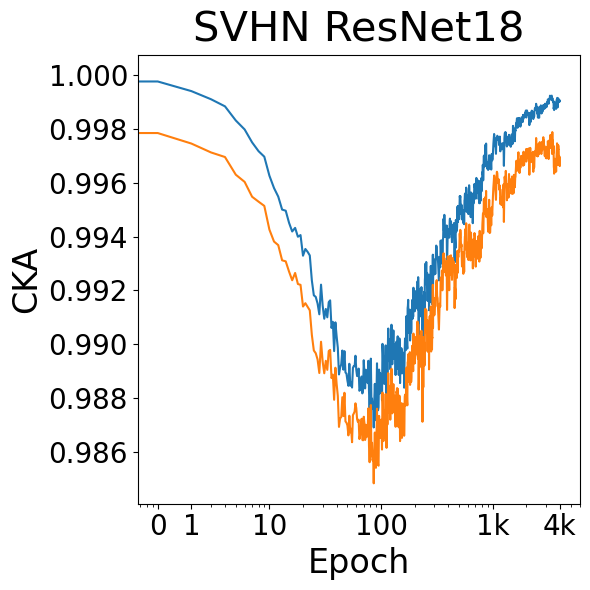}
        \label{diff_init_plot_conv1_resnet18_svhn}}
        \subfloat[]{
        \includegraphics[width=0.4\textwidth]{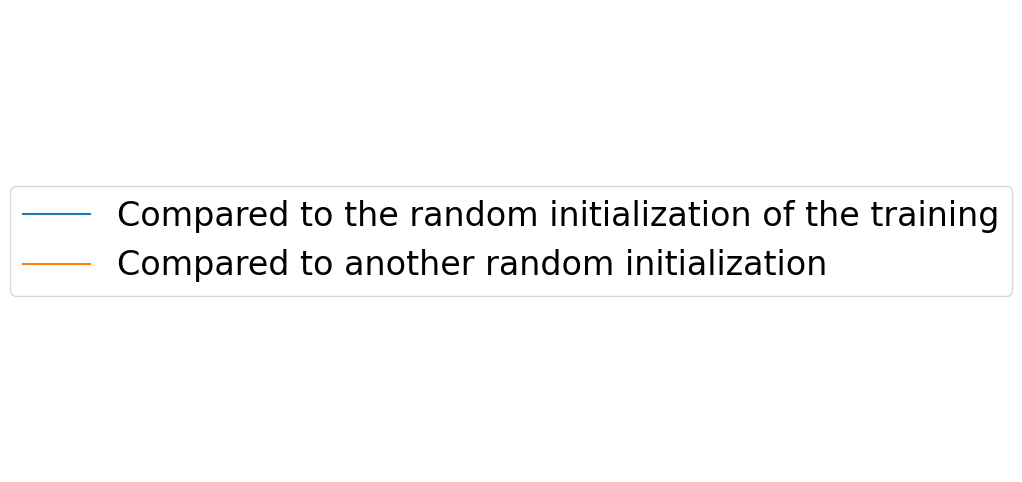}
        \label{diff_init_plot_legends}}
  \caption{
First row from the CKA diagram for the first convolutional layer (the blue line in (a) is from Fig.~\ref{Resnet18_k_64_first_conv_layer_noiseless}; the blue line in (b) is from Fig.~\ref{svhn_Resnet18_k_64_adam_first_conv_layer_noiseless}), together with the CKA similarity with respect to another random initialization (orange line). The model is ResNet-18 trained on (a) CIFAR-10, (b) SVHN.}
  \label{fig:resnet18_conv1_similarity_with_diff_init}
\end{figure*}

As mentioned in Section \ref{subsec:Analysis of the CKA Results} of the main text, the similarity of the first layer representations to the random representations (as appear, e.g., in Figs.~\ref{cka_comparison_resnet18_k_64_noise_20__first_conv_layer}, \ref{Resnet18_k_64_first_conv_layer_noiseless}) is not only to the \textit{specific} random initialization of the training process, but also to other random representations unrelated to the specific training process. 
This can be observed in Fig.~\ref{fig:resnet18_conv1_similarity_with_diff_init} where we evaluate the CKA similarity of the first layer representations during training with respect to a random representation that was not used as the initialization of the specific training process; yet, it should be noted that the CKA similarity to the specific initialization is likely to be marginally higher than the CKA similarity to another initialization. 

\subsection{The Effect of DNN Width on the First Layer Representations}
\label{appendix:subsec:The Effect of DNN Width and Weight Decay on the First Layer Representations}

Importantly, the representations of the first layer have increased similarity to the random initialization at the perfect fitting regime only \textbf{if the DNN is wide enough}. For example, see Figs.~\ref{Resnet18_k_32_first_conv_layer}, \ref{Resnet18_k_42_first_conv_layer} where we examine ResNet-18 of smaller widths (width parameters $k=32,42$) that still achieve perfect fitting (and show epoch-wise double descent) but do not have increased similarity to initialization at the end of training. When the DNN is wider (width parameter $k=52$), in Fig.~\ref{Resnet18_k_52_first_conv_layer}, the behavior is more similar to our results in Fig.~\ref{cka_comparison_resnet18_k_64_noise_20__first_conv_layer} for the standard width of ResNet-18 (width parameter $k=64$).

Figs.~\ref{fig:first_layer_grad_adam_k_42} show that the componentwise gradient magnitudes for width parameters $k=42$ training are sufficiently large compared to the standard value of the numerical stability parameter $\epsilon$, implying that Adam's adaptive normalization should apply its effect on the learning process. Yet, we show that effectively disabling Adam's adaptive normalization by increasing the numerical stability parameter $\epsilon$ disables the first layer phenomenon -- implying that the adaptive normalization of gradients is a necessary, but not sufficient, condition for the phenomenon. 

\subsection{Additional CKA Evaluations}
In Figs.~\ref{fig:Resnet18_k_64_noise_10}-\ref{fig:cka_comparison_vit_b_16_noise_20} we provide CKA evaluations for various learning settings, in addition to the results in the main paper. See Table \ref{table:cka_experiment_summary} for the summary of all training settings.

\newpage
\begin{figure*}[t]
  \centering
    \subfloat[]{
    \includegraphics[width=0.13\textwidth]{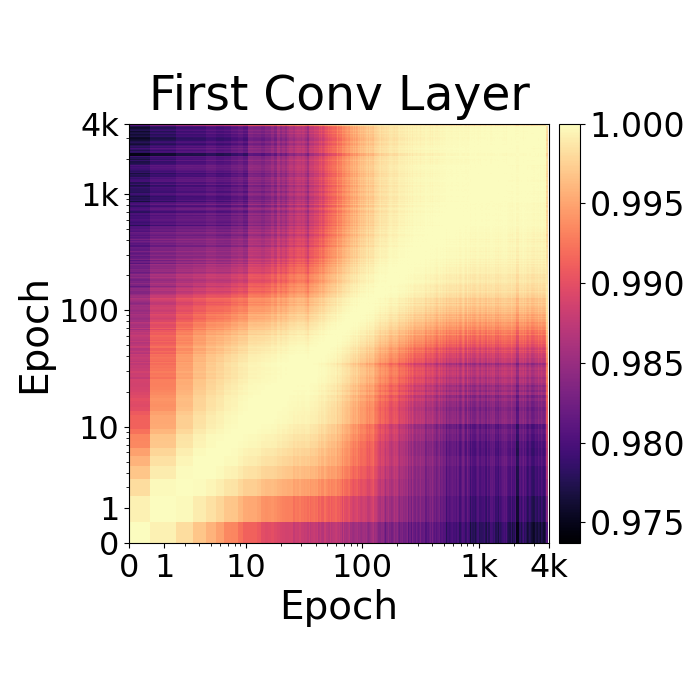}
    \label{Resnet18_k_64_first_conv_layer_tiny_imagenet_noise_20}}
    \subfloat[]{
    \includegraphics[width=0.13\textwidth]{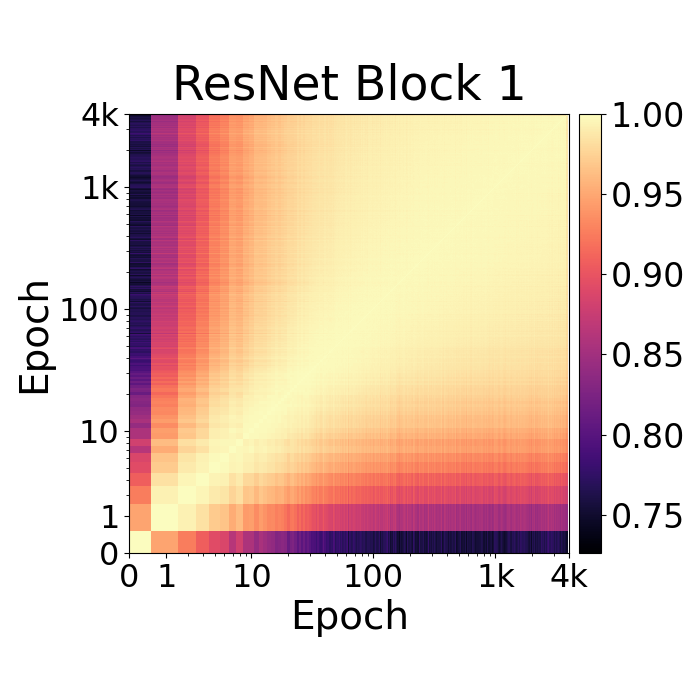}
    \label{Resnet18_k_64_block_1_tiny_imagenet_noise_20}
    }
    \subfloat[]{
    \includegraphics[width=0.13\textwidth]{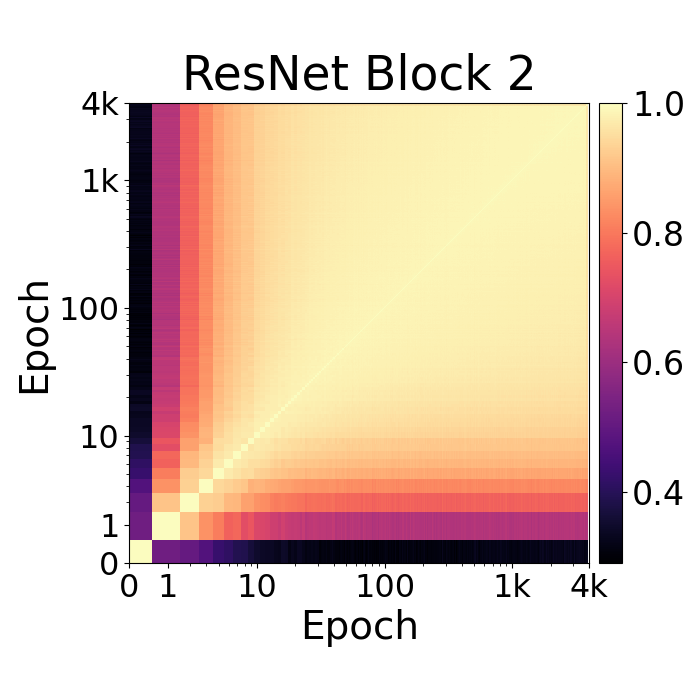}
    \label{Resnet18_k_64_block_2_tiny_imagenet_noise_20}}
    \subfloat[]{
    \includegraphics[width=0.13\textwidth]{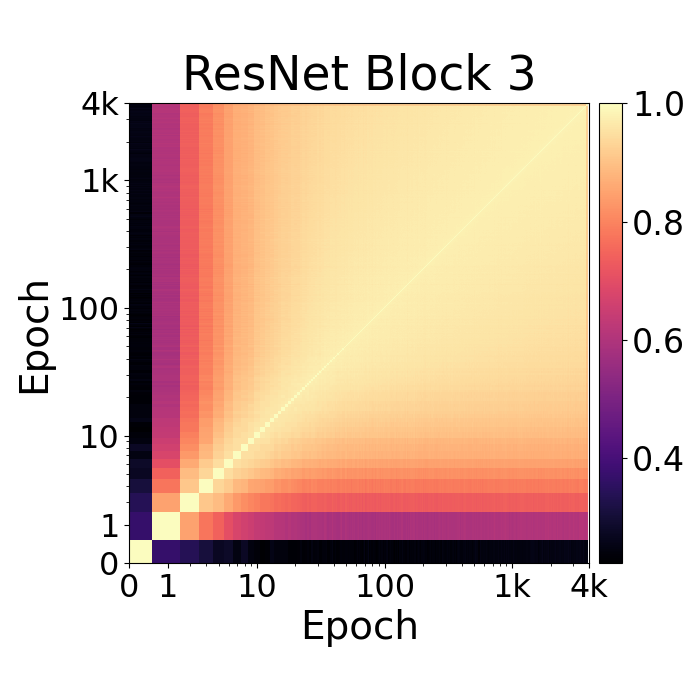}
    \label{Resnet18_k_64_block_3_tiny_imagenet_noise_20}}
    \subfloat[]{
    \includegraphics[width=0.13\textwidth]{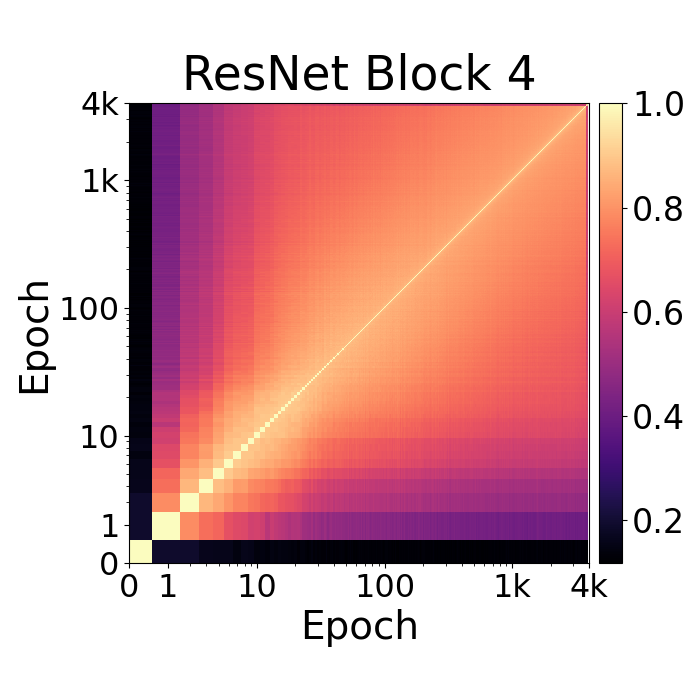}
    \label{Resnet18_k_64_block_4_tiny_imagenet_noise_20}}
     \subfloat[]{
    \includegraphics[width=0.13\textwidth]{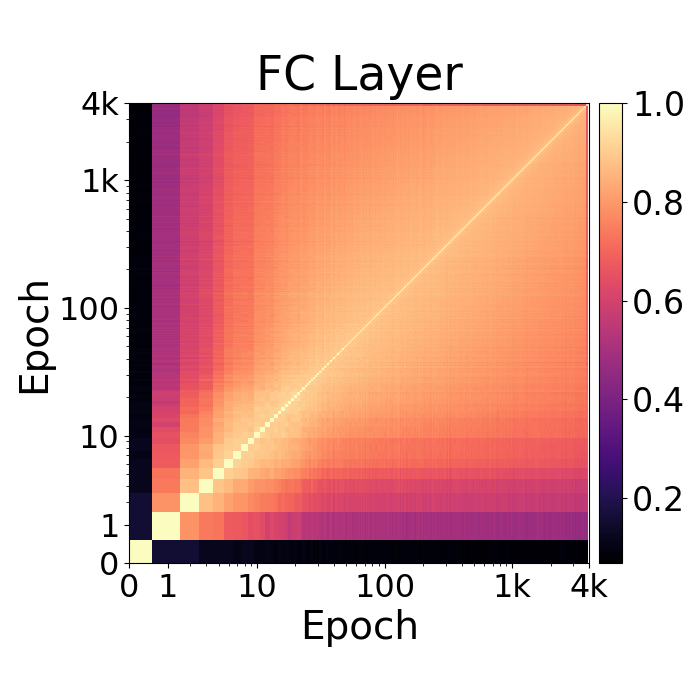}
    \label{Resnet18_k_64_FC_tiny_imagenet_noise_20}}
    \subfloat[]{
    \includegraphics[width=0.13\textwidth]{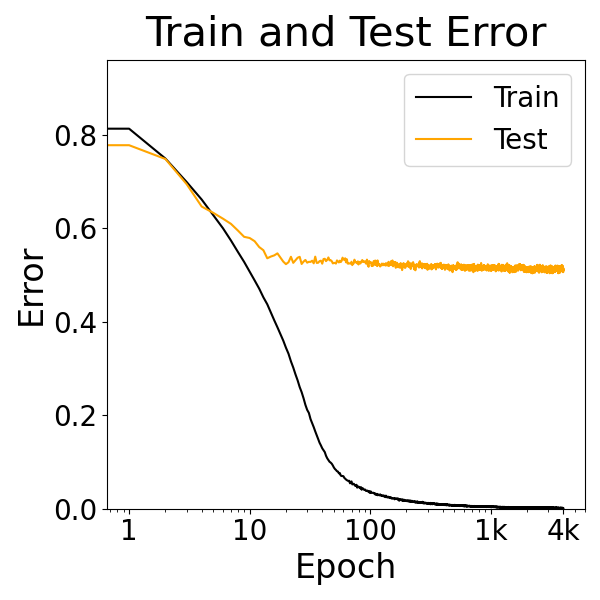}
    \label{Resnet18_k_64_error_curve_tiny_imagenet_noise_20}}
  \caption{CKA evaluations for ResNet-18 trained on Tiny ImageNet with 20\% label noise. Each of the (a)-(f) subfigures shows the CKA representational similarity of a specific layer in the ResNet-18 during its training. ResNet Block $j$ refers to the representation at the end of the $j^{\sf th}$ block of layers in the ResNet-18 architecture. (g) shows the test and train errors during training.}
  \label{fig:Resnet18_k_64_tiny_imagenet_noise_20}
\end{figure*}

\begin{figure*}[t]
  \centering
    \subfloat[]{
    \includegraphics[width=0.13\textwidth]{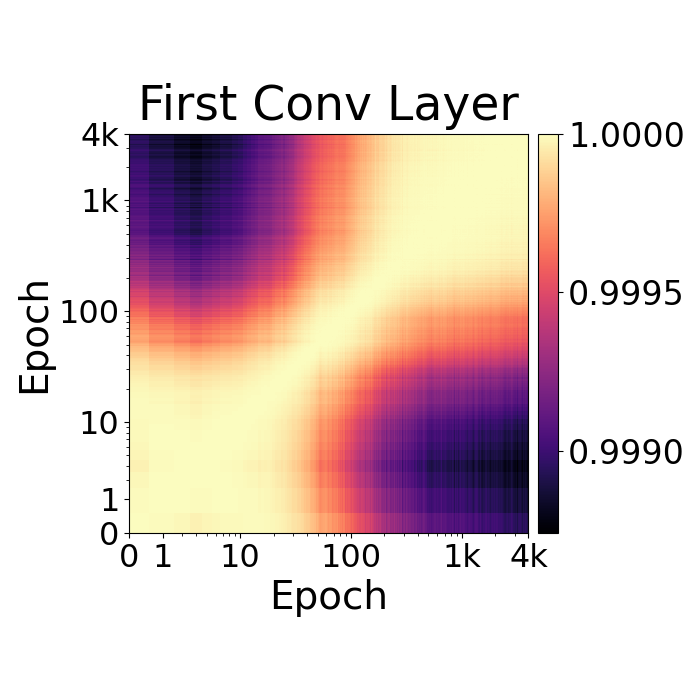}
    \label{cifar10_resnet18_noise_0_k_64_sgd_lr_0.001_momentum_0_bs_128_first_conv_layer_noiseless}}
    \subfloat[]{
    \includegraphics[width=0.13\textwidth]{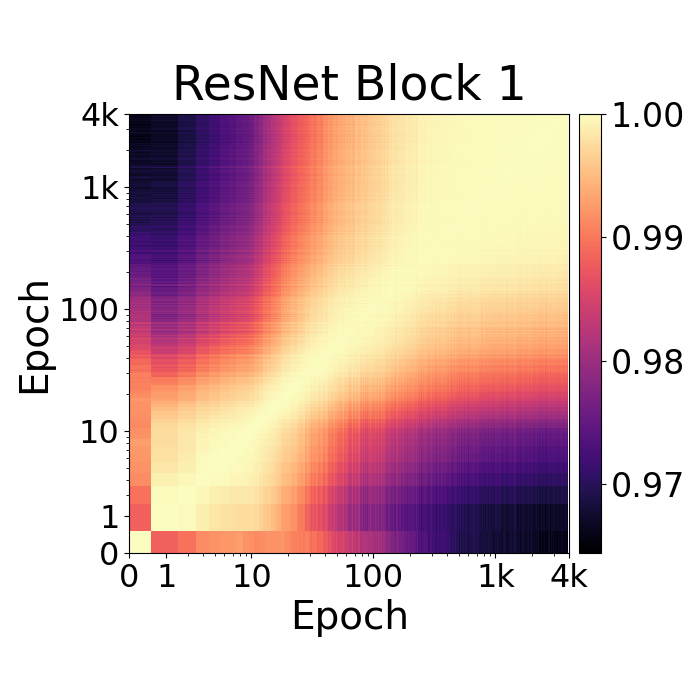}
    \label{cifar10_resnet18_noise_0_k_64_sgd_lr_0.001_momentum_0_bs_128_block_1_noiseless}
    }
    \subfloat[]{
    \includegraphics[width=0.13\textwidth]{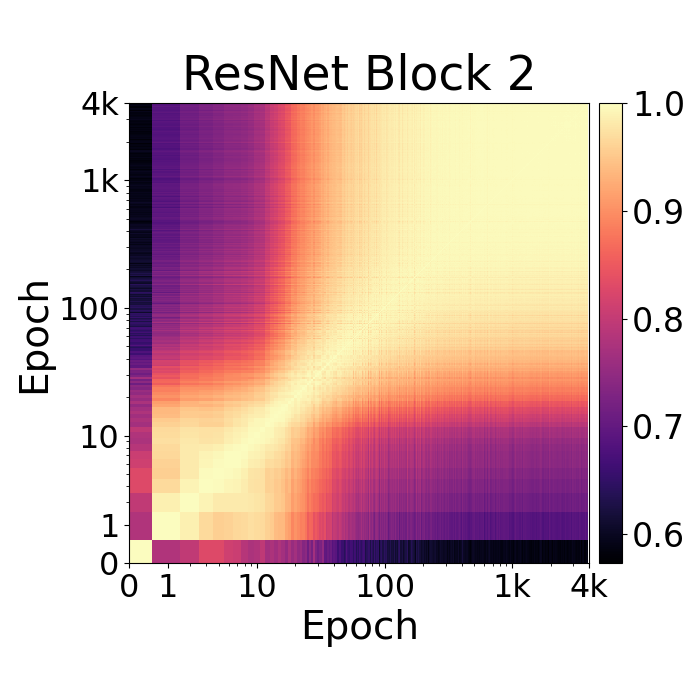}
    \label{cifar10_resnet18_noise_0_k_64_sgd_lr_0.001_momentum_0_bs_128_block_2_noiseless}}
    \subfloat[]{
    \includegraphics[width=0.13\textwidth]{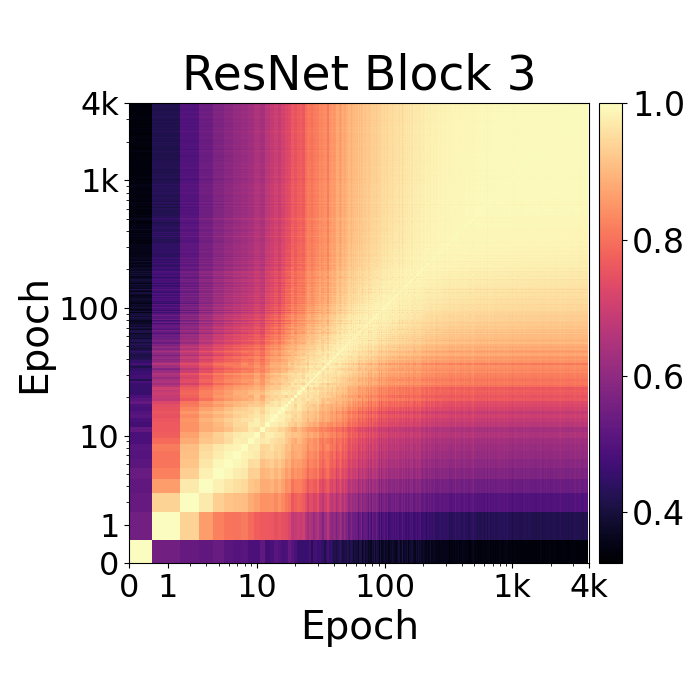}
    \label{cifar10_resnet18_noise_0_k_64_sgd_lr_0.001_momentum_0_bs_128_block_3_noiseless}}
    \subfloat[]{
    \includegraphics[width=0.13\textwidth]{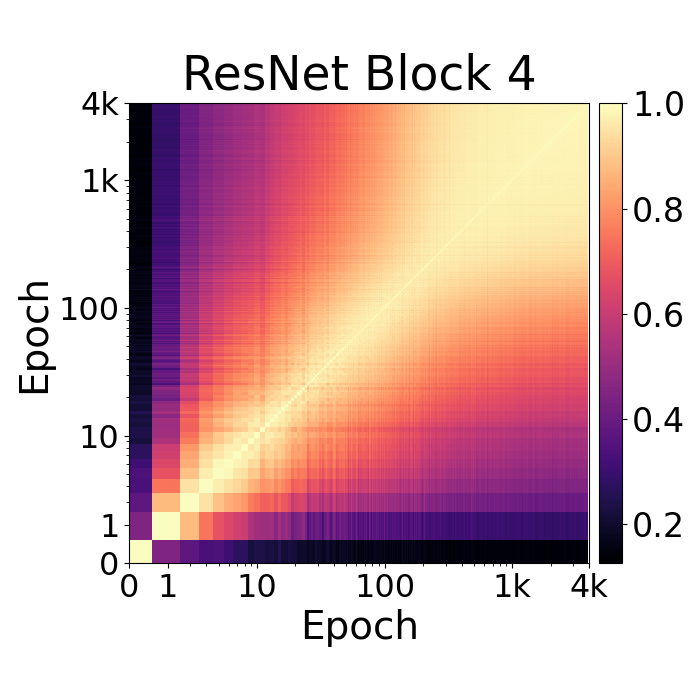} 
    \label{cifar10_resnet18_noise_0_k_64_sgd_lr_0.001_momentum_0_bs_128_block_4_noiseless}}
     \subfloat[]{
    \includegraphics[width=0.13\textwidth]{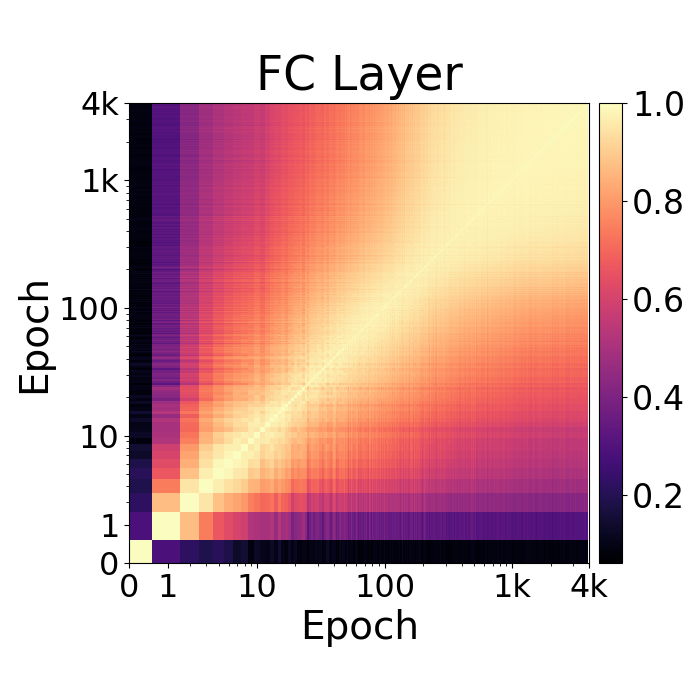}
    \label{cifar10_resnet18_noise_0_k_64_sgd_lr_0.001_momentum_0_bs_128_FC_noiseless}}
    \subfloat[]{
    \includegraphics[width=0.13\textwidth]{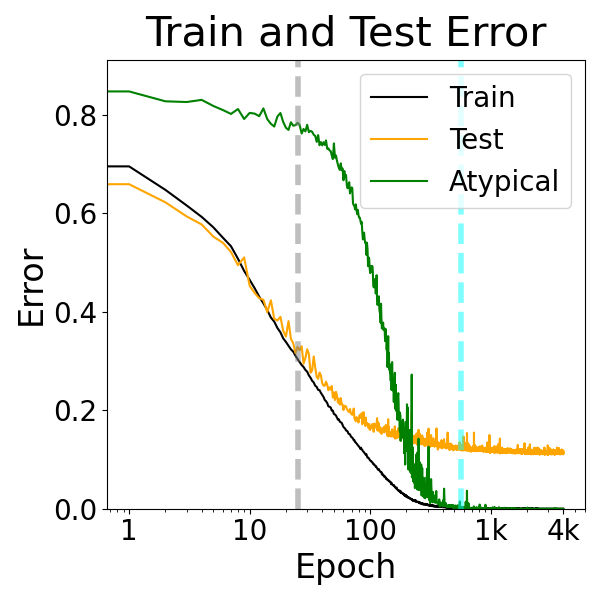}\label{cifar10_resnet18_noise_0_k_64_sgd_lr_0.001_momentum_0_bs_128_cifar10_noiseless_error_curve}
    }
  \caption{CKA evaluations for ResNet-18, trained using SGD (with learning rate of 0.001 and without momentum) on CIFAR-10 without label noise. Each of the (a)-(f) subfigures shows the CKA representational similarity of a specific layer in the ResNet-18 during its training. ResNet Block $j$ refers to the representation at the end of the $j^{\sf th}$ block of layers in the ResNet-18 architecture. (g) shows the test and train errors during training.}
  \label{fig:cifar10_resnet18_noise_0_k_64_sgd_lr_0.001_momentum_0_bs_128}
\end{figure*}

\begin{figure*}[t]
  \centering
    \subfloat[]{
    \includegraphics[width=0.13\textwidth]{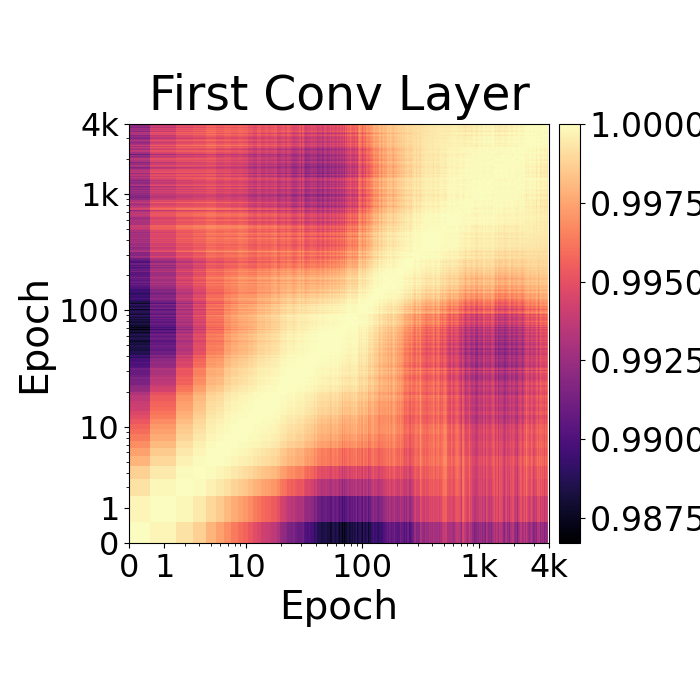}
    \label{Resnet18_k_64_first_conv_layer_tiny_imagenet_noise_20_half_train_set}}
    \subfloat[]{
    \includegraphics[width=0.13\textwidth]{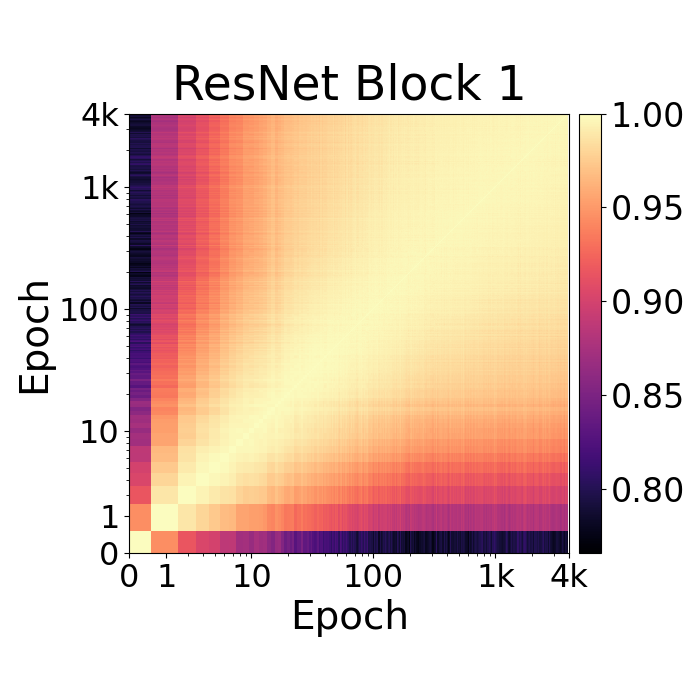}
    \label{Resnet18_k_64_block_1_tiny_imagenet_noise_20_half_train_set}
    }
    \subfloat[]{
    \includegraphics[width=0.13\textwidth]{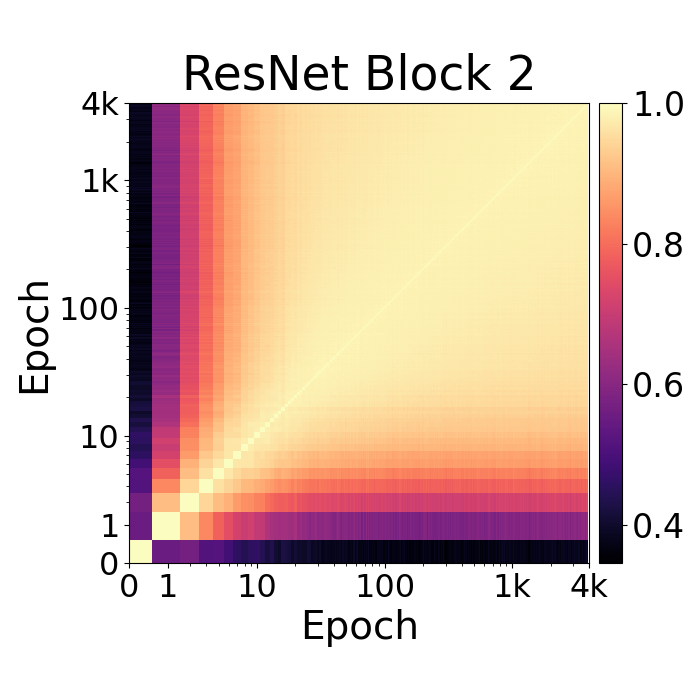}
    \label{Resnet18_k_64_block_2_tiny_imagenet_noise_20_half_train_set}}
    \subfloat[]{
    \includegraphics[width=0.13\textwidth]{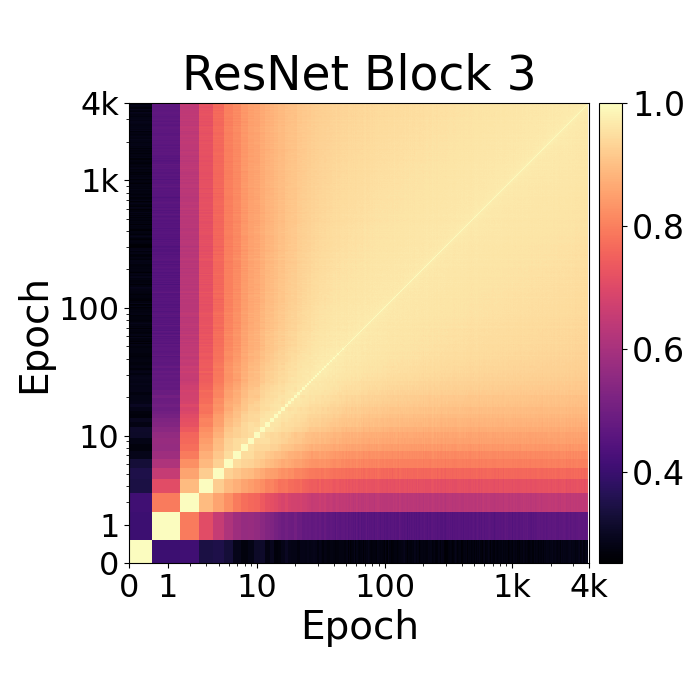}
    \label{Resnet18_k_64_block_3_tiny_imagenet_noise_20_half_train_set}}
    \subfloat[]{
    \includegraphics[width=0.13\textwidth]{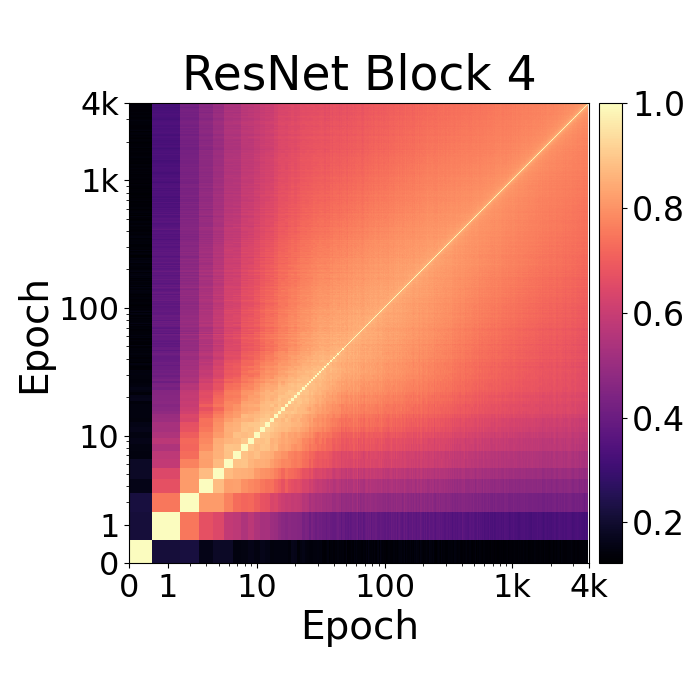}
    \label{Resnet18_k_64_block_4_tiny_imagenet_noise_20_half_train_set}}
     \subfloat[]{
    \includegraphics[width=0.13\textwidth]{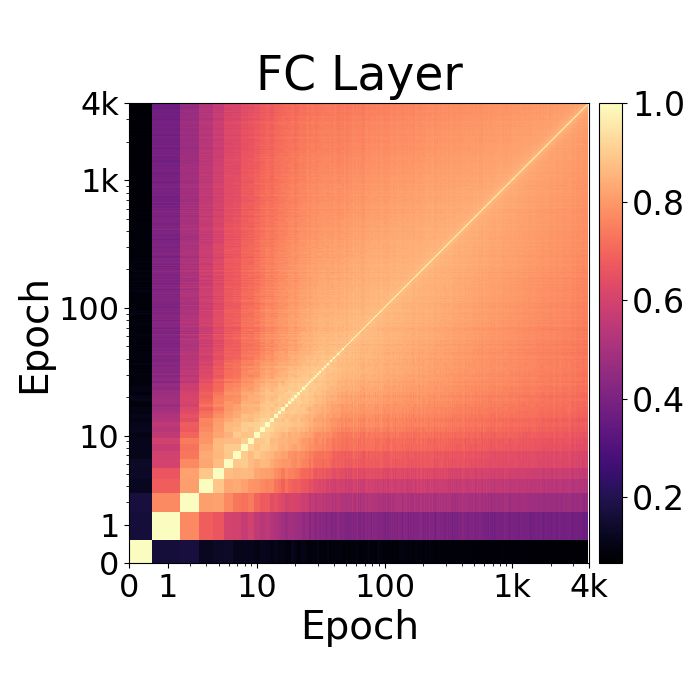}
    \label{Resnet18_k_64_FC_tiny_imagenet_noise_20_half_train_set}}
    \subfloat[]{
    \includegraphics[width=0.13\textwidth]{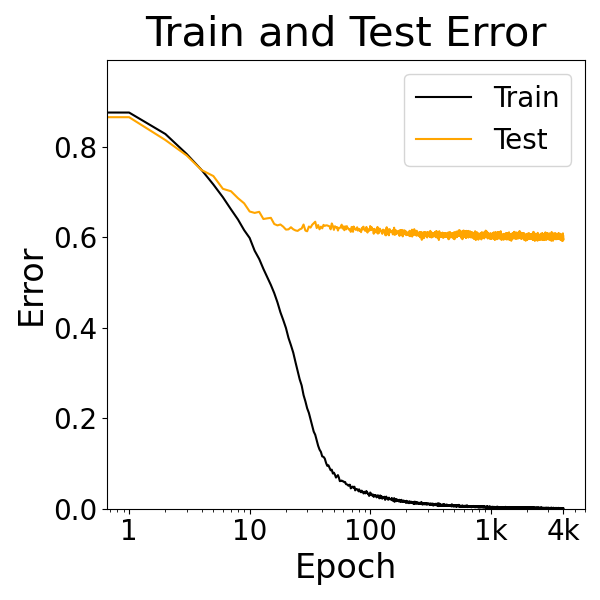}
    \label{Resnet18_k_64_error_curve_tiny_imagenet_noise_20_half_train_set}}
  \caption{CKA evaluations for ResNet-18 trained on Tiny-ImageNet with 20\% label noise. In contrast to Fig.~\ref{Resnet18_k_64_error_curve_tiny_imagenet_noise_20}, \textbf{here we used only half of the train set} (but still all the 200 classes) for training the model. Each of the (a)-(f) subfigures shows the CKA representational similarity of a specific layer in the ResNet-18 during its training. ResNet Block $j$ refers to the representation at the end of the $j^{\sf th}$ block of layers in the ResNet-18 architecture. (g) shows the test and train errors during training.}
  \label{fig:Resnet18_k_64_tiny_imagenet_noise_20_half_train_set}
\end{figure*}

\begin{figure*}[t]
  \centering
    \subfloat[]{
    \includegraphics[width=0.24\textwidth]{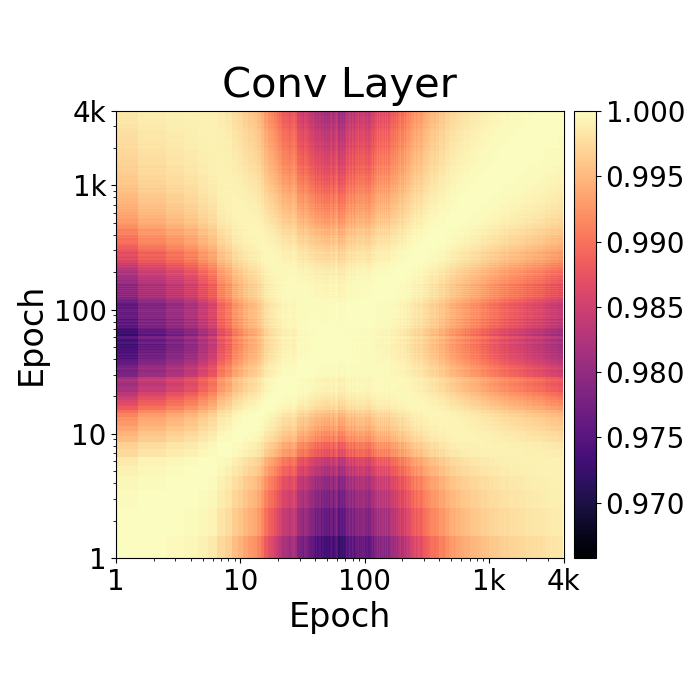}
    \label{simple_model_conv}}
     \subfloat[]{
    \includegraphics[width=0.24\textwidth]{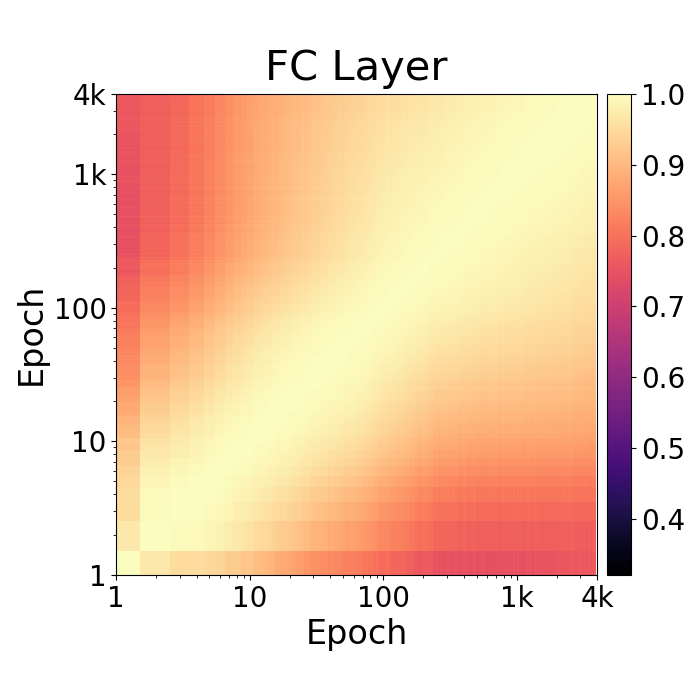}
    \label{simple_model_FC}}
    \subfloat[]{
    \includegraphics[width=0.23\textwidth]{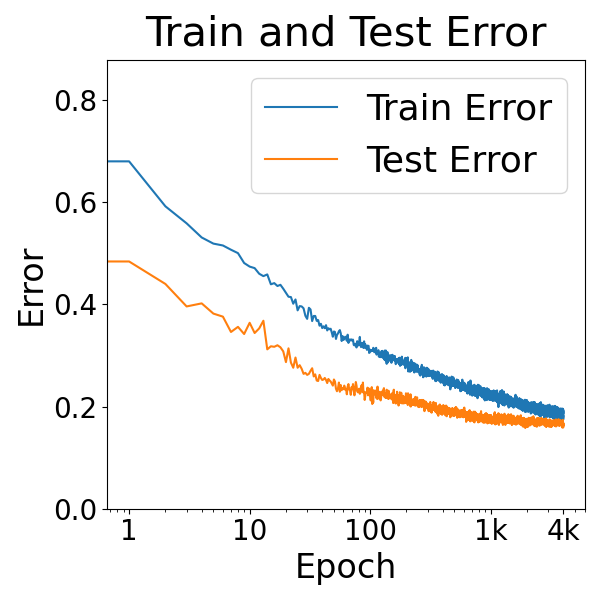}
    \label{simple_model_fashion_mnist_error_curve}}
  \caption{CKA evaluations for a simple 2 layer model, trained with Adam (with learning rate of 0.001) on 1000 samples from the Fashion MNIST dataset.}
  \label{fig:simple_model_fashion_mnist}
\end{figure*}

\begin{figure*}[t]
  \centering
    \subfloat[]{
    \includegraphics[width=0.13\textwidth]{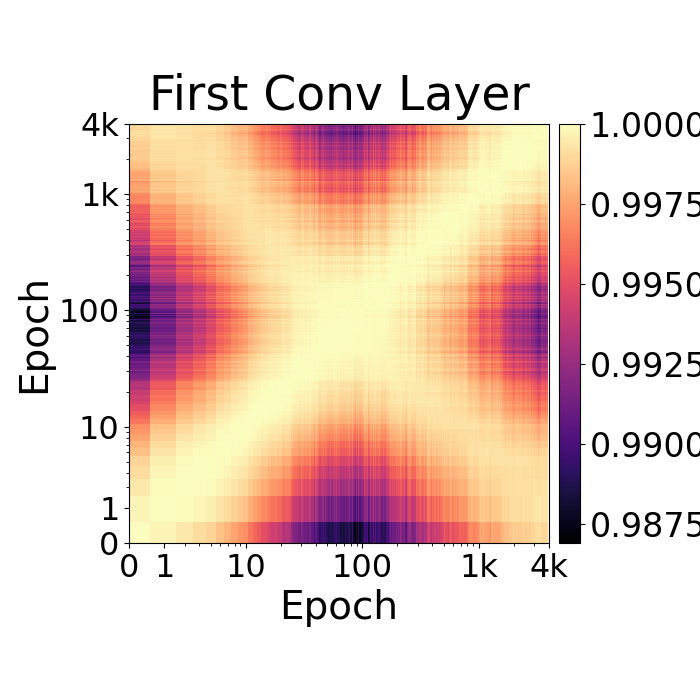}
    \label{svhn_Resnet18_k_64_adam_first_conv_layer_noiseless}}
    \subfloat[]{
    \includegraphics[width=0.13\textwidth]{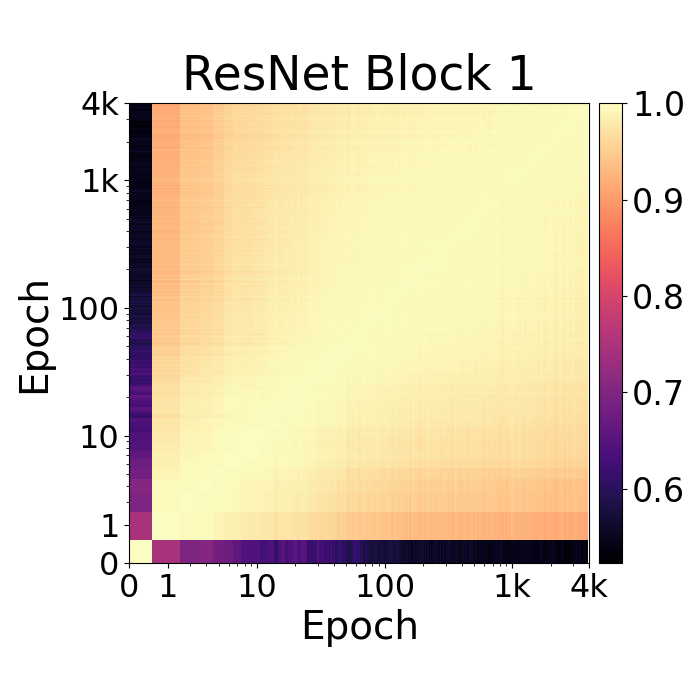}
    \label{svhn_Resnet18_k_64_adam_block_1_noiseless}
    }
    \subfloat[]{
    \includegraphics[width=0.13\textwidth]{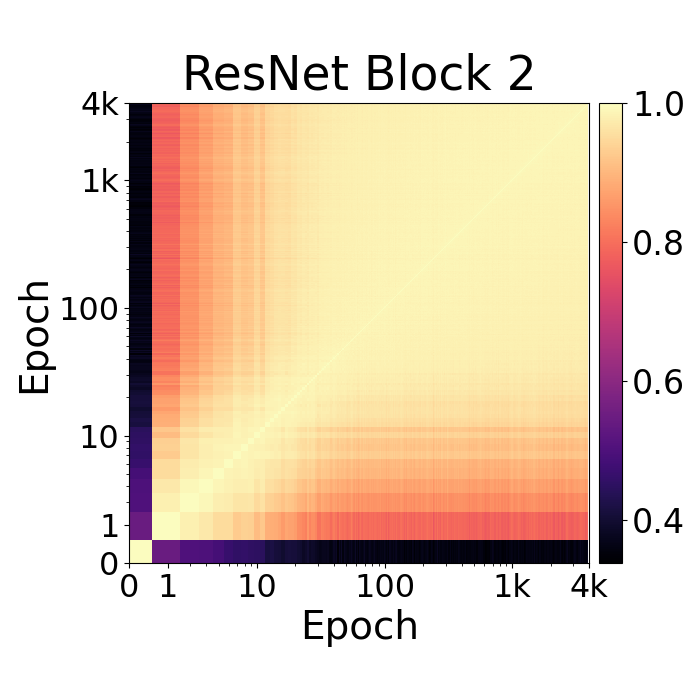}
    \label{svhn_Resnet18_k_64_adam_block_2_noiseless}}
    \subfloat[]{
    \includegraphics[width=0.13\textwidth]{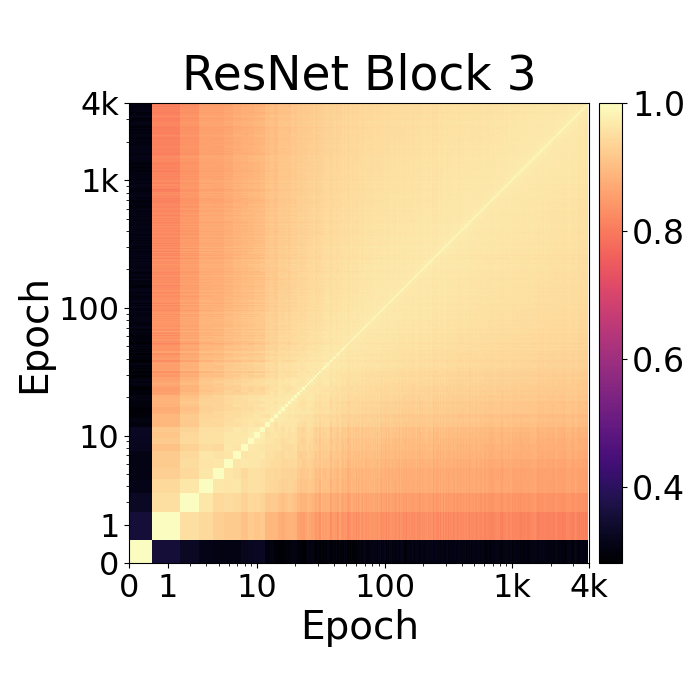}
    \label{svhn_Resnet18_k_64_adam_block_3_noiseless}}
    \subfloat[]{
    \includegraphics[width=0.13\textwidth]{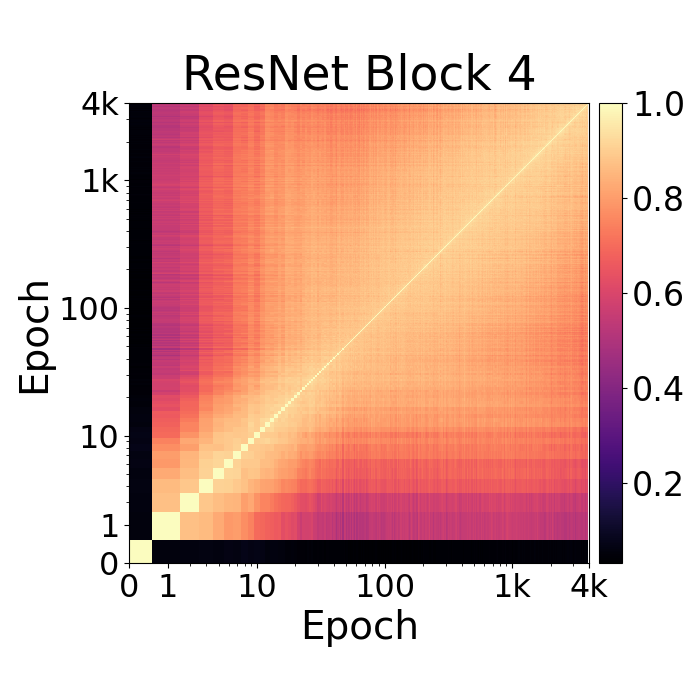}   
    \label{svhn_Resnet18_k_64_adam_block_4_noiseless}}
     \subfloat[]{
    \includegraphics[width=0.13\textwidth]{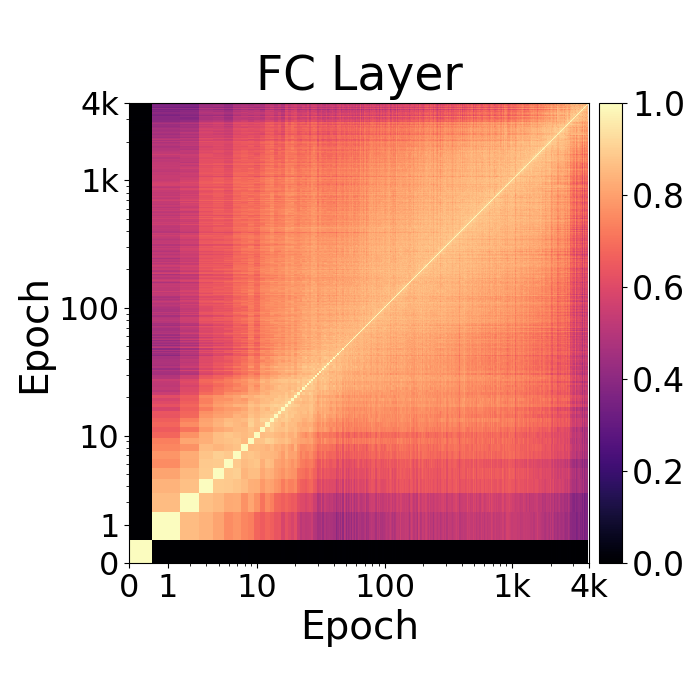}
    \label{svhn_Resnet18_k_64_adam_FC_noiseless}}
    \subfloat[]{
    {
    \includegraphics[width=0.13\textwidth]{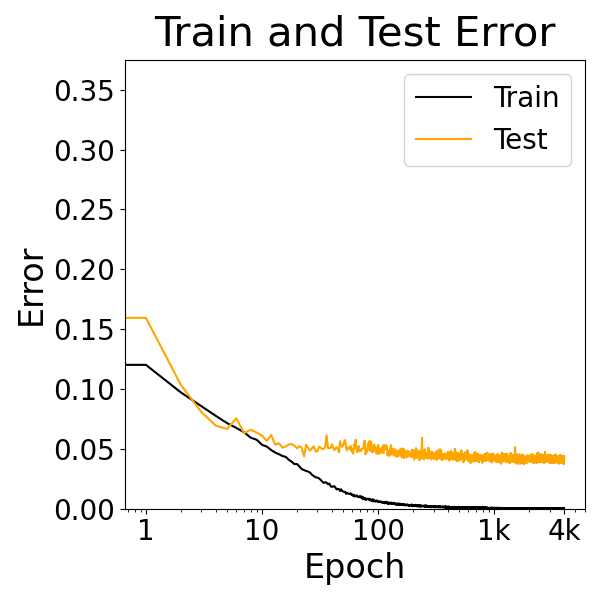}
    \label{ResNet18_k_64_adam_svhn_noiseless_error_curve}}
    }
  \caption{CKA evaluations for ResNet-18 trained on SVHN without label noise. Each of the (a)-(f) subfigures shows the CKA representational similarity of a specific layer in the ResNet-18 during its training. ResNet Block $j$ refers to the representation at the end of the $j^{\sf th}$ block of layers in the ResNet-18 architecture. (g) shows the test and train errors during training.}
  \label{fig:svhn_Resnet18_k_64_adam_noise_0}
\end{figure*}

\begin{figure*}[t]
  \centering
    \subfloat[]{
    \includegraphics[width=0.13\textwidth]{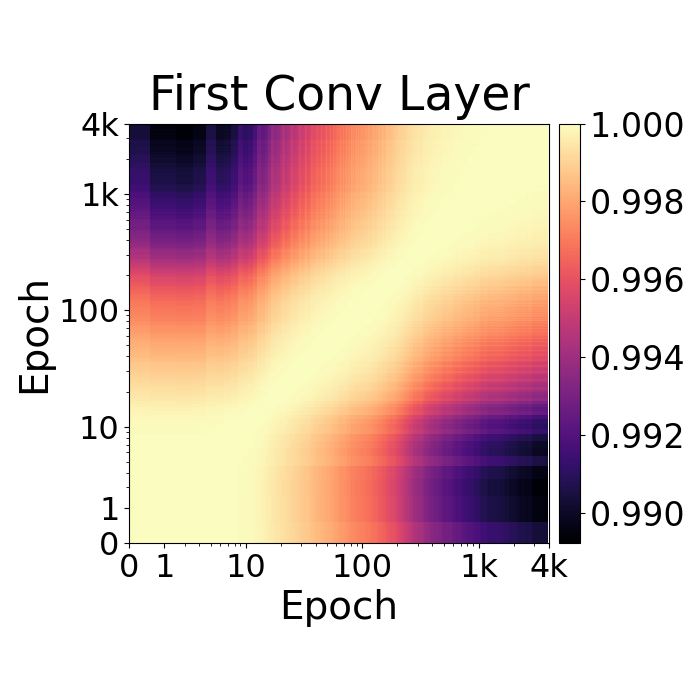}
    \label{svhn_Resnet18_k_64_sgd_first_conv_layer_noiseless}}
    \subfloat[]{
    \includegraphics[width=0.13\textwidth]{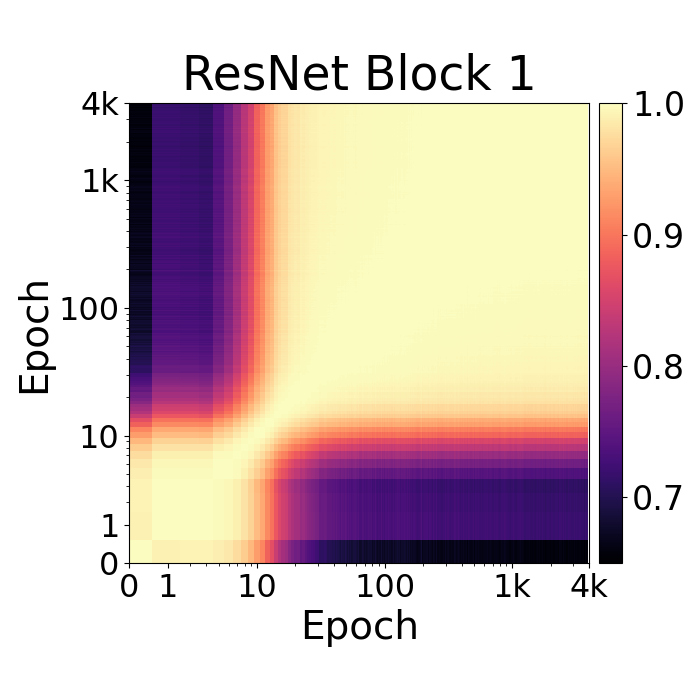}
    \label{svhn_Resnet18_k_64_sgd_block_1_noiseless}
    }
    \subfloat[]{
    \includegraphics[width=0.13\textwidth]{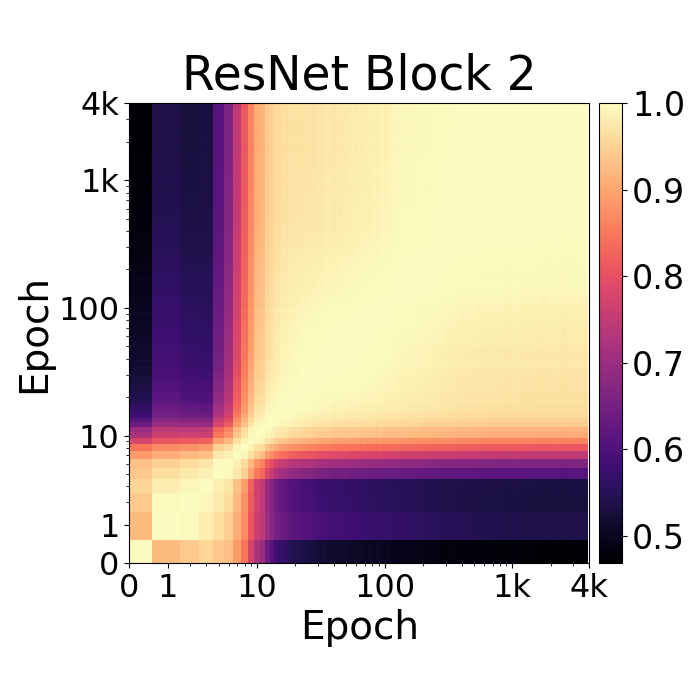}
    \label{svhn_Resnet18_k_64_sgd_block_2_noiseless}}
    \subfloat[]{
    \includegraphics[width=0.13\textwidth]{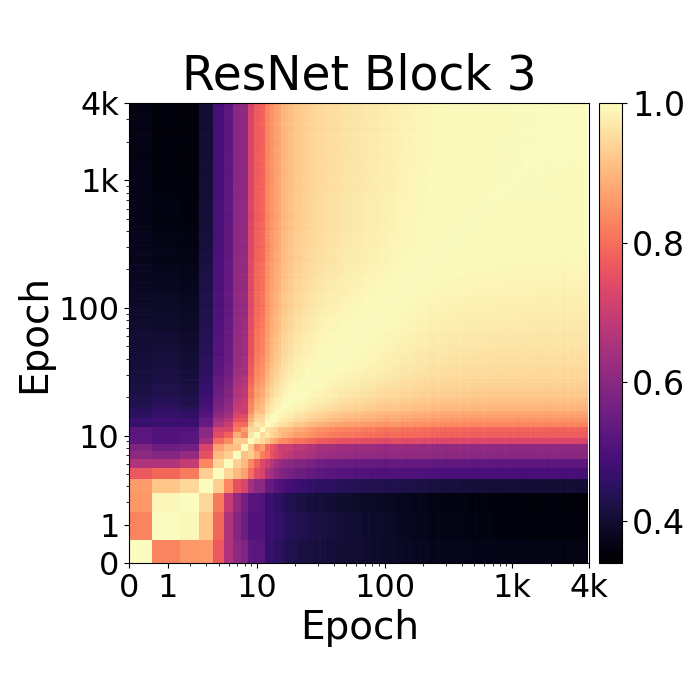}
    \label{svhn_Resnet18_k_64_sgd_block_3_noiseless}}
    \subfloat[]{
    \includegraphics[width=0.13\textwidth]{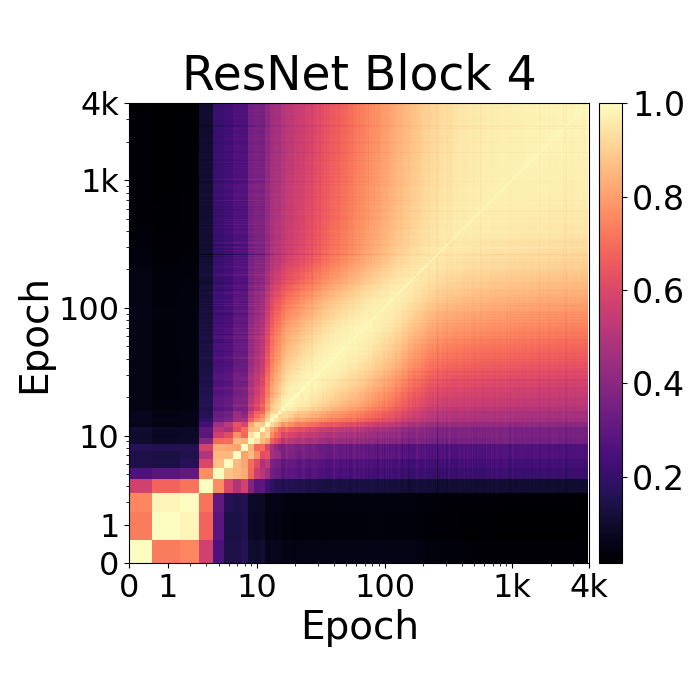}   
    \label{svhn_Resnet18_k_64_sgd_block_4_noiseless}}
     \subfloat[]{
    \includegraphics[width=0.13\textwidth]{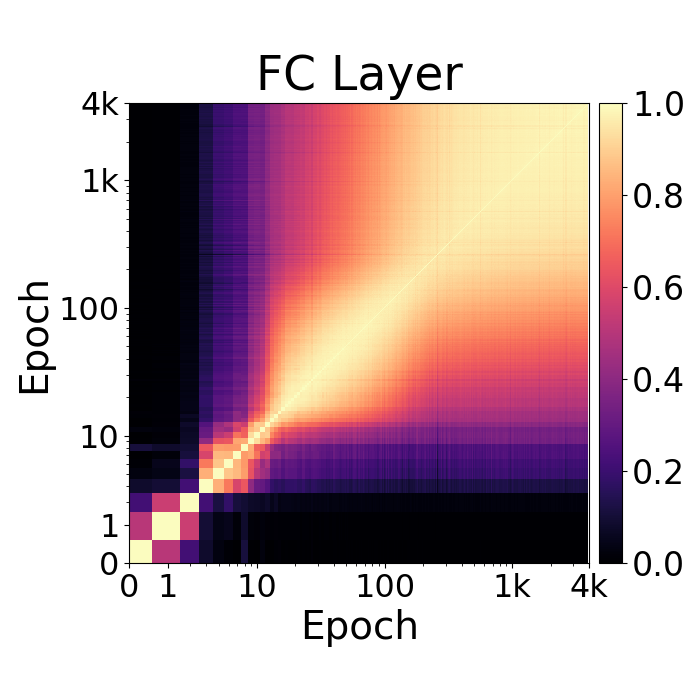}
    \label{svhn_Resnet18_k_64_sgd_FC_noiseless}}
    \subfloat[]{
    {
    \includegraphics[width=0.13\textwidth]{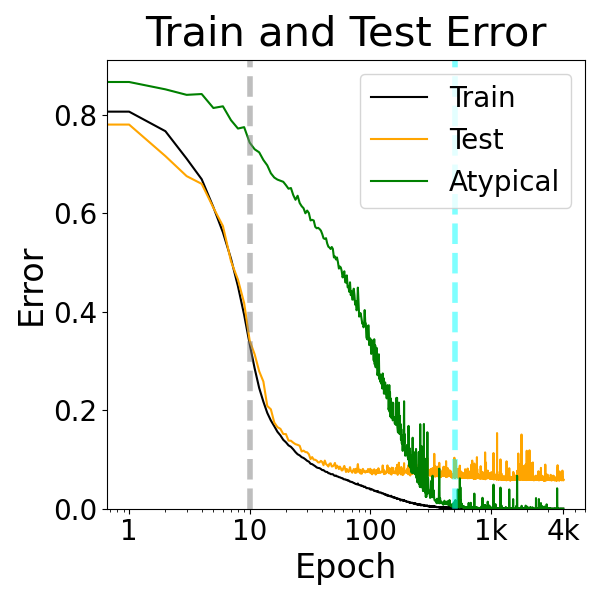}
    \label{ResNet18_k_64_sgd_svhn_noiseless_error_curve}}
    }
  \caption{CKA evaluations for ResNet-18 trained on SVHN without label noise using SGD optimizer. Each of the (a)-(f) subfigures shows the CKA representational similarity of a specific layer in the ResNet-18 during its training. ResNet Block $j$ refers to the representation at the end of the $j^{\sf th}$ block of layers in the ResNet-18 architecture. (g) shows the test and train errors during training.}
  \label{fig:svhn_Resnet18_k_64_sgd_noise_0}
\end{figure*}

\begin{figure*}[]
\centering
\subfloat[]{
\includegraphics[width=0.24\textwidth]{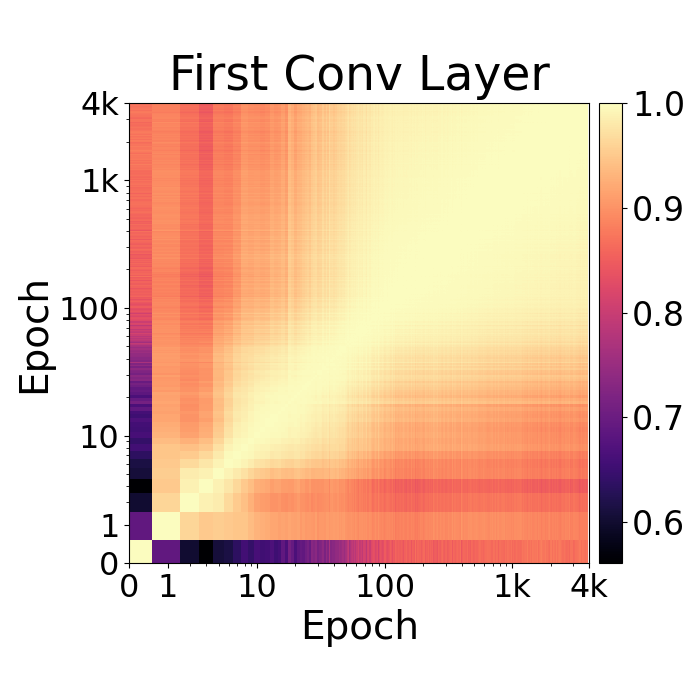}
\label{vit_first_conv_layer}}
\subfloat[]{
\includegraphics[width=0.24\textwidth]{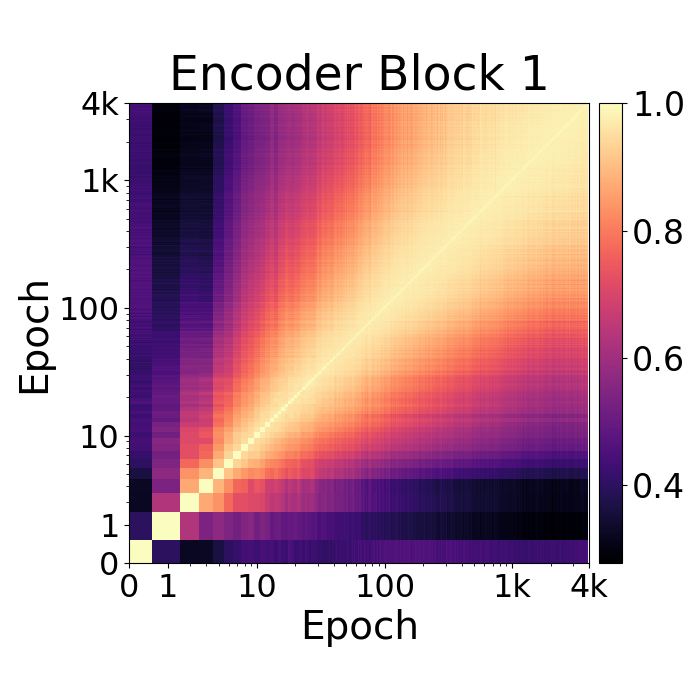}
\label{}}
\subfloat[]{
\includegraphics[width=0.24\textwidth]{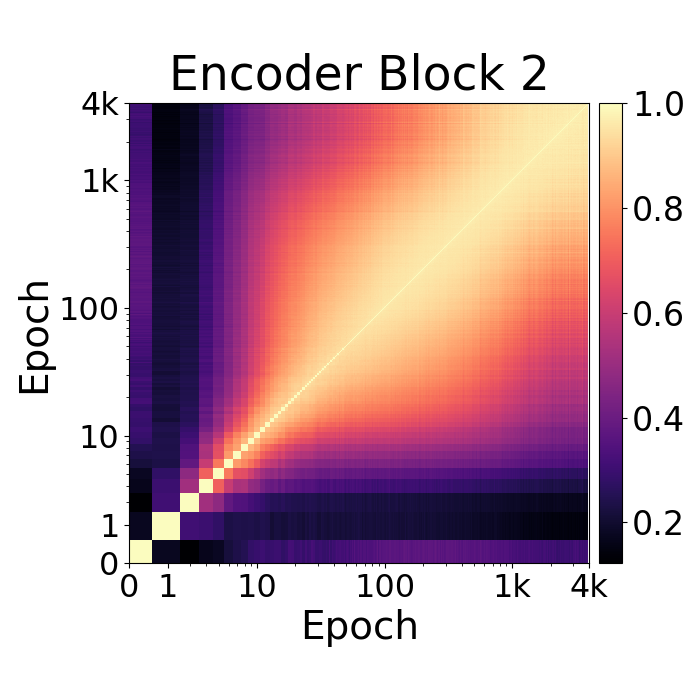}
\label{}}
\subfloat[]{
\includegraphics[width=0.24\textwidth]{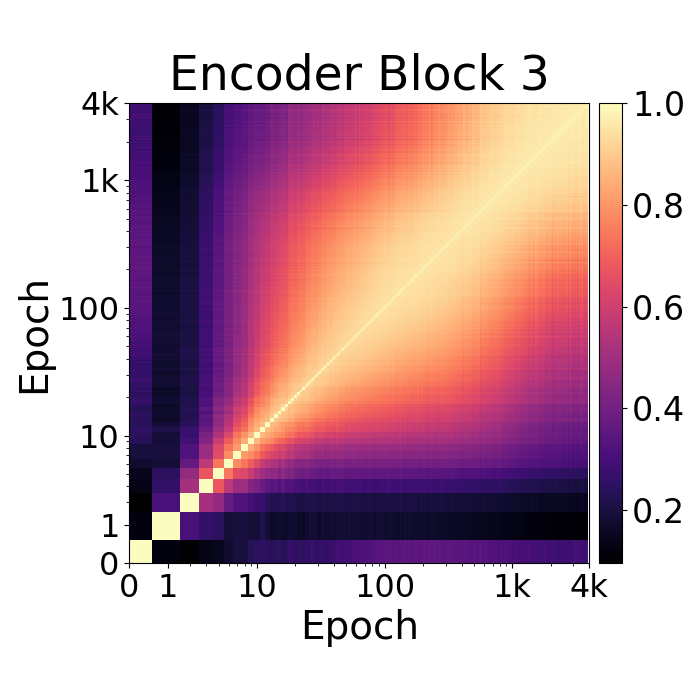}
\label{}}
\\[-3ex]
\subfloat[]{
\includegraphics[width=0.24\textwidth]{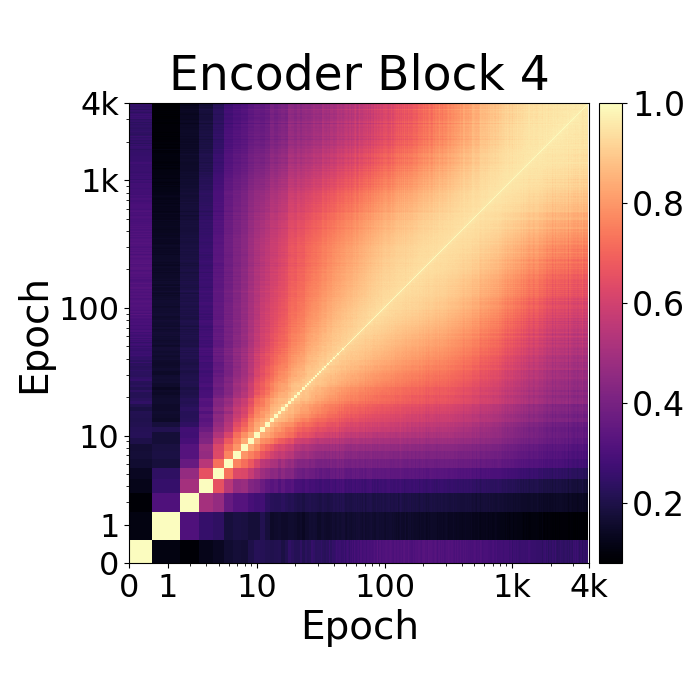}
\label{}}
\subfloat[]{
\includegraphics[width=0.24\textwidth]{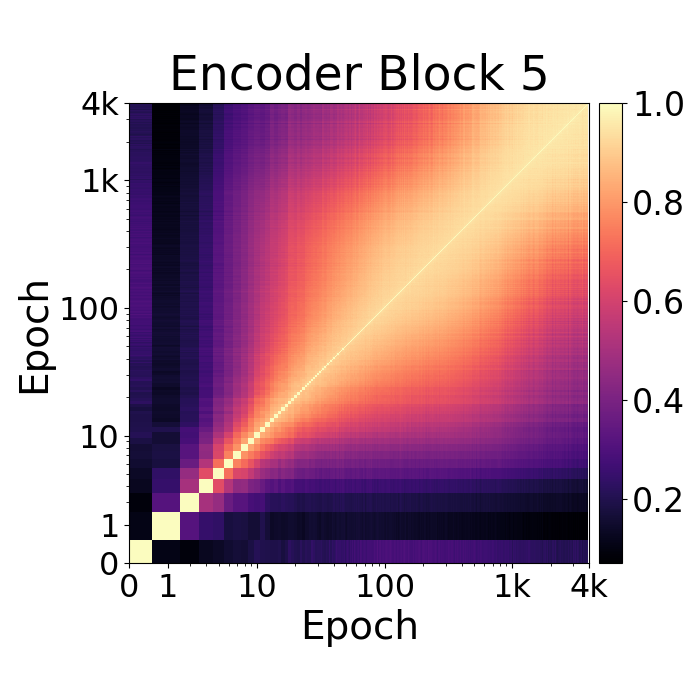}
\label{}}
\subfloat[]{
\includegraphics[width=0.24\textwidth]{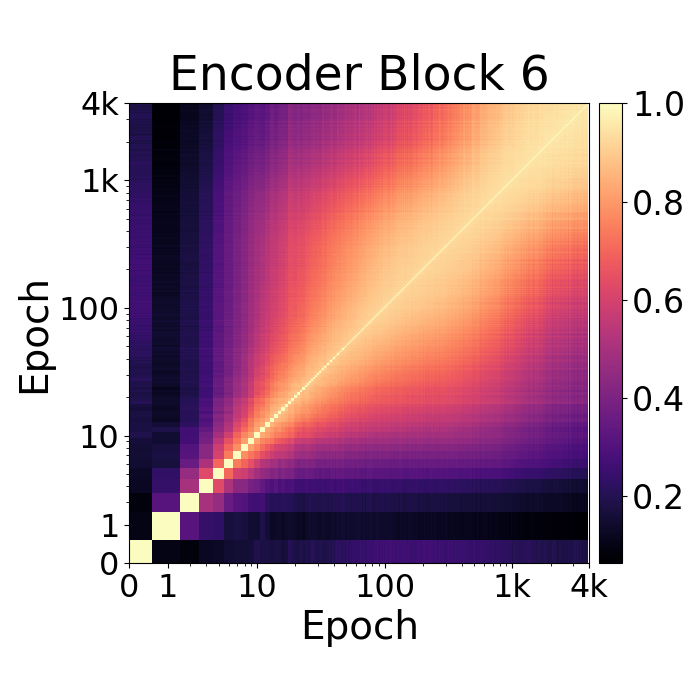}
\label{}}
\subfloat[]{
\includegraphics[width=0.24\textwidth]{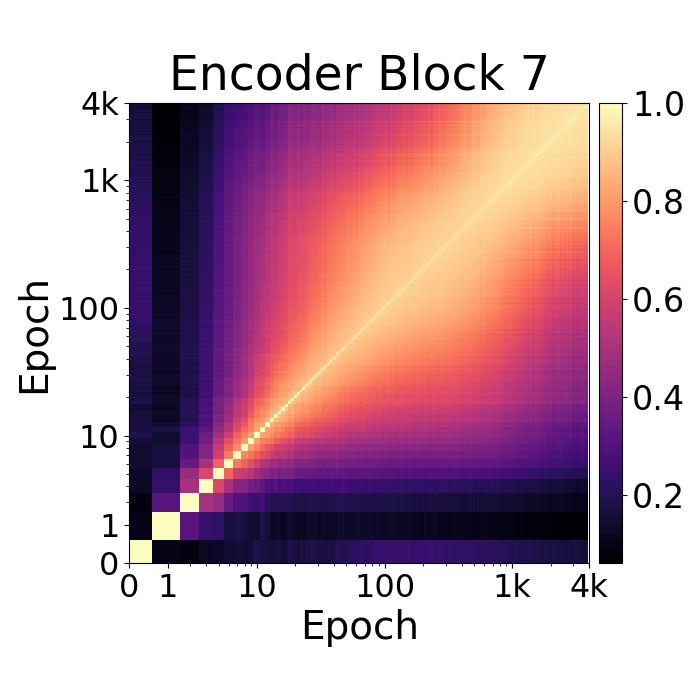}
\label{}}
\\[-3ex]
\subfloat[]{
\includegraphics[width=0.24\textwidth]{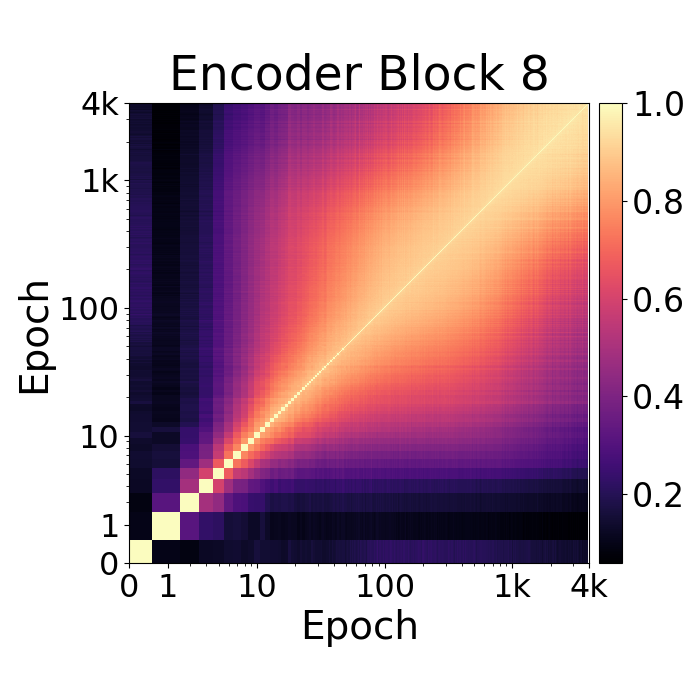}
\label{}}
\subfloat[]{
\includegraphics[width=0.24\textwidth]{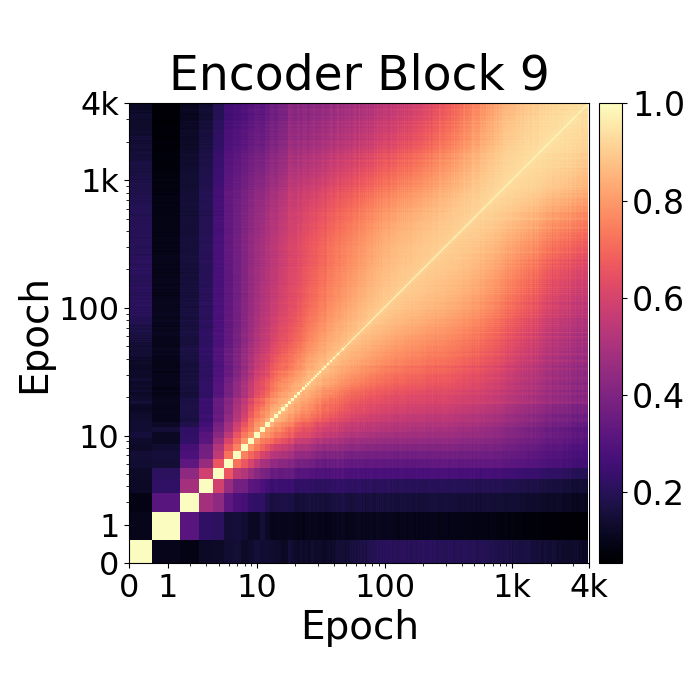}
\label{}}
\subfloat[]{
\includegraphics[width=0.24\textwidth]{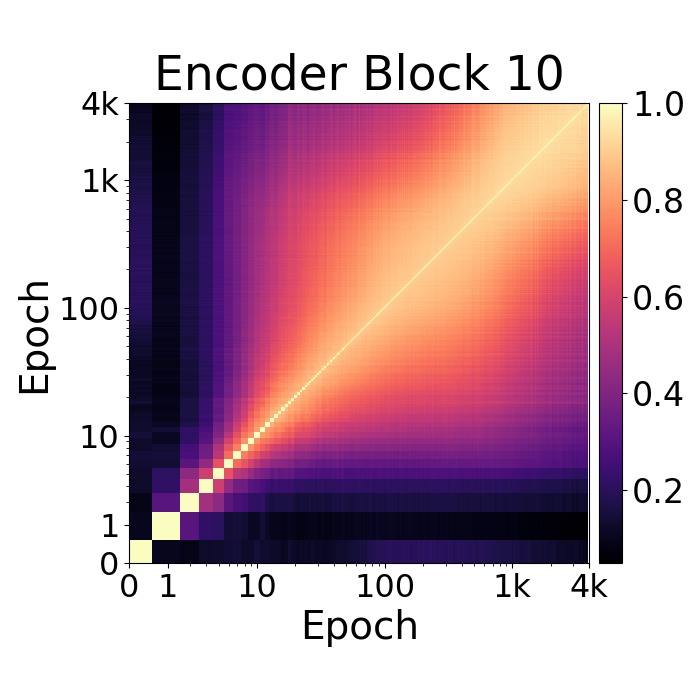}
\label{}}
\subfloat[]{
\includegraphics[width=0.24\textwidth]{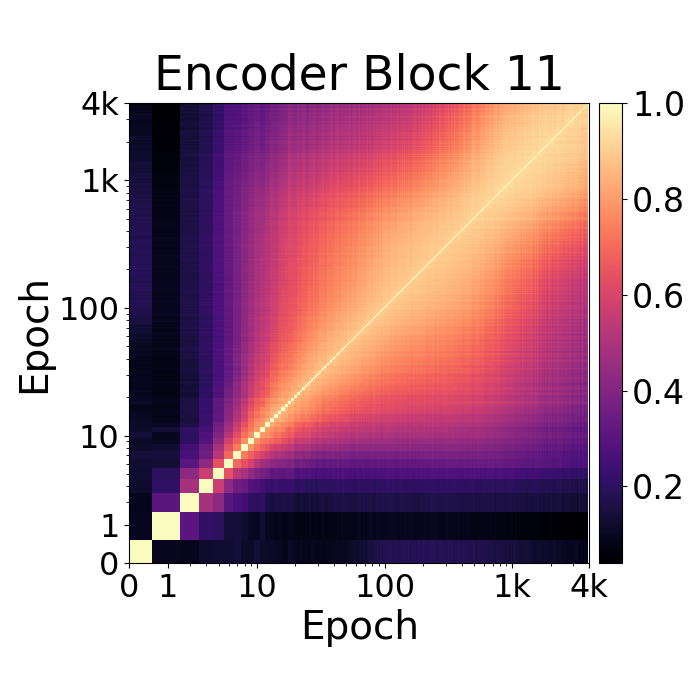}
\label{}}
\\[-3ex]
\subfloat[]{
\includegraphics[width=0.24\textwidth]{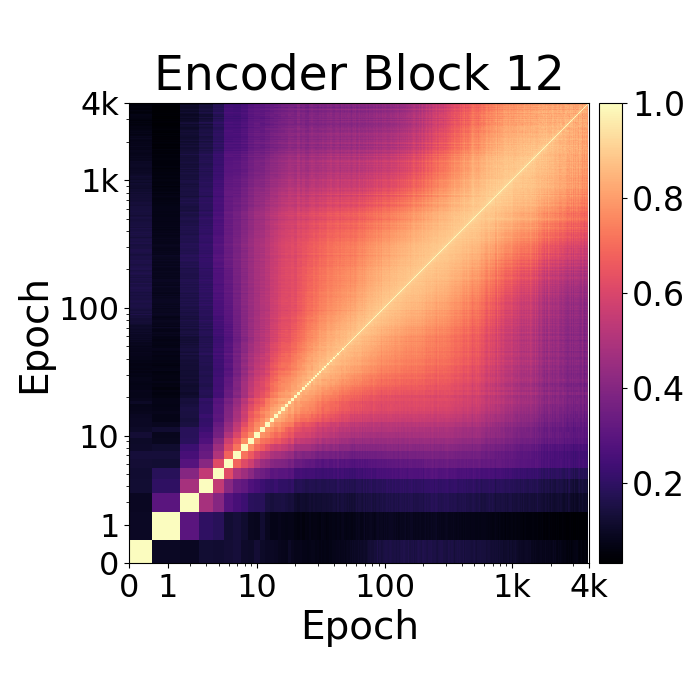}
\label{}}
\subfloat[]{
\includegraphics[width=0.23\textwidth]{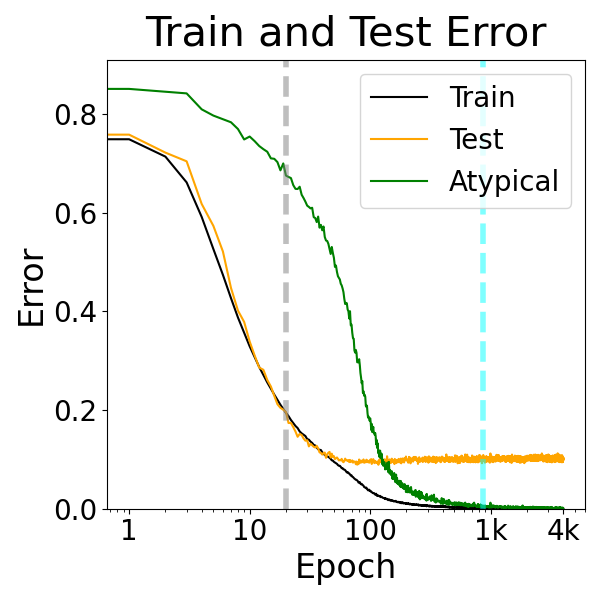}
\label{}}
\caption{CKA evaluations for ViT-B/16, trained on SVHN without label noise and with Adam optimizer.}
\label{fig:svhn_vit_b_16_noise_0_k_64_adam_lr_0.0001_momentum_0_bs_128_20250112}
\end{figure*}

\begin{figure*}[t]
  \centering
    \subfloat[]{
    \includegraphics[width=0.23\textwidth]{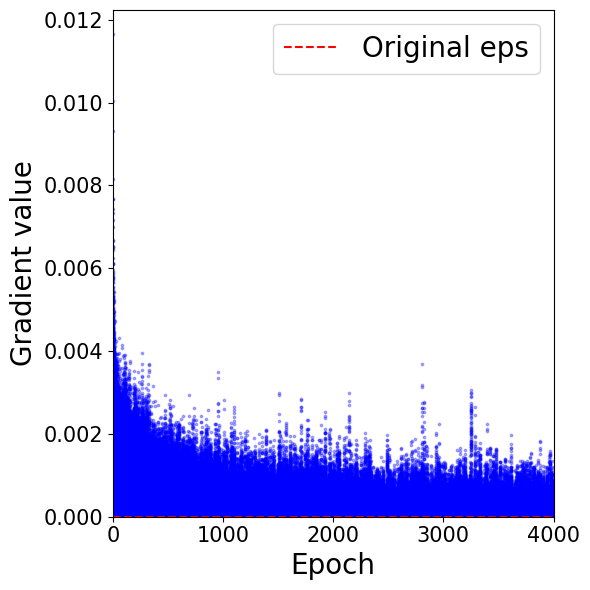}
    \label{fig:first_layer_grad_adam_standart_eps}}
    \subfloat[]{
    \includegraphics[width=0.23\textwidth]{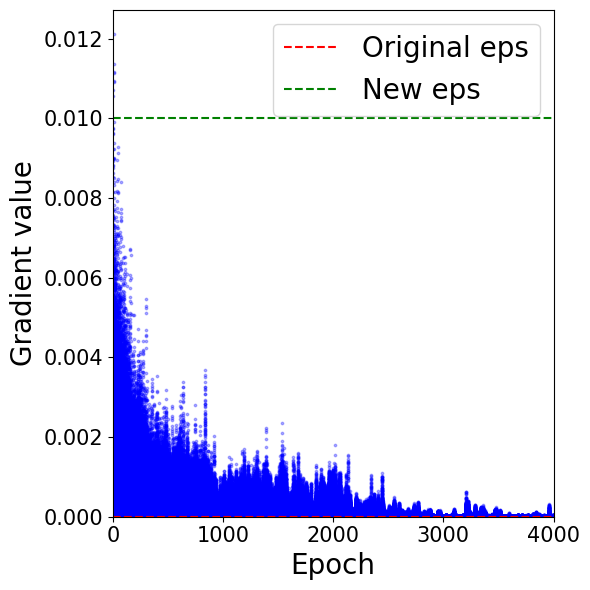}
    \label{fig:first_layer_grad_adam_low_eps}}
    \subfloat[]{
    \includegraphics[width=0.23\textwidth]{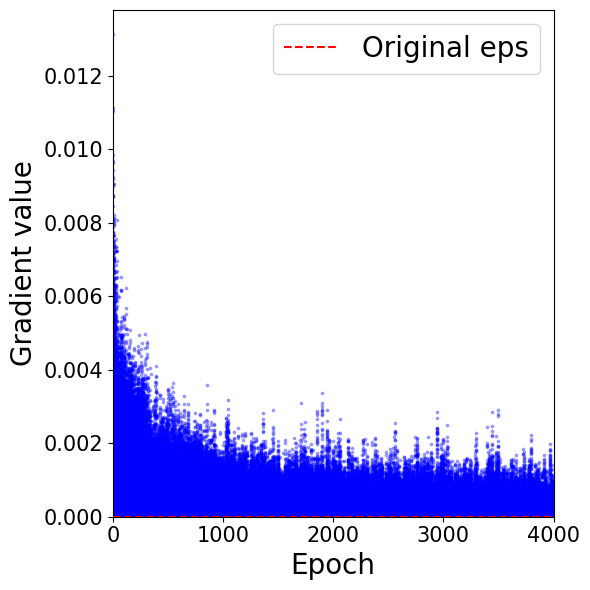}
    \label{fig:first_layer_grad_adam_k_42}}
    \subfloat[]{
    \includegraphics[width=0.23\textwidth]{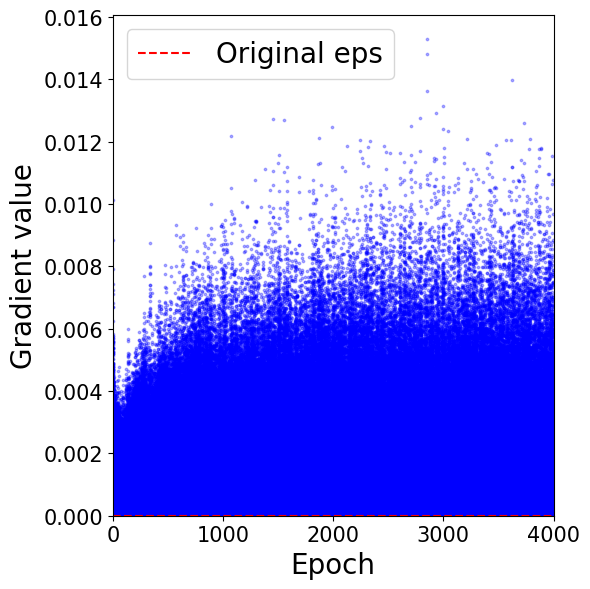}
    \label{fig:first_layer_grad_adam_wd}}
  \\
  \subfloat[]{
    \includegraphics[width=0.23\textwidth]{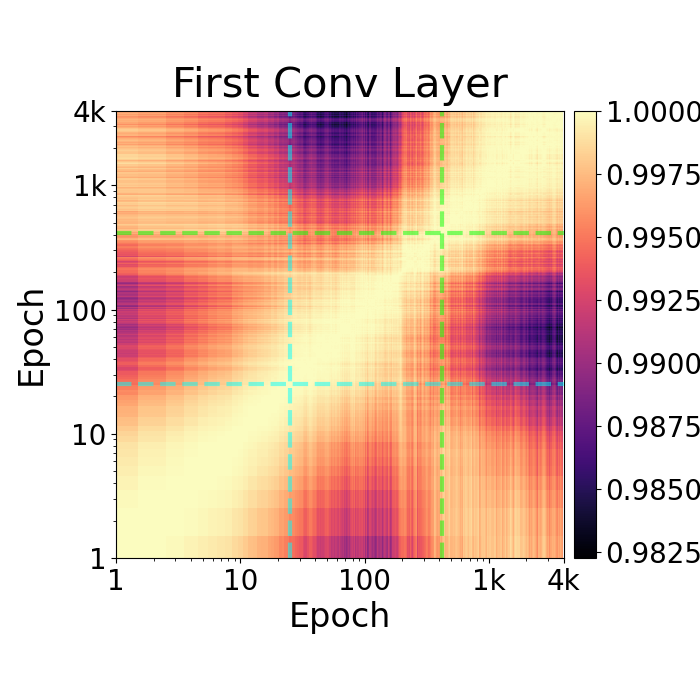}
    \label{fig:first_layer_grad_adam_standart_eps_conv1}}
    \subfloat[]{
    \includegraphics[width=0.23\textwidth]{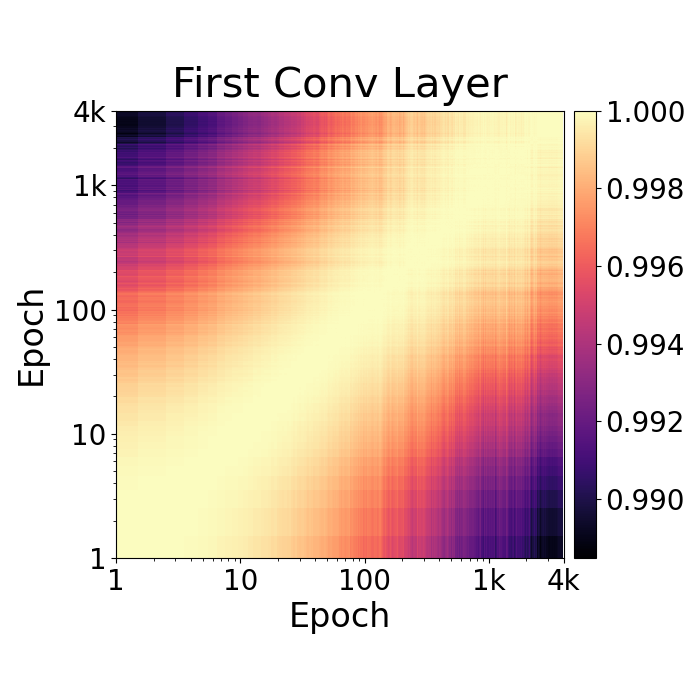}
    \label{fig:first_layer_grad_adam_low_eps_conv1}}
    \subfloat[]{
    \includegraphics[width=0.23\textwidth]{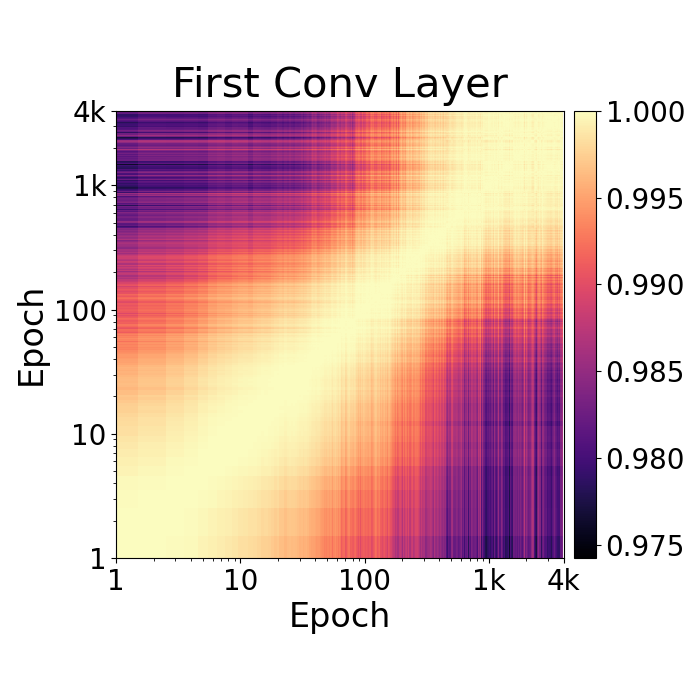}
    \label{fig:first_layer_grad_adam_k_42_conv}}
    \subfloat[]{
    \includegraphics[width=0.23\textwidth]{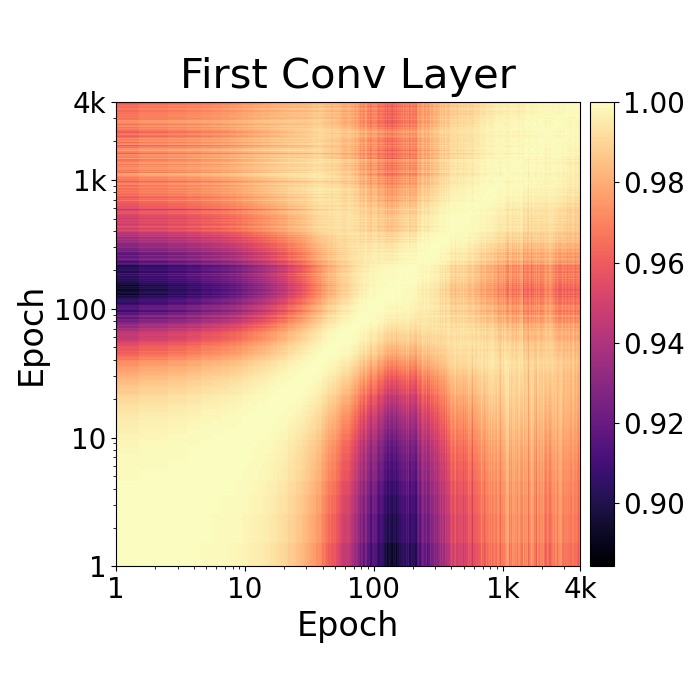}
    \label{fig:first_layer_grad_adam_wd}}
  \caption{The values of the components of the gradient of the first convolution layer in Adam training of ResNet-18 (upper row) on CIFAR-10, and the CKA heatmaps of the same layer (lower row). First column is for standard training with numerical stability parameter with the standard value of $\epsilon=1e-8$, second column is for training with $\epsilon=0.01$, third column for ResNet-18 with parameter $k=42$ (and standard value $\epsilon=1e-8$) and the the last column is for ResNet-18 trained with weight decay regularization ($\epsilon=1e-8$).}
  \label{fig:first_layer_gradient}
\end{figure*}

\begin{figure*}[]
\centering
\subfloat[]{
\includegraphics[width=0.14\textwidth]{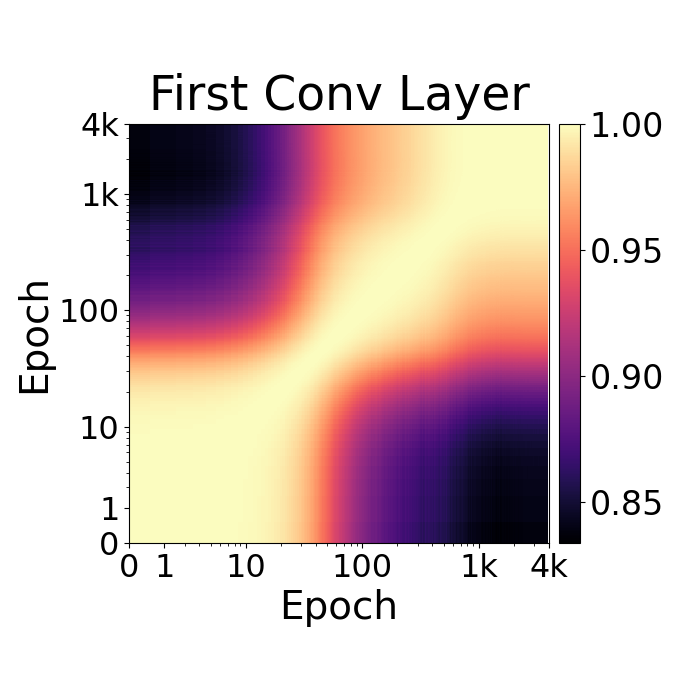}
\label{fig:vit_cifar10_sgd_first_conv_layer_noise_0}}
\subfloat[]{
\includegraphics[width=0.14\textwidth]{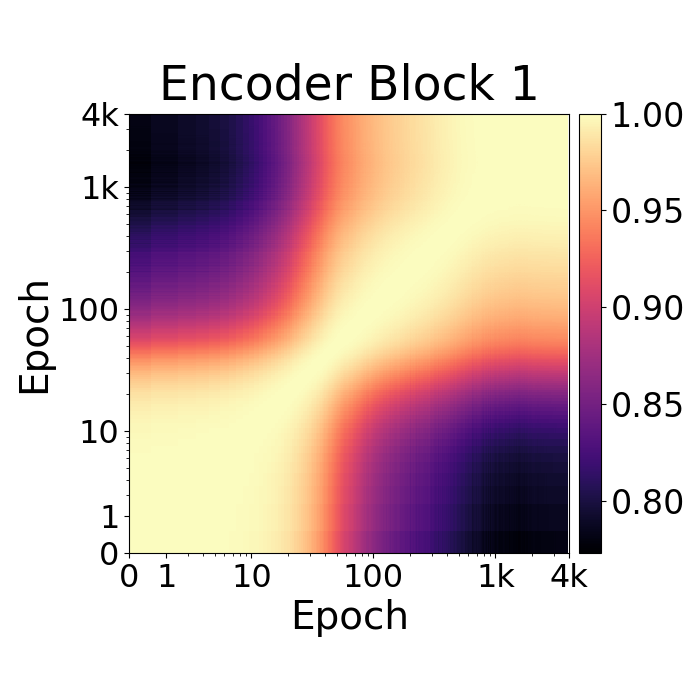}
\label{fig:vit_cifar10_sgd_Encoder_block_1_noise_0}}
\subfloat[]{
\includegraphics[width=0.14\textwidth]{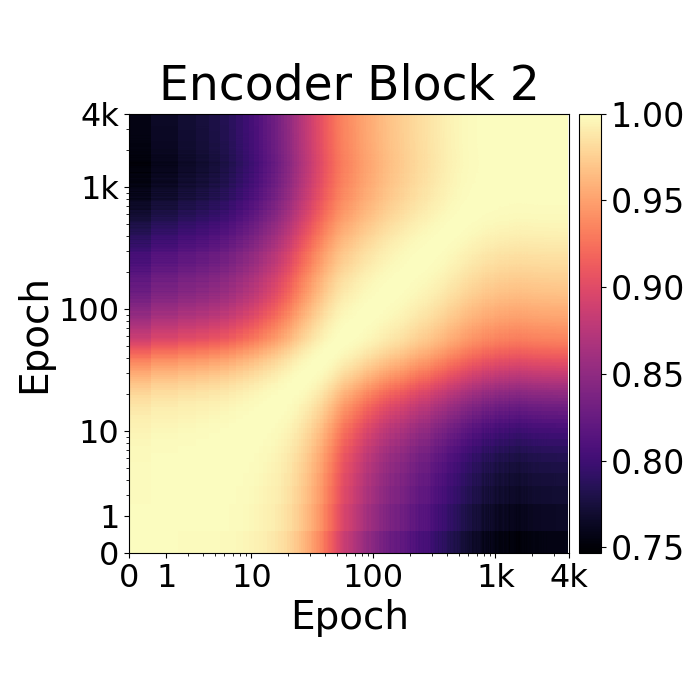}
\label{fig:vit_cifar10_sgd_Encoder_block_2}}
\subfloat[]{
\includegraphics[width=0.14\textwidth]{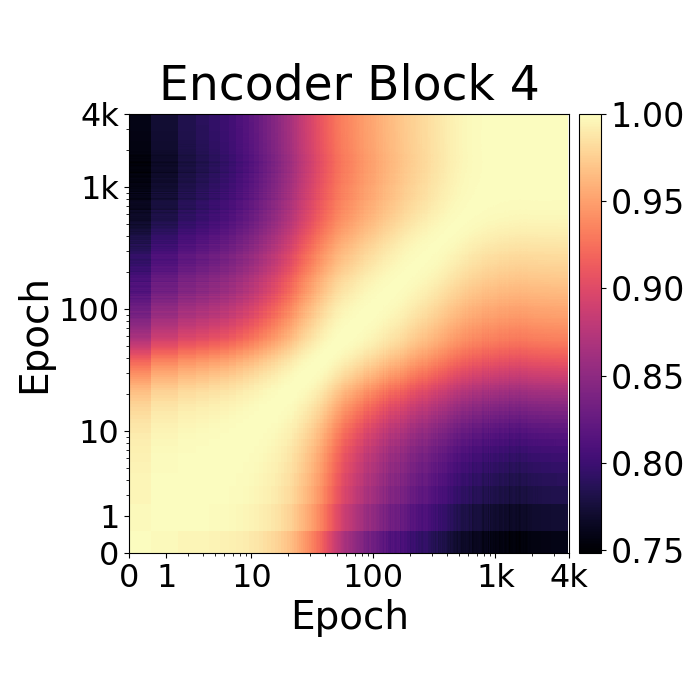}
\label{fig:vit_cifar10_sgd_Encoder_block_4_noise_0}}
\subfloat[]{
\includegraphics[width=0.14\textwidth]{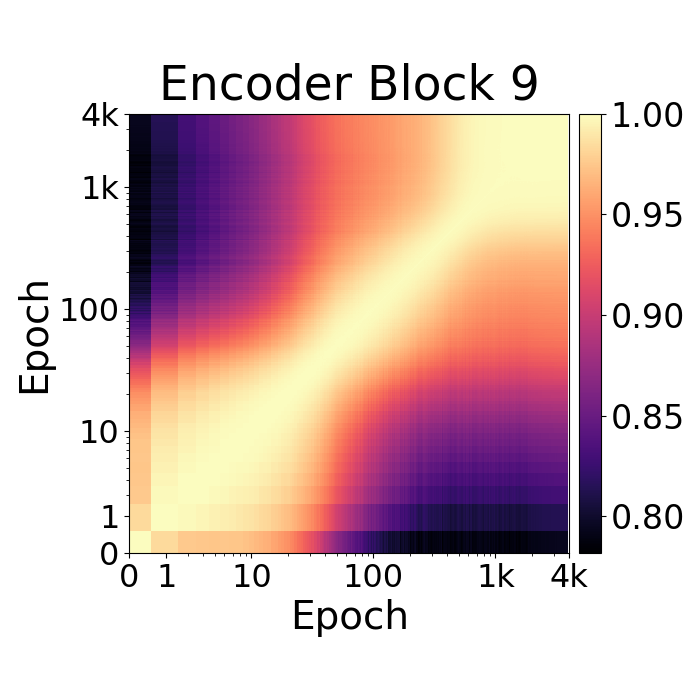}
\label{fig:vit_cifar10_sgd_Encoder_block_9_noise_0}}
\subfloat[]{
\includegraphics[width=0.14\textwidth]{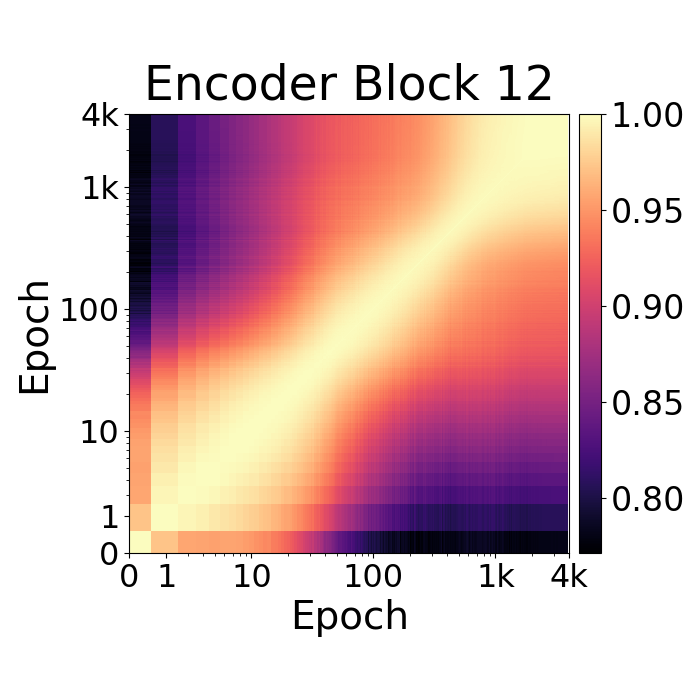}
\label{fig:vit_cifar10_sgd_Encoder_block_12_noise_0}}
\subfloat[]{
\includegraphics[width=0.14\textwidth]{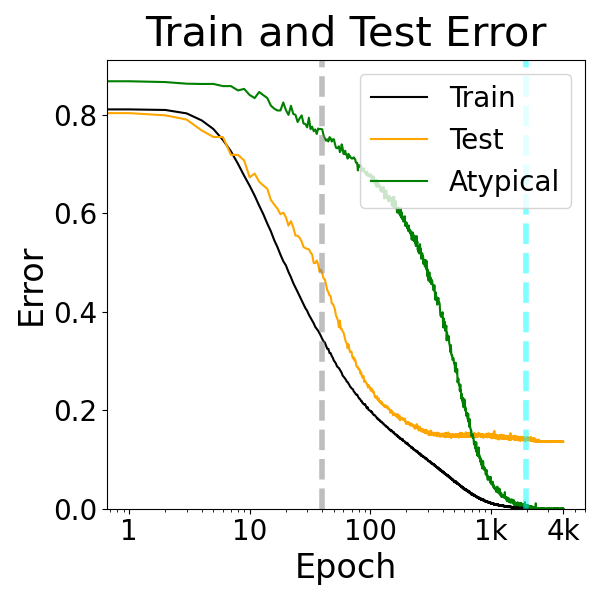}
\label{fig:vit_cifar10_sgd_vit_train_test_error_noise_0}}
\caption{CKA evaluations for ViT-B/16, trained on CIFAR-10 without label noise, using SGD optimizer with learning rate of 0.001 and without momentum}
\label{fig:vit_cifar10_sgd_cka_comparison_noise_0}
\end{figure*}

\begin{figure*}[]
\centering
\subfloat[]{
\includegraphics[width=0.14\textwidth]{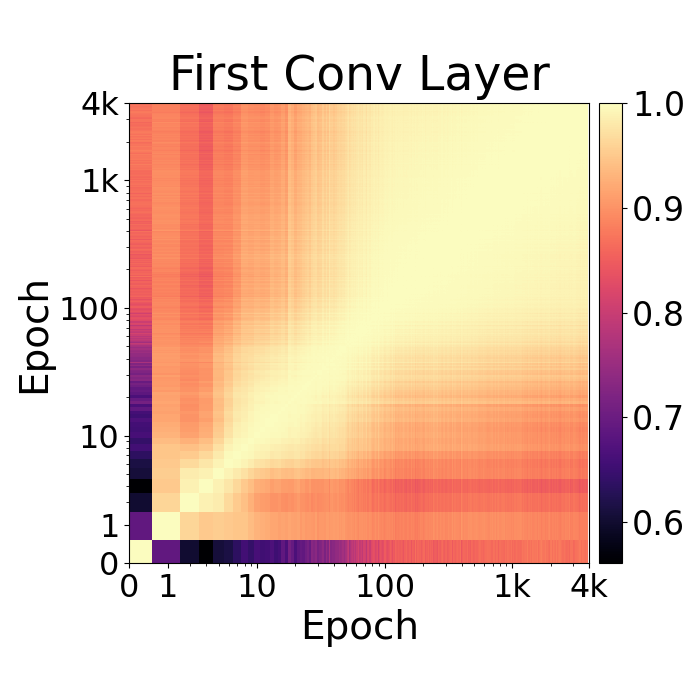}
\label{fig:vit_svhn_Adam_first_conv_layer_noise_0}}
\subfloat[]{
\includegraphics[width=0.14\textwidth]{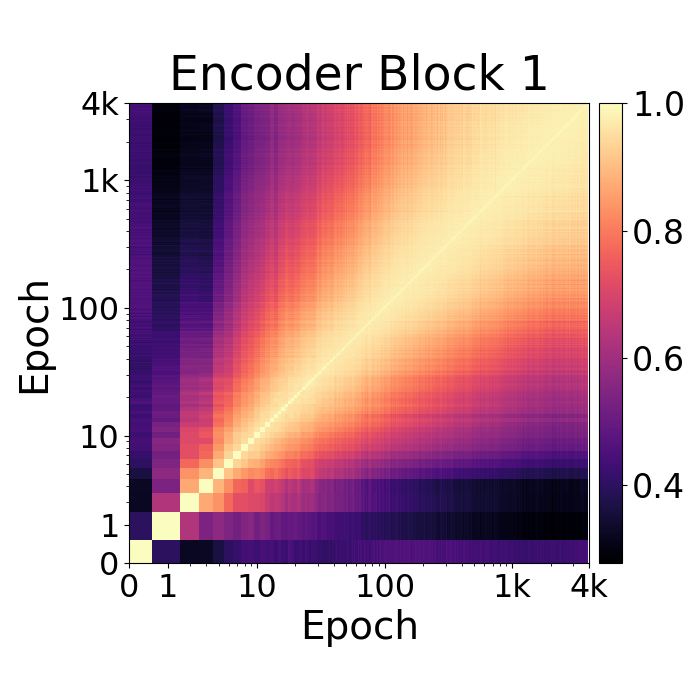}
\label{fig:vit_svhn_Adam_Encoder_block_1_noise_0}}
\subfloat[]{
\includegraphics[width=0.14\textwidth]{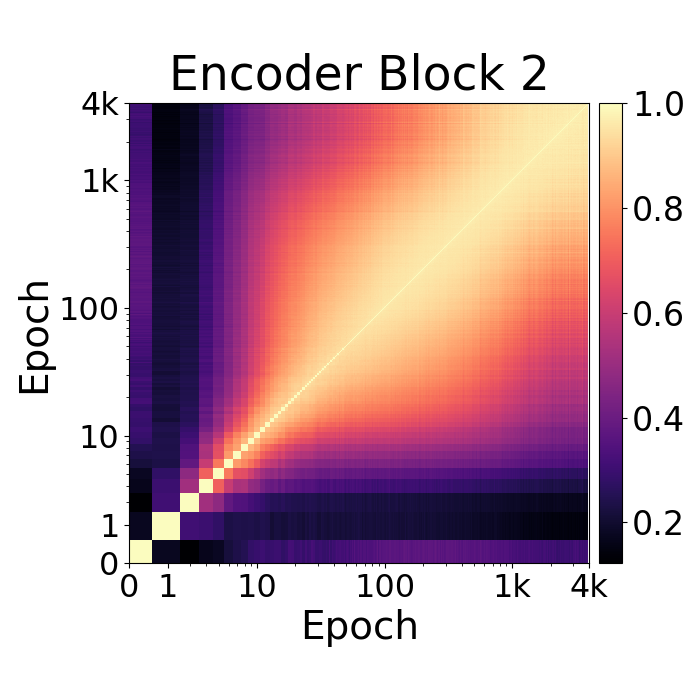}
\label{fig:vit_svhn_Adam_Encoder_block_2}}
\subfloat[]{
\includegraphics[width=0.14\textwidth]{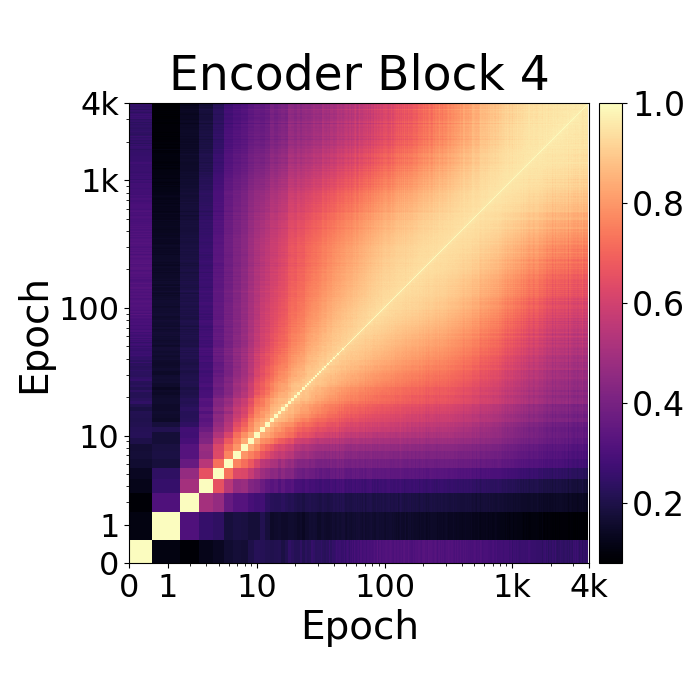}
\label{fig:vit_svhn_Adam_Encoder_block_4_noise_0}}
\subfloat[]{
\includegraphics[width=0.14\textwidth]{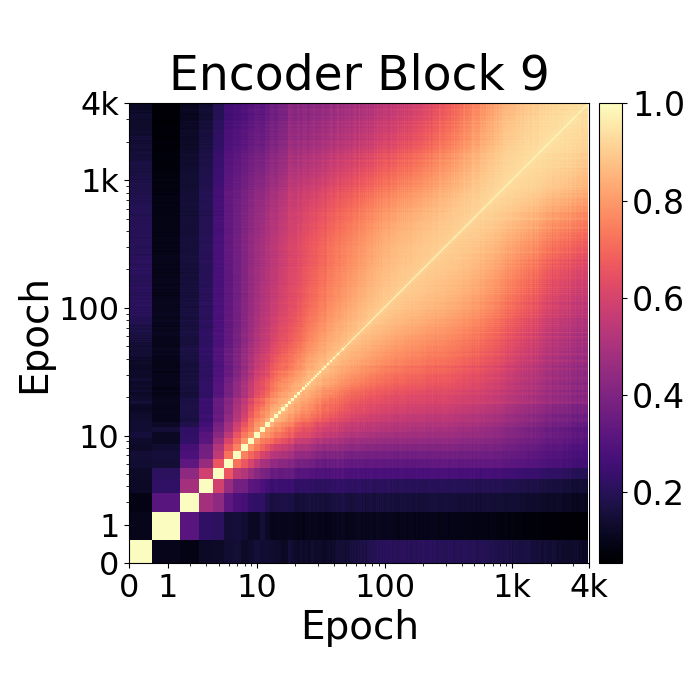}
\label{fig:vit_svhn_Adam_Encoder_block_9_noise_0}}
\subfloat[]{
\includegraphics[width=0.14\textwidth]{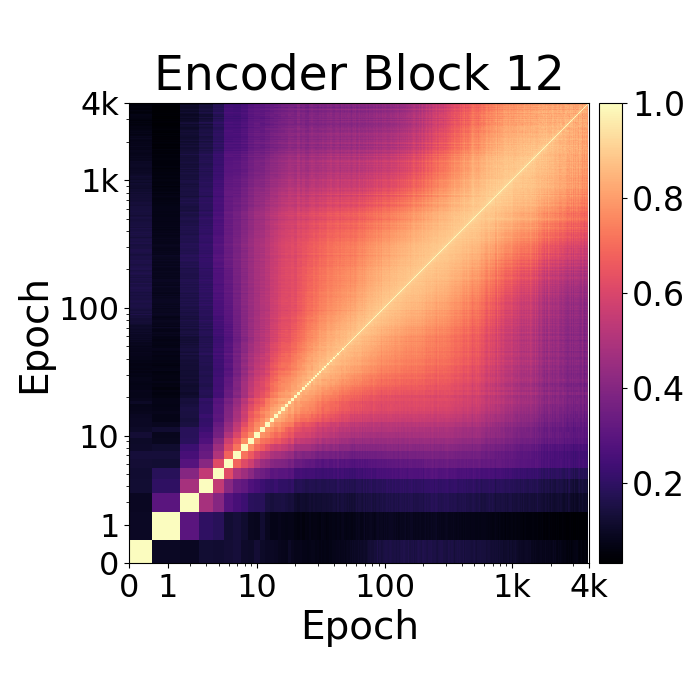}
\label{fig:vit_svhn_Adam_Encoder_block_12_noise_0}}
\subfloat[]{
\includegraphics[width=0.14\textwidth]{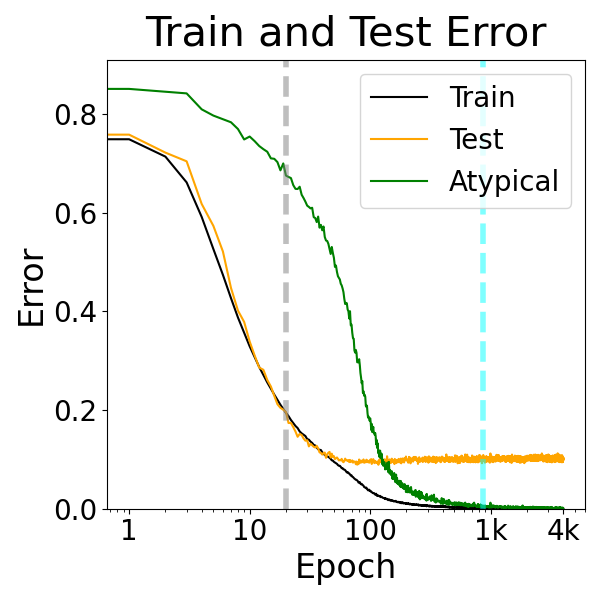}
\label{fig:vit_svhn_Adam_vit_train_test_error_noise_0}}
\caption{CKA evaluations for ViT-B/16, trained on SVHN without label noise, using Adam optimizer with learning rate of 0.0001.}
\label{fig:vit_svhn_Adam_cka_comparison_noise_0}
\end{figure*}

\begin{figure*}[]
\centering
\subfloat[]{
\includegraphics[width=0.14\textwidth]{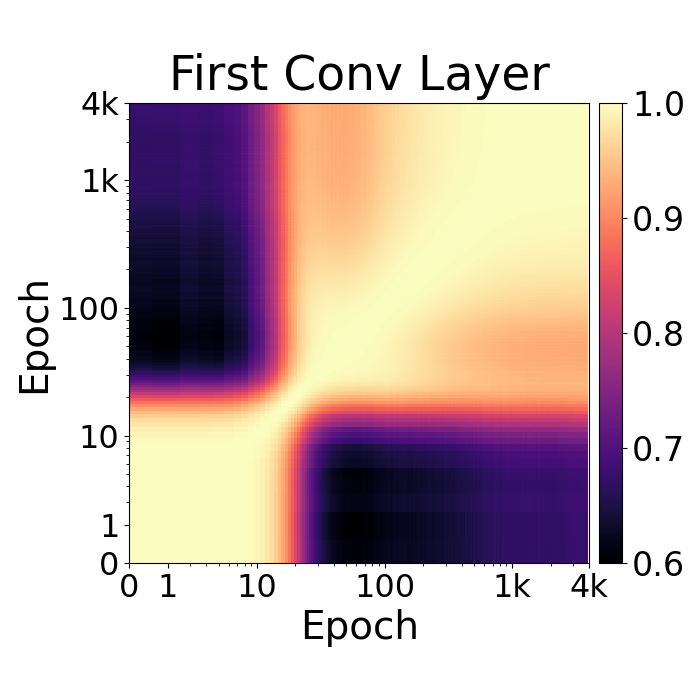}
\label{fig:vit_svhn_sgd_first_conv_layer_noise_0}}
\subfloat[]{
\includegraphics[width=0.14\textwidth]{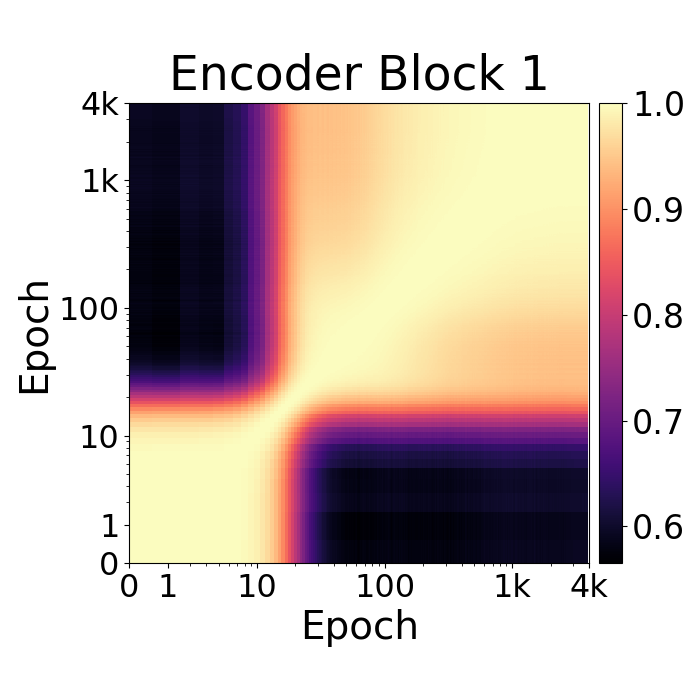}
\label{fig:vit_svhn_sgd_Encoder_block_1_noise_0}}
\subfloat[]{
\includegraphics[width=0.14\textwidth]{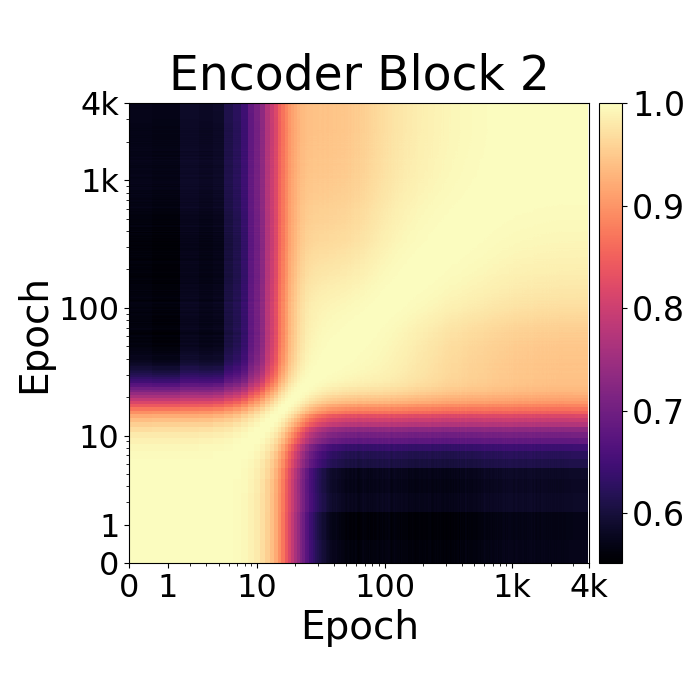}
\label{fig:vit_svhn_sgd_Encoder_block_2}}
\subfloat[]{
\includegraphics[width=0.14\textwidth]{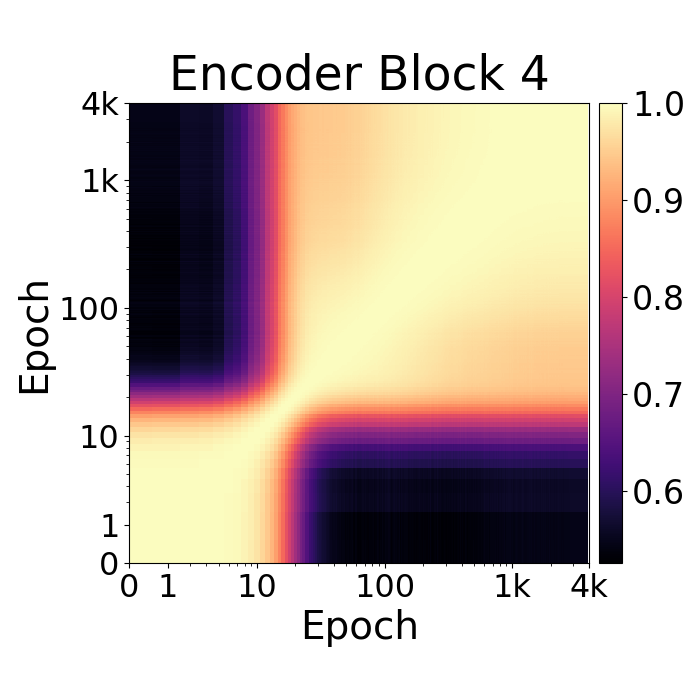}
\label{fig:vit_svhn_sgd_Encoder_block_4_noise_0}}
\subfloat[]{
\includegraphics[width=0.14\textwidth]{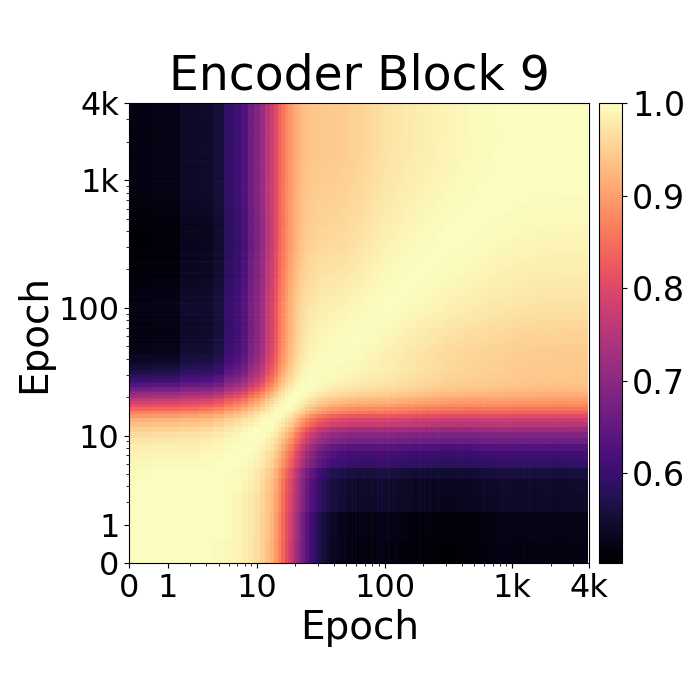}
\label{fig:vit_svhn_sgd_Encoder_block_9_noise_0}}
\subfloat[]{
\includegraphics[width=0.14\textwidth]{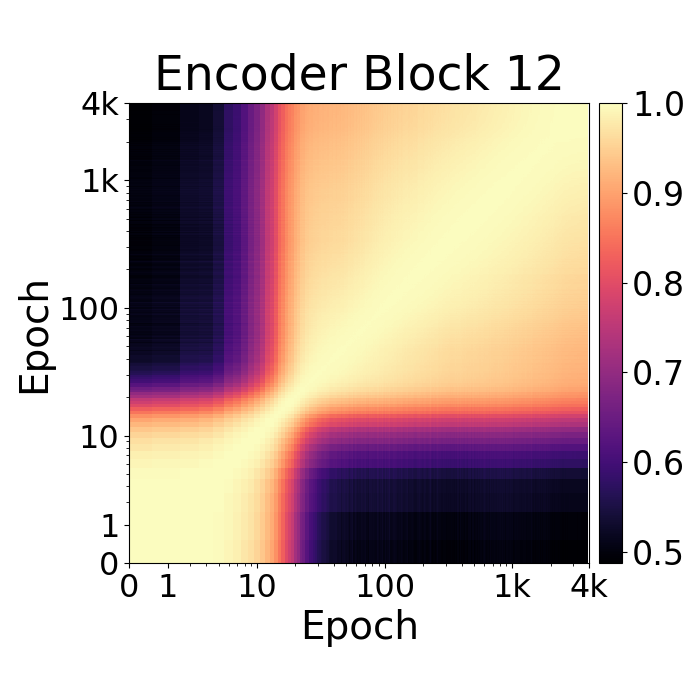}
\label{fig:vit_svhn_sgd_Encoder_block_12_noise_0}}
\subfloat[]{
\includegraphics[width=0.14\textwidth]{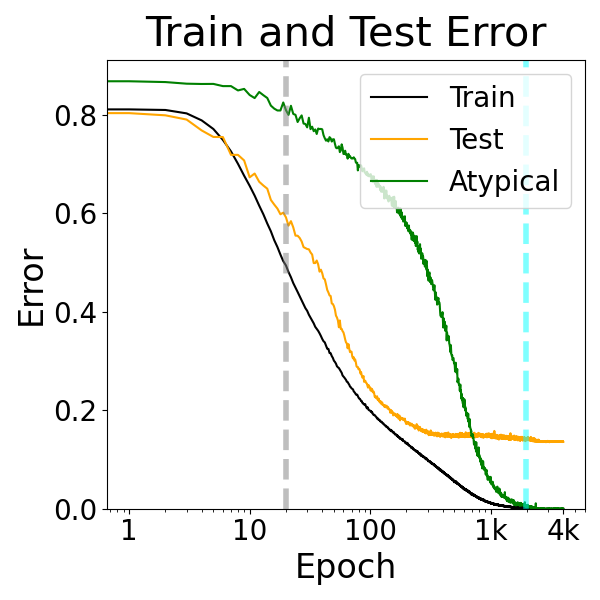}
\label{fig:vit_svhn_sgd_vit_train_test_error_noise_0}}
\caption{CKA evaluations for ViT-B/16, trained on SVHN without label noise, using SGD optimizer with learning rate of 0.001 and without momentum}
\label{fig:vit_svhn_sgd_cka_comparison_noise_0}
\end{figure*}

\subsection{Adam vs.~SGD: Direct Comparison by Representational Similarity}
\label{appendix:sec:Adam vs SGD Direct Comparison by Representational Similarity}
\begin{figure*}[]
  \centering
    \subfloat[]{
    \includegraphics[width=0.14\textwidth]{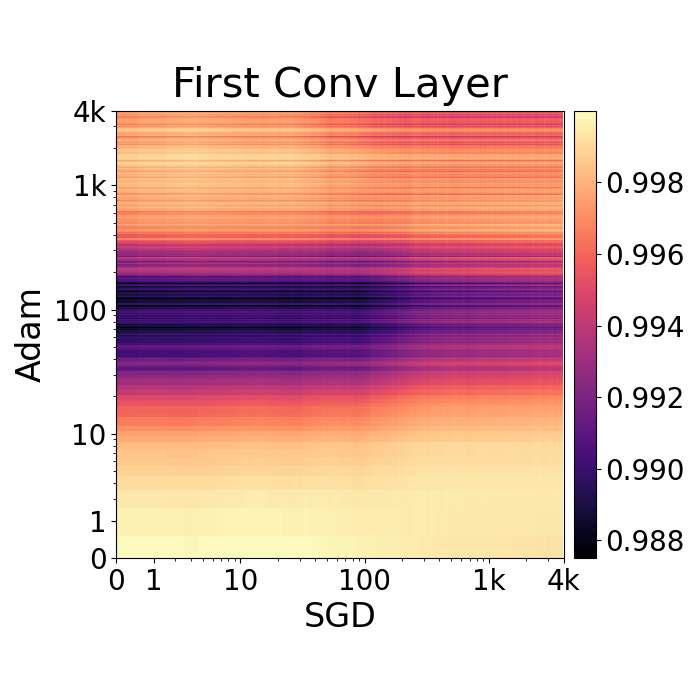}
    \label{fig:Resnet18_cifar10_adam_vs_sgd_conv1}}
    \subfloat[]{
    \includegraphics[width=0.14\textwidth]{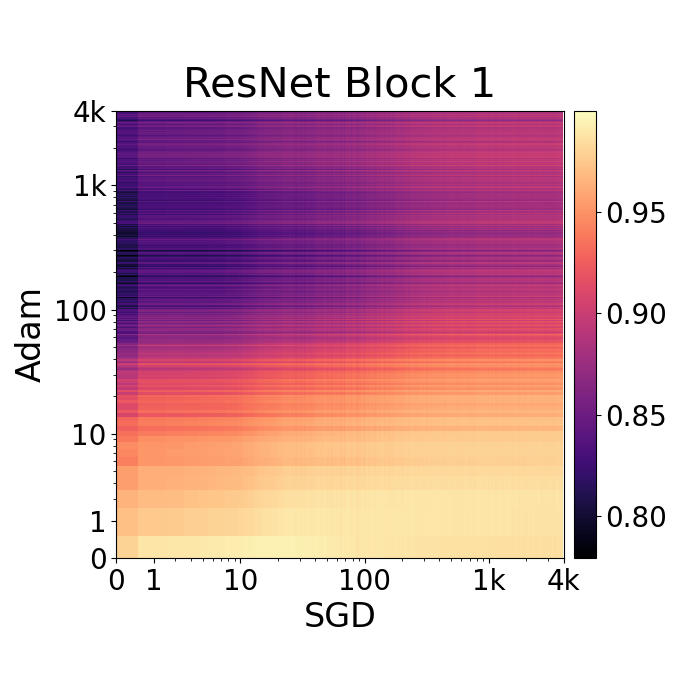}
    \label{fig:Resnet18_cifar10_adam_vs_sgd_block1}
    }
    \subfloat[]{
    \includegraphics[width=0.14\textwidth]{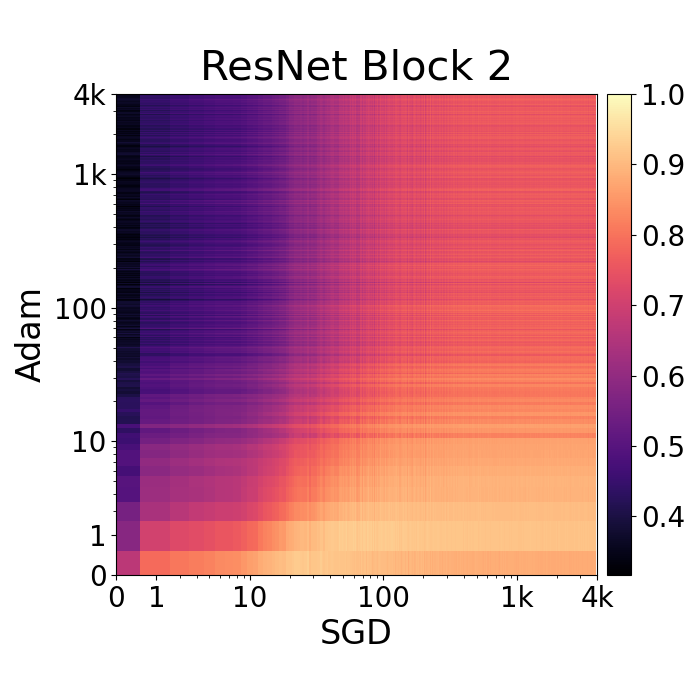}
    \label{fig:Resnet18_cifar10_adam_vs_sgd_block2}}
    \subfloat[]{
    \includegraphics[width=0.14\textwidth]{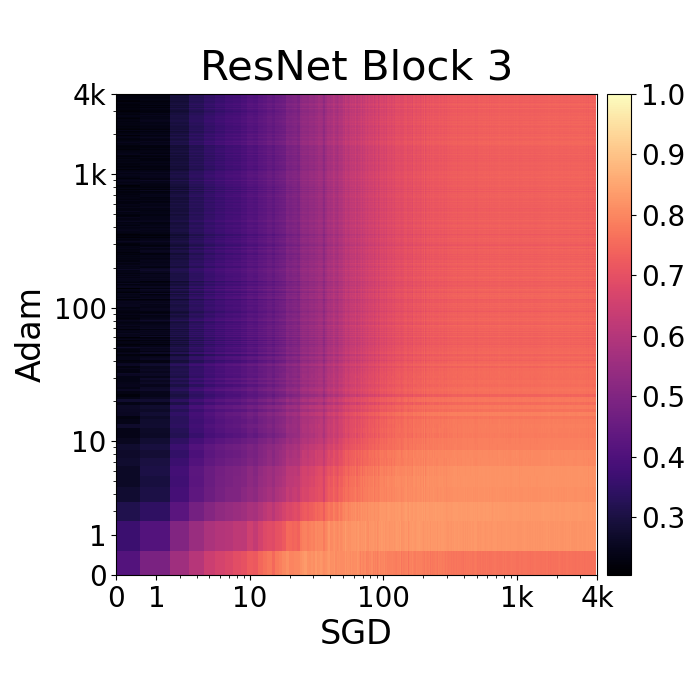}
    \label{fig:Resnet18_cifar10_adam_vs_sgd_block3}}
    \subfloat[]{
    \includegraphics[width=0.14\textwidth]{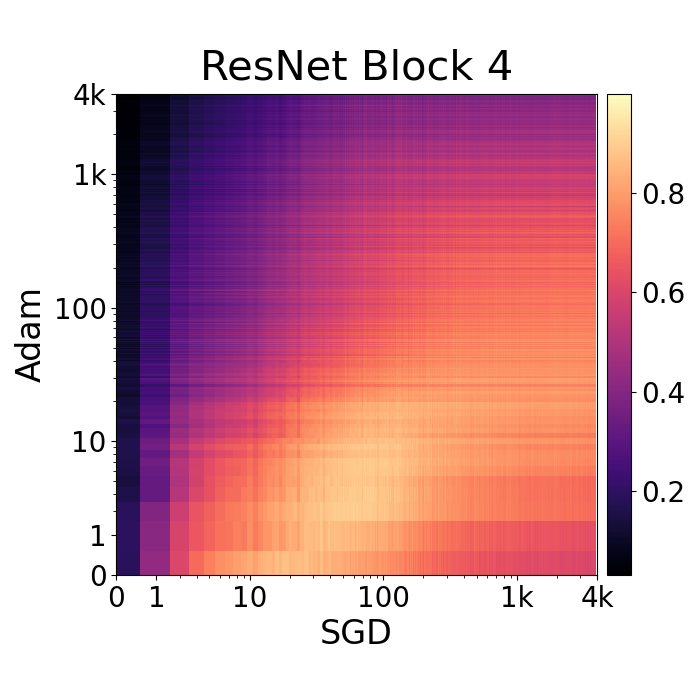}
    \label{fig:Resnet18_cifar10_adam_vs_sgd_block4}}
    \subfloat[]{
    \includegraphics[width=0.14\textwidth]{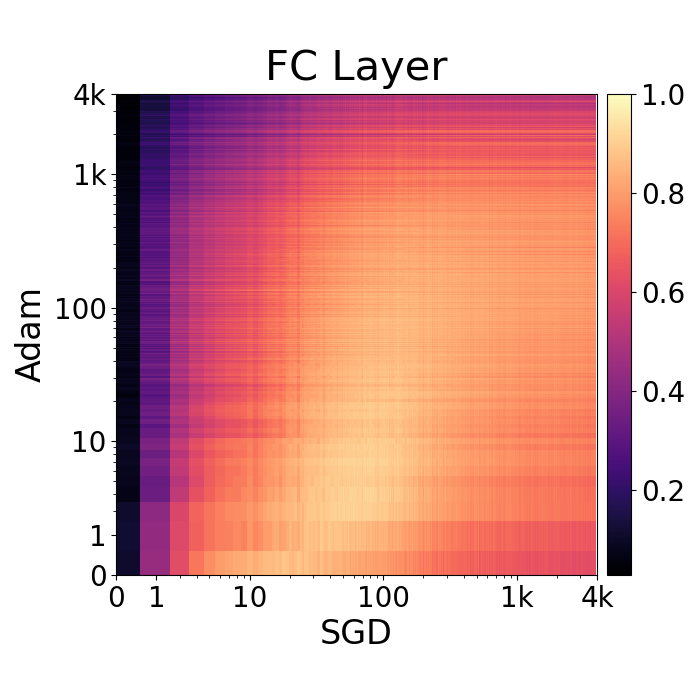}
    \label{fig:Resnet18_cifar10_adam_vs_sgd_fc}}
  \caption{CKA heatmaps of two ResNet-18 models that were separately trained on the CIFAR-10 dataset: One model was trained using Adam optimizer, and the second using SGD.}
  \label{fig:Resnet18_cifar10_adam_vs_sgd}
\end{figure*}

\begin{figure*}[]
  \centering
    \subfloat[]{
    \includegraphics[width=0.14\textwidth]{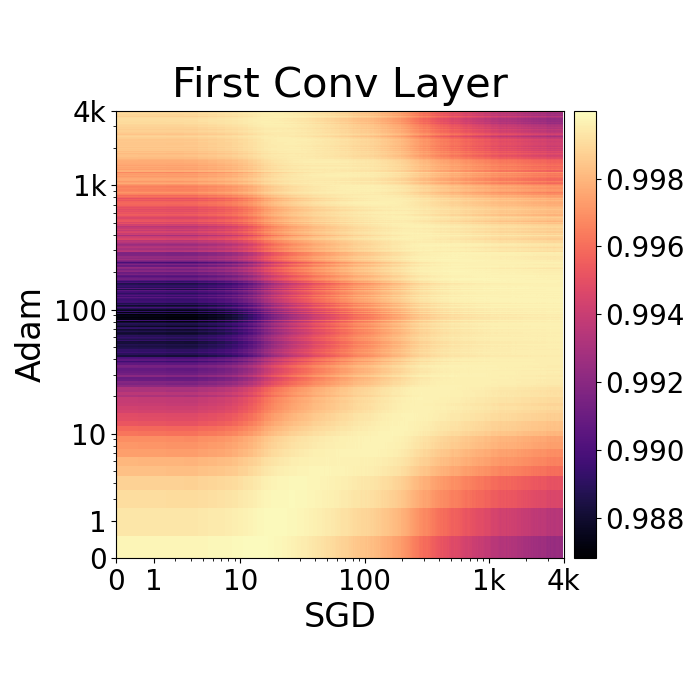}
    \label{fig:Resnet18_svhn_adam_vs_sgd_conv1}}
    \subfloat[]{
    \includegraphics[width=0.14\textwidth]{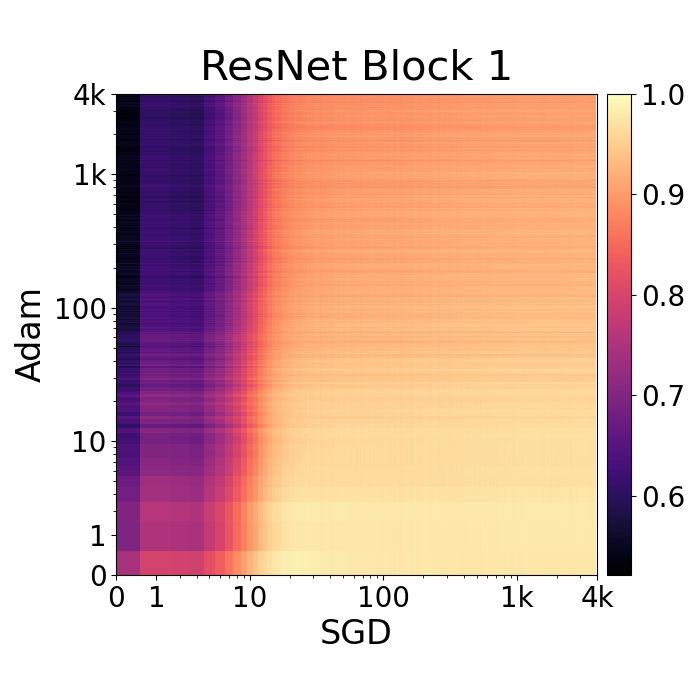}
    \label{fig:Resnet18_svhn_adam_vs_sgd_block1}
    }
    \subfloat[]{
    \includegraphics[width=0.14\textwidth]{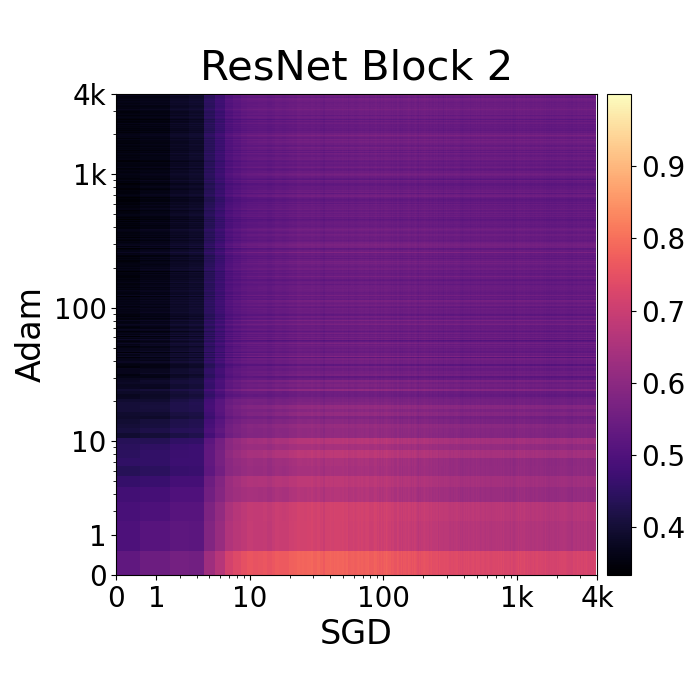}
    \label{fig:Resnet18_svhn_adam_vs_sgd_block2}}
    \subfloat[]{
    \includegraphics[width=0.14\textwidth]{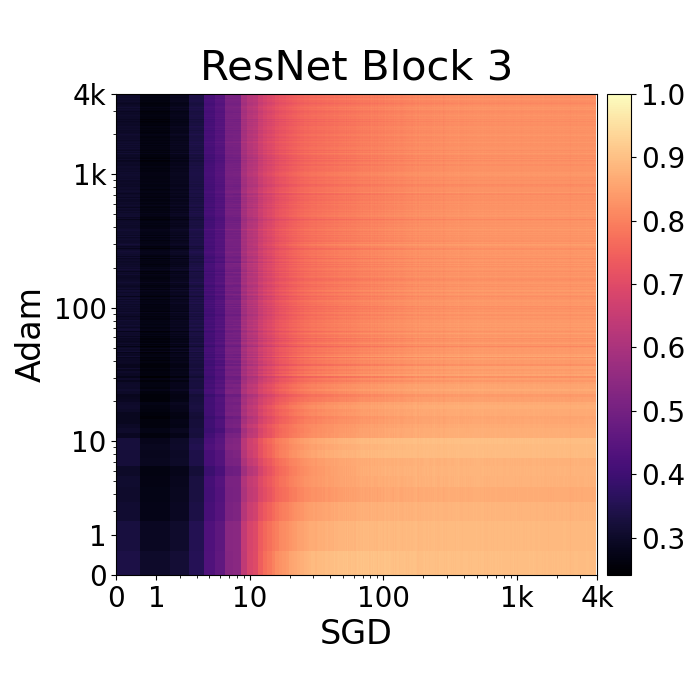}
    \label{fig:Resnet18_svhn_adam_vs_sgd_block3}}
    \subfloat[]{
    \includegraphics[width=0.14\textwidth]{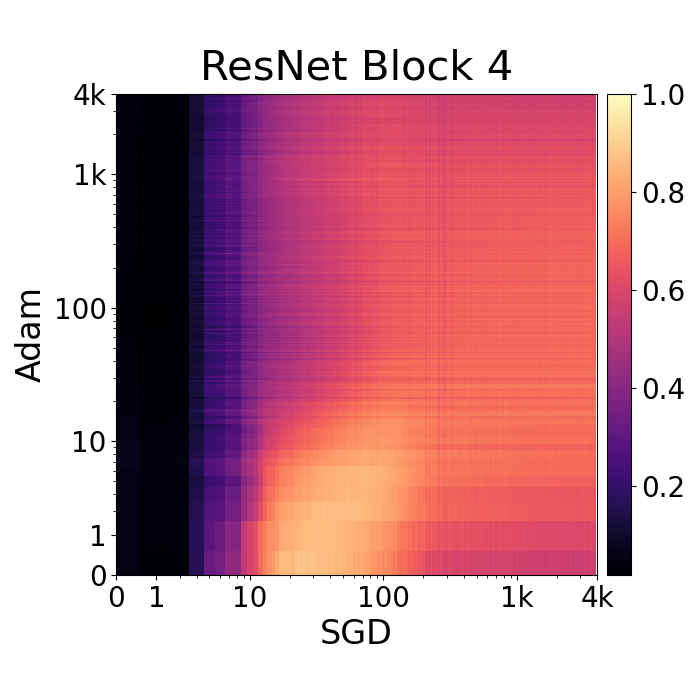}
    \label{fig:Resnet18_svhn_adam_vs_sgd_block4}}
    \subfloat[]{
    \includegraphics[width=0.14\textwidth]{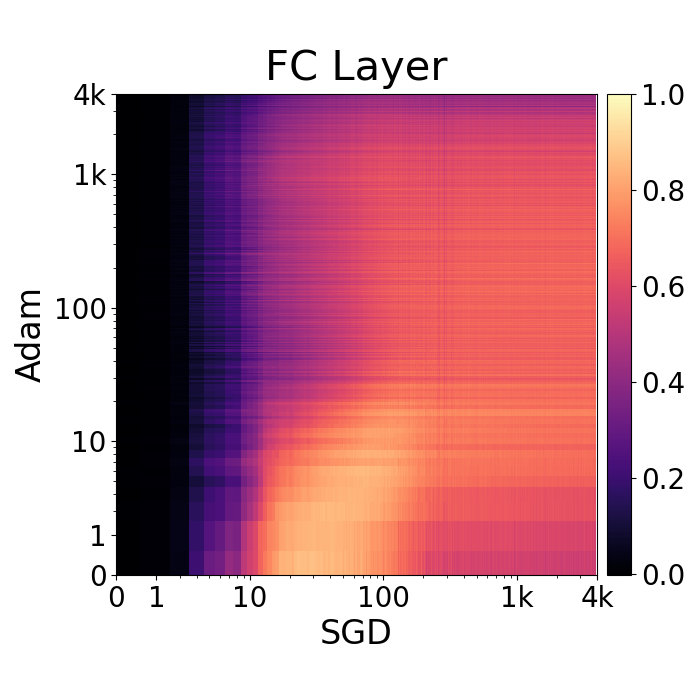}
    \label{fig:Resnet18_svhn_adam_vs_sgd_fc}}
  \caption{CKA heatmaps of two ResNet-18 models that were separately trained on the SVHN dataset: One model was trained using Adam optimizer, and the second using SGD.}
  \label{fig:Resnet18_svhn_adam_vs_sgd}
\end{figure*}

\begin{figure*}[]
  \centering
    \subfloat[]{
    \includegraphics[width=0.14\textwidth]{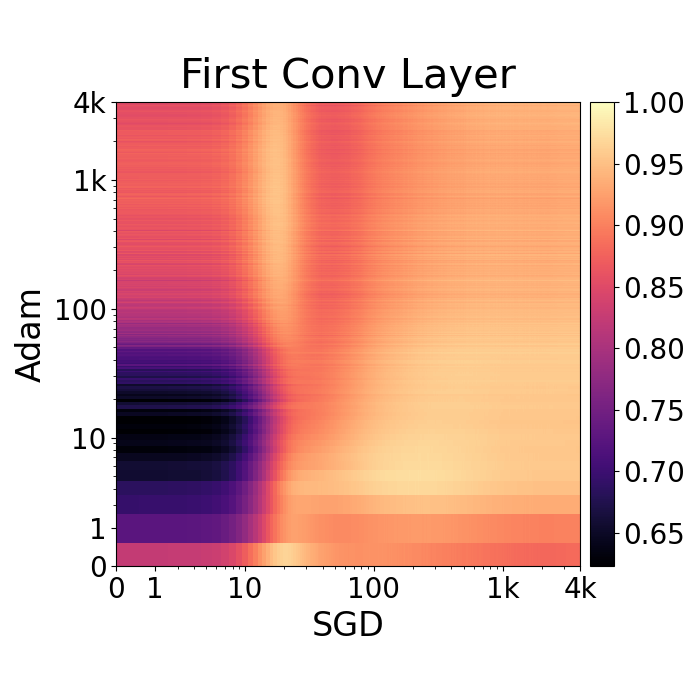}
    \label{fig:vit_b_16_cifar10_adam_vs_sgd_conv1}}
    \subfloat[]{
    \includegraphics[width=0.14\textwidth]{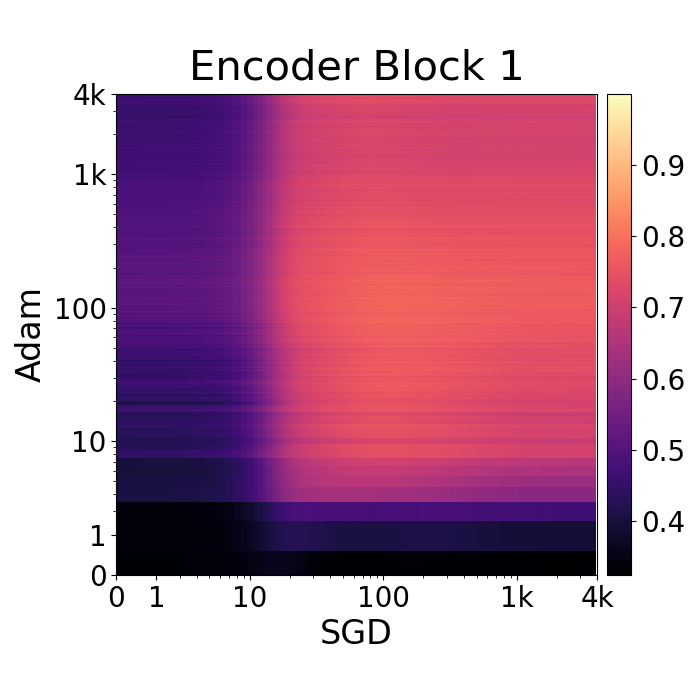}
    \label{fig:vit_b_16_cifar10_adam_vs_sgd_block1}
    }
    \subfloat[]{
    \includegraphics[width=0.14\textwidth]{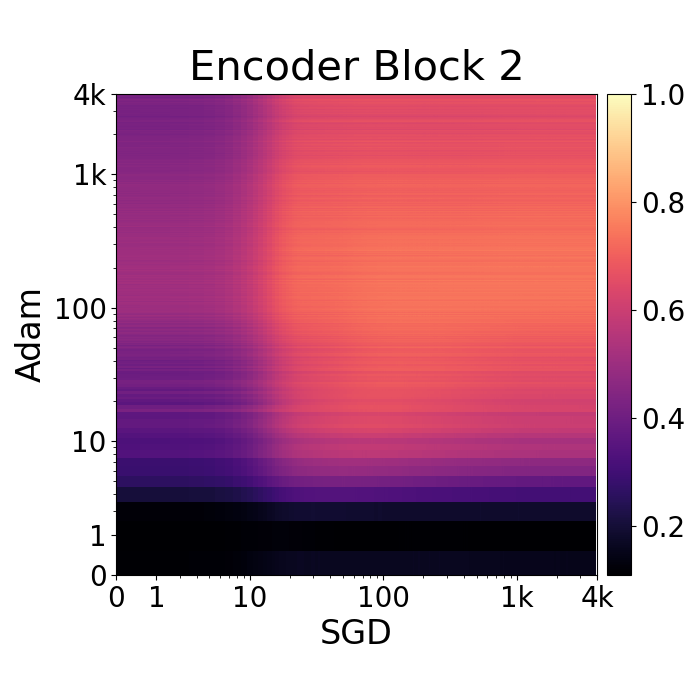}
    \label{fig:vit_b_16_cifar10_adam_vs_sgd_block2}}
    \subfloat[]{
    \includegraphics[width=0.14\textwidth]{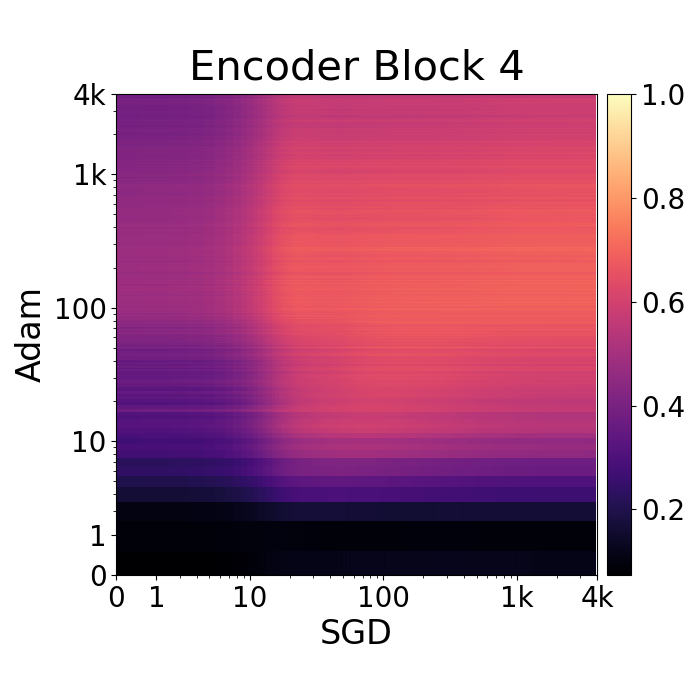}
    \label{fig:vit_b_16_cifar10_adam_vs_sgd_block4}}
    \subfloat[]{
    \includegraphics[width=0.14\textwidth]{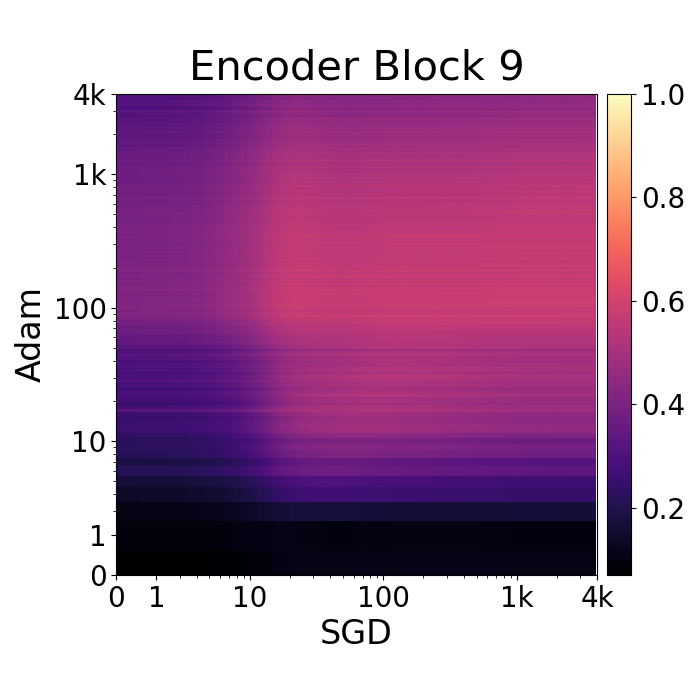}
    \label{fig:vit_b_16_cifar10_adam_vs_sgd_block9}}
    \subfloat[]{
    \includegraphics[width=0.14\textwidth]{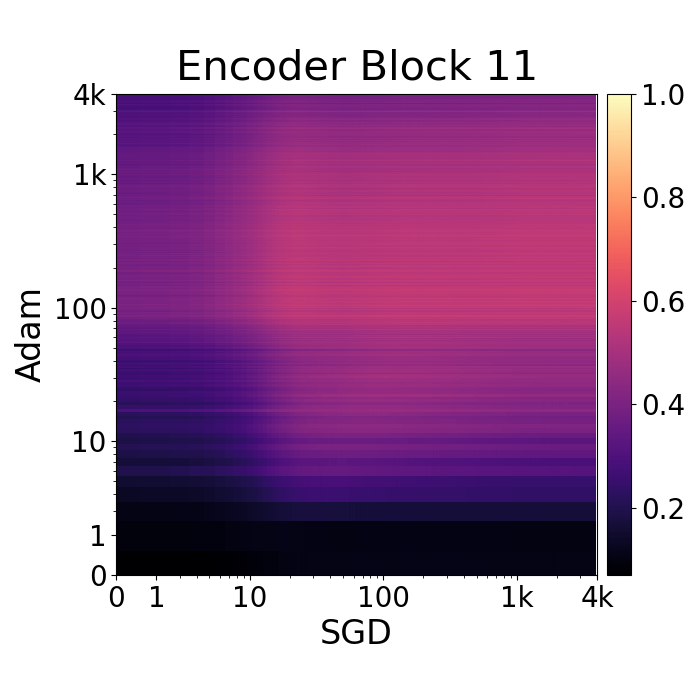}
    \label{fig:vit_b_16_cifar10_adam_vs_sgd_block11}}
  \caption{CKA heatmaps of two ViT-B/16 models that were separately trained on the CIFAR-10 dataset: One model was trained using Adam optimizer, and the second using SGD.}
  \label{fig:vit_b_16_cifar10_adam_vs_sgd}
\end{figure*}

\begin{figure*}[]
  \centering
    \subfloat[]{
    \includegraphics[width=0.14\textwidth]{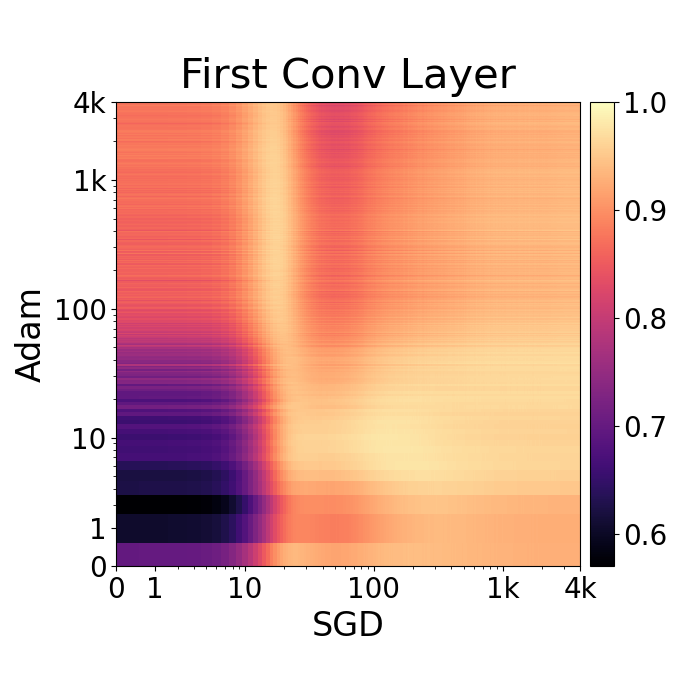}
    \label{fig:vit_b_16_svhn_adam_vs_sgd_conv1}}
    \subfloat[]{
    \includegraphics[width=0.14\textwidth]{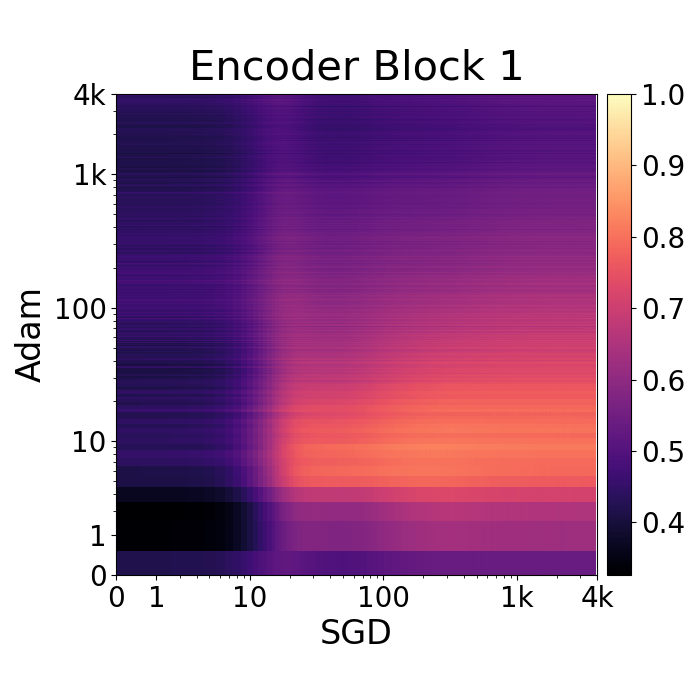}
    \label{fig:vit_b_16_svhn_adam_vs_sgd_block1}
    }
    \subfloat[]{
    \includegraphics[width=0.14\textwidth]{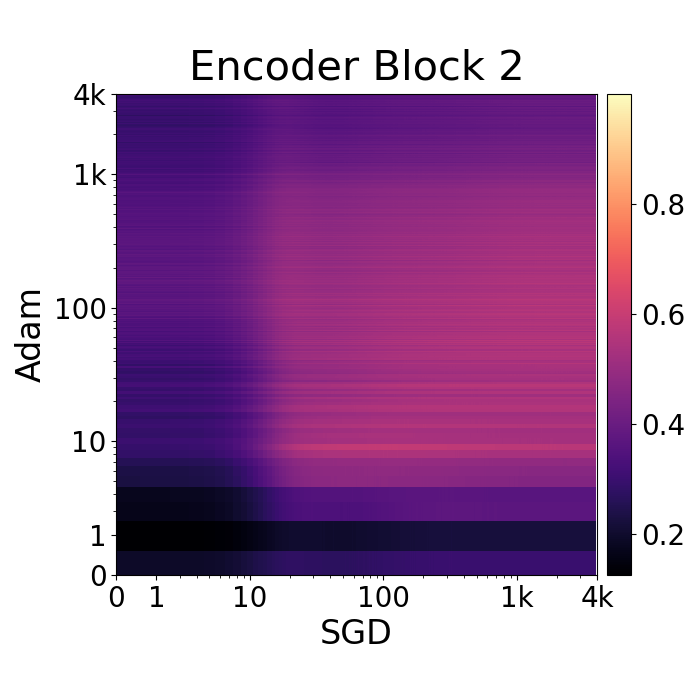}
    \label{fig:vit_b_16_svhn_adam_vs_sgd_block2}}
    \subfloat[]{
    \includegraphics[width=0.14\textwidth]{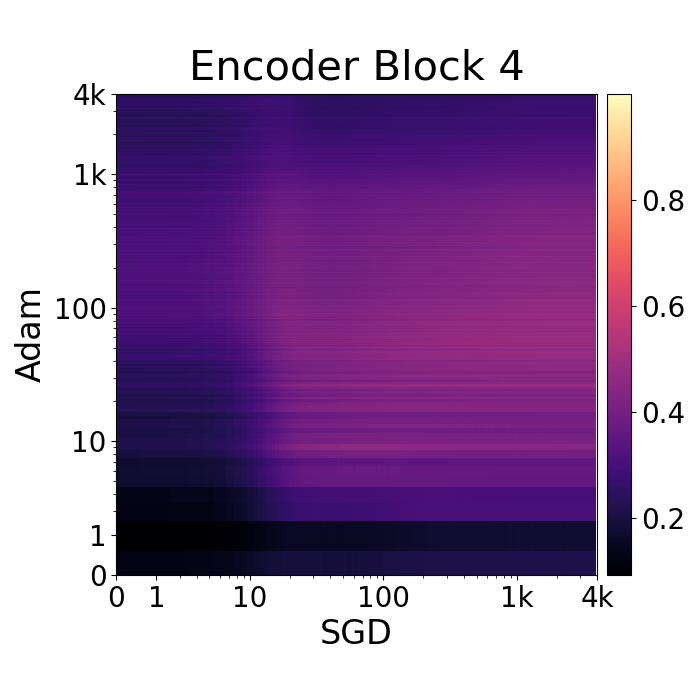}
    \label{fig:vit_b_16_svhn_adam_vs_sgd_block4}}
    \subfloat[]{
    \includegraphics[width=0.14\textwidth]{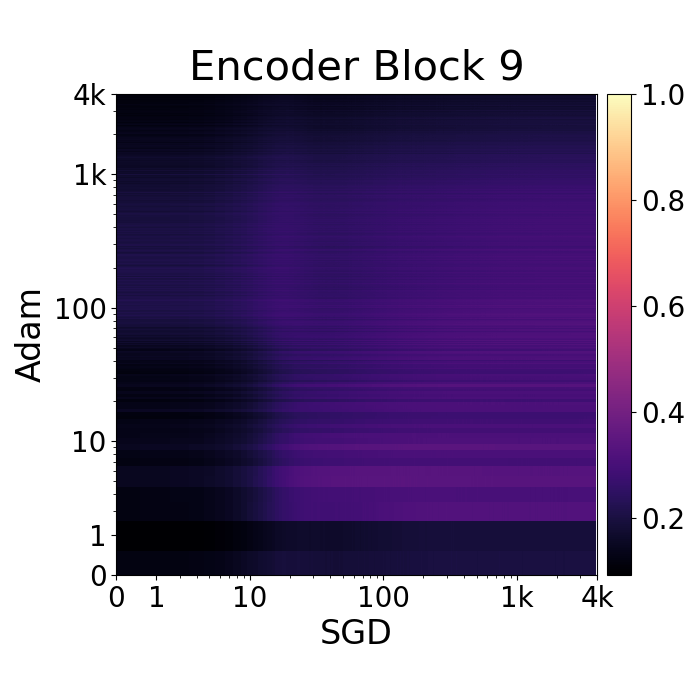}
    \label{fig:vit_b_16_svhn_adam_vs_sgd_block9}}
    \subfloat[]{
    \includegraphics[width=0.14\textwidth]{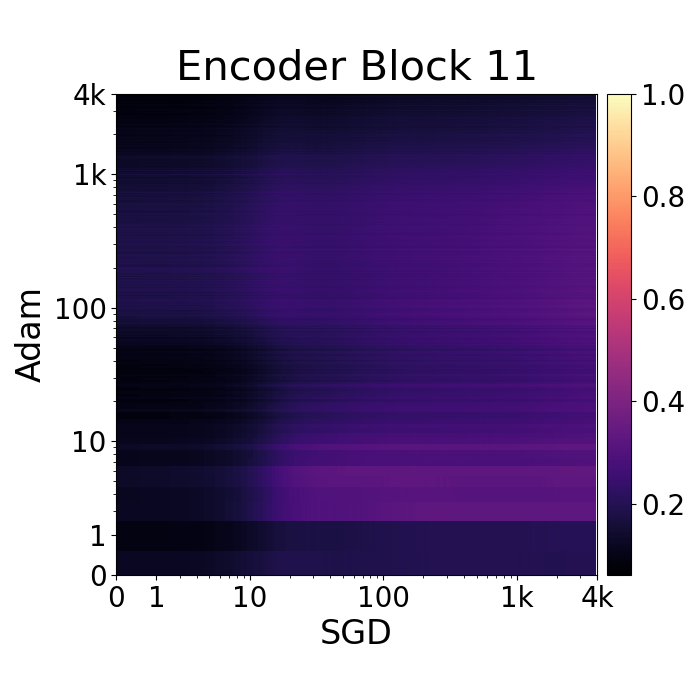}
    \label{fig:vit_b_16_svhn_adam_vs_sgd_block11}}
  \caption{CKA heatmaps of two ViT-B/16 models that were separately trained on the SVHN dataset: One model was trained using Adam optimizer, and the second using SGD.}
  \label{fig:vit_b_16_svhn_adam_vs_sgd}
\end{figure*}

In Figs.~\ref{fig:Resnet18_cifar10_adam_vs_sgd}, \ref{fig:Resnet18_svhn_adam_vs_sgd}, \ref{fig:vit_b_16_svhn_adam_vs_sgd}, \ref{fig:vit_b_16_cifar10_adam_vs_sgd}, we show representational similarity diagrams that compare pairs of representations from two different trainings of the same model -- one with Adam and one with SGD. This is in contrast to all the other analyses in this paper, where we evaluate similarity of representations at two different epochs in the same training process (hence for the same optimizer). Specifically, each of the 2D diagrams in Figs.~\ref{fig:Resnet18_cifar10_adam_vs_sgd}, \ref{fig:Resnet18_svhn_adam_vs_sgd}, \ref{fig:vit_b_16_svhn_adam_vs_sgd}, \ref{fig:vit_b_16_cifar10_adam_vs_sgd} is for a specific layer in the DNN and can be read according to its rows. Each row is associated with a different epoch in the Adam training to consider its representation as a reference to evaluate the similarity for the entire SGD training process, which corresponds to the horizontal axis of the diagram.

Figs.~\ref{fig:Resnet18_cifar10_adam_vs_sgd}, \ref{fig:Resnet18_svhn_adam_vs_sgd}, \ref{fig:vit_b_16_svhn_adam_vs_sgd}, \ref{fig:vit_b_16_cifar10_adam_vs_sgd} suggest that it is intricate to obtain significant insights from a direct comparison of the Adam and SGD representations, and therefore the qualitative comparison of the separate behaviors of Adam and SGD as we do in the rest of this paper is a more insightful approach.

\subsection{Additional DRS Evaluations}
\begin{figure*}[]
\centering
\subfloat[]{
\includegraphics[width=0.19\textwidth]{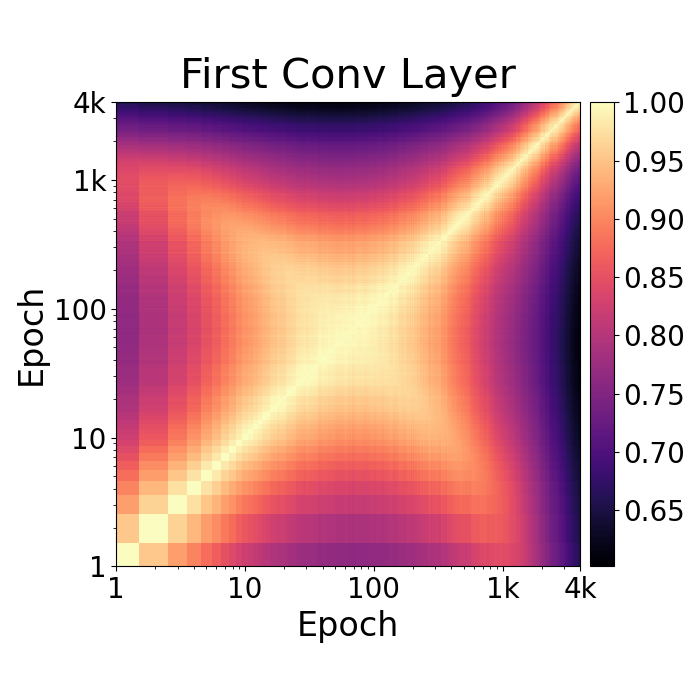}
\label{lp_resnet_noise_20_first_conv_layer}}
\subfloat[]{
\includegraphics[width=0.19\textwidth]{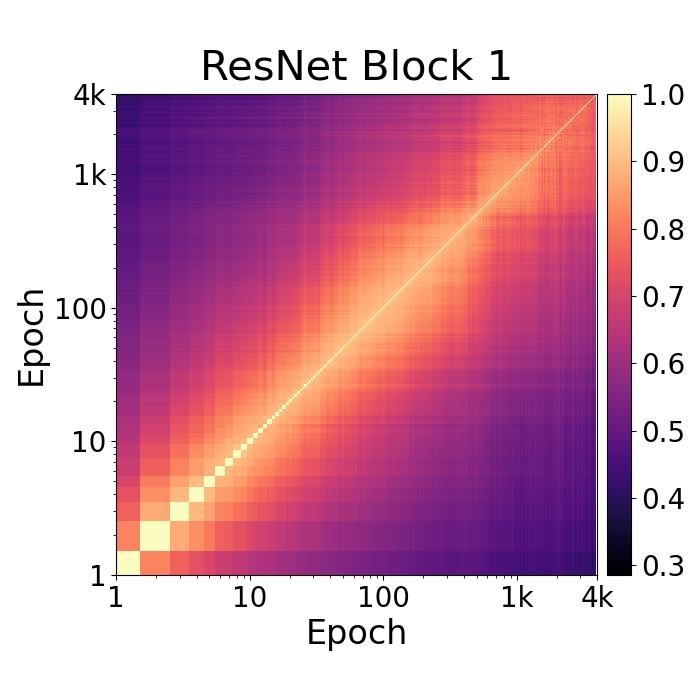}
\label{lp_resnet_noise_20_block_1}}
\subfloat[]{
\includegraphics[width=0.19\textwidth]{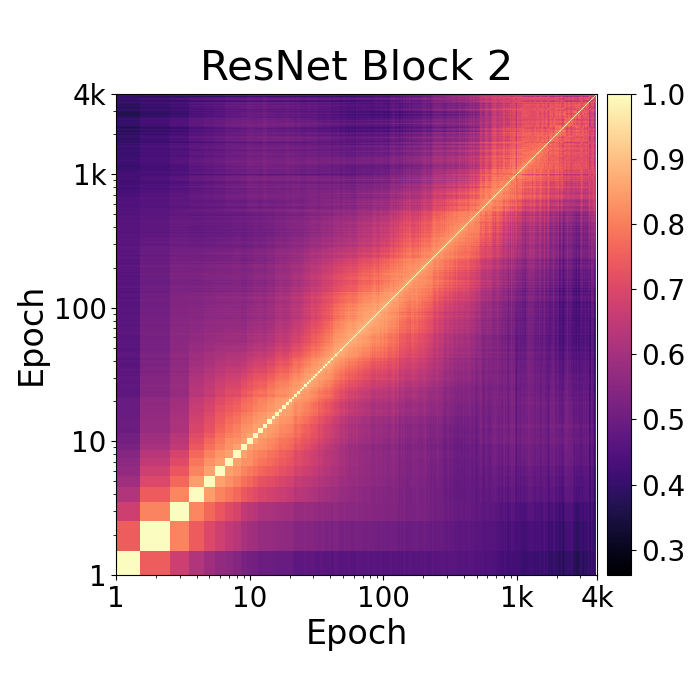}
\label{lp_resnet_noise_20_block_2}}
\subfloat[]{
\includegraphics[width=0.19\textwidth]{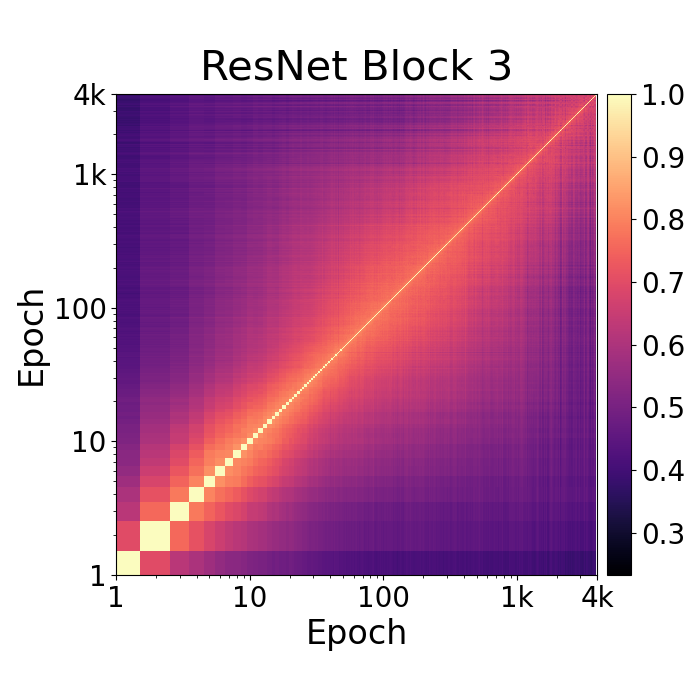}
\label{lp_resnet_noise_20_block_3}}
\subfloat[]{
\includegraphics[width=0.19\textwidth]{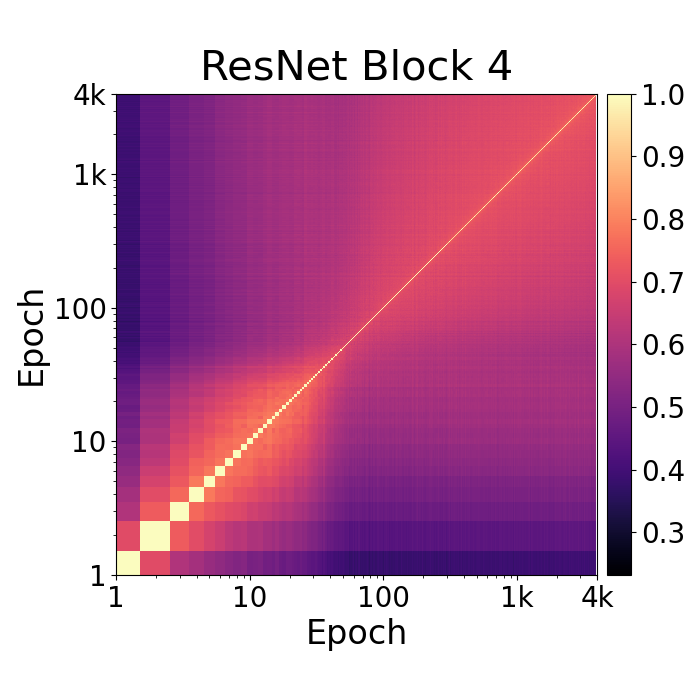}
\label{lp_resnet_noise_20_block_4}}
\caption{Decision regions similarity (DRS) of linear classifier probes. Evaluations for ResNet-18 trained on CIFAR-10 \textbf{with 20\% label noise}. Each subfigure shows the DRS of a specific layer in the ResNet-18 during its training. The test and error curves are provided in Fig.~\ref{ResNet18_k_64_cifar10_noise_20_error_curve}.}
\label{fig:lp_resnet_noise_20}
\end{figure*}

\begin{figure*}[]
\centering
\subfloat[]{
\includegraphics[width=0.24\textwidth]{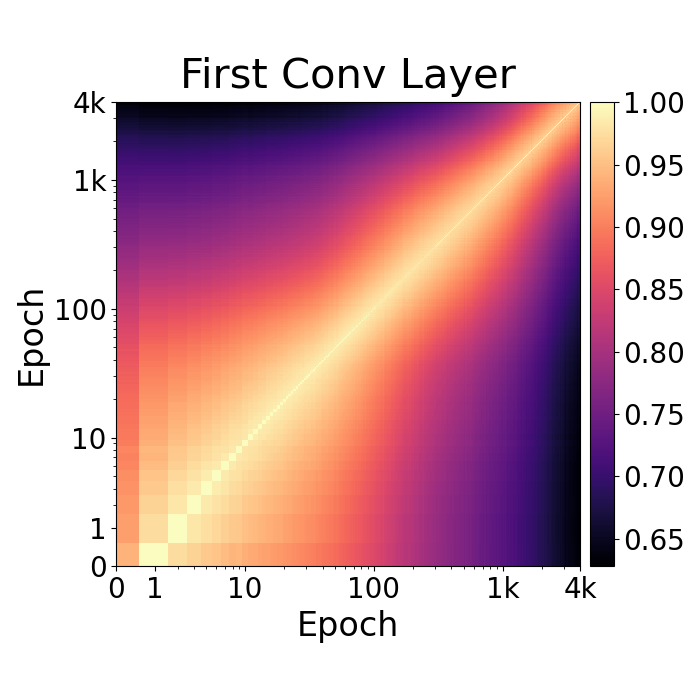}
\label{DRS_cifar10_vit_noise_0_vit_first_conv_layer}}
\subfloat[]{
\includegraphics[width=0.24\textwidth]{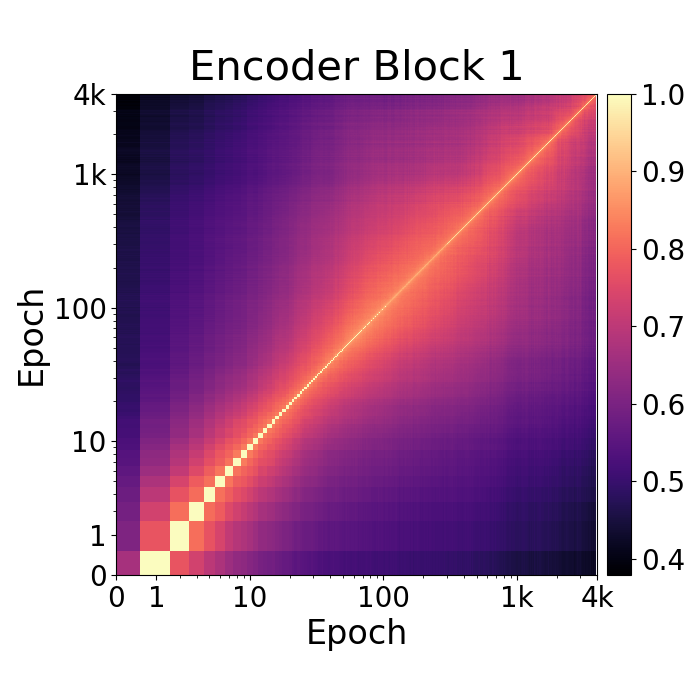}
\label{DRS_cifar10_vit_noise_0_Encoder_block_1}}
\subfloat[]{
\includegraphics[width=0.24\textwidth]{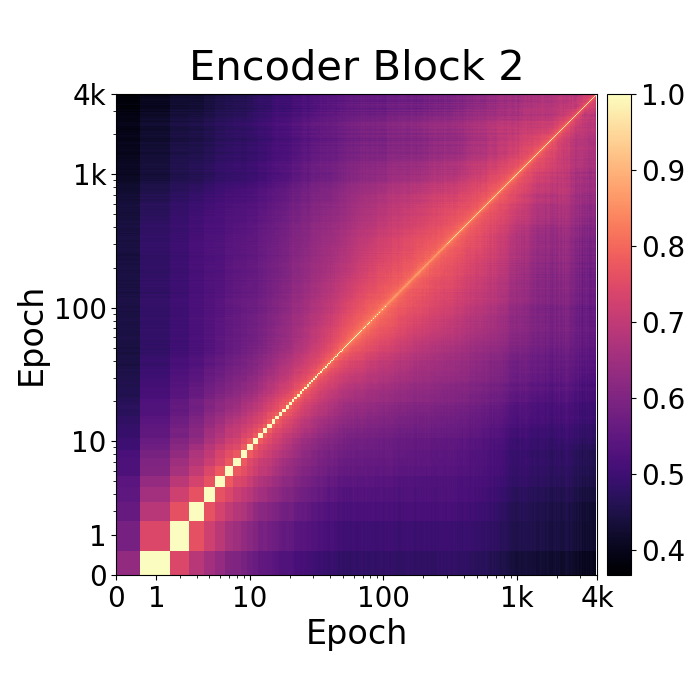}
\label{DRS_cifar10_vit_noise_0_Encoder_block_2}}
\subfloat[]{
\includegraphics[width=0.24\textwidth]{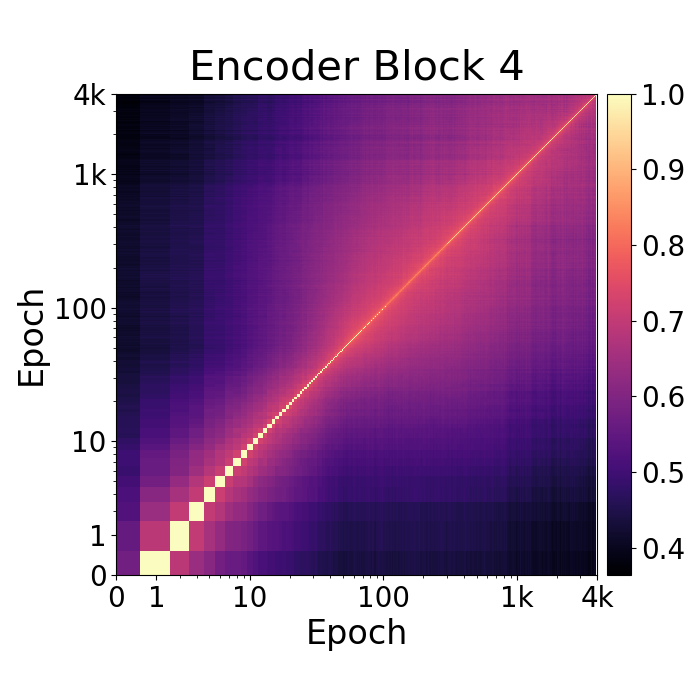}
\label{DRS_cifar10_vit_noise_0_Encoder_block_4}}
\\[-3ex]
\subfloat[]{
\includegraphics[width=0.24\textwidth]{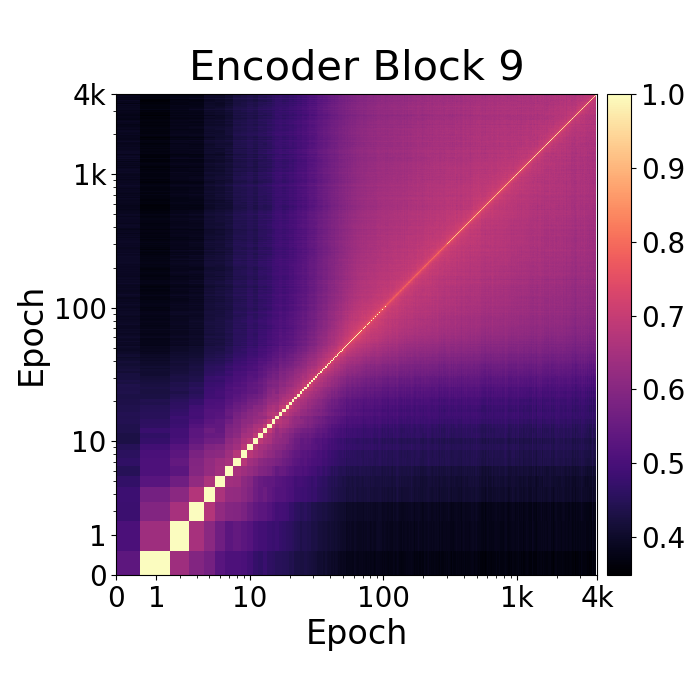}
\label{DRS_cifar10_vit_noise_0_Encoder_block_9}}
\subfloat[]{
\includegraphics[width=0.24\textwidth]{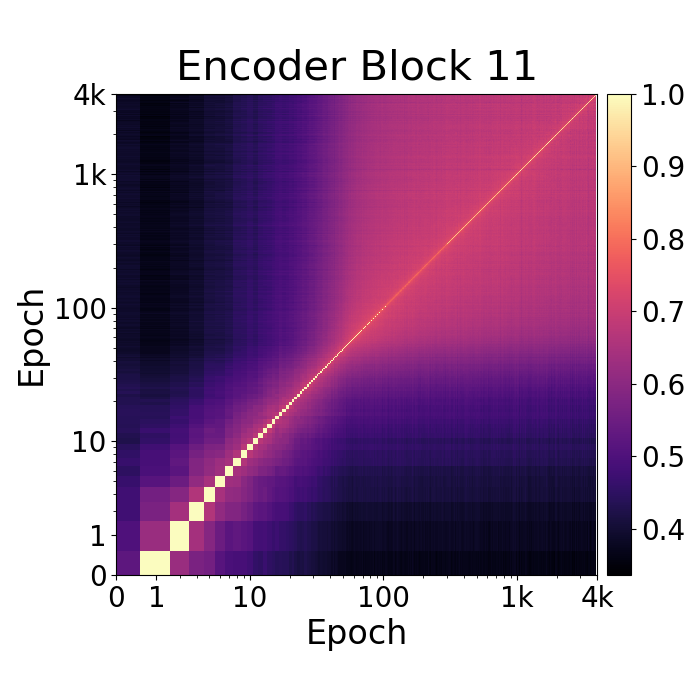}
\label{DRS_cifar10_vit_noise_0_Encoder_block_11}}
\caption{Decision regions similarity (DRS) of linear classifier probes. Evaluations for ViT-B/16, trained on CIFAR-10 without label noise. Each subfigure shows the DRS of a specific layer in the ViT-b/16 during its training. The DNN error curves are provided in Fig.~\ref{vit_train_test_error_noise_0}.}
\label{fig:DRS_cifar10_vit_noise_0}
\end{figure*}

\begin{figure*}[]
  \centering
  \subfloat[]{
\includegraphics[width=0.19\textwidth]{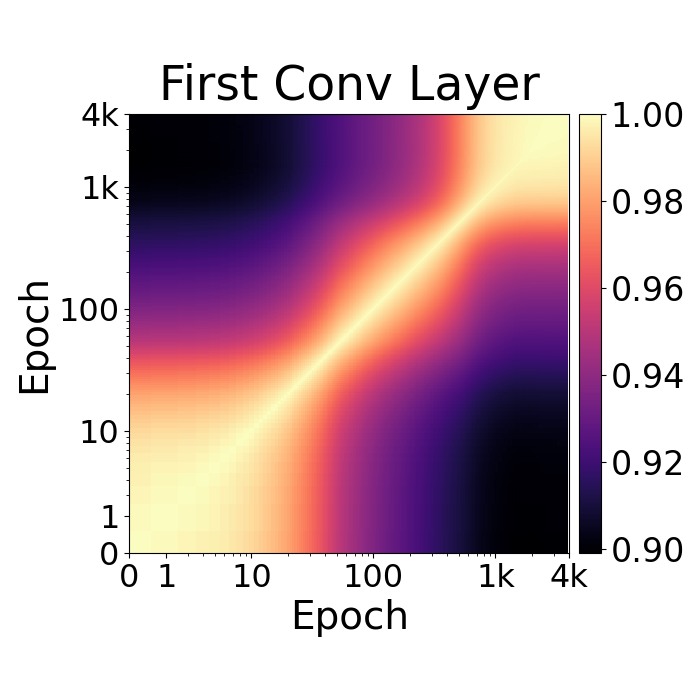}
\label{}}
\subfloat[]{
\includegraphics[width=0.19\textwidth]{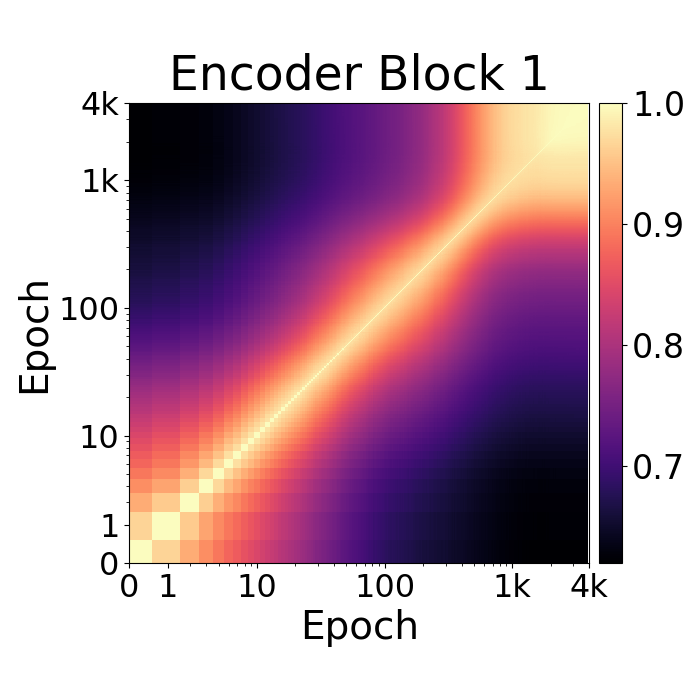}
\label{}}
\subfloat[]{
\includegraphics[width=0.19\textwidth]{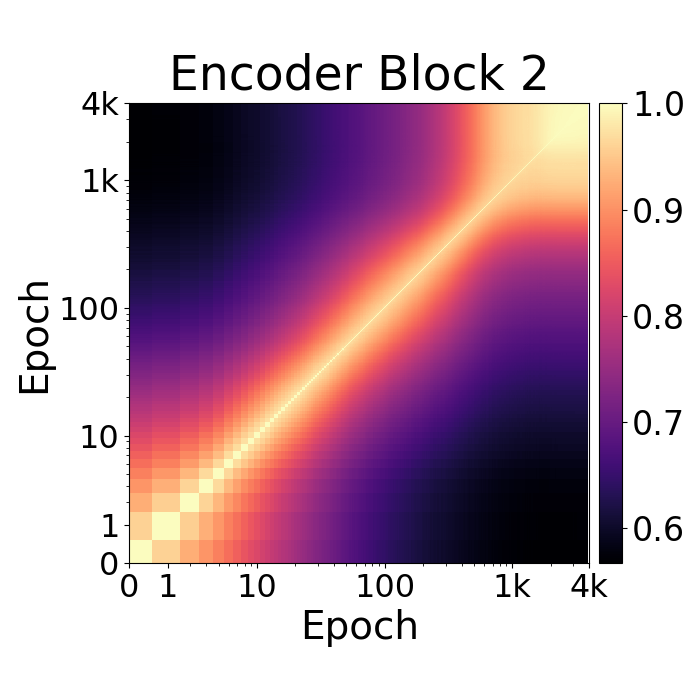}
\label{}}
\subfloat[]{
\includegraphics[width=0.19\textwidth]{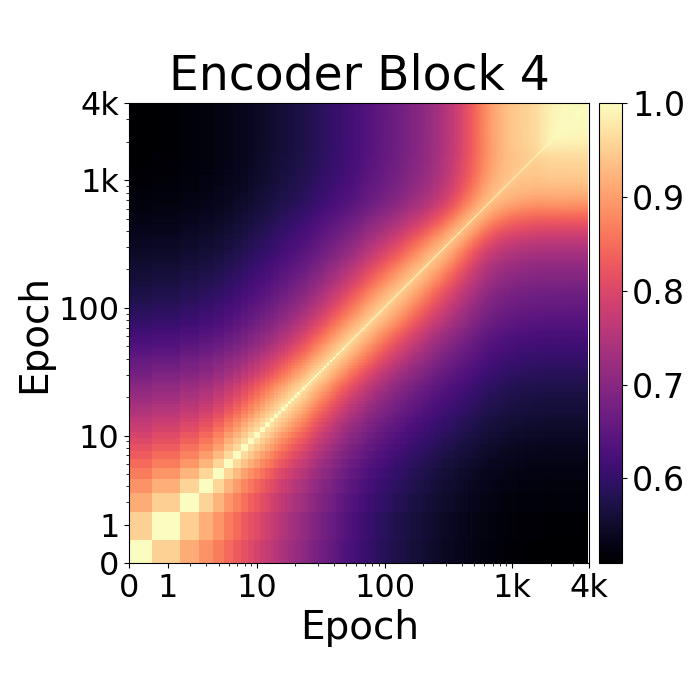}
\label{}}
\subfloat[]{
\includegraphics[width=0.19\textwidth]{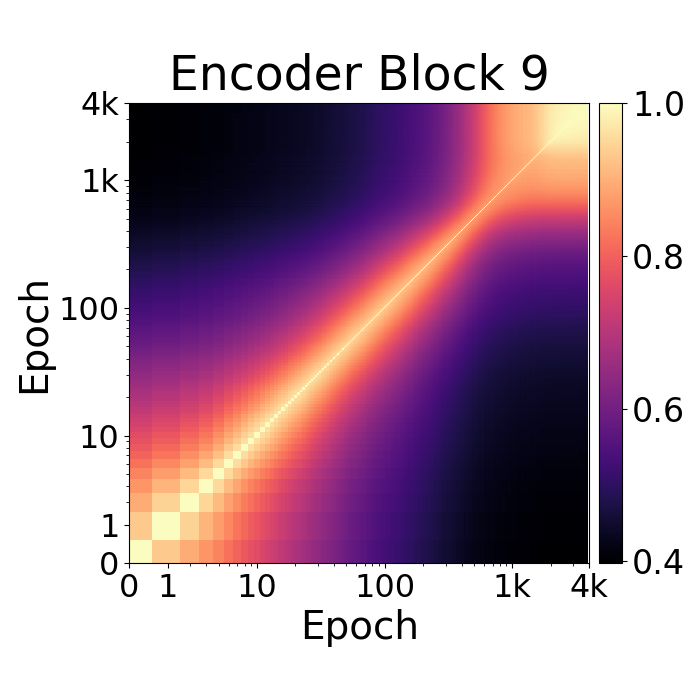}
\label{}}
\caption{Decision regions similarity (DRS) of linear classifier probes. Evaluations for ViT-B/16, trained on CIFAR-10 without label noise using SGD Optimizer with learning rate of 0.001. Each subfigure shows the DRS of a specific layer in the ViT-b/16 during its training. The DNN error curves are provided in Fig.~\ref{fig:vit_cifar10_sgd_vit_train_test_error_noise_0}.}  \label{fig:DRS_cifar10_vit_noise_0_sgd}
\end{figure*}

\begin{figure*}[]
  \centering
  \subfloat[]{
\includegraphics[width=0.19\textwidth]{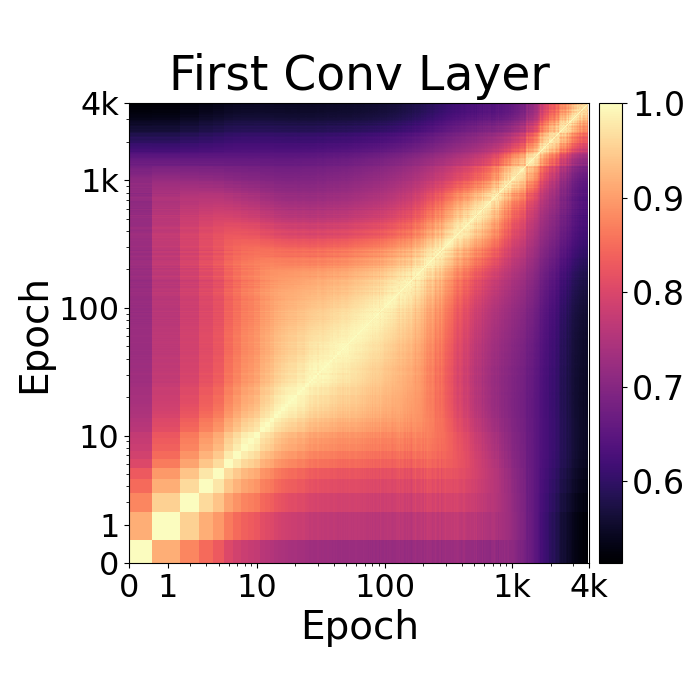}
\label{}}
\subfloat[]{
\includegraphics[width=0.19\textwidth]{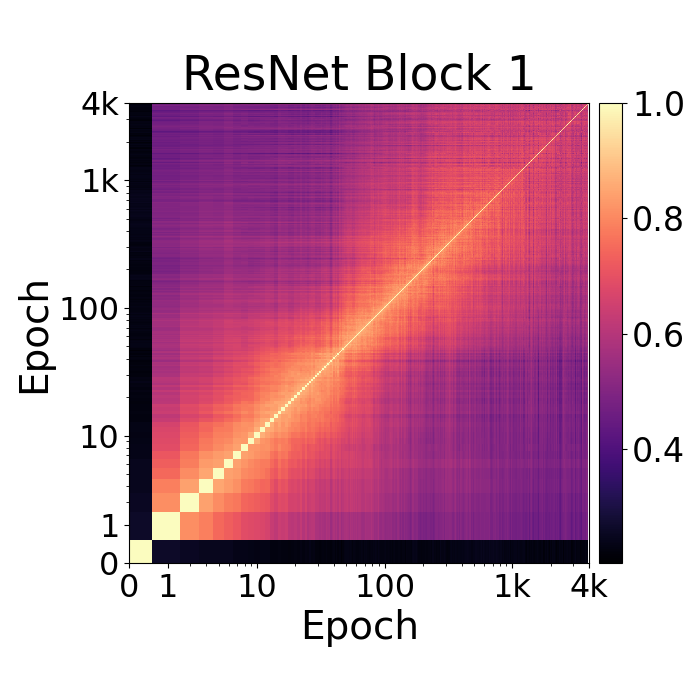}
\label{}}
\subfloat[]{
\includegraphics[width=0.19\textwidth]{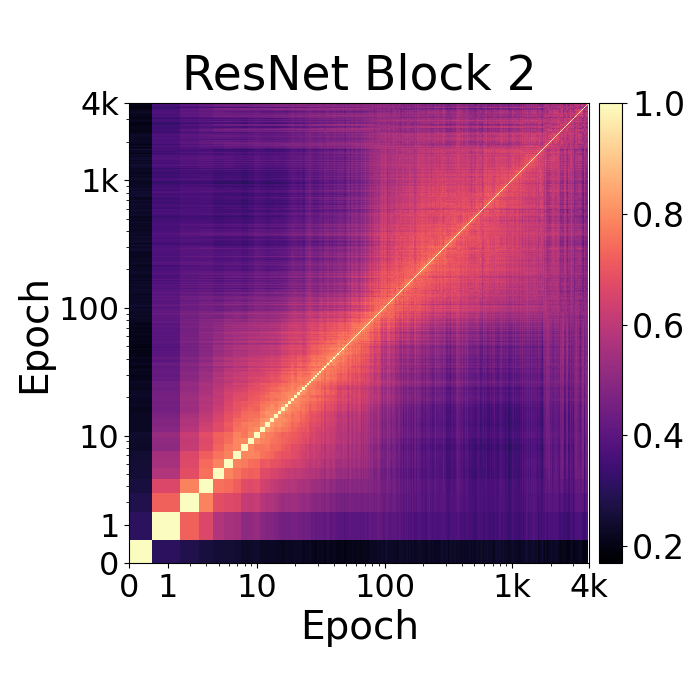}
\label{}}
\subfloat[]{
\includegraphics[width=0.19\textwidth]{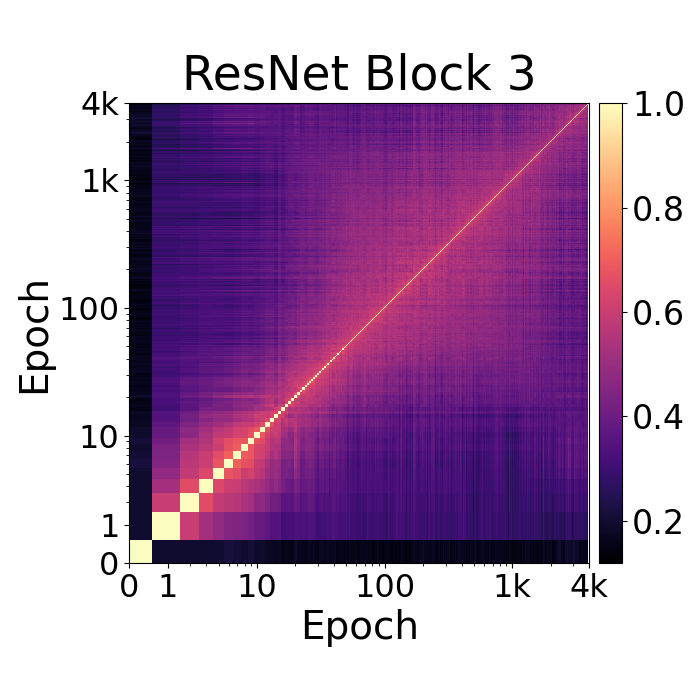}
\label{}}
\subfloat[]{
\includegraphics[width=0.19\textwidth]{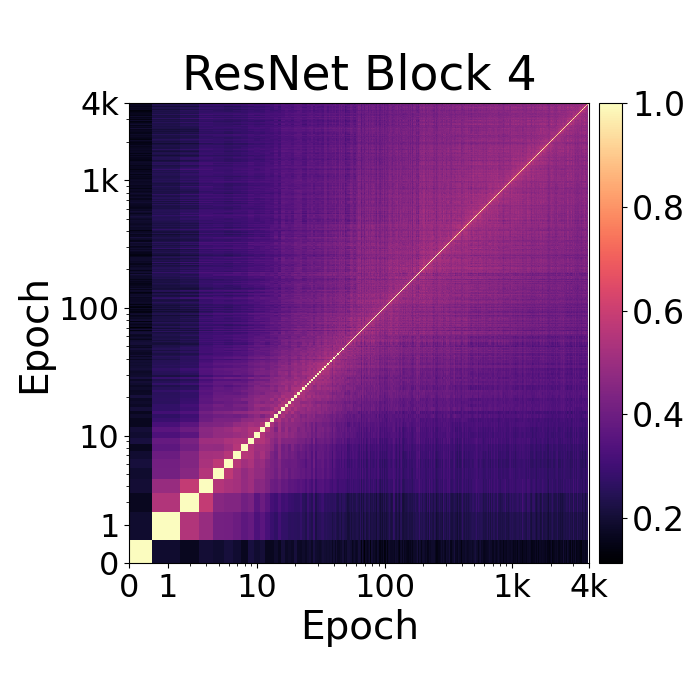}
\label{}}
\caption{Decision regions similarity (DRS) of linear classifier probes. Evaluations for ResNet-18, trained on SVHN without label noise using Adam Optimizer with learning rate of 0.0001. Each subfigure shows the DRS of a specific layer in the ResNet-18 during its training. The DNN error curves are provided in Fig.~\ref{ResNet18_k_64_adam_svhn_noiseless_error_curve}.}  \label{fig:DRS_svhn_resnet_noise_0_Adam}
\end{figure*}

\begin{figure*}[]
  \centering
  \subfloat[]{
\includegraphics[width=0.19\textwidth]{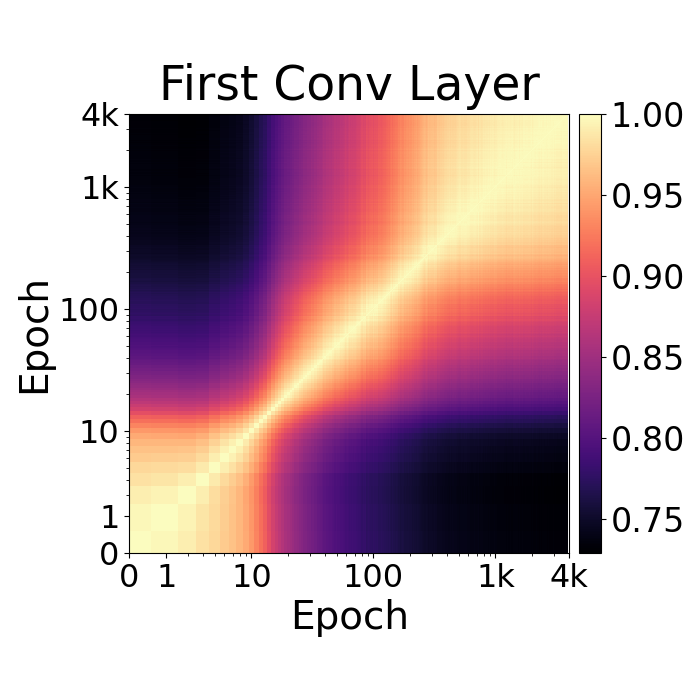}
\label{}}
\subfloat[]{
\includegraphics[width=0.19\textwidth]{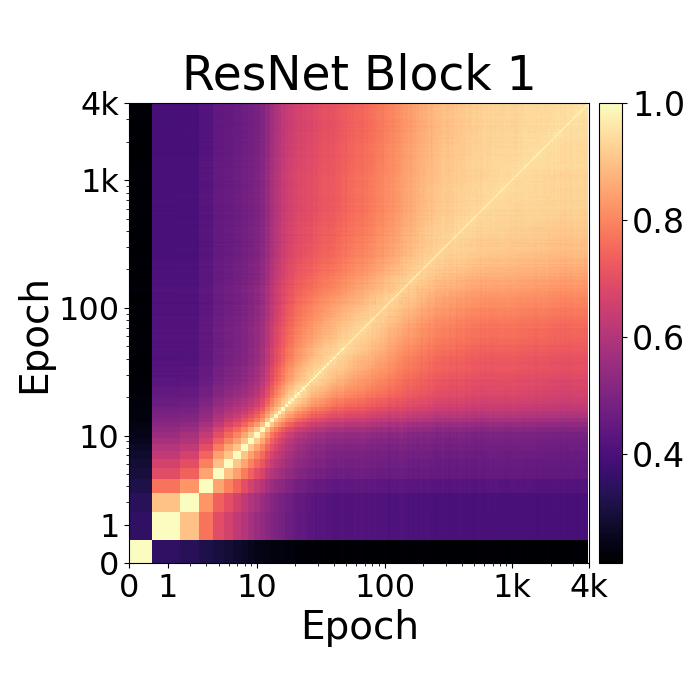}
\label{}}
\subfloat[]{
\includegraphics[width=0.19\textwidth]{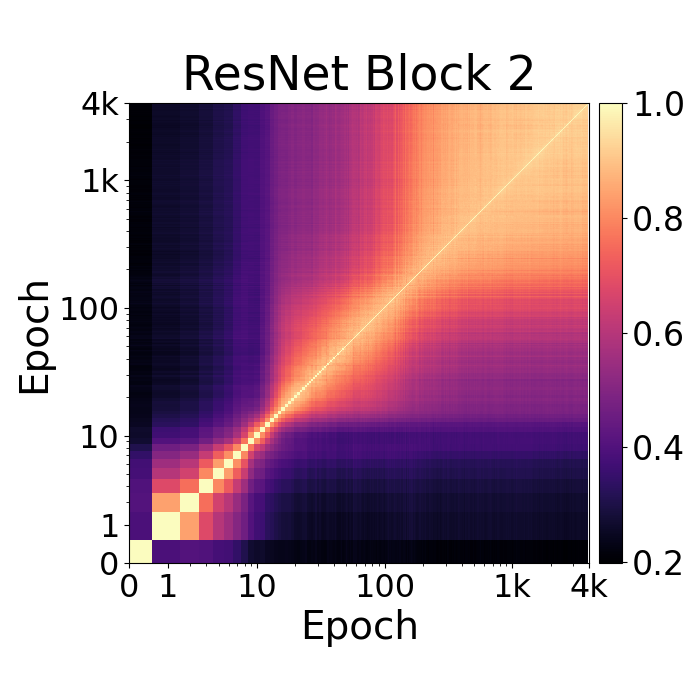}
\label{}}
\subfloat[]{
\includegraphics[width=0.19\textwidth]{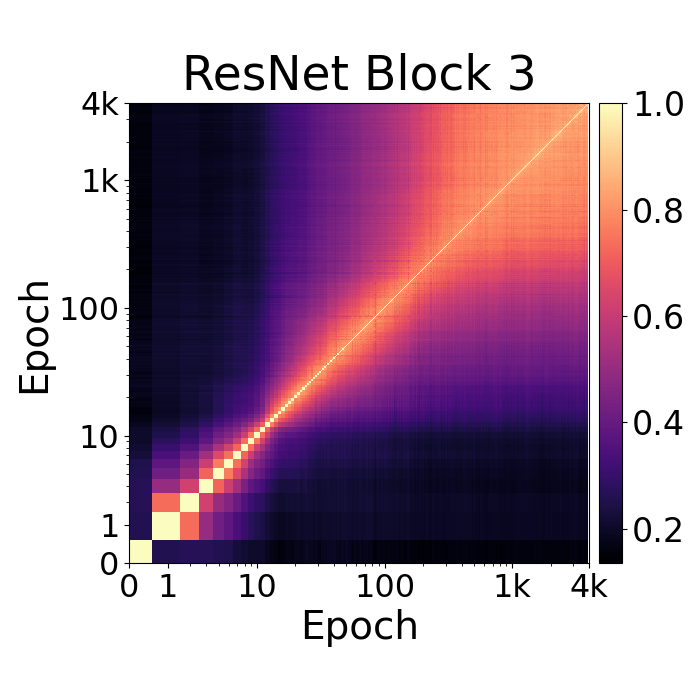}
\label{}}
\subfloat[]{
\includegraphics[width=0.19\textwidth]{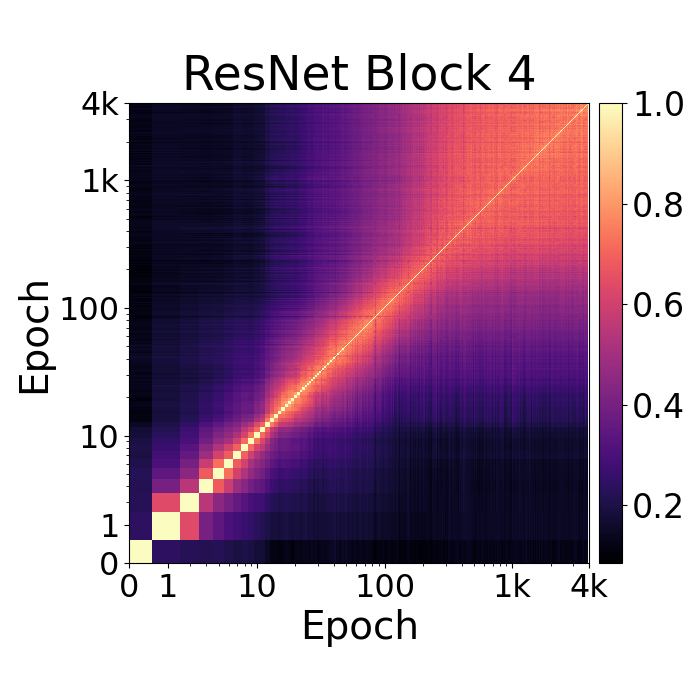}
\label{}}
\caption{Decision regions similarity (DRS) of linear classifier probes. Evaluations for ResNet-18, trained on SVHN without label noise using SGD Optimizer with learning rate of 0.001. Each subfigure shows the DRS of a specific layer in the ResNet-18 during its training. The DNN error curves are provided in Fig.~\ref{ResNet18_k_64_sgd_svhn_noiseless_error_curve}.}  \label{fig:DRS_svhn_resnet_noise_0_SGD}
\end{figure*}

In Figs.~\ref{fig:lp_resnet_noise_0}, \ref{fig:lp_resnet_noise_20}-\ref{fig:DRS_svhn_resnet_noise_0_SGD} we provide DRS evaluations in addition to the DRS evaluations on ResNet-18 in Fig.~\ref{fig:lp_resnet_noise_0} of the main paper. 

\begin{figure*}[]
  \centering
  \subfloat[]{
\includegraphics[width=0.19\textwidth]{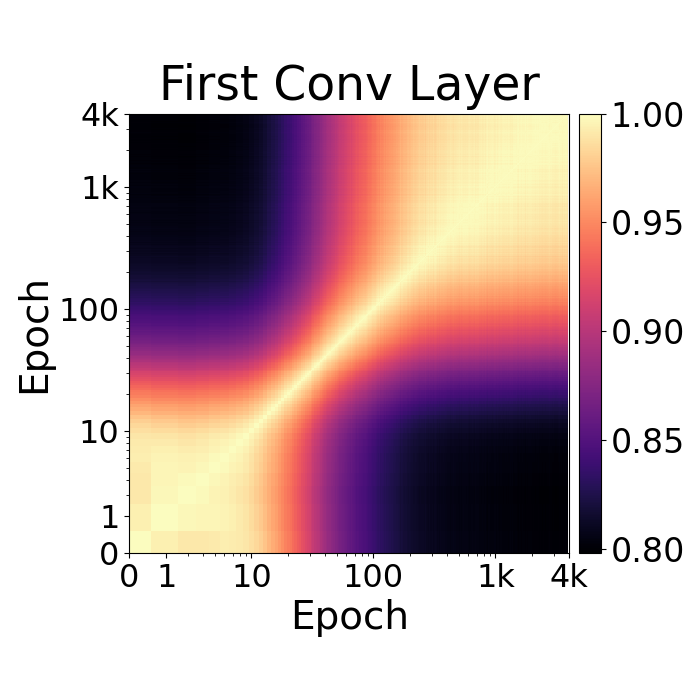}
\label{}}
\subfloat[]{
\includegraphics[width=0.19\textwidth]{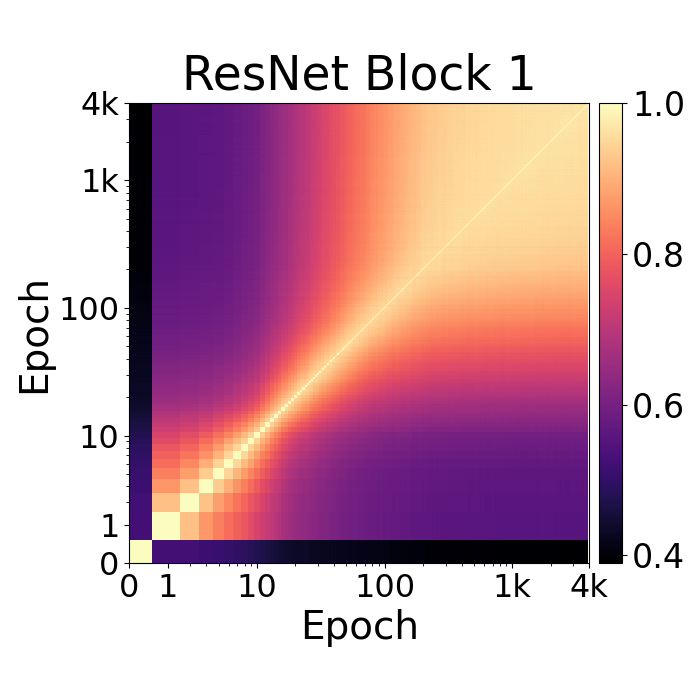}
\label{}}
\subfloat[]{
\includegraphics[width=0.19\textwidth]{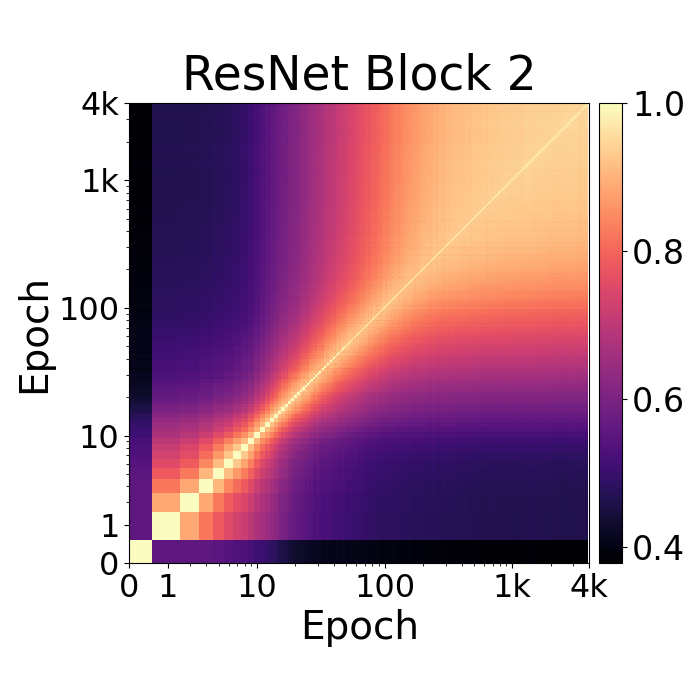}
\label{}}
\subfloat[]{
\includegraphics[width=0.19\textwidth]{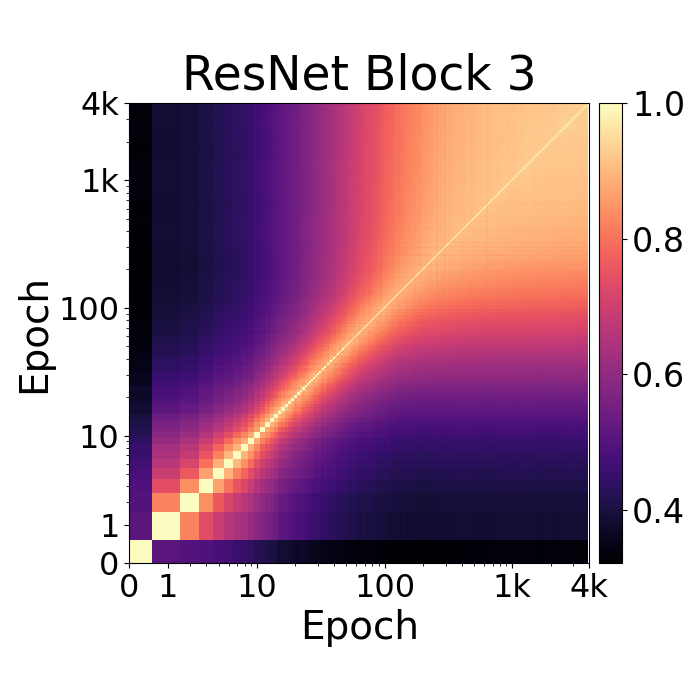}
\label{}}
\subfloat[]{
\includegraphics[width=0.19\textwidth]{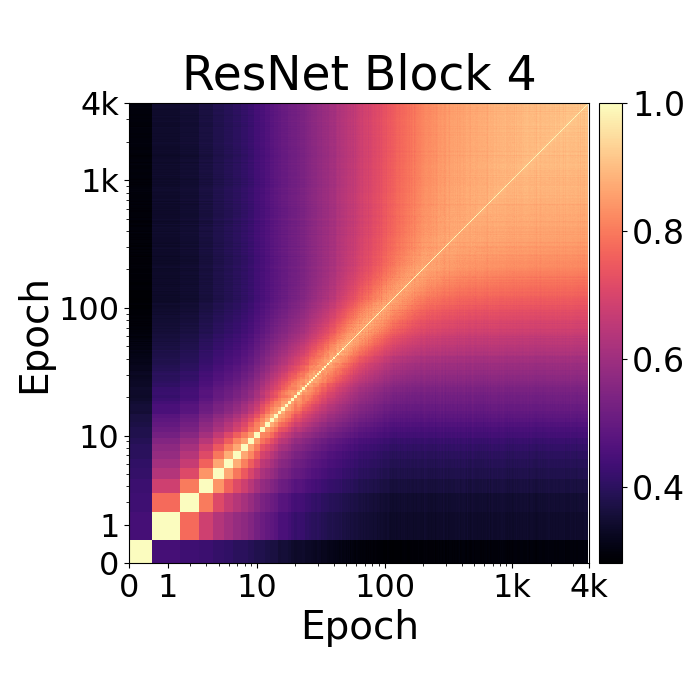}
\label{}}
\caption{Decision regions similarity (DRS) of linear classifier probes. Evaluations for ResNet-18, trained on CIFAR-10 without label noise using SGD Optimizer with learning rate of 0.001. Each subfigure shows the DRS of a specific layer in the ResNet-18 during its training. The DNN error curves are provided in Fig.~\ref{cifar10_resnet18_noise_0_k_64_sgd_lr_0.001_momentum_0_bs_128_cifar10_noiseless_error_curve}.}  \label{fig:DRS_cifar_resnet_noise_0_sgd}
\end{figure*}

\section{Quantitative Evaluation of Decision Region Fragmentation of Layer Probes and DNN Output}
\label{appendix:sec:Quantitative Evaluation of Decision Region Fragmentation of Layer Probes and DNN Output}

\begin{figure*}[]
  \centering
    \subfloat[]{
    \includegraphics[width=0.23\textwidth]{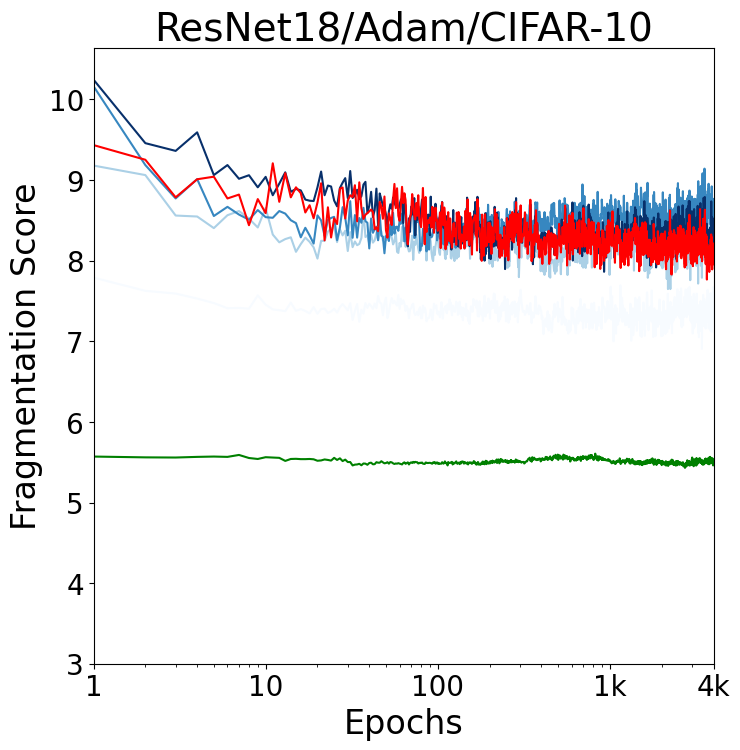}
    \label{fig:frag_score_adam_cifar10}}
    \subfloat[]{
    \includegraphics[width=0.23\textwidth]{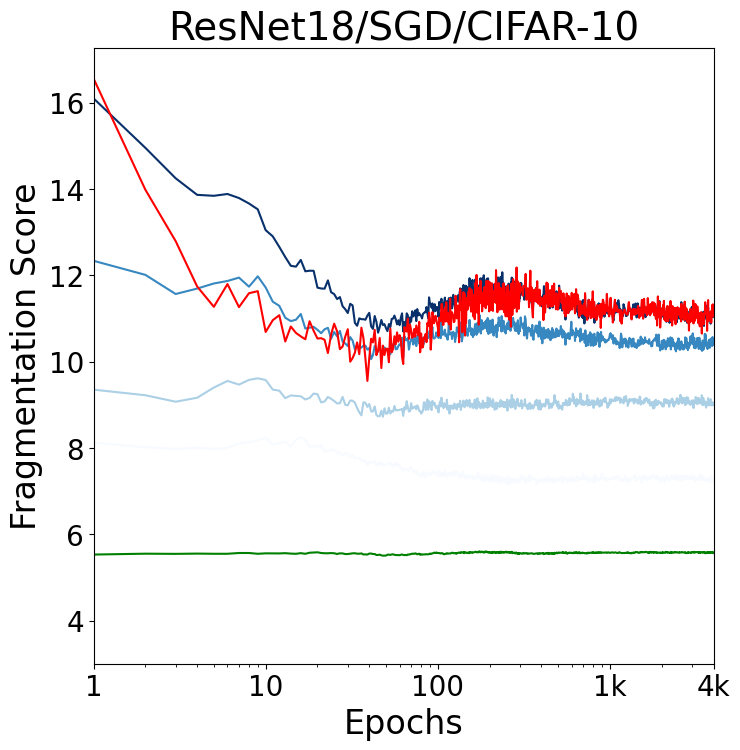}
    \label{fig:frag_score_sgd_no_momentum_cifar10}}
    \subfloat[]{
    \includegraphics[width=0.24\textwidth]{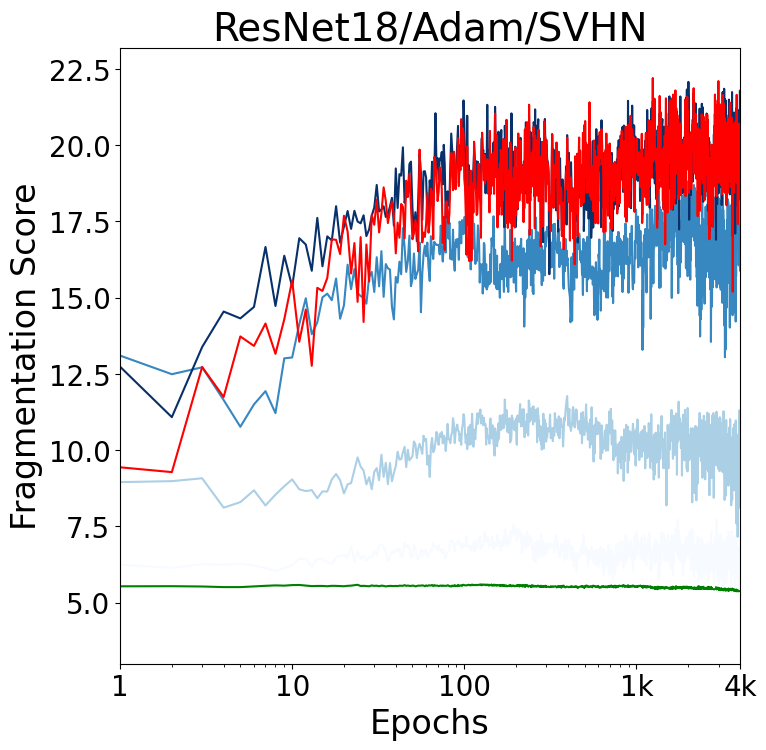}
    \label{fig:frag_score_adam_svhn}}
    \subfloat[]{
    \includegraphics[width=0.23\textwidth]{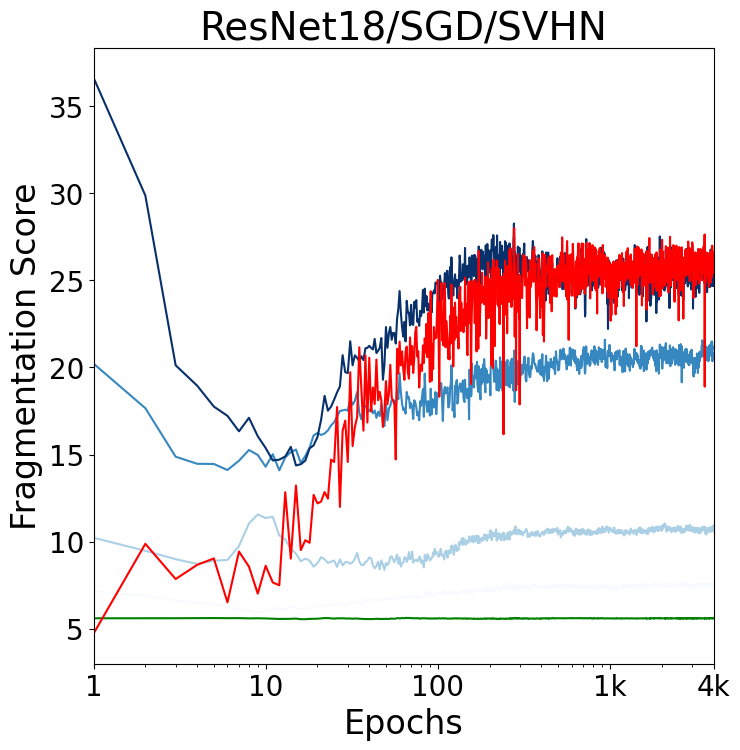}
    \label{fig:frag_score_SGD_no_momentum_svhn}}\\
\includegraphics[width=0.22\textwidth]{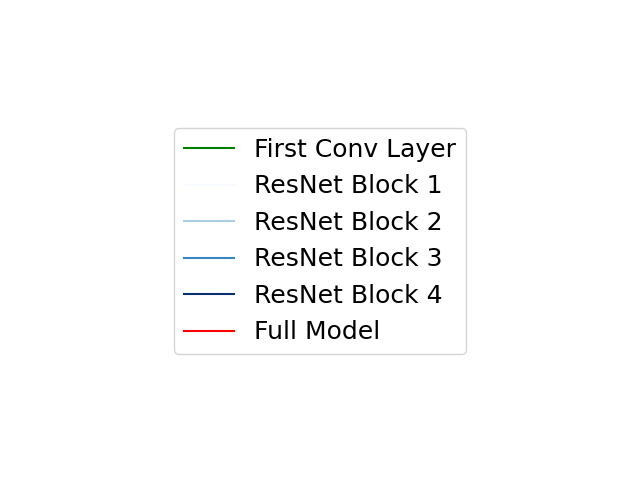}
\caption{Average number of fragments per plane (``fragmentation score'') of the decision regions of the linear probes and DNN prediction output, for ResNet-18 trained on noiseless CIFAR-10 and SVHN datasets.}
  \label{fig:fragmentation_score_resnet}
\end{figure*}
\begin{figure*}[]
  \centering
  \subfloat[]{
    \includegraphics[width=0.23\textwidth]{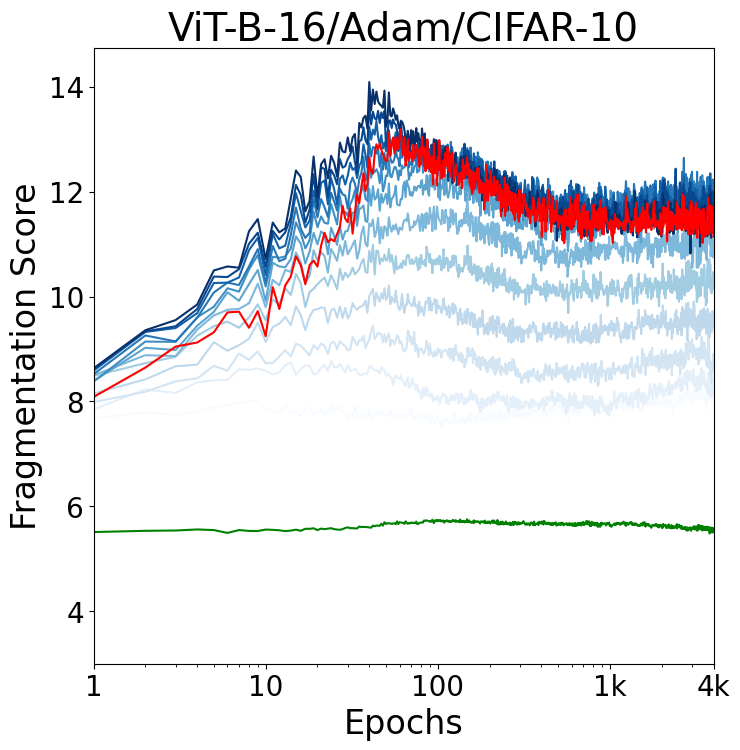}
    \label{fig:frag_score_adam_cifar10_vit}}
  \subfloat[]{
    \includegraphics[width=0.23\textwidth]{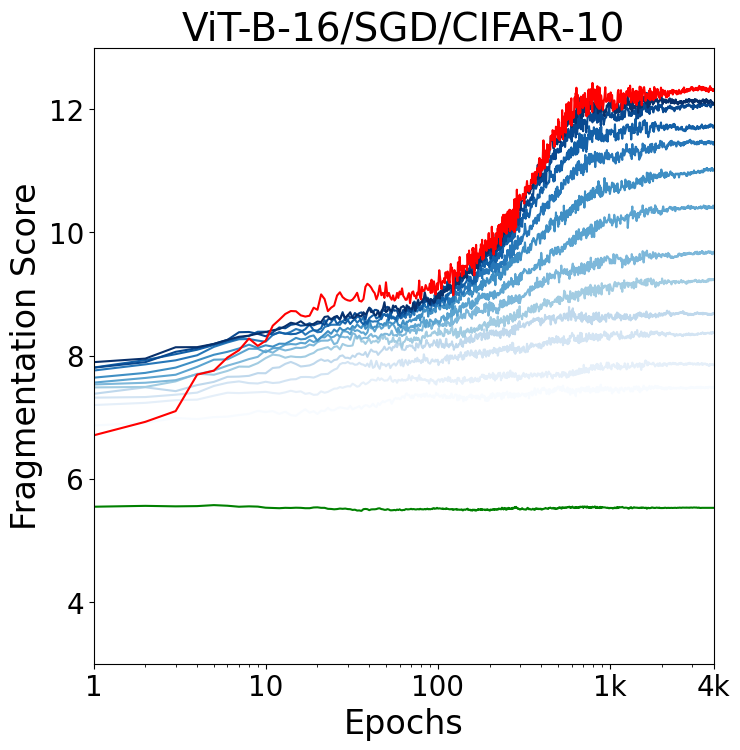}
    \label{fig:frag_score_sgd_no_momentum_vit_cifar10_vit}}
    \includegraphics[width=0.23\textwidth]{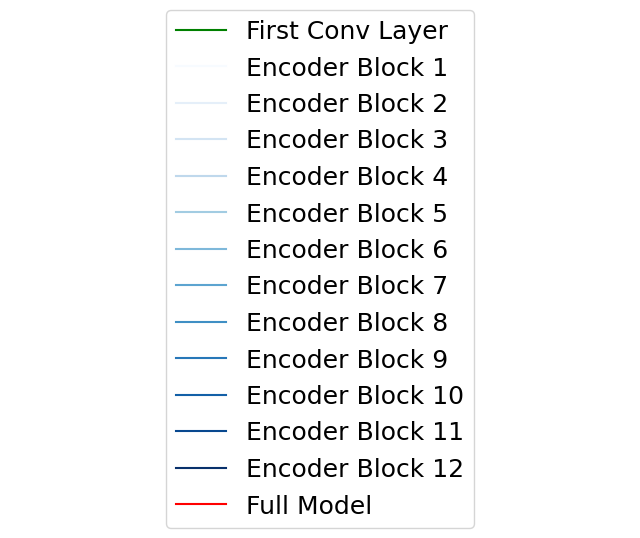}
\caption{Average number of fragments per plane (``fragmentation score'') of the decision regions of the linear probes and DNN prediction output, for ViT-B/16 trained on noiseless CIFAR-10 dataset.}
  \label{fig:fragmentation_score_vit}
\end{figure*}

A main insight from our CKA similarity analysis is that the three training phases of representation learning are more distinguishable in ViT than in ResNet, and in SGD than in Adam training. Now, we give a closer look on this behavior using the decision regions \textit{geometry} of the linear classifier probes, which reflects the task-specific features. 

We propose to evaluate the fragmentation level of the decision regions of the layerwise linear probes during training. Namely, in each training epoch of interest, the fragmentation level is the average number of decision regions in the plane segments that were defined in Section \ref{subsubsec:The DRS Computation Procedure}. The computation technique is based on counting the connected regions in the examined planes (which are actually truncated planes according to their definition) as proposed by \citet{somepalli2022can} for evaluating fragmentation-level of \textit{fully-trained} DNNs in their standard classification functionality of the DNN prediction output (i.e., not its layers and without probes, and not during training like we do here).

The evaluation of fragmentation level for intermediate layers and the overall DNN output \textit{during training} is a new experiment that we propose here. The fragmentation level elucidates the complexity of the learned representations \textit{during training}, which can further emphasize the difference among the training phases.

In Figs.~\label{fig:fragmentation_score_resnet}, \ref{fig:fragmentation_score_vit} we show the fragmentation evaluations for ResNet-18 and ViT, each trained using both Adam and SGD on CIFAR-10 without label noise; the ResNet-18 was examined also for training on the SVHN dataset. 

For ResNet-18 training on noiseless CIFAR-10, the fragmentation score does not change significantly during training (especially for Adam compared) (Figs.~\ref{fig:frag_score_adam_cifar10}, \ref{fig:frag_score_sgd_no_momentum_cifar10}). This is in contrast to the training of ResNet-18 on noiseless SVHN (Figs.~\ref{fig:frag_score_adam_svhn}, \ref{fig:frag_score_SGD_no_momentum_svhn}) and training of ViT on noiseless CIFAR-10 (Figs.~\ref{fig:frag_score_adam_cifar10_vit}, \ref{fig:frag_score_sgd_no_momentum_vit_cifar10_vit}), for which the intermediate/deep layers and the DNN output have increasing fragmentation score during the training phase of memorizing atypical examples, and then in the perfect fitting training phase the fragmentation score stabilizes (after some reduction in the case of Adam training of ViT, see Fig.~\ref{fig:frag_score_adam_cifar10_vit}). This shows that the complexity of the learned representations behave differently in each of the training phases, depending on the optimizer and architecture, which in turn translates also to the highly distinguishable training phases in the representation similarity analysis in this paper. 

\section{Additional Visualization of The Linear Probes Decision Regions}
\label{appendix:sec:Additional Visualization of The Linear Probes Decision Regions}
\begin{figure*}[t]
    \centering
        \includegraphics[width=0.8\textwidth]{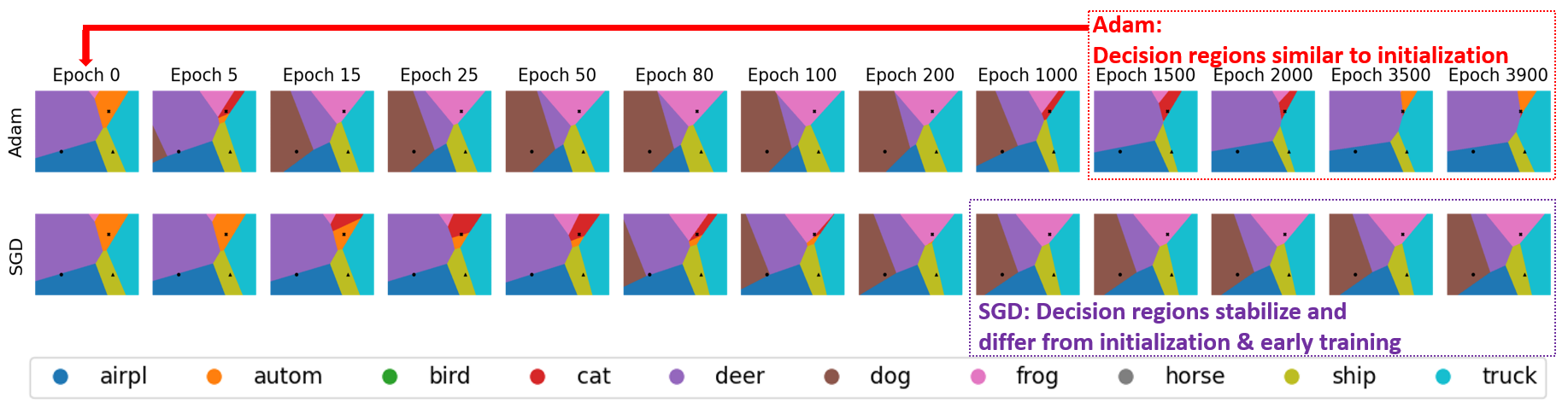}
\caption{Visualization of decision regions of the linear probe of the first convolution layer in ResNet-18 (trained on CIFAR-10).}
\label{fig:cifar10_resnet_first_layer_probe_vis_appendix_1}
\end{figure*}

\begin{figure*}[t]
    \centering
        \includegraphics[width=0.8\textwidth]{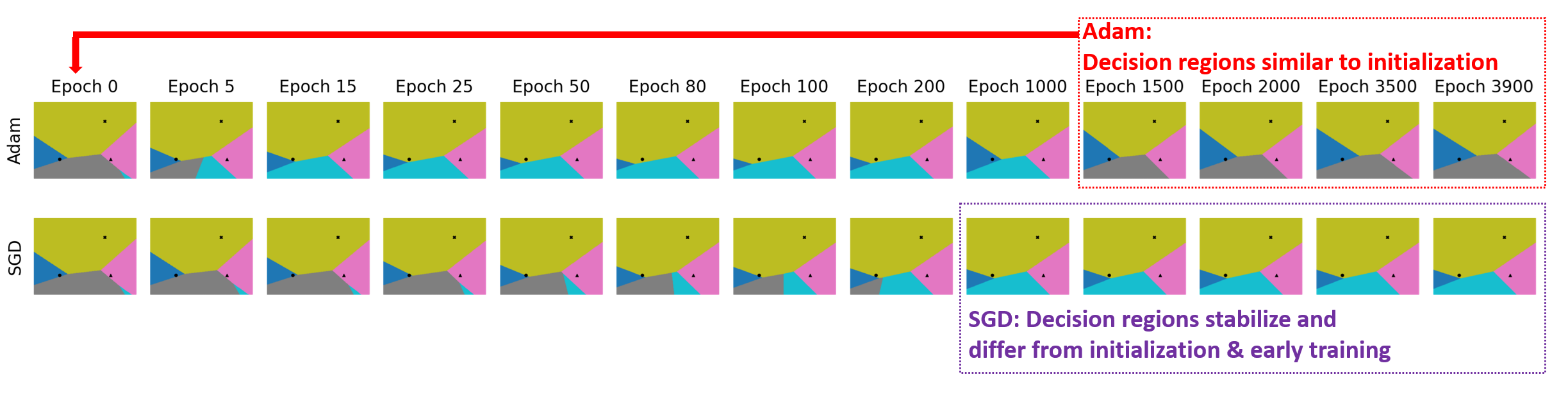}
\caption{Visualization of decision regions of the linear probe of the first convolution layer in ResNet-18 (trained on CIFAR-10).}
\label{fig:cifar10_resnet_first_layer_probe_vis_appendix_2}
\end{figure*}

\begin{figure*}[]
    \centering
        \includegraphics[width=0.9\textwidth]{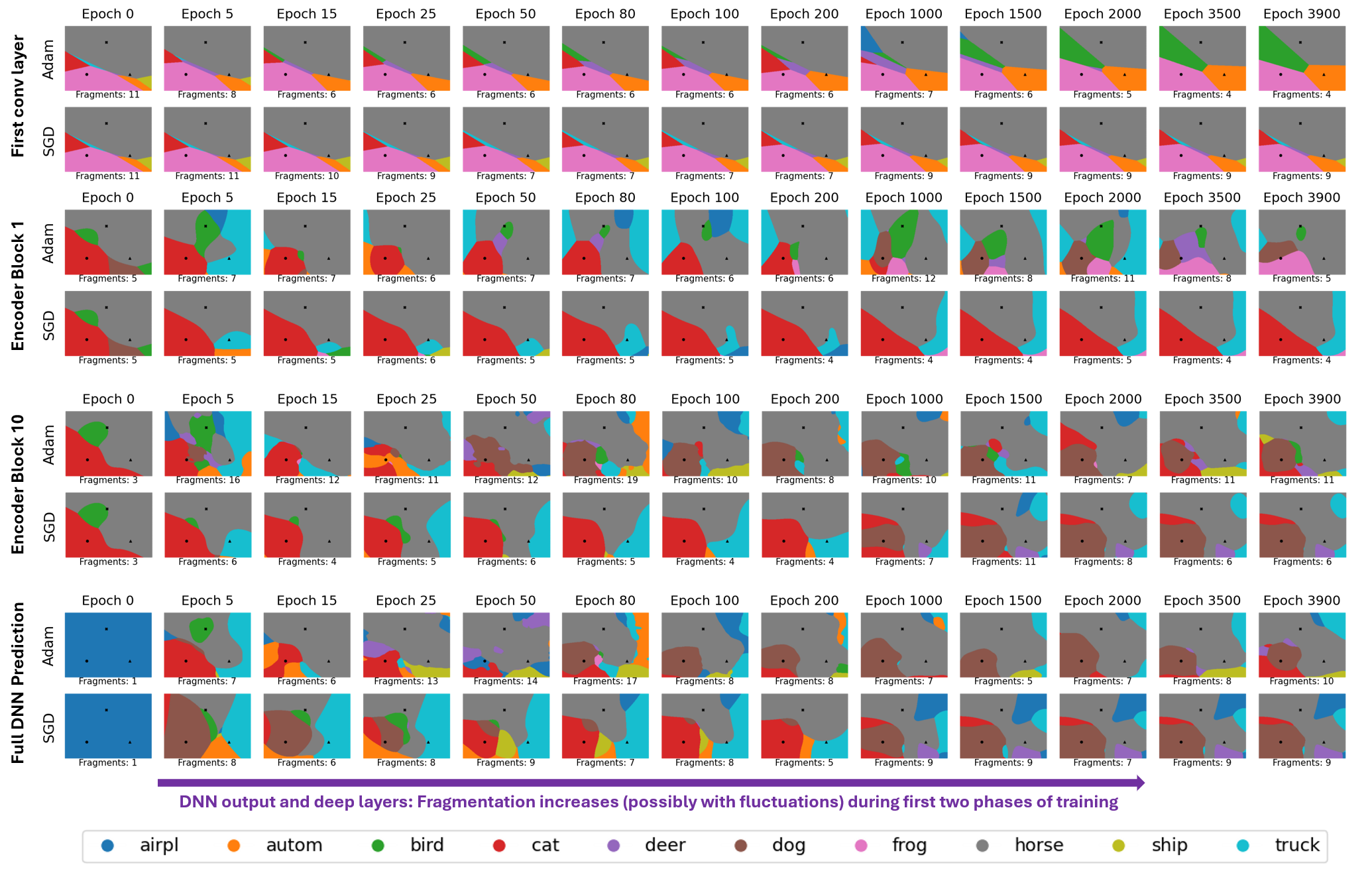}
\caption{Visualization of the decision regions of the layerwise linear probes and full DNN prediction in ViT-B/16 (trained on CIFAR-10).}
   \label{fig:ViT_CIFAR10_plane_visual_for_appendix_1}
\end{figure*}

\begin{figure*}[]
  \centering
    \includegraphics[width=0.9\textwidth]{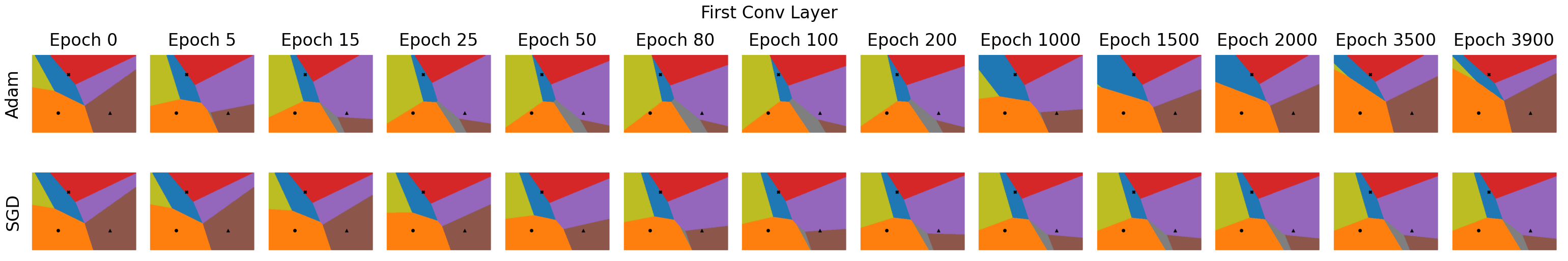}
    \\
    ~
    \\
    \includegraphics[width=0.9\textwidth]{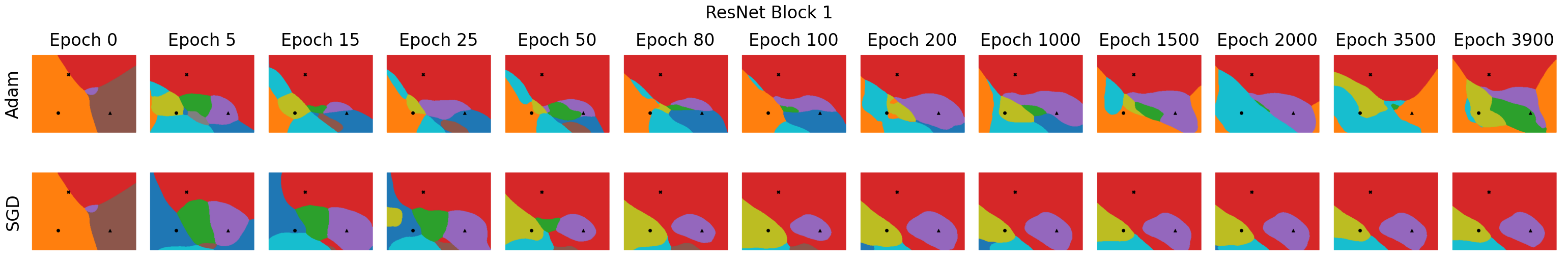}
    \\
    ~
    \\
    \includegraphics[width=0.9\textwidth]{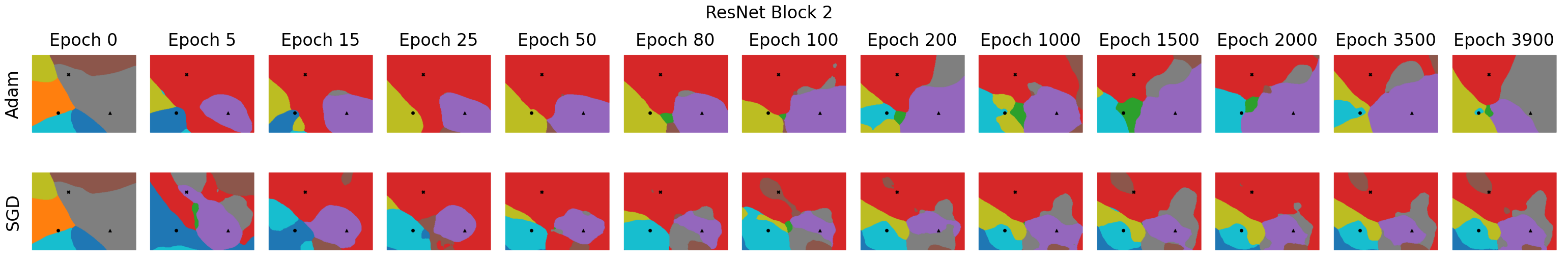}
    \\
    ~
    \\
    \includegraphics[width=0.9\textwidth]{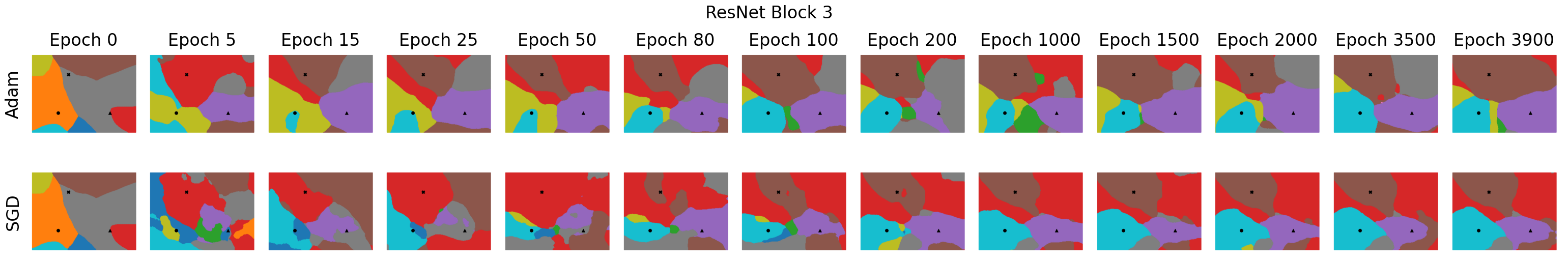}
    \\
    ~
    \\
    \includegraphics[width=0.9\textwidth]{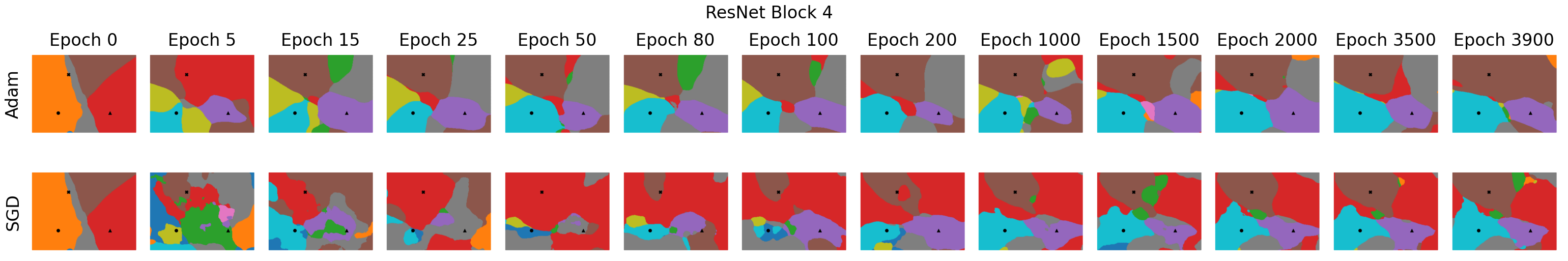}
    \\
    ~
    \\
    \includegraphics[width=0.9\textwidth]{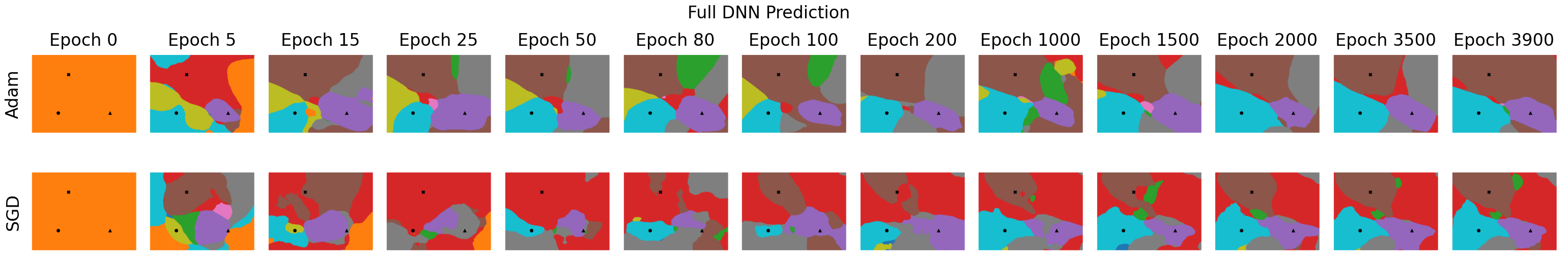}
    \\
    \includegraphics[width=0.9\textwidth]{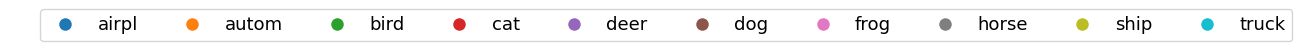}
\caption{Visualization of the decision regions of all layers in ResNet18 trained on CIFAR-10 dataset.}
  \label{fig:resnet_full_vis_triplet_494}
\end{figure*}

Here we provide additional visualizations of decision regions: First layer of ResNet-18 (Figs.~\ref{fig:cifar10_resnet_first_layer_probe_vis_appendix_1}, \ref{fig:cifar10_resnet_first_layer_probe_vis_appendix_2}), ViT-B/16 layers and DNN output (Fig.~\ref{fig:ViT_CIFAR10_plane_visual_for_appendix_1}), and ResNet-18 layers and DNN output (Fig.~\ref{fig:resnet_full_vis_triplet_494}).

\end{document}